\documentclass[twoside]{article}

\usepackage[accepted]{aistats2022}

\usepackage[round]{natbib}

\usepackage[utf8]{inputenc} %
\usepackage[T1]{fontenc}    %
\usepackage[colorlinks=true,citecolor=gray,linkcolor=blue]{hyperref}  %
\usepackage{url}            %
\usepackage{booktabs}       %
\usepackage{amsfonts}       %
\usepackage{nicefrac}       %
\usepackage{microtype}      %
\usepackage{xcolor}         %
\usepackage{ifthen}
\usepackage[figuresright]{rotating}

\usepackage{graphicx}
\usepackage{subfigure}

\usepackage{verbatim}
\usepackage{esvect}
\usepackage{amsmath}
\usepackage{graphicx}
\usepackage{breakurl}

\usepackage{enumitem}
\usepackage{multirow}
\usepackage{graphicx}
\usepackage{amsthm}
\usepackage{amssymb}
\usepackage{bbm}
\usepackage{overpic}
\usepackage[rightcaption]{sidecap}
\usepackage{pdflscape}

\usepackage{dsfont}

\usepackage{todonotes}

\newcommand{\ignore}[1]{}

\newcommand{\N}[0]{{\mathds N}}
\newcommand{\R}[0]{{\mathds R}}
\newcommand{\Psymb}[0]{{\mathds P}}
\newcommand{\Ex}[0]{{\mathds E}}

\newcommand{\charfct}{\mathds{1}}

\newtheorem{lemma}{Lemma}
\newtheorem{proposition}{Proposition}
\newtheorem{theorem}{Theorem}

\DeclareMathOperator{\MAE}{MAE}
\DeclareMathOperator{\DPviol}{DP-viol}
\DeclareMathOperator{\EOviol}{EO-viol}
\DeclareMathOperator{\Fairviol}{Fair-viol}
\DeclareMathOperator{\sign}{sgn}
\DeclareMathOperator{\Obj}{Obj}
\DeclareMathOperator{\Cost}{Cost}

\begin{document}

\renewcommand{\thefootnote}{\fnsymbol{footnote}}

\runningtitle{Pairwise Fairness for Ordinal Regression}
\runningauthor{M. Kleindessner, S. Samadi, M. B. Zafar, K. Kenthapadi, C. Russell }

\twocolumn[

\aistatstitle{Pairwise Fairness for Ordinal Regression}

\aistatsauthor{Matthäus Kleindessner \And Samira Samadi \And Muhammad Bilal Zafar }

\aistatsaddress{ Amazon \And  MPI-IS Tübingen \And Amazon } 

\aistatsauthor{Krishnaram Kenthapadi\footnotemark[1] \And Chris Russell}

\aistatsaddress{ Fiddler AI \And Amazon } ]

\begin{abstract}
We initiate 
the study of fairness for ordinal regression. We 
adapt 
two fairness notions previously  considered in %
fair ranking and propose a strategy for training a 
predictor 
that is 
approximately 
fair according to 
either %
notion. 
Our predictor 
has the form of 
a threshold model, composed of a scoring function and a set of thresholds, and our strategy 
is based on a reduction to fair binary classification for learning the scoring function and 
local search 
for 
choosing 
the thresholds. 
 We provide generalization guarantees on the error and fairness violation of our predictor,  and we illustrate  the effectiveness of our approach in extensive~experiments.
\end{abstract}

\footnotetext[1]{Work done at Amazon.}
\renewcommand{\thefootnote}{\arabic{footnote}}

\section{INTRODUCTION}\label{section_intro}
As machine learning (ML) algorithms have become an integral part 
of numerous human-centric domains, 
they have shown a 
wide 
range of concerning behaviors:
criminal recidivism tools 
mislabeling 
black low-risk defendants as high-risk 
\citep{Angwin2018}; 
word2vec embeddings encoding 
stereotypes such as ``a father is to a doctor as a mother is to a nurse'' \citep{bolukbasi2016man}; 
and 
facial recognition systems 
having 
lower accuracy on darker-skinned or female faces 
\citep{Buolamwini2018},   
 to name just the most prominent examples. These \mbox{observations} have led~to~the~study of fairness in ML \citep{barocas-hardt-narayanan}, and in the past 
 years 
 numerous 
 ML 
 tasks 
 have been studied~from 
 a fairness perspective. 
 While most
 works consider 
 (binary) 
 classification \citep[e.g.,][]{hardt2016equality}, fair algorithms have also been developed for regression \citep[e.g.,][]{berk2017,agarwal2019}
 and several unsupervised learning tasks \citep[e.g., ][]{fair_clustering_Nips2017,celis2018,celis_fair_ranking,ekstrand2018,samira2018,fair_k_center_2019,fair_SC_2019,ghadiri2021}.
 
While  
ordinal regression is a widespread task in ML and data science, this is the first work to address the problem of fair ordinal regression.
Ordinal regression 
\citep[aka ordinal classification---see Sec.~\ref{section_ordinal_reg} for a formal description;][]{gutierrez2016ordinal}
refers to 
multiclass classification
 over an ordered  label set.  
Consider a hiring scenario, where given a job applicant's features, such as 
education, we want to predict a label in 
$\{$bad, okay, good, excellent$\}$. 
Clearly, 
it is less critical to misclassify an excellent 
applicant 
as good than misclassifying  
an 
excellent 
one
as bad. Algorithms for ordinal~regression take such order information into account and  
typically assign different costs to  different misclassifications.
However, 
the order information not only entails that different 
kinds of 
misclassifications should be weighted differently;
it 
also carries fairness implications. For example, an 
applicant that should be scored as okay 
would feel treated unfairly if misclassified as bad, but probably not if they were misclassified as good. Or it might be acceptable to misclassify all excellent 
applicants 
from a minority group as good as long 
as 
the 
excellent 
applicants from other (majority) groups are 
 misclassified 
 in the same way.

In this paper, we initiate the study of fairness for ordinal regression by making the following \textbf{contributions}:

\begin{itemize}[leftmargin=*]%
    \item We propose 
    two 
    pairwise fairness notions for ordinal regression, which we 
    adapt  
    from 
    the literature on 
    fair ranking~(Sec.~\ref{subsection_fairness_notions}; see Sec.~\ref{section_related_work} 
    for related work). We also sketch other possible fairness notions (Sec.~\ref{section_discussion}).
    
    \item Focusing on the two pairwise notions, 
    we propose a  
    strategy to learn a 
    predictor that is 
    accurate and approximately fair according to either notion (Sec.~\ref{subsection_learning_scoring_function}~\& Sec.~\ref{subsection_choosing_thresholds}). 
    Our approach  is  based on a reduction to fair binary classification and some local search procedure. It 
    allows us to control the trade-off between accuracy and fairness that typically exists via a parameter. 
    
    \item
    We provide generalization bounds on the error and the fairness violation of 
    our predictor (Sec.~\ref{subsection_generalization}).

    \item
    We shed light on computational 
    aspects of our strategy  (Sec. \ref{subsection_choosing_thresholds}), discuss its limitations (Sec.~\ref{subsection_limitations}), 
    and prove that some simpler alternatives can perform arbitrarily worse (Sec.~\ref{subsection_both_steps_necessary}). 
    
    \item We perform extensive experiments 
    and 
    compare to 
    ``unfair'' state-of-the-art 
    methods in oder to illustrate the effectiveness of our approach (Sec.~\ref{section_experiments} \& App.~\ref{appendix_partB}).
\end{itemize}

\section{SETUP \& FAIRNESS NOTIONS}\label{section_ordinal_reg}

\paragraph{Setup in Ordinal Regression (Without Fairness)} 
Given an input point 
$x\in \mathcal{X}$, we want to accurately predict 
its label $y\in \mathcal{Y}$, where the label set $\mathcal{Y}$ is  totally ordered 
(such as  bad $\prec$ okay $\prec$ good $\prec$ excellent in the example of Sec.~\ref{section_intro}). 
W.l.o.g., we 
identify $\mathcal{Y}$ with $[k]:=\{1,\ldots,k\}$, that is $\mathcal{Y}=[k]$.
Accuracy is measured 
according to 
a cost matrix~$C\in \R_{\geq 0}^{k\times k}$, where $C_{i,j}$ is the cost that we incur when misclassifying a point with true label~$i$ as having label~$j$. 
The order on $\mathcal{Y}$ entails 
information about the proximity of labels, and we assume that the misclassification cost~$C_{i,j}$ 
can only increase
as 
$j$ moves away from $i$: 
formally, we assume $C$ to have V-shaped rows, that is $C_{i,j-1}\geq C_{i,j}$ for $2\leq j\leq i$ and $C_{i,j}\leq C_{i,j+1}$ for $i\leq j\leq k-1$, with $C_{i,i}=0$, $i\in[k]$. 
One 
choice for $C$ is the binary cost matrix~$C_{i,j}=\charfct\{i\neq j\}$,   used in
 standard 
multiclass classification, 
but not taking 
label order into account. A 
popular 
choice in ordinal regression is the absolute cost matrix~$C_{i,j}=|i- j|$.

Datapoints~$(x,y)$ are 
assumed to be 
drawn i.i.d. from a joint distribution~$\Psymb$ on 
$\mathcal{X}\times [k]$. 
Given a set of 
training points $((x_i,y_i))_{i=1}^n$, 
our goal is to learn a predictor 
$f:\mathcal{X}\rightarrow [k]$  
with small expected cost~$\Ex_{(x,y)\sim\Psymb}~C_{y,f(x)}$. If 
$C$ is the absolute cost matrix,
we refer to the expected cost as mean absolute~error~(MAE). 

\subsection{Pairwise Fairness Notions}\label{subsection_fairness_notions}

When concerned about fairness, 
we assume that datapoints come with~a~protected attribute~$a\in\mathcal{A}$ and consider $(x,y,a)\sim \Psymb$, where $\Psymb$ is now a distribution on %
$\mathcal{X}\times [k] \times \mathcal{A}$. 
The attribute~$a$  encodes sensitive information such as gender or race,  
and we assume $\mathcal{A}$ is finite.\footnote{
Our 
notions fall into the category of group fairness 
as opposed to individual fairness 
\citep{friedler2016}.
 }  
As before, 
we want to 
learn a predictor 
$f:\mathcal{X}\rightarrow [k]$ 
with small expected cost~$\Ex_{(x,y,a)\sim\Psymb}~C_{y,f(x)}$, but~now~$f$ 
should also be fair
with respect to
the protected attribute
$a$ 
and
avoid discrimination against 
any protected group 
(i.e., all individuals that share a value of $a$). 
Although satisfying our proposed fairness notions would 
be 
easier if we granted 
$f$ 
access to $a$,  
we assume  
 that $f$ does not 
 use 
 $a$.\footnote{Technically, our formulation 
 allows 
 $f$ to use $a$ by encoding $a$ as part of $x$. Using $a$ would make the problem easier 
 since 
 it would allow us to learn group-dependent thresholds (cf. Section~\ref{section_our_strategy}).}
 This  
 is a common assumption in the literature on fair ML, 
  because otherwise we would commit disparate treatment \citep{barocas-hardt-narayanan} or 
 the attribute 
 $a$ 
 might not 
 even  
 be available at test time. 
 At training time or when assessing the fairness of a predictor, we assume~access~to~$a$.

There are several plausible ways to define what it means that 
$f$ 
is fair. 
In this paper we focus on two pairwise fairness notions as we  formalize them in the following. 
We 
sketch 
other 
options~in~\mbox{Section~\ref{section_discussion}}.

\vspace{3pt}
\textbf{Pairwise demographic parity (DP)}~~ 
The predictor~$f$ satisfies pairwise DP if for all $\tilde{a},\hat{a}\in\mathcal{A}$ 
\begin{align}\label{fairness_notion_DP}
\begin{split}
&\Psymb[f(x_1)> f(x_2) | a_1=\tilde{a},a_2=\hat{a}]=\\
&~~~~~~~~~~~~\Psymb[f(x_1)< f(x_2) | a_1=\tilde{a},a_2=\hat{a}],
\end{split}
\end{align}
where the probability is over 
$(x_1,y_1,a_1)$ and $(x_2,y_2,a_2)$ 
being independent samples from  $\Psymb$  and potentially the randomness of the predictor $f$. 

\vspace{4pt}
\textbf{Pairwise equal opportunity (EO)}~~ 
Similarly, we say that $f$ satisfies pairwise EO if for all $\tilde{a},\hat{a}\in\mathcal{A}$ 
\begin{align}\label{fairness_notion_EO}
\begin{split}
&\Psymb[f(x_1)> f(x_2) | a_1=\tilde{a},a_2=\hat{a},y_1>y_2]=\\
&~~~~~~~~~~\Psymb[f(x_1)< f(x_2) | a_1=\tilde{a},a_2=\hat{a},y_1<y_2].  
\end{split}
\end{align}

To motivate these two 
definitions, assume that for 
a point~$x$ 
to be classified (e.g., a job applicant) some predictions~$f(x)$ are %
preferable to 
others and that the order of preference coincides with the order on $\mathcal{Y}$ (e.g., a prediction 
``good'' 
is preferred to 
``okay''). 
Pairwise~DP 
asks 
that it is as likely for a point sampled from one protected group to be preferred over a point sampled from a second group as it is for the converse to happen  
(e.g., a female applicant being considered better than a male applicant happens just as likely as a male applicant being considered better than a female one). This is a pairwise analogue of the standard fairness notion of DP \citep{kamiran2011}, which requires the prediction to be 
independent of the protected attribute, that is  $\Psymb[f(x)=\tilde{y} | a=\tilde{a}]=\Psymb[f(x)=\tilde{y} | a=\hat{a}]$ for all $\tilde{y}\in\mathcal{Y}$ and $\tilde{a},\hat{a}\in\mathcal{A}$, in the sense that pairwise DP requires the order of the predicted labels for an input pair to be independent of whether the first point is from group~$\tilde{a}$ and the second one from group~$\hat{a}$, or the other way round.
Pairwise EO 
asks for the same condition as pairwise DP, 
but conditioned on the order of the ground-truth labels 
(e.g., a female applicant that is de facto better than a male competitor 
being 
considered better happens just as likely as a male, de facto better applicant being considered better than a female competitor).
This is a pairwise analogue of the standard notion of 
EO 
\citep{hardt2016equality}, which considers $y=1$ to be the preferred 
outcome 
and requires
$\Psymb[f(x)=1 | a=\tilde{a},y=1]=\Psymb[f(x)=1 | a=\hat{a},y=1]$ for all $\tilde{a},\hat{a}\in\mathcal{A}$.
The analogue of $y=1$ being the preferred outcome is having a higher label in~our~case.  

One might wonder whether there is also a pairwise analogue of the %
notion of equalized odds \citep{hardt2016equality}, which 
is a stricter notion than standard EO and 
requires that  $\Psymb[f(x)=\hat{y} | a=\tilde{a},y=\tilde{y}]=\Psymb[f(x)=\hat{y} | a=\hat{a},y=\tilde{y}]$,  
$\tilde{a},\hat{a}\in\mathcal{A}$,  $\tilde{y},\hat{y}\in\mathcal{Y}$. 
 A pairwise analogue yields only a slightly 
 stronger notion than pairwise EO (see App.~\ref{appendix_pairwise_equalized_odds}), and we do not consider~it~here.

Fairness notions 
similar to pairwise DP / EO 
as defined in \eqref{fairness_notion_DP} 
or \eqref{fairness_notion_EO} 
were recently introduced in the context of ranking (where $y\in[n]$ and $f:\mathcal{X}\rightarrow[n]$ if the dataset to be ranked comprises $n$ elements; 
\citealp{beutel2019fairness}), 
bipartite ranking ($y\in\{0,1\}$ and $f:\mathcal{X}\rightarrow\R$; \citealp{kallus2019}) and standard regression ($y\in\R$ and $f:\mathcal{X}\rightarrow\R$; \citealp{narasimhan2020pairwise}). However, %
they have not been studied in the context of ordinal regression. %
We 
discuss related work in Section~\ref{section_related_work}.

We make some remarks on pairwise DP and EO as defined in \eqref{fairness_notion_DP} and \eqref{fairness_notion_EO} (proofs 
can be found 
in App.~\ref{appendix_proofs_remarks}):

\begin{itemize}[leftmargin=*]%
\item Any constant predictor~$f(x)=i$,
for some $i\in[k]$, 
satisfies 
both fairness notions. 
The perfect predictor~$f(x)=y$ satisfies pairwise EO, but
not necessarily 
pairwise DP.
    
\item 
When $k=2$, pairwise DP and standard DP 
are equivalent. For general $k$, standard DP implies pairwise~DP, but not the other way round.

\item  
Standard EO is neither sufficient nor necessary for pairwise EO, even for $k=2$. 
The 
notion of equalized odds  implies pairwise EO for $k=2$, but not 
for~$k>2$.
\end{itemize}

Consequently, pairwise DP is a less restrictive 
fairness 
notion than standard DP, 
 and pairwise EO and standard EO~/ equalized odds are incomparable. We believe that 
 each of these notions can be the most appropriate one in a 
 given 
 scenario 
 (cf. Sec.~\ref{section_discussion}). For example, assume we predict the quality of 
 cars offered on a marketplace on a scale from one to ten, and we want to be fair 
 with respect to 
 different vendors. Vendors would care more that their cars do not unjustifiably loose the comparisons with the cars offered by their competitors (thus requesting pairwise EO), rather than caring whether the probability of a car with true quality~``3'' being predicted a quality of~``7'' is the same for all vendors (corresponding to equalized odds).   
Standard EO, which is primarily designed for binary classification, is not appropriate 
in this scenario since it requires declaring a single preferred outcome and treats all other outcomes as equal, and both standard and pairwise DP do not take the different ground-truth qualities into account.
In Appendix~\ref{appendix_example_tshirts} we 
present 
a similar example in more~detail.

In the next section we propose a strategy to learn a predictor that is 
accurate and approximately satisfies either pairwise DP or EO. We measure the amount by which a predictor $f$ violates the two 
notions by $\DPviol$ or $\EOviol$, 
defined as the 
maximum 
absolute difference between the left and right sides of \eqref{fairness_notion_DP}~or~\eqref{fairness_notion_EO}; e.g., 
\begin{align*}%
    &\DPviol(f;\Psymb)=
    \max_{\tilde{a},\hat{a}\in\mathcal{A}} |\Psymb[f(x_1)> f(x_2) | a_1=\tilde{a},a_2=\hat{a}]\\
    &~~~~~~~~~~~~~~~~-\Psymb[f(x_1)< f(x_2) | a_1=\tilde{a},a_2=\hat{a}]|.
\end{align*}
On a dataset~$\mathcal{D}$ of size $n$ we can evaluate $\DPviol(f;\mathcal{D})$ in time~$\mathcal{O}(n+k|\mathcal{A}|^2)$ and 
$\EOviol(f;\mathcal{D})$ in time~$\mathcal{O}(n+k^2|\mathcal{A}|^2)$, assuming evaluating $f$ on 
$x$ 
takes time 
$\mathcal{O}(1)$.
If $\mathcal{D}$ is an i.i.d. sample from $\Psymb$, for a given~$f$, we have 
$|\Fairviol(f;\Psymb)-\Fairviol(f;\mathcal{D})|\leq  M \sqrt{{(\log\frac{|\mathcal{A}|^2}{\delta})}/{n}}$ 
with probability $1-\delta$ over the sample~$\mathcal{D}$, 
where $M$ is some constant  
(see App.~\ref{appendix_convergnce}).

\section{LEARNING A FAIR 
MODEL 
}\label{section_our_strategy}

Threshold models  are a common approach to ordinal regression  
\citep{gutierrez2016ordinal}. 
They consist of a scoring function $s: x\mapsto s(x)\in \R$ and $k-1$ thresholds $\theta_1\leq\ldots\leq\theta_{k-1}$
 and predict label~$i$ for 
 input point~$x$ 
 if $s(x)\in(\theta_{i-1},\theta_i]$ (with $\theta_0=-\infty$ and $\theta_k=+\infty$). 
 Our proposed 
  strategy 
 consists of learning 
 an accurate and approximately fair threshold model 
 in a two-step approach: first, we learn a 
 scoring function~$s$ that 
 approximately 
 satisfies pairwise DP or EO 
 (i.e., $s$ 
 satisfies \eqref{fairness_notion_DP} or \eqref{fairness_notion_EO} with $f$ replaced by $s$) via a reduction to fair binary classification. %
 Next, 
 we choose thresholds that result in an approximately  fair predictor~$f$. 
 Although our definitions of pairwise DP and EO allow for a randomized 
 $f$, 
 we 
 aim to 
 learn a deterministic~$f$ since in 
  decisions 
  strongly 
  affecting 
 humans' lives 
 (such as hiring) 
 randomization is 
 often seen as problematic \citep{cotter_neurips2019}.~From~now~on,~we~assume~that~$\mathcal{X}\subseteq\R^d$. The proofs of all 
 statements are~in~Appendix~\ref{appendix_proof_reduction_to_binary_classification}-\ref{appendix_local_search}.

\subsection{Learning a Fair Scoring Function}\label{subsection_learning_scoring_function}

Ideally, 
$s$ 
satisfies $s(x_1)< s(x_2) \Leftrightarrow y_1< y_2$.  
Considering a linear scoring function $s(x)=w\cdot x$ for some $w\in\R^d$, we have 
$s(x_1)< s(x_2)
\Leftrightarrow\sign (w\cdot(x_1-x_2))=-1$. 
Given training data~$\mathcal{D}=((x_i,y_i))_{i=1}^n$,
the well-known approach of \citet{herbrich_ordinal_regression} 
 to ordinal regression
exploits this equivalence and learns 
a linear scoring function, parameterized by $w$, by learning a linear classifier, also parameterized by $w$, that aims to solve the binary classification problem with training dataset $\mathcal{D}'=\{
(x',y')
=(x_i-x_j,\sign(y_i-y_j)):i,j\in[n],y_i\neq y_j\}$.
Concretely, the SVM-based algorithm of \citeauthor{herbrich_ordinal_regression} chooses $w$ to minimize the hinge loss on $\mathcal{D}'$. We adapt their approach by learning a linear classifier on $\mathcal{D}'$ that 
approximately 
satisfies 
 a fairness 
 constraint 
 closely related to 
standard DP or EO (cf. Sec.~\ref{subsection_fairness_notions}) on $\mathcal{D}'$ 
with respect to some attribute~$a'$.  
As we prove, this implies 
that the scoring function approximately satisfies 
pairwise DP or EO on $\mathcal{D}=((x_i,y_i,a_i))_{i=1}^n$:

\begin{proposition}[Reduction to fair binary classification]\label{proposition_reduction_to_fair_binary_classification}
Let $\mathcal{D}=((x_i,y_i,a_i))_{i=1}^n\subseteq \R^d\times[k]\times\mathcal{A}$ and $\mathcal{D}'=\{(x',y',a')=(x_i-x_j,\sign(y_i-y_j),(a_i,a_j)):i,j\in[n],y_i\neq y_j\}\subseteq \R^d\times\{-1,1\}\times \mathcal{A}^2$. For $w\in\R^d$, let $c_w$ be the binary classifier 
$c_w(x)=\sign(w\cdot x)$
and $s_w$ be the scoring function $s_w(x)=w\cdot x$. 
We have, for arbitrary $\varepsilon$, 
\begin{align}\label{eq_reduction}
\begin{split}
 &\max_{\tilde{a},\hat{a}\in\mathcal{A}} |\Psymb_{(x',y',a')\sim\mathcal{D}'}[c_w(x')=1|a'=(\tilde{a},\hat{a})]-\\
 &~~~~~~~~~\Psymb_{(x',y',a')\sim\mathcal{D}'}[c_w(x')=1|a'=(\hat{a},\tilde{a})]|\leq \varepsilon
\end{split}
\end{align}
if and only if 
$s_w$ satisfies $\DPviol(s_w;\mathcal{D})\leq\varepsilon$  
conditioned on $y_1\neq y_2$, and 
we have 
\eqref{eq_reduction} with the probabilities conditioned on $y'=1$ 
if and only if 
$s_w$ satisfies $\EOviol(s_w;\mathcal{D})\leq \varepsilon$.
\end{proposition}

When 
aiming for 
pairwise DP, 
having 
$\DPviol(s_w)\leq\varepsilon$ conditioned on $y_1\neq y_2$ is sufficient for our purposes 
since
we can 
hope for 
$s_w(x_1)\approx s_w(x_2)$ for $y_1=y_2$. This allows us to construct a predictor~$f$ with  $f(x_1)=f(x_2)$ for $y_1=y_2$ and $\DPviol(f)\leq\varepsilon$.
We note that 
the fairness constraint~\eqref{eq_reduction} is 
easier to satisfy than standard DP since it only compares probabilities conditioned on $a'=(\tilde{a},\hat{a})$ and $\bar{a}'=(\hat{a},\tilde{a})$, respectively, but not arbitrary $a',\bar{a}'\in\mathcal{A}^2$. 
For learning 
$c_w$ that satisfies \eqref{eq_reduction} (or the related constraint in case of EO) on $\mathcal{D}'$, we can utilize the existing approaches 
for fair binary linear classification 
from the literature, 
such as the reduction approach of \citet{agarwal_reductions_approach} 
or the various relaxation-based approaches \citep[e.g.,][]{donini2018,wu2019,zafar2019}.
These methods allow us to choose any of the standard convex 
loss functions for performing constrained
regularized   
empirical risk minimization 
(ERM) 
to learn $c_w$, 
and they also allow us to control 
the trade-off between optimizing for accuracy and satisfying the fairness~constraint~(i.e.,~controlling~$\varepsilon$~in~\eqref{eq_reduction}).

\subsection{Learning Fair Thresholds}\label{subsection_choosing_thresholds}

Once 
$s$ 
is learned, we require thresholds $\theta_1\leq\ldots\leq\theta_{k-1}$. We 
propose to 
choose thresholds that minimize a weighted combination of the misclassification cost and the fairness violation 
of the resulting 
predictor 
on the training data~$\mathcal{D}=((x_i,y_i,a_i))_{i=1}^n$: %
if $f(\cdot\,;s,\boldsymbol\theta)$ denotes the threshold model-predictor with scoring function~$s$ and thresholds~$\boldsymbol\theta=(\theta_1,\ldots,\theta_{k-1})$, we solve
\begin{align}\label{threshold_objective}
\min_{\boldsymbol\theta} \frac{1}{n}\sum_{i=1}^n C_{y_i,f(x_i;s,\boldsymbol\theta)}+
\lambda\cdot \Fairviol(f(\cdot\,;s,\boldsymbol\theta);\mathcal{D}),
\end{align} 
where $\Fairviol$ is 
a 
generic 
notation 
for $\DPviol$ or $\EOviol$ as defined 
in 
Section~\ref{subsection_fairness_notions} and    
$\lambda\geq 0$ 
controls   
the extent to which we optimize for accuracy (i.e., small cost) and fairness, respectively. In our experiments, we 
choose $\lambda$ aligned with $\varepsilon$ from the previous section.

Perhaps surprisingly, the 
problem~\eqref{threshold_objective} can be solved in 
polynomial time 
using dynamic programming:
\begin{proposition}[\eqref{threshold_objective} can be solved in poly-time]\label{proposition_dynamic_programming}
If $|\mathcal{A}|=2$, then for any $\lambda\geq 0$ and $\Fairviol\in\{\DPviol,\EOviol\}$, we can find optimal thresholds solving \eqref{threshold_objective} in time~$\mathcal{O}(n^6k^3)$.
\end{proposition}

The dynamic programming approach of Proposition~\ref{proposition_dynamic_programming} can be generalized to $|\mathcal{A}|>2$, but then has  running time exponential in $|\mathcal{A}|$. 
Even when $|\mathcal{A}|=2$, the running time is prohibitively high 
in practice.
Instead, we 
have to 
make use of heuristics to efficiently find local minima of \eqref{threshold_objective}. 
We propose 
to perform a local search,
 moving one threshold~$\theta_i$ at a time. 
We can implement 
such 
local search 
with running time 
$\mathcal{O}(n|\mathcal{A}|^2)$ per iteration in case of pairwise DP or  
$\mathcal{O}(nk|\mathcal{A}|^2)$
in case of pairwise EO 
(see 
App.~\ref{appendix_local_search}
for details).

\subsection{Generalization Guarantees}\label{subsection_generalization}

We provide generalization bounds on the $\MAE$ and the fairness violation of the predictor learned by our strategy 
($B_R(0)$ denotes the 
ball $\{x\in\R^d:\|x\|_2\leq R\}$):

\begin{theorem}[Generalization bounds]\label{proposition_generalization}
Assume that the training data $\mathcal{D}$ is 
an i.i.d. sample 
from 
a 
distribution~$\Psymb$ on $B_R(0)\times[k]\times\mathcal{A}$ 
with $\Psymb[a=\tilde{a}]\geq\beta>0$ for all $\tilde{a}\in\mathcal{A}$. 
Assume that we learn a scoring function~$s$ 
as outlined in 
Section~\ref{subsection_learning_scoring_function}  
by means of constrained regularized ERM 
with Ivanov $l_2$-regularization $\|w\|_2^2\leq \nu$, and thresholds as outlined in %
Section~\ref{subsection_choosing_thresholds}. 
There exists a constant~$M$ 
such that for any $\gamma>0$ and  $0<\delta<1$, 
our learned predictor $f=f(\cdot\,;s,\boldsymbol\theta)$ satisfies with probability at least $1-\delta$ over the training sample~$\mathcal{D}$ of size $n$, for $n$ sufficiently~large, 
\begin{align*}
    &\MAE(f;\Psymb)\leq \widehat{L}_{\mathcal{D}}^{\gamma}(s,\boldsymbol\theta)+(k-1)\cdot\\
    &~~~~~\sqrt{\frac{M}{n}\left(\frac{R^2\nu}{\gamma^2}\ln(n) +\ln\left(2\frac{(k-1)R\sqrt{\nu}}{\gamma\delta}\right) \right)}
 \end{align*}
and
 \begin{align*}
       &\DPviol(f;\Psymb)\leq \DPviol(f;\mathcal{D})+ M k^2\cdot\\ &~~~~\sqrt{\left({d+\log\frac{4|\mathcal{A}|}{\delta}}\right)\cdot\left[{\left(1-\sqrt{\frac{2\log\frac{4|\mathcal{A}|}{\delta}}{n\beta}}\right)n\beta}\right]^{-1}}.
\end{align*}
$\widehat{L}_{\mathcal{D}}^{\gamma}(s,\boldsymbol\theta)$ is the empirical $\gamma$-margin loss 
(see App.~\ref{appendix_generalization}).
\end{theorem}

The statement for $\EOviol$ 
is provided in Appendix~\ref{appendix_generalization}.

\subsection{Limitations of our Approach}\label{subsection_limitations}

We briefly discuss limitations of our approach 
and potential remedies. 

\begin{itemize}[leftmargin=*]%
\item 
Jointly optimizing the objective function in \eqref{threshold_objective} over the scoring function~$s$ and thresholds~$\boldsymbol\theta$ could potentially yield better results 
than our 
two-step approach.
This is an interesting direction for future~work.
\item 
Our approach learns a linear scoring function, which might be too restrictive in some settings. If the method for learning the fair binary classifier~$c_w$ can be kernelized,
then our approach can be kernelized 
similarly 
to the unfair approach of \citet{herbrich_ordinal_regression}. 
Alternatively, 
random features \citep{rahimi2007random} could be used to increase expressiveness.
\item
Starting with a training set $\mathcal{D}$ of size $n$
our approach constructs a training set $\mathcal{D}'$ of size $n^2$, which 
seems natural in light of the pairwise nature of our fairness notions, but  
might be infeasible for large~$\mathcal{D}$.  
 If so, 
 we 
 subsample a training set for learning the classifier~$c_w$. 
\end{itemize}

\subsection{Is Fairness Needed in Both Steps?}\label{subsection_both_steps_necessary}

One might wonder whether we must 
enforce 
fairness on the scoring function~$s$, or whether~it~would~suffice to enforce fairness only when choosing the thresholds. Indeed, since any constant predictor~is perfectly fair (cf. Sec.~\ref{subsection_fairness_notions}), no matter what $s$ is, we can always choose thresholds~so~as~to~achieve any desired level of fairness. However, 
if $s$ is fair to some extent, 
there tend to be 
more choices of thresholds that yield a fair predictor
(see App.~\ref{appendix_simulations_fairness_Score_vs_threshold}
 for some simulations)  
and hence we can 
get 
a more accurate predictor.   
Similarly, one can wonder whether it 
suffices 
to enforce fairness only when learning the scoring function. The next lemma shows that an approach in which~we~enforce~fairness in only one of the two 
learning 
steps can perform arbitrarily~worse~compared~to~our~two-step~approach:

\begin{lemma}[Enforcing fairness in both steps can be  necessary]\label{lemma_fairness_in_both_steps_necessary}
There exist datasets in $\R^2\times [4]\times \{0,1\}$ and a class~$S$ of scoring functions $s:\R^2\rightarrow\R$ that~show:

\vspace{-3pt}
\begin{enumerate}[leftmargin=*]%
    \item Choosing a scoring function $s$ 
    that minimizes the number of label flips, but is not required to satisfy pairwise DP (EO), and subsequently thresholds that minimize the $\MAE$ 
    under the constraint that the 
    resulting 
    predictor satisfies pairwise DP (EO) can result in a $\MAE$ that is arbitrarily high compared to %
    enforcing pairwise DP (EO) both when choosing 
    $s$
    and 
    the~\mbox{thresholds}.

    \item Choosing 
    a scoring function 
    $s$ to minimize the number of label flips under the constraint that $s$ satisfies pairwise DP (EO) and subsequently thresholds that minimize the $\MAE$, 
    but are not required to yield a predictor satisfying pairwise DP (EO), 
    can result in a predictor with $\DPviol=0.5$ ($\EOviol=1$). In contrast, thresholds required to yield a predictor satisfying pairwise DP (EO), 
    will always give $\DPviol=0$ ($\EOviol=0$).
\end{enumerate}
\end{lemma}

The examples 
used 
to prove Lemma~\ref{lemma_fairness_in_both_steps_necessary} are worst-case examples that might not occur in practice.
However, 
in Section~\ref{section_experiments} we 
also 
present experiments on 
real-world datasets 
where 
our two-step approach clearly outperforms an approach that is fair in only 
one step. 
The second statement of  Lemma~\ref{lemma_fairness_in_both_steps_necessary} 
is slightly different for pairwise DP and EO ($\DPviol=0.5$ vs $\EOviol=1$). 
The next lemma shows that this is not an artefact of our proof:  
for pairwise DP, the 
fairness violation 
of the predictor can 
be upper bounded by the 
fairness violation 
of the scoring function for any choice of thresholds, 
such that $\DPviol=0.5$ is the worst we can get 
 using 
 a perfectly~fair~scoring~function.

\begin{lemma}
[Fairness of scoring function vs fairness of predictor for DP]
\label{lemma_fairness_scoring_vs_threshold_model}
Let $s:\mathcal{X}\rightarrow \R$ be a scoring function and $f:\mathcal{X}\rightarrow [k]$ be a predictor that is obtained from thresholding $s$. No matter what the thresholds are, we have 
$\DPviol(f) \leq 1/2+\DPviol(s)/2$.
\end{lemma}

\section{RELATED WORK}\label{section_related_work}

\paragraph{Ordinal Regression}
A review of classical 
techniques,  
such as the proportional odds model (POM; \citealp{mccullagh1980}), 
is provided 
by \citet{OConnell2006}. 
\citet{gutierrez2016ordinal} provide a 
more recent 
survey. 
They 
categorize 
ordinal regression methods into 
naive approaches, 
binary decomposition, and threshold models.
Our approach falls into the latter category. 
It is based on the 
SVM pairwise approach of \citet{herbrich_ordinal_regression}, which has been refined by SVM pointwise approaches \citep{shashua2002,chu2007}. 
The pointwise approaches avoid the issue of 
transforming 
a training set of size~$n$ to one of size~$n^2$. However, it is unclear how to incorporate 
fairness 
into these approaches.  
Some newer methods are the deep learning-based ones by \citet{niu2016}, \citet{liu2017}, \citet{polania2019} and \citet{cao2020}, which are 
designed for image data 
and age or body mass index estimation. 
None of the existing methods for ordinal regression takes~fairness~into~account.

\paragraph{Related Fairness Notions}
\citet{beutel2019fairness} 
introduce 
a fairness notion similar to pairwise EO as defined in \eqref{fairness_notion_EO} for a ranking function used in a recommendation system. 
They 
train a fair model by  
adding 
a correlation penalty term to the loss function.
\citet{kuhlman2019} propose fairness criteria 
equivalent 
to pairwise DP or EO for general rankings, and an auditing mechanism that evaluates the criteria on several subparts of a ranking. 
\citet{kallus2019} study a fairness notion similar to pairwise EO for bipartite ranking.
They are mainly concerned with evaluating their 
notion, but also  propose a 
simple 
post-processing mechanism to obtain a fair 
group-dependent
scoring function from an unfair one. 
The notion of \citeauthor{kallus2019} and some closely related notions by \citet{borkan2019} have been generalized by \citet{vogel2020}, 
who 
provide generalization guarantees for a scoring function learned under their fairness constraints   
and train a fair model by adding a highly non-convex surrogate of the fairness violation to the 
loss
  function. 
Pairwise DP and EO as well as 
related 
variants, 
including the case of a continuous protected attribute, 
are  
also 
introduced 
by 
\citet{narasimhan2020pairwise} for 
a ranking or regression function~$f:\mathcal{X}\rightarrow\R$. 
They 
train a fair model using the 
algorithm 
of \citet{cotter2019} for optimization problems with non-differentiable constraints. 
When 
\citeauthor{narasimhan2020pairwise} 
consider ranking, 
their approach is 
similar to the first step in our two-step approach. 
Note that there is no obvious way how to formulate our problem of fair ordinal regression as a fair ranking / bipartite ranking / regression problem. For example, when trying to formulate it as a fair regression problem 
and 
use the techniques of \citeauthor{narasimhan2020pairwise}, we would have to carefully 
choose the encoding of 
a training point's label~$y\in\mathcal{Y}$ as a label $y\in\R$. An arbitrary monotone encoding does not work since \citeauthor{narasimhan2020pairwise} only consider the squared loss on $\R$. 
Furthermore, 
we would have to decide how to map a prediction~$f(x)\in\R$~to~one~in~$\mathcal{Y}$.

Other existing, non-pairwise fairness notions for 
ranking \citep[e.g.,][]{zehlike2017,biega2018,celis_fair_ranking,singh2018}  
or 
regression \citep[e.g.,][]{johnson2016,berk2017,fair_kernel_learning2017,agarwal2019}
are very different from our proposed notions.

\section{EXPERIMENTS}\label{section_experiments}

To illustrate the 
relevance of our approach, we first consider two  
real-world 
datasets for which fairness 
is of concern. 
We then 
investigate the performance 
of our approach 
by comparing it to ``unfair'' state-of-the-art methods for ordinal regression on 
numerous 
benchmark datasets. These benchmark datasets do not come with a meaningful protected attribute,  
and we instead treat other features as ``protected''.
Although not ideal, such evaluation is widely common in the literature on fair ML. 
See Appendix~\ref{appendix_detail_datasets} for details about all datasets.

We implemented our proposed approach in Python.\footnote{Code available on \url{https://github.com/amazon-research/fair-ordinal-regression}.} 
  ~In the first step of 
our 
approach, that is 
when  
learning the scoring function 
via the reduction to fair binary classification as described in Section~\ref{subsection_learning_scoring_function},  
we built on 
the 
approach of \citet{agarwal_reductions_approach}, 
whose implementation is available as part of \textsc{Fairlearn} %
  (\url{https://fairlearn.github.io/}). 
We used the \textsc{GridSearch}-method (corresponding to Sec.~3.4 of  \citeauthor{agarwal_reductions_approach}), which returns a deterministic rather than a randomized classifier. 
We provide some details in Appendix~\ref{appendix_detail_hyperparameters}. 
For solving the second problem in our 
approach, that is choosing the thresholds, we ran the local search as explained in Section~\ref{subsection_choosing_thresholds} with random initialization for ten times and kept the 
solution with the lowest objective value (evaluated 
on training~data).

The two main parameters in our strategy correspond to $\varepsilon$ in \eqref{eq_reduction} and $\lambda$ in \eqref{threshold_objective} and govern the accuracy-vs-fairness trade-off when learning the scoring function and thresholds, respectively.  The \textsc{GridSearch}-method is not parameterized in $\varepsilon$, but in $\mu\in[0,1]$ such that the method aims to minimize $(1-\mu)\cdot 
\text{error}(c_w)+ 
\mu\cdot\Fairviol(c_w)$. We reparameterized~\eqref{threshold_objective} 
using 
$\lambda'=\frac{\lambda}{k+\lambda}\in[0,1)$ so that $\mu$ and $\lambda'$ have the same interpretation, and 
in the following we 
study our approach for various choices of $\mu=\lambda'$ (without systematically searching for an optimal correspondence between $\mu$ and $\lambda'$, we found that choosing $\mu=\lambda'$ generally works~well).
Throughout, 
we identify labels in $\mathcal{Y}$ 
with 
$1,\ldots,k$, set $C$ to the absolute cost matrix and aim for~a~small~$\MAE$ as defined in 
Section~\ref{section_ordinal_reg} (in App.~\ref{appendix_asymmetric_cost} we present an experiment with an asymmetric cost matrix).

\subsection{Experiments on Real-World Datasets}
\label{subsection_experiment_motivation}

We applied our strategy to the Drug Consumption dataset 
\citep[referred to as DC dataset;][]{drug_consumption_data} 
and the Communities and Crime dataset 
\citep[C\&C dataset;][]{comm_and_crime_data}. 
On the 
DC
dataset, we predicted 
when an individual has last consumed cannabis (never or more than a decade ago - last decade - last year - 
last month~- 
last day). The input features are a person's age, \mbox{education} and seven features measuring personality traits,  
and we used~a~person's gender (f or m) as protected attribute.
In such a setting, fairness 
can be 
of central concern,  
for example, when a security firm  makes 
this kind of predictions
for its job applicants. 
On the 
C\&C  
dataset, we predicted the total number of crimes per $10^5$ inhabitants 
(discretized to 8 classes) 
for communities represented by 95 input features such as the median 
income. 
As protected attribute~$a\in\{\text{white},\text{diverse}\}$, we~used whether a community is predominantly (i.e., $>80\%$) inhabited by Caucasians 
(in App.~\ref{appendix_comm_and_crime_finer_grid} we consider 
$a\in\{\text{white},\text{African-American},\text{Hispanic or Asian}\}$).
Also in this scenario, fairness 
is 
highly relevant \citep{predictive_policing}.
The datasets comprise 1885 
and 1994 records, respectively, which we randomly split into a training set of size~1500 and a test set of size~385 or 494. We report results on the test sets, 
averaged~over~20~\mbox{random~splits}. 

The plots in the top row of Figure~\ref{fig_drug_consumption} show the results for the 
DC 
dataset and the plots in the bottom row 
for the 
C\&C 
dataset. 
They show
on the $x$-axis the fairness 
violation~$\DPviol$ (left) or $\EOviol$ (right)  
and on the $y$-axis the $\MAE$ of 
various predictors:
\textbf{(i)}~the predictors that  we obtained from running our approach for 
$\mu=\lambda'\in\{0,0.1,0.2,0.3,0.33,0.36,0.4,0.5,\ldots,0.9\}$.
For the left plots we enforce approximate pairwise DP 
and for the right plots pairwise EO.  
\textbf{(ii)}~the predictor obtained from fitting a POM model 
\citep{mccullagh1980}, 
a classical 
threshold model 
technique 
that does not require any hyperparameter tuning and is among the most widely used ordinal regression methods 
in practice 
\citep[][Sec.~4.4]{gutierrez2016ordinal}. 
This is a simple unfair baseline. 
\textbf{(iii)}~the best constant predictor; since we want to minimize the $\MAE$, this is $f(x)= \text{median}(y_1,\ldots,y_n)$ for training data~$\mathcal{D}=((x_i,y_i,a_i))_{i=1}^n$. This is the most simple fair baseline. 
\textbf{(iv)}~randomized predictors that predict $\text{median}(y_1,\ldots,y_n)$ with probability~$p$ and the prediction of the POM-predictor with probability~$1-p$, for $p=\frac{j}{50}$, $j\in[49]$.  
For these randomized predictors, the plots show the performance obtained from averaging over 
100 times of applying such a predictor 
to 
the test set. 
This is a simple approximately fair baseline. 
\textbf{(v)}~the 
predictors that we obtained by choosing 
approximately fair 
thresholds for the scoring function of the POM model 
based on $\eqref{threshold_objective}$. 
In doing so, we used the same values of $\lambda'$ 
and ran the local search strategy in the same way 
as for our approach. This is supposed to show the need of 
incorporating fairness when learning~the~scoring~function~(cf. Sec.~\ref{subsection_both_steps_necessary}).

\newcommand{\scaleparameterAdrug}{0.2}
\newcommand{\abstAdrug}{0pt}

\newcommand{\scaleparameterAcomm}{0.205}

\begin{figure}[t]
\centering
    
    \includegraphics[width=\linewidth]{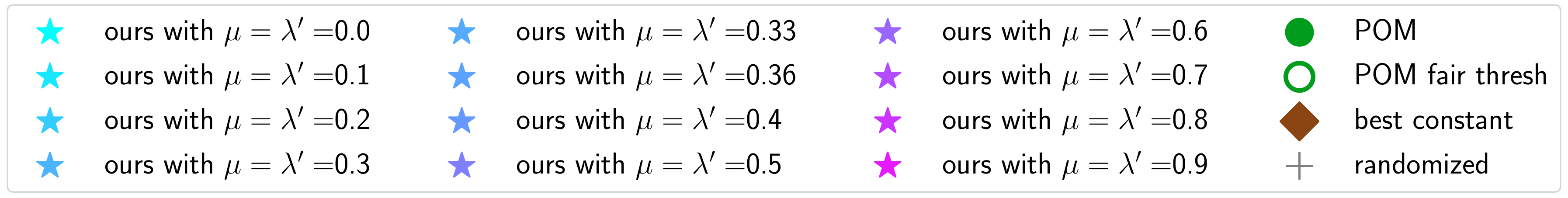}
    
    \vspace{8pt}
    \includegraphics[scale=\scaleparameterAdrug]{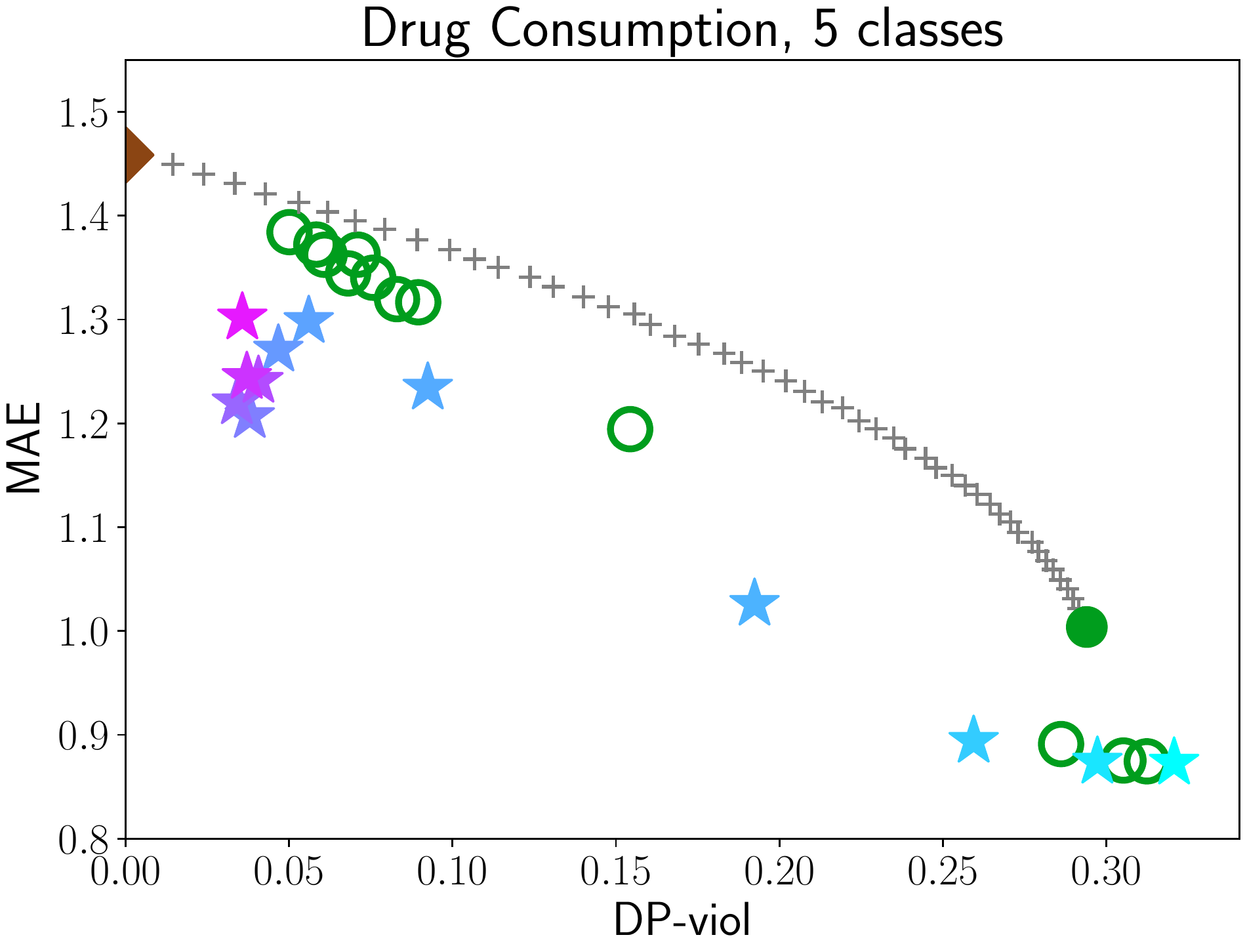}
    \hspace{\abstAdrug}
    \includegraphics[scale=\scaleparameterAdrug]{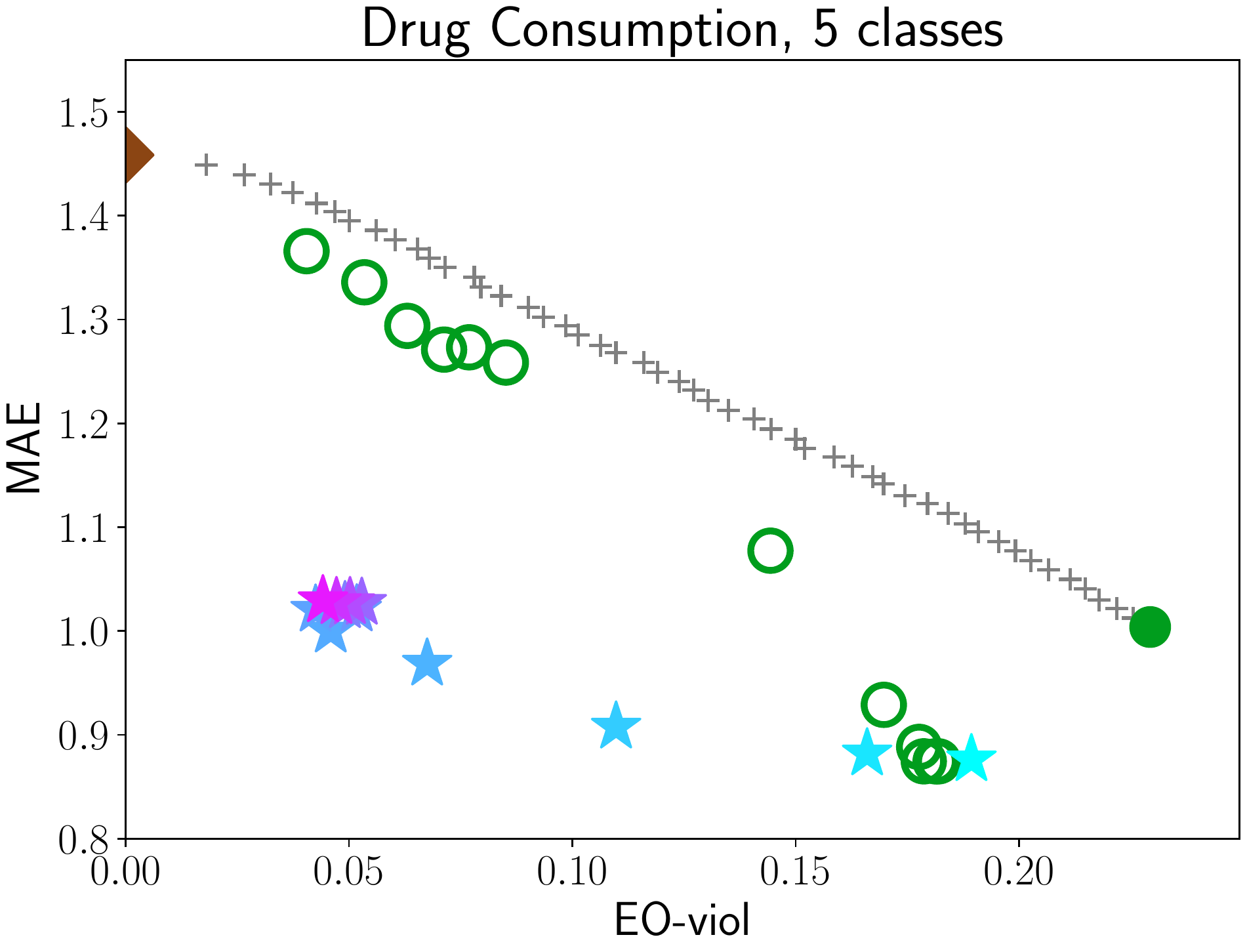}

\vspace{4pt}
    \includegraphics[scale=\scaleparameterAcomm]{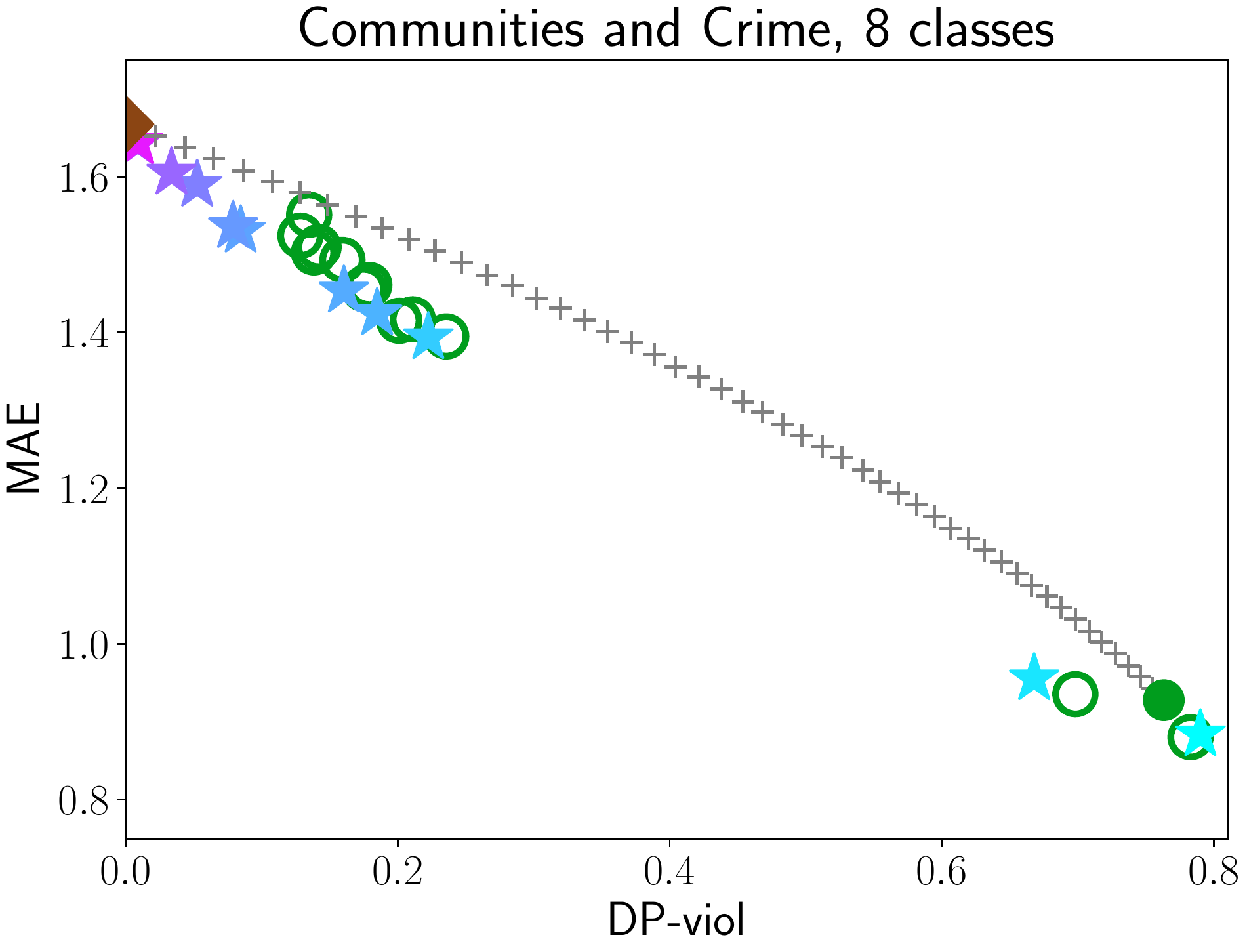}
    \hspace{\abstAdrug}
    \includegraphics[scale=\scaleparameterAcomm]{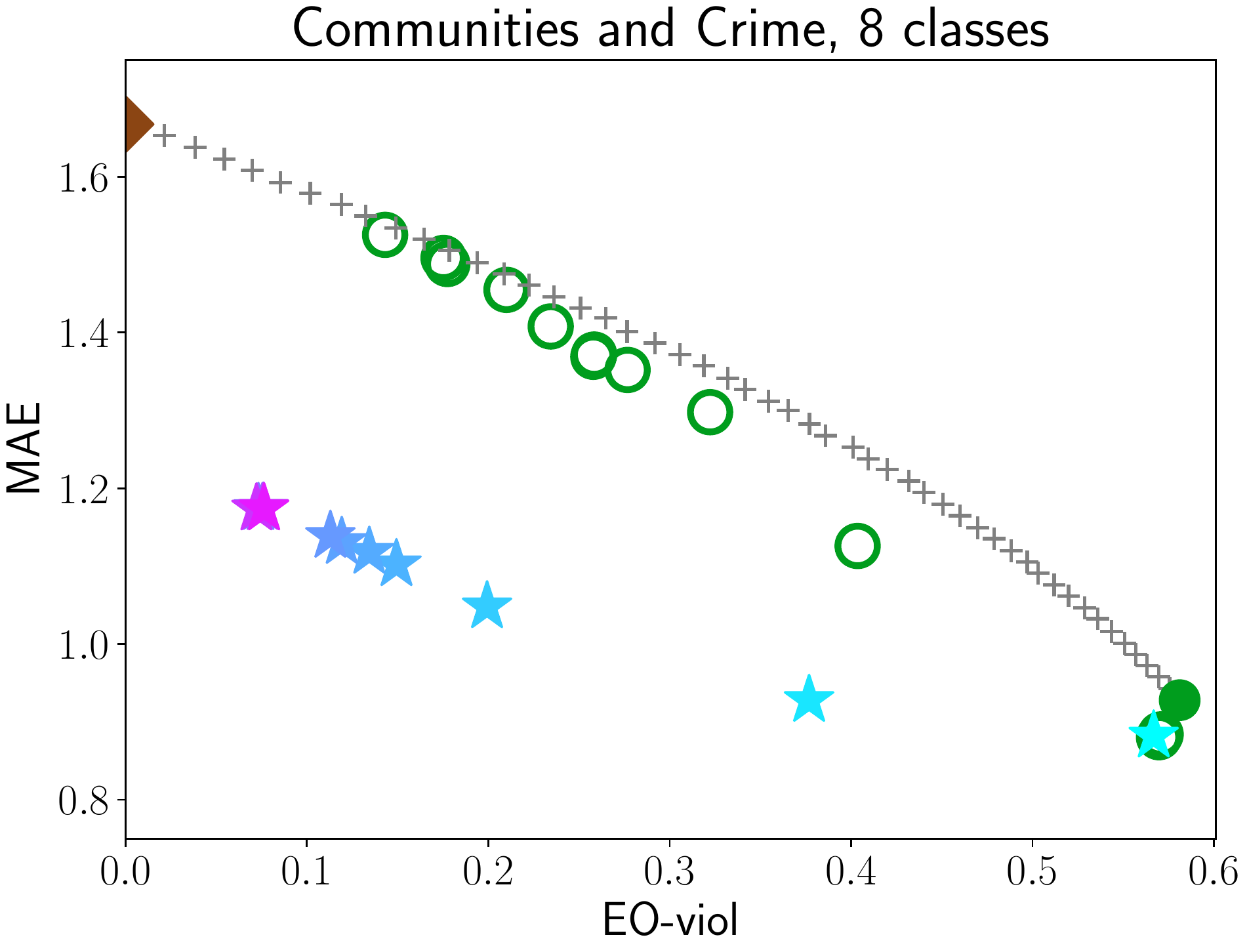}

    \caption{$\MAE$ vs $\DPviol$ \textbf{(left)} and $\MAE$ vs $\EOviol$ \textbf{(right)} for various predictors: the predictors produced by our approach (cyan to purple stars), by the POM algorithm (green filled circle), by solving \eqref{threshold_objective} for the scoring function of the POM predictor for various values of $\lambda$ (green circles), and by
    randomly mixing the POM predictor with the best constant one (grey crosses);  the best constant predictor is represented by the brown diamond. See Appendix~\ref{appendix_comm_and_crime_finer_grid} for a version of the bottom left plot where we fill the empty area by considering additional values of $\mu=\lambda'$, and another version where we consider three protected groups.}
    \label{fig_drug_consumption}
    
\end{figure}

\newcommand{\scaleparameterA}{0.21}

\newcommand{\abstA}{1pt}

\begin{figure*}
    \centering
    \includegraphics[width=\linewidth]{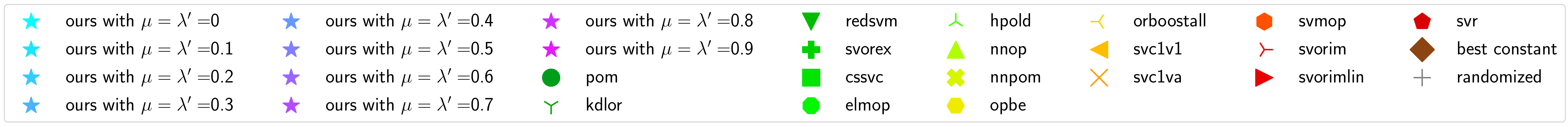}

    \vspace{4pt}

      \includegraphics[scale=\scaleparameterA]{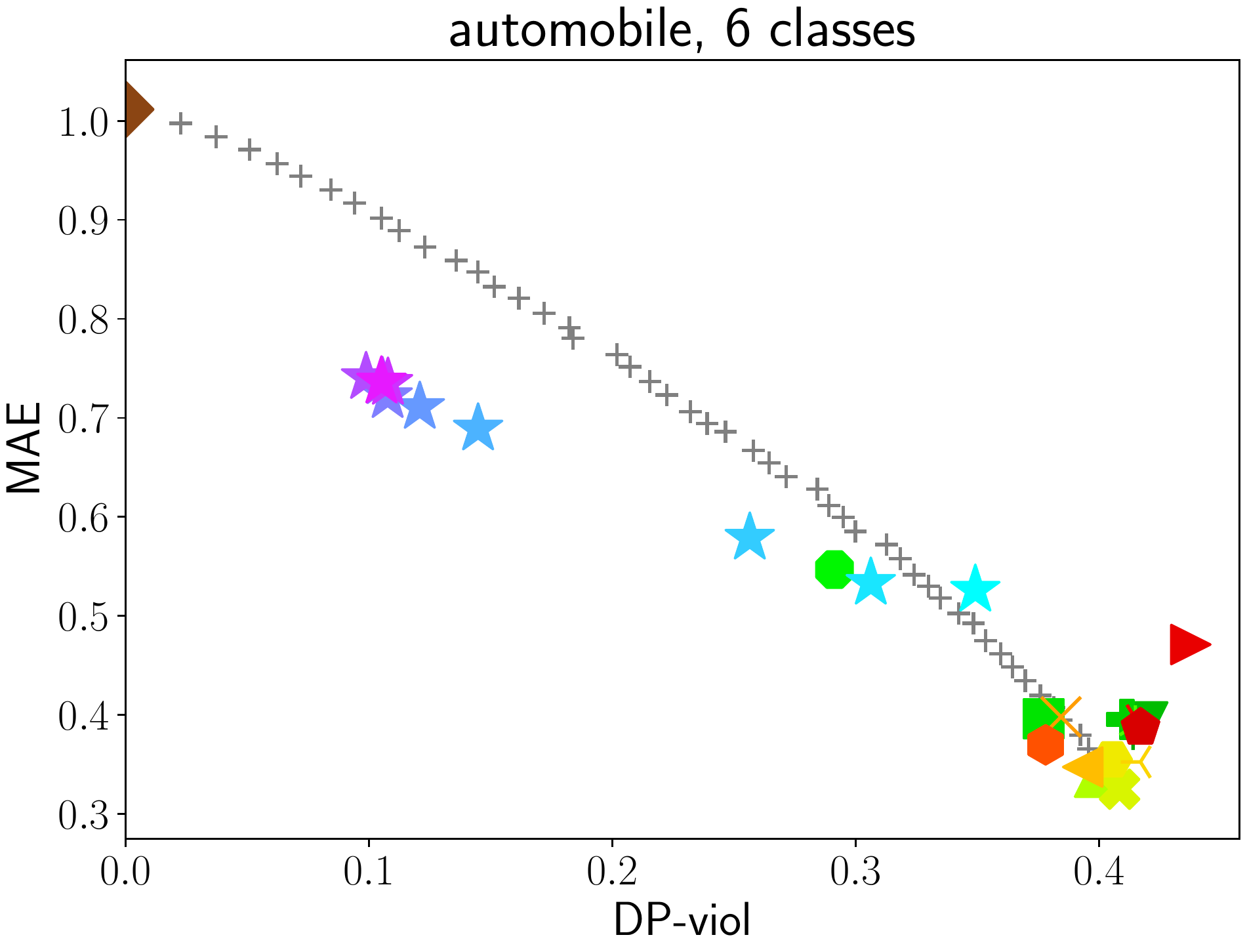}
    \hspace{\abstA}
   \includegraphics[scale=\scaleparameterA]{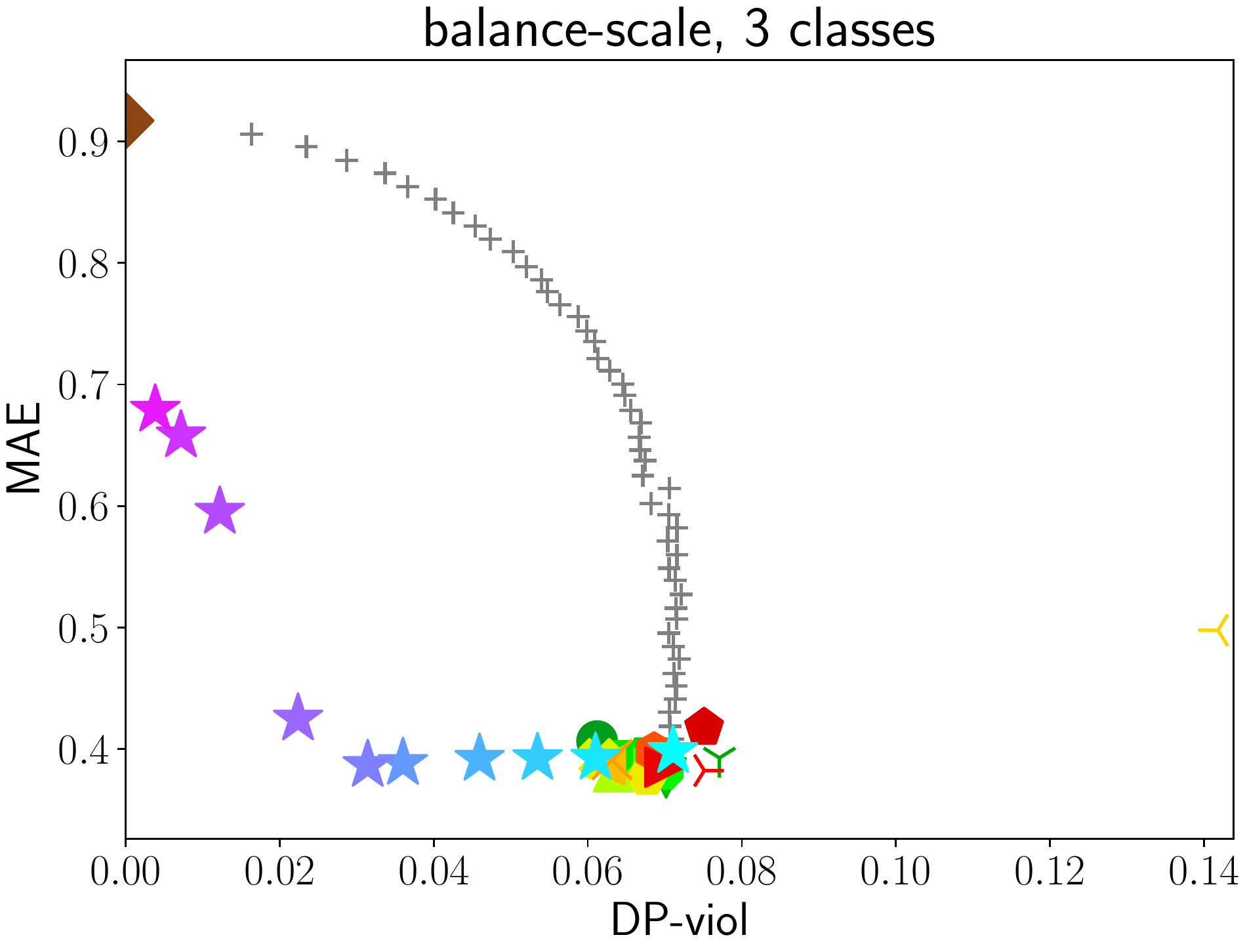}
   \hspace{\abstA}
   \includegraphics[scale=\scaleparameterA]{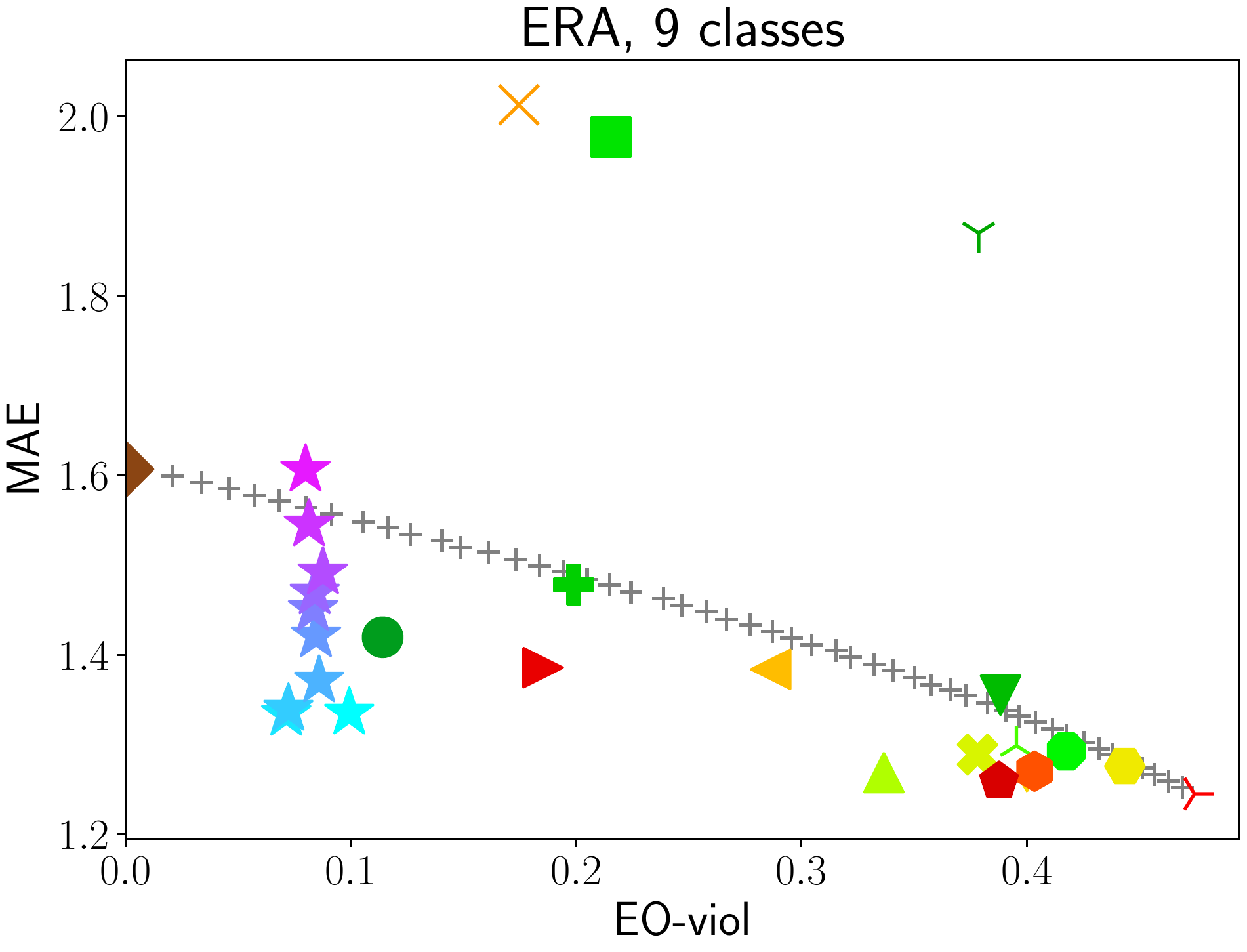}
   \hspace{\abstA}
   \includegraphics[scale=\scaleparameterA]{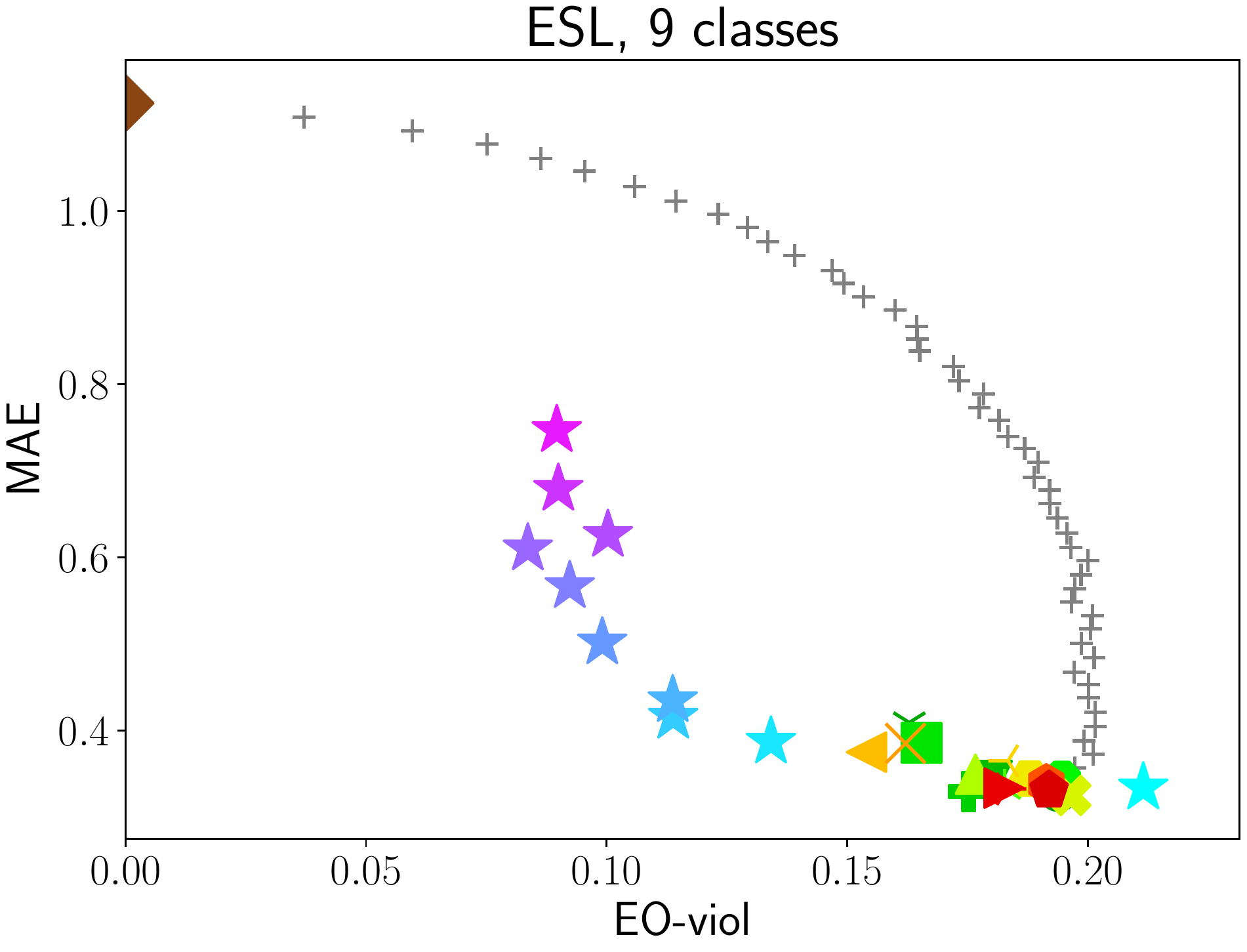}

    \includegraphics[scale=\scaleparameterA]{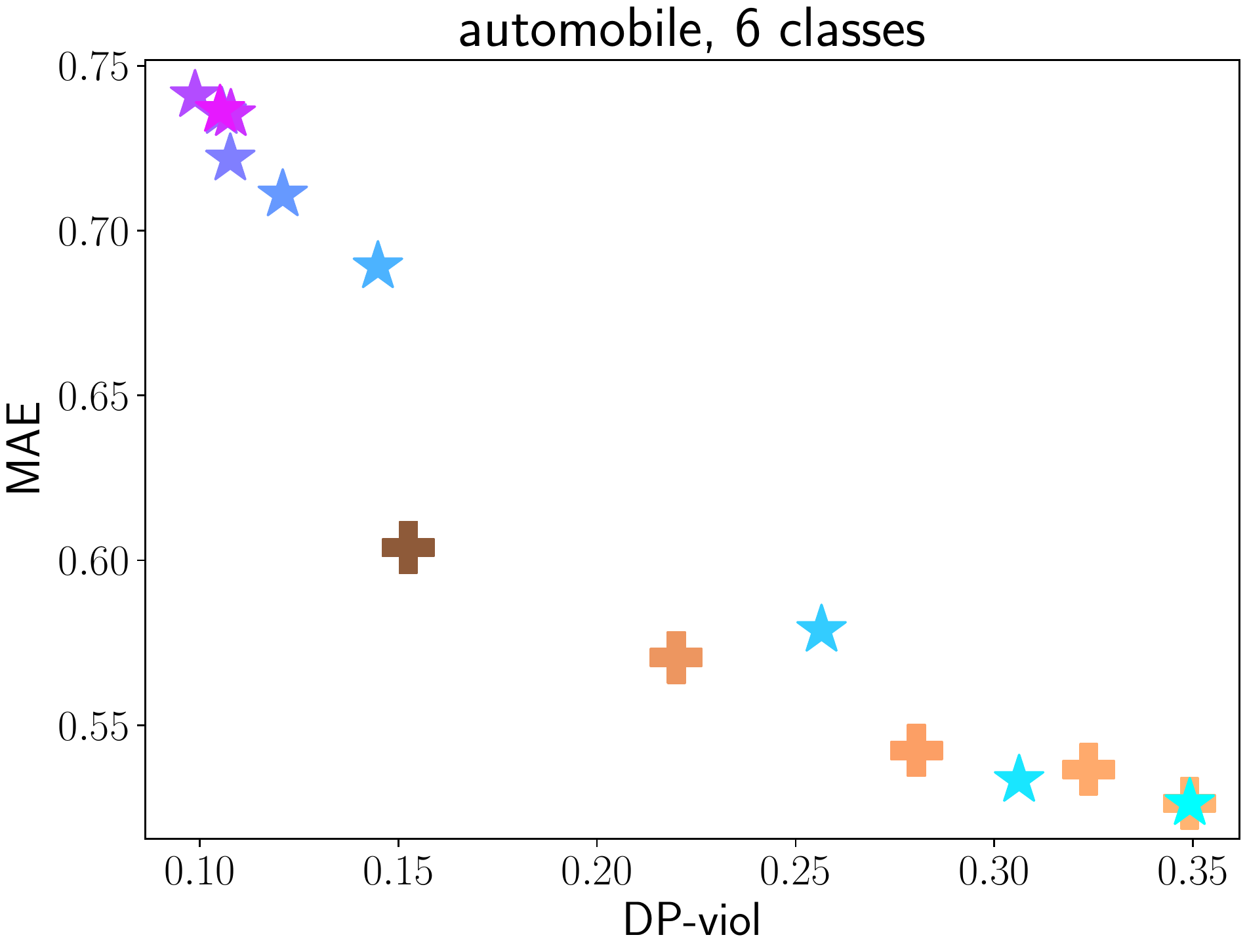}
    \hspace{\abstA}
    \includegraphics[scale=\scaleparameterA]{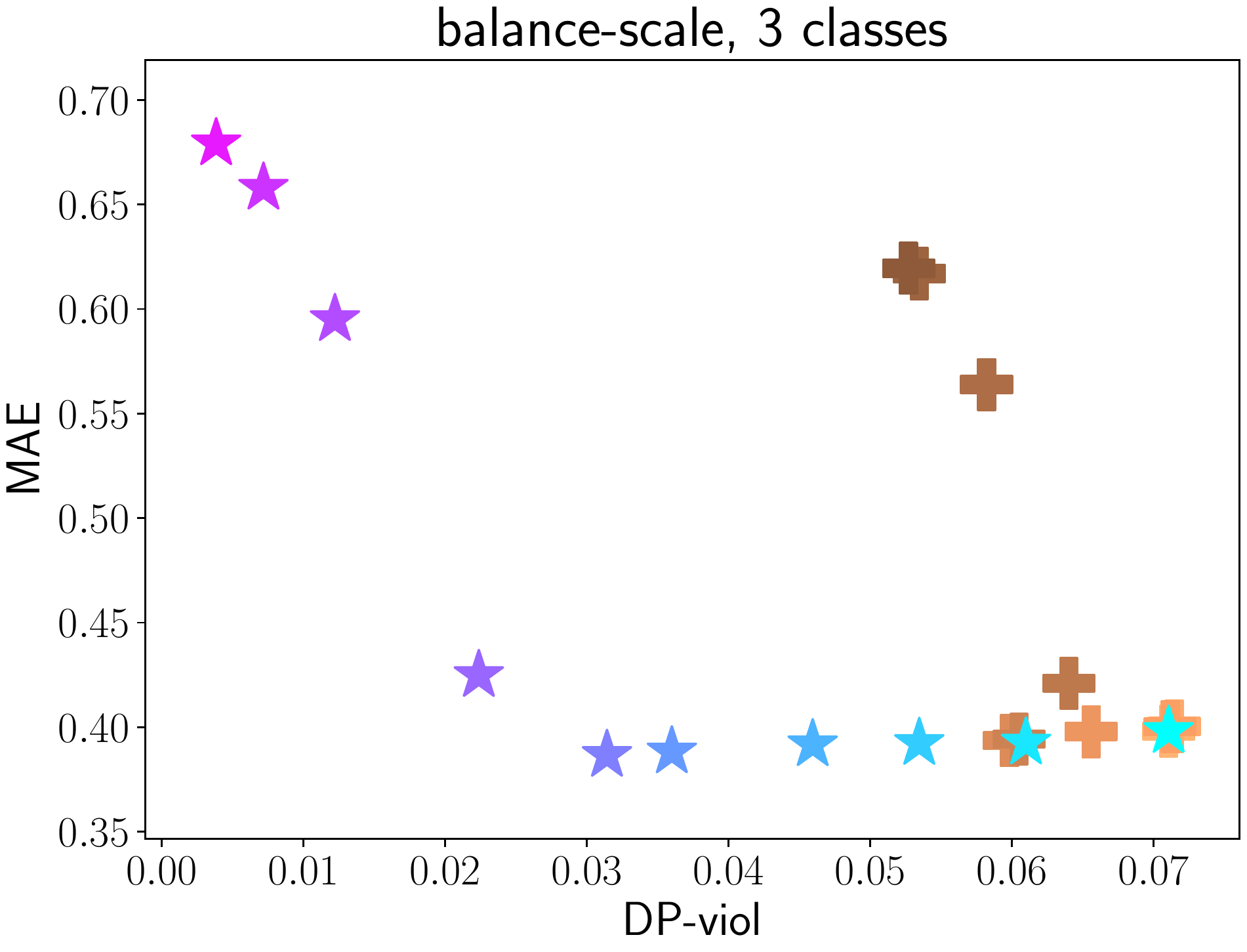}
    \hspace{\abstA}
    \includegraphics[scale=\scaleparameterA]{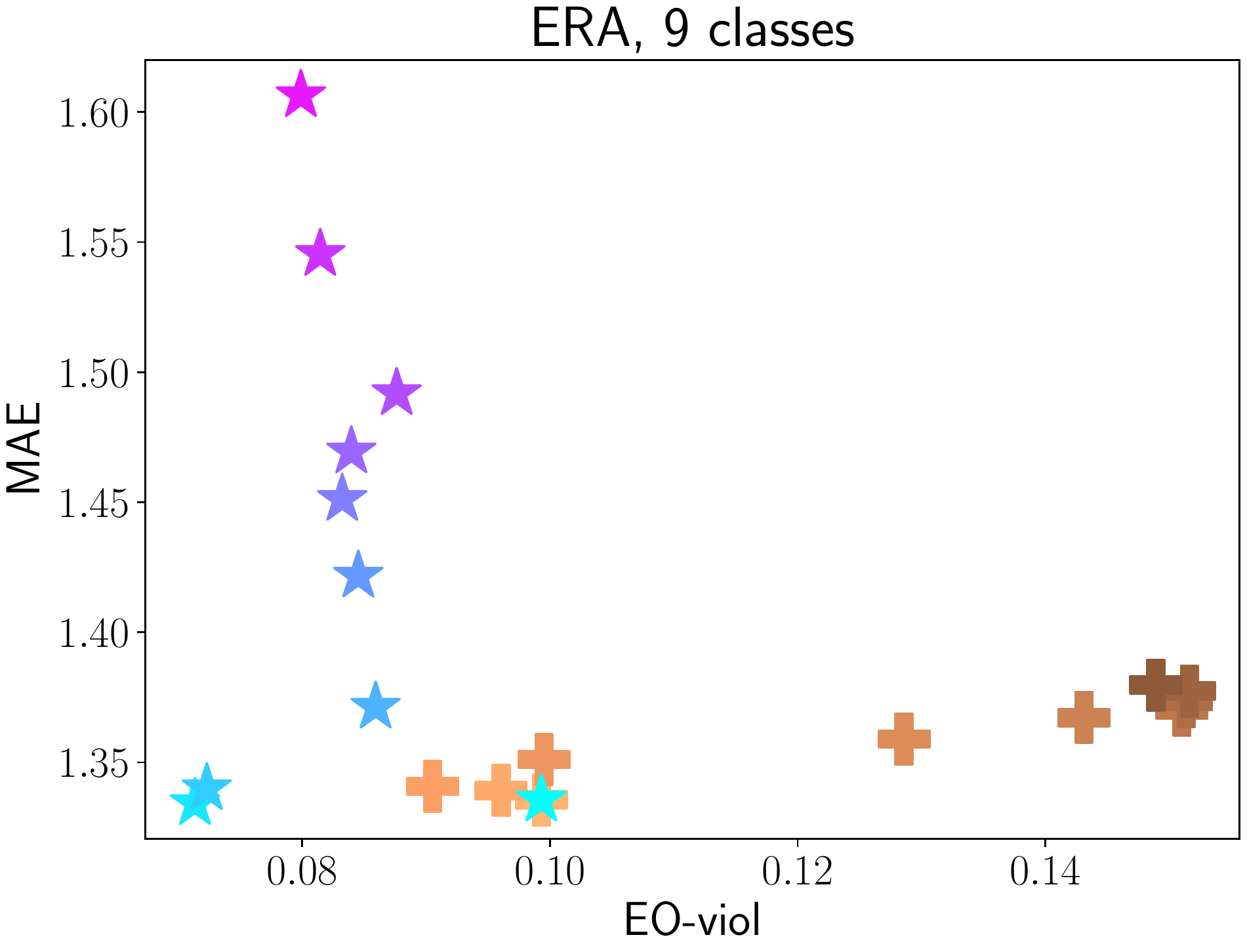}
    \hspace{\abstA}
    \includegraphics[scale=\scaleparameterA]{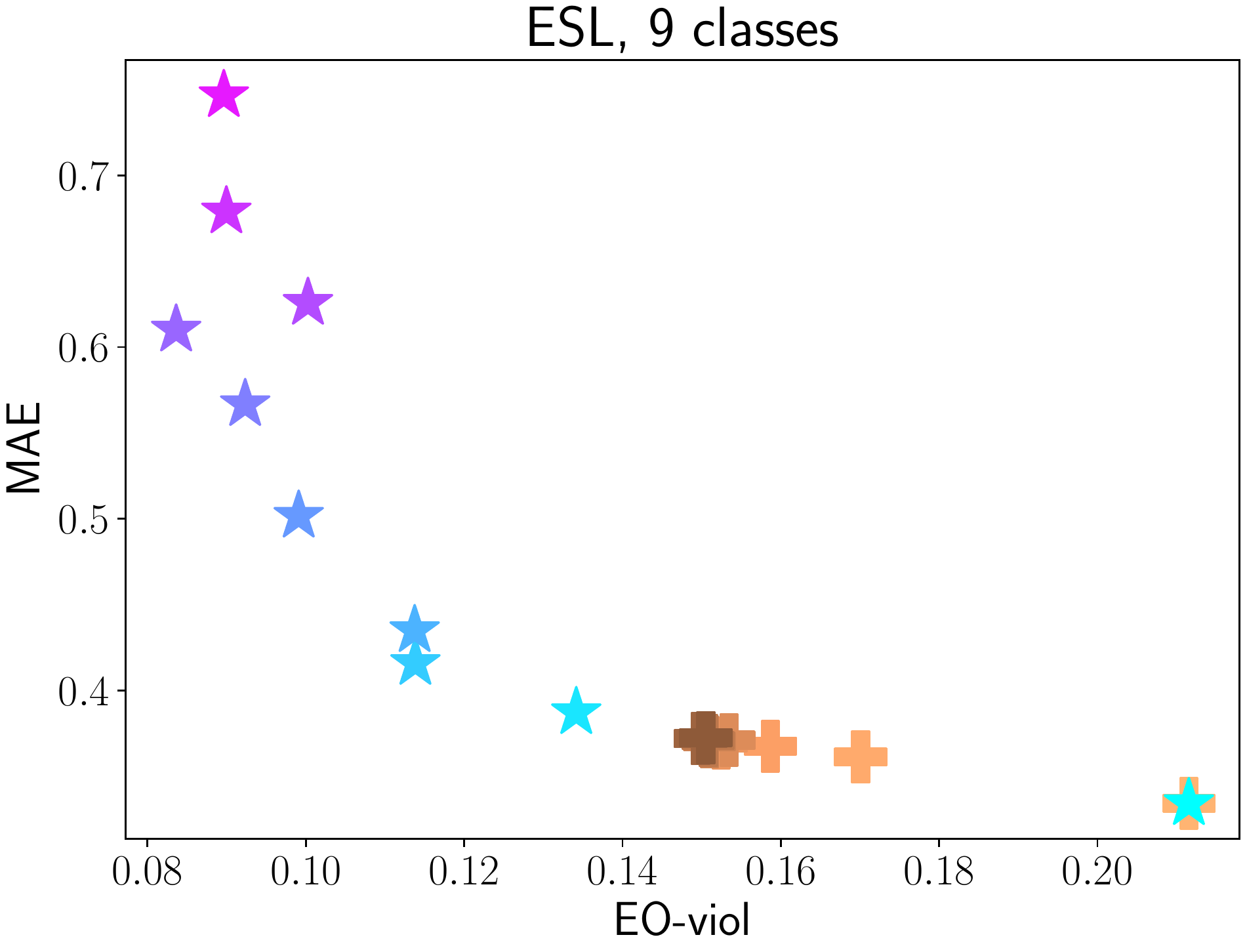}
    \hspace{\abstA}

    \caption{\textbf{Top row:} $\MAE$ vs $\DPviol$ (1st + 2nd plot) or $\MAE$ vs $\EOviol$ (3rd + 4th plot) for the predictors produced by our approach in comparison to the algorithms
    in 
    the \textsc{Orca} toolbox 
    for four of the datasets. 
    The 
    plots for the other datasets 
    are in
    Appendix~\ref{appendix_further_experiments}.
    \textbf{Bottom row:} $\MAE$ vs $\DPviol$ or $\MAE$ vs $\EOviol$ for the predictors produced by our approach with $\mu=\lambda'$ 
    and with 
    the same values of $\mu$, but 
    $\lambda'=0$ (light 
    to dark~brown~crosses).}
    \label{fig:exp_comparison}
    
\end{figure*}

\renewcommand{\scaleparameterA}{0.2}
\renewcommand{\abstA}{5pt}

We make the following observations: 
\textbf{(i)}~the POM 
model 
can be highly unfair with $\DPviol=0.29$ / $\EOviol=0.23$ on the
DC 
dataset and 
$\DPviol=0.76$ / $\EOviol=0.58$ on the 
C\&C 
dataset. 
This shows that we cannot expect a standard method for ordinal regression to be fair and that there is a need for an approach 
that explicitly takes fairness into account. 
\textbf{(ii)}~The predictors 
produced 
by our approach, in general, nicely explore the trade-off between accuracy and fairness. As expected, the larger the value of $\mu=\lambda'$, the more fair and the less accurate is the predictor. However, on the C\&C dataset and when aiming for pairwise DP (bottom left), there is an area in the $\MAE$-vs-$\DPviol$ plot (in the range of $\DPviol\in[0.25,0.65]$) that is unexplored. 
One can try to fill that area by running our approach with additional values of $\mu=\lambda'\in[0.1,0.2]$, 
and indeed by doing so we succeed. The corresponding plot is shown in Appendix~\ref{appendix_comm_and_crime_finer_grid}. We deliberately chose not to show that plot instead of the one 
shown 
in Figure~\ref{fig_drug_consumption} 
to illustrate that we have to choose the parameter $\mu=\lambda'$ in an adaptive way 
if we want 
to fully explore the accuracy-vs-fairness trade-off 
that is 
achievable by our method 
on 
a given dataset. 
\textbf{(iii)}~For small $\mu=\lambda'$, our predictors have 
even 
smaller $\MAE$ than the POM predictor. 
Since $|\Psymb[y_1> y_2 | a_1=\text{f},a_2=\text{m}]-\Psymb[y_1< y_2 | a_1=\text{f},a_2=\text{m}]|=0.35$ on the 
DC 
dataset and $|\Psymb[y_1> y_2 | a_1=\text{white},a_2=\text{diverse}]-\Psymb[y_1< y_2 | a_1=\text{white},a_2=\text{diverse}]|=0.72$ on the 
C\&C 
dataset, it is 
not surprising 
that when aiming for pairwise DP and for larger values of $\mu=\lambda'$,  
our predictors have higher $\MAE$. 
When aiming for pairwise EO, even for large values of $\mu=\lambda'$ the $\MAE$ of our predictors 
is comparable to the $\MAE$ of the POM predictor, 
but our predictors are significantly more fair.
\textbf{(iv)}~Our predictors, which are deterministic, clearly outperform the randomized predictors. 
Note that the performance of the randomized predictors interpolates between 
 the performance of the POM predictor and the performance of the best constant predictor 
in a non-linear way 
due to the pairwise nature of $\Fairviol$.  
Once more, 
we 
emphasize that applying a randomized predictor to human subjects might be problematic 
(cf. Sec.~\ref{section_our_strategy}). 
\textbf{(v)}~Choosing thresholds for the scoring function of the POM model based on $\eqref{threshold_objective}$ improves the $\MAE$ of the POM predictor for small values of $\lambda$. For larger values of~$\lambda$, it improves the fairness of the POM predictor, but also 
increases its $\MAE$. 
In particular when aiming for pairwise EO, its $\MAE$ is much larger than the $\MAE$ of our predictors.  
This confirms our findings of Section~\ref{subsection_both_steps_necessary}, and in particular the first claim of Lemma~\ref{lemma_fairness_in_both_steps_necessary}, where we showed 
that solving $\eqref{threshold_objective}$ for an unfair scoring function can result in a $\MAE$ that is arbitrarily high compared to 
when
using 
a fair scoring function.

\subsection{
Comparison 
to 
Unfair State-Of-The-Art Methods on Benchmark Datasets}\label{subsection_experiment_comparison}

\citet{gutierrez2016ordinal} performed an 
extensive 
 evaluation of 16 algorithms for ordinal regression, including classical methods (such as the POM model used 
in the previous section) 
as well as 
state-of-the-art methods (such as the SVM-based algorithms of \citealp{chu2007}, or the neural network approach of \citealp{cheng2008}). They applied these algorithms to 41 
benchmark 
datasets, 17 of which 
come with an actual 
ordinal regression task 
(referred to as real ordinal regression datasets) and 24 of which originally come with a standard regression task,
that is $y\in\R$, 
and
for which the label~$y$ has been discretized 
to either five or ten classes 
(referred to as discretized regression datasets).  
 The algorithms and the datasets are publicly available as part of the 
\textsc{Orca}  toolbox for \textsc{Matlab} \citep{orca_jmlr},  
which also comprises the hierarchical model of \citet{monedero2018}; hence, there are 17~algorithms~in~total.

We applied the 17 
\textsc{Orca} 
algorithms and our strategy 
for $\mu=\lambda'\in\{0,0.1,0.2,\ldots,0.9\}$ 
to the 33 datasets containing at least 200 datapoints. The datasets do not have a meaningful protected attribute, and 
we treat an  arbitrary binary feature as the protected attribute $a\in\{0,1\}$, or chose some real-valued feature~$x_r$ and set $a=\charfct\{x_r\geq \text{median}(x_r)\}$ for those 
datasets that do not have any  
binary features. %
 We ran all \textsc{Orca} algorithms in the same way as \citeauthor{gutierrez2016ordinal} did; in particular, we performed 5-fold cross validation over the same sets of hyperparameters with the goal of minimizing the $\MAE$.
However, no
algorithm observes the protected attribute~$a$  
(or the feature~$x_r$ if applicable) as part of the input~$x$. 
Hence, 
our experiments do not exactly replicate the ones by \citeauthor{gutierrez2016ordinal}.
We 
also 
used the same splits into 30 or 20 training and test sets 
and report all results on the test sets, averaged over the 
various 
splits. 
In Appendix~\ref{appendix_further_experiments}, we 
show 
results that include the standard~deviation~over~the~splits.

 The plots in the top row of Figure~\ref{fig:exp_comparison} show the  results for the first four of the real ordinal regression datasets and when aiming for pairwise DP or EO. The results for all datasets, both for pairwise DP and EO,  
are 
in Appendix~\ref{appendix_further_experiments}. 
The plots in the bottom row show the performance of  predictors obtained 
from 
our approach with $\mu=\lambda'$ 
in comparison with 
the performance of 
predictors obtained from our approach with $\lambda'=0$. In the latter case, we 
enforce fairness only when learning the scoring function.  
The 
main findings 
are similar to the ones from 
Figure~\ref{fig_drug_consumption} 
(we provide some more interpretation in App.~\ref{appendix_further_experiments}): \textbf{(i)} the state-of-the-art methods can be highly unfair. \textbf{(ii)} Our predictors explore the accuracy-vs-fairness trade-off, but sometimes there are unexplored areas and we would need to adaptively choose additional values of $\mu=\lambda'$ that we run our strategy with. \textbf{(iii)} In particular when aiming for pairwise EO, our predictors are often significantly more fair, but only slightly less accurate than the competitors.~\textbf{(iv)}~We 
outperform the predictors that we obtain by randomly mixing the predictor with the smallest $\MAE$ with the best constant one (similarly as we did in the previous section; shown by the grey crosses).  
\textbf{(v)} 
Our predictors  can potentially produce 
a much better accuracy-vs-fairness trade-off than the  predictors obtained by running our approach with $\lambda'=0$. For example, on the balance-scale dataset shown in the second plot of the bottom row  of Figure~\ref{fig:exp_comparison},  
the predictor corresponding to $\mu=\lambda'=0.5$ has $\MAE=0.39$ and $\DPviol=0.03$, while the best predictor 
that corresponds to some $\mu\geq 0$ and $\lambda'=0$ 
and has 
the same value of $\MAE$ 
has twice as high a value of  $\DPviol$. 
This 
supports 
our findings of Section~\ref{subsection_both_steps_necessary}
and in particular
the second claim of Lemma~\ref{lemma_fairness_in_both_steps_necessary}.

\section{DISCUSSION \& FUTURE WORK}\label{section_discussion}

This paper initiates the study of fair ordinal regression. We adapted two pairwise fairness notions from 
the literature on 
fair ranking and  proposed a two-step strategy for  training a 
threshold model 
that allows us to control the trade-off between accuracy and fairness.

There are 
two main 
directions for future work: 
we  designed an algorithm 
for training 
an 
accurate and 
approximately fair predictor. It would be interesting to understand how we can do so in an optimal way (cf. Sec.~\ref{subsection_limitations}, first item) and 
what is the best 
accuracy-vs-fairness trade-off that we can achieve, both in principle and for a given model class. 
The latter question 
is still 
receiving considerable attention even in the binary classification setting \citep[e.g.,][]{dutta2020,kim2020}. 
On the other hand, it~would~be interesting to study other 
fairness notions for ordinal regression. Two examples of 
other 
possible 
notions 
are to require that,   for all $\tilde{a}\in\mathcal{A}$, 
$\Ex [C_{y,f(x)}|f(x)<y,a=\tilde{a}]=\Ex [C_{y,f(x)}|f(x)<y]$ or 
$\Ex[C_{y,f(x)}\charfct\{f(x)<y\}|a=\tilde{a}]=\Ex[C_{y,f(x)}\charfct\{f(x)<y\}]$.
The motivation behind these two fairness notions is that an input point~$x$  suffers some harm whenever its predicted label~$f(x)$ is less preferable than its actual label~$y$. 
One could also consider a notion of individual fairness \citep{fta} for ordinal regression, requiring that similar datapoints are treated similarly. 
In contrast 
to standard multiclass classification, in case of ordinal regression, the order on the label set~$\mathcal{Y}$ 
already provides some similarity information about $\mathcal{Y}$. 
Of course, once we have several fairness notions for ordinal regression at our disposal, it is a non-trivial question to choose the most appropriate one for a given task, just as it is the case in standard binary \mbox{classification~\citep{Makhlouf2020}}.

\bibliography{ref,mybibfile_fair_clustering}
\bibliographystyle{plainnat}

\clearpage
\onecolumn

\appendix
\section*{APPENDIX}

\section{PROOFS AND DETAILED EXPLANATIONS}\label{appendix_partA}

\subsection{Pairwise Equalized Odds}\label{appendix_pairwise_equalized_odds}

One might wonder whether there is also a pairwise analogue of the standard fairness notion of equalized odds \citep{hardt2016equality}, which 
is a stricter notion than standard EO and 
requires that  $\Psymb[f(x)=\hat{y} | a=\tilde{a},y=\tilde{y}]=\Psymb[f(x)=\hat{y} | a=\hat{a},y=\tilde{y}]$ for all $\tilde{a},\hat{a}\in\mathcal{A}$ and $\tilde{y},\hat{y}\in\mathcal{Y}$. 
When interpreting pairwise EO as a pairwise analogue of standard EO (cf. Section~\ref{subsection_fairness_notions}), we considered having a higher label, in a pair of points, as an analogue of $y=1$ being the preferred outcome in standard EO. To derive a pairwise analogue of standard equalized odds, we 
can 
consider 
having a non-higher label (i.e., a smaller or equal label), in a pair of points,  as an analogue of $y=0$ being the 
unpreferred outcome in standard equalized odds for binary classification (with $y\in\{0,1\}$).
We then require 
\begin{align}\label{pairw_fairness_notion_eq_odds}
\Psymb[f(x_1)\leq f(x_2) | a_1=\tilde{a},a_2=\hat{a},y_1\leq y_2]=\Psymb[f(x_1)\geq f(x_2) | a_1=\tilde{a},a_2=\hat{a},y_1\geq y_2]
\end{align}
in addition to \eqref{fairness_notion_EO}. But \eqref{pairw_fairness_notion_eq_odds} is the same as \eqref{fairness_notion_EO}, except that the strict inequalities are replaced by non-strict ones.

\subsection{Proofs of the Remarks Made in Section~\ref{subsection_fairness_notions}}\label{appendix_proofs_remarks}

\begin{itemize}[leftmargin=*]
    \item When $f(x)=i$, then $\Psymb[f(x_1)> f(x_2) | a_1=\tilde{a},a_2=\hat{a}]=0$ and $\Psymb[f(x_1)> f(x_2) | a_1=\tilde{a},a_2=\hat{a},y_1>y_2]=0$ for all $\tilde{a},\hat{a}\in\mathcal{A}$. It is clear that the perfect predictor~$f(x)=y$ satisfies pairwise EO, but
not 
necessarily 
pairwise DP.
    
    \item  If $\Psymb[f(x)=\tilde{y} | a=\tilde{a}]=\Psymb[f(x)=\tilde{y} | a=\hat{a}]$ for all $\tilde{y}\in\mathcal{Y}$ and $\tilde{a},\hat{a}\in\mathcal{A}$, then 
    \begin{align*}
\Psymb[f(x_1)> f(x_2) | a_1=\tilde{a},a_2=\hat{a}]&=\sum_{i=1}^{k-1}\sum_{j=i+1}^k \Psymb[f(x_1)=j, f(x_2)=i | a_1=\tilde{a},a_2=\hat{a}]\\
&=\sum_{i=1}^{k-1}\sum_{j=i+1}^k \Psymb[f(x_1)=j | a_1=\tilde{a}]\Psymb[f(x_2)=i | a_2=\hat{a}]\\
&=\sum_{i=1}^{k-1}\sum_{j=i+1}^k \Psymb[f(x_1)=j | a_1=\hat{a}]\Psymb[f(x_2)=i | a_2=\tilde{a}]\\
&=\sum_{i=1}^{k-1}\sum_{j=i+1}^k \Psymb[f(x_1)=j, f(x_2)=i | a_1=\hat{a},a_2=\tilde{a}]\\
&=\Psymb[f(x_1)> f(x_2) | a_1=\hat{a},a_2=\tilde{a}]\\
&=\Psymb[f(x_1)< f(x_2) | a_1=\tilde{a},a_2=\hat{a}],
\end{align*}
which shows that standard DP implies pairwise DP.

When $k=2$ and   $\Psymb[f(x_1)> f(x_2) | a_1=\tilde{a},a_2=\hat{a}]=\Psymb[f(x_1)< f(x_2) | a_1=\tilde{a},a_2=\hat{a}]$, then $\Psymb[f(x_1)=2,f(x_2)=1 | a_1=\tilde{a},a_2=\hat{a}]=\Psymb[f(x_1)=1,f(x_2)=2 | a_1=\tilde{a},a_2=\hat{a}]$ and 
\begin{align*}
    \Psymb[f(x_1)=2| a_1=\tilde{a}]\cdot 
    \Psymb[f(x_2)=1 | a_2=\hat{a}]=\Psymb[f(x_1)=1 | a_1=\tilde{a}]\cdot 
    \Psymb[f(x_2)=2 | a_2=\hat{a}]&~~\Leftrightarrow\\[5pt]
    \Psymb[f(x_1)=2| a_1=\tilde{a}]\cdot 
    (1-\Psymb[f(x_2)=2 | a_2=\hat{a}])=(1-\Psymb[f(x_1)=2 | a_1=\tilde{a}])\cdot
    \Psymb[f(x_2)=2 | a_2=\hat{a}]&~~\Leftrightarrow\\[5pt]
    \Psymb[f(x_1)=2| a_1=\tilde{a}]=\Psymb[f(x_2)=2 | a_2=\hat{a}]&,
\end{align*}
    which shows that for $k=2$ pairwise DP implies standard DP.
 
 Let $k=3$, $\mathcal{D}=\{(x_1,y_1,0),(x_2,y_2,1),(x_3,y_3,0)\}\subseteq \mathcal{X}\times[3]\times\{0,1\}$, and $f(x_1)=1$, $f(x_2)=2$, $f(x_3)=3$. Then $f$ satisfies pairwise DP on $\mathcal{D}$, but not standard DP, which shows that for general $k$ the two fairness notions are not equivalent.
 
    \item Let $k=2$ and $\mathcal{D}=\{(x_1,1,0),(x_2,1,0),(x_3,1,1),(x_4,2,0),(x_5,2,1),(x_6,2,0)\}\subseteq\mathcal{X}\times[2]\times\{0,1\}$. If $f(x_1)=2$, $f(x_2)=1$, $f(x_3)=1$, $f(x_4)=2$, $f(x_5)=2$, $f(x_6)=1$,  then $f$ satisfies pairwise EO, but not standard EO on $\mathcal{D}$. If $f(x_1)=1$ instead of $f(x_1)=2$,  then $f$ satisfies standard EO, but not pairwise EO on $\mathcal{D}$.
    
    If $k=2$ and 
    $\Psymb[f(x)=1 | a=\tilde{a},y=\tilde{y}]=\Psymb[f(x)=1 | a=\hat{a},y=\tilde{y}]$ for all $\tilde{a},\hat{a}\in\mathcal{A}$ and $\tilde{y}\in\mathcal{Y}=\{0,1\}$, then 
    \begin{align*}
        \Psymb[f(x_1)> f(x_2) | a_1=\tilde{a},a_2=\hat{a},y_1>y_2]&=        \Psymb[f(x_1)=1, f(x_2)=0 | a_1=\tilde{a},a_2=\hat{a},y_1=1,y_2=0]\\
        &=\Psymb[f(x_1)=1 | a_1=\tilde{a},y_1=1]\cdot \Psymb[f(x_2)=0 | a_2=\hat{a},y_2=0]\\
        &=\Psymb[f(x_1)=1 | a_1=\hat{a},y_1=1]\cdot \Psymb[f(x_2)=0 | a_2=\tilde{a},y_2=0]\\
        &=\Psymb[f(x_1)=1, f(x_2)=0 | a_1=\hat{a},a_2=\tilde{a},y_1=1,y_2=0]\\
        &=\Psymb[f(x_1)> f(x_2) | a_1=\hat{a},a_2=\tilde{a},y_1>y_2]\\
         &=\Psymb[f(x_1)< f(x_2) | a_1=\tilde{a},a_2=\hat{a},y_1<y_2],
        \end{align*}
        which shows that for $k=2$ equalized odds implies pairwise EO.

Let $k=3$, $\mathcal{D}=\{(x_1,1,0),(x_2,2,0),(x_3,3,0),(x_4,3,0),(x_5,1,1),(x_6,2,1),(x_7,2,1),(x_8,3,1),(x_9,3,1)\}\subseteq \mathcal{X}\times[3]\times\{0,1\}$, and $f(x_1)=1$, $f(x_2)=2$, $f(x_3)=3$, $f(x_4)=2$, $f(x_5)=1$, $f(x_6)=2$, $f(x_7)=2$, $f(x_8)=3$, $f(x_9)=2$. Then $f$ satisfies equalized odds on $\mathcal{D}$, but not pairwise EO. \hfill$\square$

\end{itemize}

\vspace{5mm}

\begin{figure}[t]
    \centering
    \includegraphics[height=5.3cm]{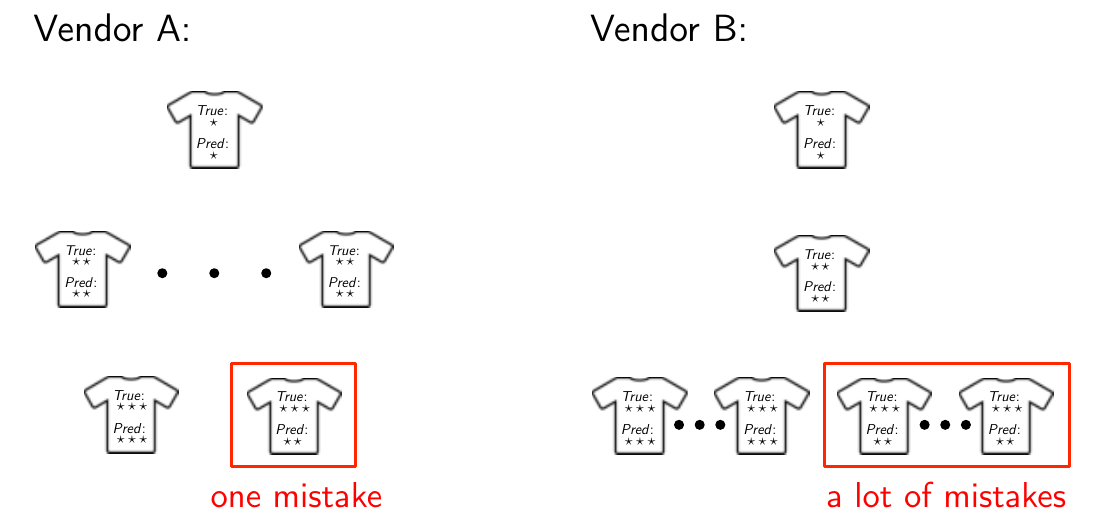}
    
    \caption{Example of a classification scenario in which we deem the predictions unfair to Vendor~B. While the predictions satisfy the fairness notion of equalized odds (and hence also standard EO), our notion of pairwise EO is heavily violated.}\label{fig:appendix_tshirts_example}
\end{figure}

\subsection{Detailed Example Comparing Standard Fairness Notions and our Pairwise Fairness Notions}\label{appendix_example_tshirts}

Consider the scenario that there are two vendors selling T-shirts with ground-truth (quality)  labels in $\{\star,\star\star,\star\star\star\}$, which we want to predict. Assume that Vendor~A sells one T-shirt with a ground-truth label of one star, $n$ T-shirts with a ground-truth label of two stars, and two T-shirts with a ground-truth label of three stars. Assume that Vendor~B sells one T-shirt with a ground-truth label of one star, one T-shirt with a ground-truth label of two stars, and $2n$ T-shirts with a ground-truth label of three stars. 
Finally, 
assume 
that all our predictions are correct, except that for half of the 3-star T-shirts of each vendor we only predict two stars. A sketch of the 
scenario 
is provided in Figure~\ref{fig:appendix_tshirts_example}. Our predictions satisfy equalized odds (and hence also standard EO), but 
our predictions 
appear to be 
very unfair to vendor~B since they incorrectly downgrade a lot of Vendor~B's high-quality T-shirts. Other than equalized odds, 
our notion of pairwise EO is heavily violated by our predictions, 
thus  identifying them as unfair:
\begin{align*}
   & \Psymb[\text{prediction for T-shirt~$i$ $>$ prediction for T-shirt~$j$} ~|~ \text{T-shirt~$i$ from Vendor~A, T-shirt~$j$ from Vendor~B,}\\
   & ~~~~~~~~~~~~~\text{ground-truth label of  T-shirt~$i$ $>$ ground-truth label of T-shirt~$j$} ]=\frac{n+3}{n+4}
   \rightarrow 1~(\text{as }n\rightarrow\infty),\\
   & \Psymb[\text{prediction for T-shirt~$i$ $<$ prediction for T-shirt~$j$} ~|~ \text{T-shirt~$i$ from Vendor~A, T-shirt~$j$ from Vendor~B,}\\
   & ~~~~~~~~~~~~~\text{ground-truth label of  T-shirt~$i$ $<$ ground-truth label of T-shirt~$j$} ]=\frac{1+2n+n^2}{1+2n+2n^2}\rightarrow \frac{1}{2}~(\text{as }n\rightarrow\infty),\\
   &\text{and }\EOviol \rightarrow \frac{1}{2}.
\end{align*}
Our predictions neither satisfy standard DP nor pairwise DP. However, 
both of these 
fairness 
notions do not take the ground-truth labels and the fact that most of Vendor~B's T-shirts are of higher quality than the ones of Vendor~A into account, and 
hence 
they 
are not 
desirable in this scenario.

\subsection{Convergence of $\Fairviol(f;\mathcal{D})$ to $\Fairviol(f;\Psymb)$}\label{appendix_convergnce}

We only consider the case that $\Fairviol=\DPviol$. The case $\Fairviol=\EOviol$ can be treated in an analogous~way.

It is not hard to see that 
    \begin{align*}
        |\DPviol(f;\Psymb)-\DPviol(f;\mathcal{D})|\leq 
        2 \max_{\tilde{a},\hat{a}\in\mathcal{A}}|\Psymb[f(x_1)> f(x_2) | a_1=\tilde{a},a_2=\hat{a}]-\widehat{\Psymb}_n[f(x_1)> f(x_2) | a_1=\tilde{a},a_2=\hat{a}]|, 
    \end{align*}
    where $\widehat{\Psymb}_n$ is the empirical distribution on $\mathcal{D}$, which is an i.i.d. sample from $\Psymb$, and $(x_1,y_1,a_1)$, $(x_2,y_2,a_2)$ are independent samples from $\Psymb$ and $\widehat{\Psymb}_n$, respectively. Let  $\mathcal{D}=((x_i,y_i,a_i))_{i=1}^n$ and fix $\tilde{a},\hat{a}$. Then 
\begin{align*}
	\widehat{\Psymb}_n[f(x_1)> f(x_2) | a_1=\tilde{a},a_2=\hat{a}]=\frac{
		\sum_{i,j\in[n]}
		\charfct[f(x_i)>f(x_j),a_i=\tilde{a},a_j=\hat{a}]}{
		\sum_{i,j\in[n]} \charfct[a_i=\tilde{a},a_j=\hat{a}]}.
\end{align*}

The random variables $\charfct[f(x_i)>f(x_j),a_i=\tilde{a},a_j=\hat{a}]_{i,j\in[n]}$ are a read-$n$ family \citep[][Definition~1]{gavinsky_tail_bound}. It is $\Psymb[\charfct[f(x_i)>f(x_j),a_i=\tilde{a},a_j=\hat{a}]=1]=\Psymb[f(x_i)>f(x_j),a_i=\tilde{a},a_j=\hat{a}]$ for $i\neq j$ and $\Psymb[\charfct[f(x_i)>f(x_j),a_i=\tilde{a},a_j=\hat{a}]=1]=0$ for $i=j$. 
It follows from  
\cite[Theorem 1.1]
{gavinsky_tail_bound} that for any $\varepsilon>0$,  
with 
probability $1-2e^{-2\varepsilon^2n}$ over the sample $\mathcal{D}$ we have
\begin{align*}
	\frac{\sum_{i,j\in[n]}\charfct[f(x_i)>f(x_j),a_i=\tilde{a},a_j=\hat{a}]}{n^2}
	\leq \Psymb[f(x_1)>f(x_2),a_1=\tilde{a},a_2=\hat{a}]+\varepsilon
\end{align*}
as well as 
\begin{align*}
	\frac{\sum_{i,j\in[n]}\charfct[f(x_i)>f(x_j),a_i=\tilde{a},a_j=\hat{a}]}{n^2}\geq \Psymb[f(x_1)>f(x_2),a_1=\tilde{a},a_2=\hat{a}]-\frac{1}{n}-\varepsilon.
\end{align*}
Similarly, we have with 
probability $1-2e^{-2\varepsilon^2n}$ over the sample $\mathcal{D}$ that 
\begin{align*}
\Psymb[a_1=\tilde{a},a_2=\hat{a}]-\varepsilon -\frac{1}{n}\leq 	\frac{\sum_{i,j\in[n]}\charfct[a_i=\tilde{a},a_j=\hat{a}]}{n^2}
	\leq \Psymb[a_1=\tilde{a},a_2=\hat{a}]+\varepsilon.
\end{align*}
It follows that if $\Psymb[a_i=\tilde{a},a_j=\hat{a}]$ is lower bounded by some positive constant, for any $\delta\in(0,1)$, with 
probability $1-\delta$ over the sample $\mathcal{D}$ we have
\begin{align*}
    |\Psymb[f(x_1)> f(x_2) | a_1=\tilde{a},a_2=\hat{a}]-\widehat{\Psymb}_n[f(x_1)> f(x_2) | a_1=\tilde{a},a_2=\hat{a}]|\leq M \sqrt{\frac{\log\frac{1}{\delta}}{n}}.
\end{align*}
for some constant~$M$.  
This implies that with 
probability $1-\delta$  over the sample $\mathcal{D}$ we have
\begin{align*}
    \max_{\tilde{a},\hat{a}\in\mathcal{A}}|\Psymb[f(x_1)> f(x_2) | a_1=\tilde{a},a_2=\hat{a}]-\widehat{\Psymb}_n[f(x_1)> f(x_2) | a_1=\tilde{a},a_2=\hat{a}]|\leq M \sqrt{\frac{\log\frac{|\mathcal{A}|^2}{\delta}}{n}}.
\end{align*}

\subsection{Proof of Proposition~\ref{proposition_reduction_to_fair_binary_classification}}\label{appendix_proof_reduction_to_binary_classification}

It is for all $\tilde{a},\hat{a}\in \mathcal{A}$
\begin{align*}
&\Psymb_{(x',y',a')\sim\mathcal{D}'}[c_w(x')=1|a'=(\tilde{a},\hat{a})]=\\[10pt]
&~~~~~~\frac{1}{|\{(x',y',a')\in\mathcal{D}':a'=(\tilde{a},\hat{a})\}|}\sum_{(x',y',a')\in\mathcal{D}':a'=(\tilde{a},\hat{a})}\charfct\{\sign(w\cdot x')=1\}=\\[10pt]
&~~~~~~\frac{\sum_{((x_i,y_i,a_i),(x_j,y_j,a_j))\in\mathcal{D}\times\mathcal{D}:\, y_i\neq y_j,a_i=\tilde{a},a_j=\hat{a}}\charfct\{w\cdot (x_i-x_j)> 0\}}{|\{((x_i,y_i,a_i),(x_j,y_j,a_j))\in\mathcal{D}\times\mathcal{D}:\, y_i\neq y_j,a_i=\tilde{a},a_j=\hat{a}\}|}=\\[10pt]
&~~~~~~\Psymb_{((x_i,y_i,a_i),(x_j,y_j,a_j))\sim\mathcal{D}^2}[s_w(x_i)> s_w(x_j)|a_i=\tilde{a},a_j=\hat{a},y_i\neq y_j],
\end{align*}
which implies the statement for DP.

\vspace{4mm}
Similarly, for all $\tilde{a},\hat{a}\in \mathcal{A}$ we have
\begin{align*}
&\Psymb_{(x',y',a')\sim\mathcal{D}'}[c_w(x')=1|a'=(\tilde{a},\hat{a}),y'=1]=\\[10pt]
&~~~~~~\frac{1}{|\{(x',y',a')\in\mathcal{D}':a'=(\tilde{a},\hat{a}),y'=1\}|}\sum_{(x',y',a')\in\mathcal{D}':a'=(\tilde{a},\hat{a}),y'=1}\charfct\{\sign(w\cdot x')=1\}=\\[10pt]
&~~~~~~\frac{\sum_{((x_i,y_i,a_i),(x_j,y_j,a_j))\in\mathcal{D}\times\mathcal{D}:\, y_i> y_j,a_i=\tilde{a},a_j=\hat{a}}\charfct\{w\cdot (x_i-x_j)> 0\}}{|\{((x_i,y_i,a_i),(x_j,y_j,a_j))\in\mathcal{D}\times\mathcal{D}:\, y_i>y_j,a_i=\tilde{a},a_j=\hat{a}\}|}=\\[10pt]
&~~~~~~\Psymb_{((x_i,y_i,a_i),(x_j,y_j,a_j))\sim\mathcal{D}^2}[s_w(x_i)> s_w(x_j)|a_i=\tilde{a},a_j=\hat{a},y_i> y_j],
\end{align*}
which implies the statement for EO.
\hfill$\square$

\subsection{Proof of Proposition~\ref{proposition_dynamic_programming}}\label{appendix_proof_proposition_dynamic_programming}

We write $s_i=s(x_i)$, $i\in[n]$, for the values of the scoring function~$s$ on the training data~$\mathcal{D}=((x_i,y_i,a_i))_{i=1}^n$ and assume the values to be sorted, that is $s_i\leq s_{i+1}$, $i\in[n-1]$, and given, that is we do not take the time for evaluating $s$ into account. 
Here we only consider the case of a binary protected attribute:  let $\mathcal{A}=\{\tilde{a},\hat{a}\}$, and let $\tilde{G}=\{i\in[n]:a_i=\tilde{a}\}$ and $\hat{G}=\{i\in[n]:a_i=\hat{a}\}$.

\vspace{2mm}
\underline{When aiming for pairwise DP:}

\vspace{1mm}
We build a table $T\in (\R_{\geq 0}\cup\{\infty\})^{(n+1)\times k\times k  \times (2\lfloor \frac{n^2}{4}\rfloor +1)}$ with
\begin{align*}%
T(i,f,l,v)=\min_{\hat{y}\in \mathcal{H}_{i,f,l,v}}\sum_{j=1}^i C_{y_j,\hat{y}_j}
\end{align*}
for $i\in\{0\}\cup[n]$, $f,l\in[k]$ and $v\in\{-\lfloor \frac{n^2}{4}\rfloor,\ldots,\lfloor \frac{n^2}{4}\rfloor\}$, where
\begin{align*}
\mathcal{H}_{i,f,l,v}&=\Big\{\hat{y}=(\hat{y}_1,\ldots,\hat{y}_i)\in[k]^i: \hat{y}~\text{are predictions for %
$x_1,\ldots,x_i$
that are sorted, that is $y_r\leq y_{r+1}$ for $r\in [i-1]$,}\\
&~~~~~~~~~~~~\text{with $\hat{y}_i=l$, take at most $f$ different values and satisfy $\hat{y}_r=\hat{y}_{r'}$ for $s_r=s_{r'}$, and for which}\\  
&~~~~~~~~~~~~~\text{$\sum_{1\leq r, r'\leq i:\, r\in\tilde{G}, r'\in\hat{G}} \big[\charfct\{\hat{y}_r>\hat{y}_{r'}\}-\charfct\{\hat{y}_r<\hat{y}_{r'}\}\big]=v$} \Big\}
\end{align*}
and $T(i,f,l,v)=\infty$ if $\mathcal{H}_{i,f,l,v}=\emptyset$. Note that $\sum_{1\leq r, r'\leq i:\, r\in\tilde{G}, r'\in\hat{G}} 1\leq |\tilde{G}|\cdot|\hat{G}|\leq\lfloor \frac{n^2}{4}\rfloor$. 
Given the table~$T$, we can compute the optimal value of \eqref{threshold_objective} as 
\begin{align*}
\min_{l\in[k],v\in \{-\lfloor \frac{n^2}{4}\rfloor,\ldots,\lfloor \frac{n^2}{4}\rfloor\}}\frac{1}{n}\cdot T(n,k,l,v)+\frac{\lambda}{|\tilde{G}|\cdot |\hat{G}|}\cdot |v|.
\end{align*}

\vspace{4mm}
It is
\begin{align*}
T(0,f,l,0)&=0,\quad f,l\in[k], \qquad T(0,f,l,v)=\infty \quad f,l\in[k],v \neq 0,\\
T(1,f,l,0)&=C_{y_1,l} \quad f,l\in[k], \qquad T(1,f,l,v)=\infty \quad f,l\in[k],v \neq 0, \\
T(i,1,l,0)&=\sum_{j=1}^iC_{y_j,l} \quad i\in[n],l\in[k], \qquad T(i,1,l,v)=\infty \quad i\in[n],l\in[k], v\neq 0, \\
T(i,f,1,0)&=\sum_{j=1}^iC_{y_j,1} \quad i\in[n],f\in[k], \qquad T(i,f,1,v)=\infty \quad i\in[n],f\in[k], v\neq 0, 
\end{align*}
and 
\begin{align*}
T(i,f,l,v)&=\min_{i'<i,l'<l}\left[T(i',f-1,l',v^*)+\infty\cdot\charfct\{i'>0 \wedge (s_{i'}=s_{i'+1})\}+\sum_{r=i'+1}^i C_{y_r,l}\right],
\end{align*}
where $v^*=v-|\{{i'+1},\ldots,i\}\cap \tilde{G}|\cdot |[i']\cap \hat{G}|+|\{{i'+1},\ldots,i\}\cap \hat{G}|\cdot |[i']\cap \tilde{G}|$.

\vspace{3mm}
We can build $T$ in time $\mathcal{O}(n^4 k^3)$. If we store the minimizing $(i',l')$ with $T(i,f,l,v)$, we can easily construct an optimal solution from $T$.

\vspace{5mm}
\underline{When aiming for pairwise EO:}

\vspace{1mm}
We build a table $T\in (\R_{\geq 0}\cup\{\infty\})^{(n+1)\times k\times k  \times (\lfloor \frac{n^2}{4}\rfloor +1)\times (\lfloor \frac{n^2}{4}\rfloor +1)}$ with
\begin{align*}%
T(i,f,l,v_1,v_2)=\min_{\hat{y}\in \mathcal{H}_{i,f,l,v_1,v_2}}\sum_{j=1}^i C_{y_j,\hat{y}_j}
\end{align*}
for $i\in\{0\}\cup[n]$, $f,l\in[k]$ and $v_1,v_2\in\{0\}\cup[\lfloor \frac{n^2}{4}\rfloor]$, where
\begin{align*}
&\mathcal{H}_{i,f,l,v_1,v_2}=
\Big\{\hat{y}=(\hat{y}_1,\ldots,\hat{y}_i)\in[k]^i: \hat{y}~\text{are predictions for %
$x_1,\ldots,x_i$
that are sorted, that is $y_r\leq y_{r+1}$ for} \\
&~~~~~~~~~\text{$r\in [i-1]$, with $\hat{y}_i=l$, take at most $f$ different values and satisfy $\hat{y}_r=\hat{y}_{r'}$ for $s_r=s_{r'}$, and for which}\\  
&~~~~~~~~~~~~\text{$\sum_{1\leq r, r'\leq i:\, r\in\tilde{G}, r'\in\hat{G}}
\charfct\{\hat{y}_r\geq \hat{y}_{r'} \wedge y_r<y_{r'}\}=v_1$ and $\sum_{1\leq r, r'\leq i:\, r\in\tilde{G}, r'\in\hat{G}}\charfct\{\hat{y}_r\leq \hat{y}_{r'} \wedge y_r>y_{r'}\}=v_2$} \Big\}
\end{align*}
and $T(i,f,l,v)=\infty$ if $\mathcal{H}_{i,f,l,v_1,v_2}=\emptyset$. 
Note that, just as before, $\sum_{1\leq r, r'\leq i:\, r\in\tilde{G}, r'\in\hat{G}} 1\leq \lfloor \frac{n^2}{4}\rfloor$. 
Given the table~$T$, we can compute the optimal value of \eqref{threshold_objective} as 
\begin{align*}
\min_{l\in[k],v_1,v_2\in \{0\}\cup [\lfloor \frac{n^2}{4}\rfloor]}\frac{1}{n}\cdot T(n,k,l,v_1,v_2)+\lambda\cdot \left|\frac{v_1}{|\{(r,r')\in \tilde{G}\times \hat{G}: y_r<y_{r'}\}|}
-\frac{v_2}{|\{(r,r')\in \tilde{G}\times \hat{G}: y_r>y_{r'}\}|}\right|.
\end{align*}

\vspace{4mm}
It is for all $i\in[n]$ and $f,l\in[k]$
\begin{align*}
T(0,f,l,0,0)&=0, ~~~\qquad T(0,f,l,v_1,v_2)=\infty, \quad (v_1,v_2) \neq (0,0),\\
T(1,f,l,0,0)&=C_{y_1,l}, ~~~\qquad T(1,f,l,v_1,v_2)=\infty,\quad(v_1,v_2) \neq (0,0), \\
T(i,1,l,v_1,v_2)&=
\begin{cases}
\sum_{j=1}^i C_{y_j,l} & (v_1,v_2)=\\
&~~~~(\sum_{1\leq r, r'\leq i:\, r\in\tilde{G}, r'\in\hat{G}}
\charfct\{y_r<y_{r'}\},\sum_{1\leq r, r'\leq i:\, r\in\tilde{G}, r'\in\hat{G}}
\charfct\{\hat{y}_r\geq \hat{y}_{r'} \wedge y_r>y_{r'}\})\\
\infty & \text{else}
\end{cases},\\
T(i,f,1,v_1,v_2)&=
\begin{cases}
\sum_{j=1}^i C_{y_j,1} & (v_1,v_2)=\\
&~~~~(\sum_{1\leq r, r'\leq i:\, r\in\tilde{G}, r'\in\hat{G}}
\charfct\{y_r<y_{r'}\},\sum_{1\leq r, r'\leq i:\, r\in\tilde{G}, r'\in\hat{G}}
\charfct\{\hat{y}_r\geq \hat{y}_{r'} \wedge y_r>y_{r'}\})\\
\infty & \text{else}
\end{cases}
\end{align*}
and 
\begin{align*}
T(i,f,l,v_1,v_2)&=\min_{i'<i,l'<l}\left[T(i',f-1,l',v^*_1,v^*_2)+\infty\cdot\charfct\{i'>0 \wedge (s_{i'}=s_{i'+1})\}+\sum_{r=i'+1}^i C_{y_r,l}\right],
\end{align*}
where 
\begin{align*}
v^*_1=v_1-|\{(r,r')\in(\tilde{G}\cap\{{i'+1},\ldots,i\})\times (\hat{G}\cap[i]): y_a<y_b\}|,\\
v^*_2=v_2-|\{(r,r')\in (\tilde{G}\cap[i])\times (\hat{G} \cap\{{i'+1},\ldots,i\}): y_a>y_b\}|.
\end{align*}

\vspace{3mm}
Using two helper tables $H_1,H_2\in (\N\cup\{0\})^{n\times n}$ with 
\begin{align*}
H_1(i,j)=|\{(r,r')\in(\tilde{G}\cap[i])\times (\hat{G}\cap[j]): y_r<y_{r'}\}|,\\
H_2(i,j)=|\{(r,r')\in(\tilde{G}\cap[i])\times (\hat{G}\cap[j]): y_r>y_{r'}\}|,
\end{align*}
which we can certainly build in time $\mathcal{O}(n^6)$, we can build $T$ in time $\mathcal{O}(n^6 k^3)$. If we store the minimizing $(i',l')$ with $T(i,f,l,v)$, we can easily construct an optimal solution from $T$.
\hfill$\square$

\subsection{Addendum to Section~\ref{subsection_generalization}}\label{appendix_generalization}

We first provide the statement for $\EOviol$.

\vspace{2mm}
\emph{Under the same assumptions as in Theorem~\ref{proposition_generalization}, but instead of $\Psymb[a=\tilde{a}]\geq \beta>0$ for all $\tilde{a}\in\mathcal{A}$ requiring that $\Psymb[a=\tilde{a},y=j']\geq \beta>0$ for all $\tilde{a}\in\mathcal{A},j'\in[k]$, our learned predictor $f=f(\cdot\,;s,\boldsymbol\theta)$ satisfies with probability at least $1-\delta$ over the training sample~$\mathcal{D}$ of size $n$, for $n$ sufficiently~large, the bound on $\MAE(f;\Psymb)$ from Theorem~\ref{proposition_generalization} and
\begin{align*}
    |\EOviol(f;\Psymb)-\EOviol(f;\mathcal{D})| \leq M k^4  \sqrt{\frac{d+\log\frac{M|\mathcal{A}|^2k}{\delta}}{\left(1-\sqrt{\frac{2\log\frac{M|\mathcal{A}|^2k}{\delta}}{n\beta}}\right)n\beta}}+\frac{M}{\beta^2} \sqrt{\frac{\log\frac{M|\mathcal{A}|^2k}{\delta}}{2n}}.
\end{align*}
}

We prove the bounds on $\MAE$, $\DPviol$ and $\EOviol$ separately. Theorem~\ref{proposition_generalization} and the statement above then follow from a simple union bound.

\vspace{2mm}
\begin{itemize}
    \item \textbf{Bound on $\MAE(f;\Psymb)$}

\vspace{1mm}
According to Theorem~4 of \citet{zhang02_covering_numbers}, the covering number $\mathcal{N}_{\infty}(\gamma,\mathcal{F},m)$, where $\mathcal{F}$ is the class of linear scoring functions $s_w(x)=w\cdot x$ with $\|w\|_2\leq \sqrt{\nu}$ 
and domain 
$B_R(0)$, satisfies
\begin{align*}
\log_2 \mathcal{N}_{\infty}(\gamma,\mathcal{F},m) \leq 36 \frac{R^2\nu}{\gamma^2}
\log_2\left(2\left\lceil\frac{4R\sqrt{\nu}}{\gamma}+2\right\rceil m+1\right).
\end{align*}
Furthermore, since $|s_w(x)|\leq \|w\|_2\cdot\|x\|_2$, our strategy learns thresholds in $[-R\sqrt{\nu},+R\sqrt{\nu}]$. 
The bound on $\MAE(f;\Psymb)$ is now a simple corollary of Theorem~7 of \citet{agarwal2008_generalization}.

The empirical $\gamma$-margin loss $\widehat{L}_{\mathcal{D}}^{\gamma}(s,\boldsymbol\theta)$, with $\mathcal{D}=((x_i,y_i,a_i))_{i=1}^n$, is defined as 
\begin{align*}
    \widehat{L}_{\mathcal{D}}^{\gamma}(s,\boldsymbol\theta)=\frac{1}{n}\sum_{i=1}^n l_\gamma(s,\boldsymbol\theta,(x_i,y_i))
\end{align*}
with
\begin{align*}
    l_\gamma(s,\boldsymbol\theta,(x,y))=\sum_{j=1}^{k-1}\charfct\{y^j(s(x)-\theta_j)\leq \gamma\},
\end{align*}
where
\begin{align*}
y^j=\begin{cases}
+1 & \text{if }j\in\{1,\ldots,y-1\}  \\
-1 & \, \text{if }j\in\{y,\ldots,k-1\}
\end{cases}.    
\end{align*}

\vspace{2mm}
\item \textbf{Bound on $\DPviol(f;\Psymb)$}

\vspace{1mm}
It is not hard to see that 
    \begin{align*}
        &|\DPviol(f;\Psymb)-\DPviol(f;\mathcal{D})|\leq \\
        &~~~~~~~~~~~~~~~~2 \max_{\tilde{a},\hat{a}\in\mathcal{A}}|\Psymb[f(x_1)> f(x_2) | a_1=\tilde{a},a_2=\hat{a}]-\widehat{\Psymb}_n[f(x_1)> f(x_2) | a_1=\tilde{a},a_2=\hat{a}]|, 
    \end{align*}
    where $\widehat{\Psymb}_n$ is the empirical distribution on $\mathcal{D}$.
For now, let us consider fixed $\tilde{a},\hat{a}$. We have   
\begin{align*}
\Psymb[f(x_1)> f(x_2) | a_1=\tilde{a},a_2=\hat{a}]=\sum_{i=1}^{k-1}\sum_{j=i+1}^k \Psymb[f(x_1)=j | a_1=\tilde{a}]\cdot\Psymb[f(x_2)=i | a_2=\hat{a}]
\end{align*}
and 
\begin{align*}
   \widehat{\Psymb}_n[f(x_1)> f(x_2) | a_1=\tilde{a},a_2=\hat{a}]=\sum_{i=1}^{k-1}\sum_{j=i+1}^k\left[\frac{1}{\tilde{n}}\sum_{s=1}^{\tilde{n}}\charfct[f(\tilde{x}_s)=j]\cdot \frac{1}{\hat{n}}\sum_{t=1}^{\hat{n}} \charfct[f(\hat{x}_t)=i]\right], 
\end{align*}
where we write $(\tilde{x}_i,\tilde{y}_i,\tilde{a}_i)_{i=1}^{\tilde{n}}$ for those training sample points with $a_i=\tilde{a}$ and  $(\hat{x}_i,\hat{y}_i,\hat{a}_i)_{i=1}^{\hat{n}}$ for those  with $a_i=\hat{a}$. 

We assume $f$ to be a threshold model with a linear scoring function $s(x)=w\cdot x$, that is 
\begin{align*}
    f(x)=j \quad \Leftrightarrow \quad \theta_{j-1}< w\cdot x\leq \theta_j.
\end{align*}
 Let $\Psymb^{\tilde{mod}}$ be a distribution on $\mathcal{X}\times \{0,1\}$, corresponding to random variables $x$ and $y$, with $\Psymb^{\tilde{mod}}[y=0]=1$ and $\Psymb^{\tilde{mod}}[x]=\Psymb[x|a=\tilde{a}]$. Note that $(\tilde{x}_i,0)_{i=1}^{\tilde{n}}$ is a sample from $\Psymb^{\tilde{mod}}$. We have
 \begin{align*}
     \Psymb[f(x)=j | a=\tilde{a}]&=\Psymb^{\tilde{mod}}[\charfct[\theta_{j-1}< w\cdot x\leq \theta_j]=1]=\Ex_{(x,y)\sim \Psymb^{\tilde{mod}}}[\text{01-loss}(y,\charfct[\theta_{j-1}< w\cdot x\leq \theta_j])] 
 \end{align*}
and
\begin{align*}
    \frac{1}{\tilde{n}}\sum_{s=1}^{\tilde{n}}\charfct[f(\tilde{x}_s)=j]&=\frac{1}{\tilde{n}}\sum_{s=1}^{\tilde{n}}\charfct[\theta_{j-1}< w\cdot \tilde{x}_s\leq \theta_j]=\frac{1}{\tilde{n}}\sum_{s=1}^{\tilde{n}}\text{01-loss}(0,\charfct[\theta_{j-1}< w\cdot \tilde{x}_s\leq \theta_j]).
\end{align*}

We assume $x\in\R^d$. The class of halfspace functions $\{x\mapsto \charfct[w\cdot x\geq b]:w\in\R^d,b\in\R\}$ has VC-dimension $d+1$ \citep[e.g.,][Theorem 9.3]{shalev2014understanding}, and it follows from a theorem in \citet{blumer1989} (also stated as Theorem~A in \citealp{csikos2019}) that there exists a constant~$M$ such that the class of functions $\{x\mapsto \charfct[\theta_{j-1}< w\cdot x\leq \theta_j]:w\in\R^d,\theta_{j-1}, \theta_j\in\R\}$ has VC-dimension at most $M d$ (in the following, we write $M$ for an absolute constant that may change from line to line).
According to the fundamental theorem of statistical learning \citep[e.g.,][Theorem~6.8]{shalev2014understanding}, for fixed $\delta\in(0,1)$, we have with probability of at least $1-\delta$ over the sample $(\tilde{x}_i,\tilde{y}_i,\tilde{a}_i)_{i=1}^{\tilde{n}}$ (with $\tilde{n}$ fixed)
\begin{align*}
    \sup_{w\in\R^d,\theta_{j-1}, \theta_j\in\R} \left|\Psymb[f(x)=j | a=\tilde{a}]-\frac{1}{\tilde{n}}\sum_{s=1}^{\tilde{n}}\charfct[f(\tilde{x}_s)=j]\right|\leq M\sqrt{\frac{d+\log\frac{1}{\delta}}{\tilde{n}}}.  
\end{align*}

Assuming that $\Psymb[a=\tilde{a}]\geq \beta$ 
it follows from Chernoff's bound and a simple union bound that for any $\delta\in(0,1)$ with probability at least  $1-2\delta$ over the sample $(x_i,y_i,a_i)_{i=1}^n$ we have
\begin{align*}
 &   \tilde{n}\geq \left(1-\sqrt{\frac{2\log\frac{1}{\delta}}{n\beta}}\right)n\beta\qquad\text{and}\\
  &   \forall j:    \sup_{w\in\R^d,\theta_{j-1}, \theta_j\in\R} \left|\Psymb[f(x)=j | a=\tilde{a}]-\frac{1}{\tilde{n}}\sum_{s=1}^{\tilde{n}}\charfct[f(\tilde{x}_s)=j]\right|\leq M \sqrt{\frac{d+\log\frac{1}{\delta}}{\left(1-\sqrt{\frac{2\log\frac{1}{\delta}}{n\beta}}\right)n\beta}}.
\end{align*}
Assuming that $\Psymb[a=\tilde{a}]\geq \beta$ for all $\tilde{a}\in\mathcal{A}$, it follows that   with probability $1-2|\mathcal{A}|\delta$ over the sample $(x_i,y_i,a_i)_{i=1}^n$ we have 
\begin{align*}
\forall \tilde{a}\in\mathcal{A}, \forall j:    \sup_{w\in\R^d,\theta_{j-1}, \theta_j\in\R} \left|\Psymb[f(x)=j | a=\tilde{a}]-\frac{1}{\tilde{n}}\sum_{s=1}^{\tilde{n}}\charfct[f(\tilde{x}_s)=j]\right|\leq M \sqrt{\frac{d+\log\frac{1}{\delta}}{\left(1-\sqrt{\frac{2\log\frac{1}{\delta}}{n\beta}}\right)n\beta}}.
\end{align*}
Since
$|uv-\hat{u}\hat{v}|\leq 3\max\{|u-\hat{u}|,|v-\hat{v}|\}$ for $u,v,\hat{u},\hat{v}\in[0,1]$,
it follows that with probability $1-2|\mathcal{A}|\delta$ 
we have
\begin{align*}
    \max_{\tilde{a},\hat{a}\in\mathcal{A}}\sup_f |\Psymb[f(x_1)> f(x_2) | a_1=\tilde{a},a_2=\hat{a}]-\widehat{\Psymb}_n[f(x_1)> f(x_2) | a_1=\tilde{a},a_2=\hat{a}]|\leq M k^2  \sqrt{\frac{d+\log\frac{1}{\delta}}{\left(1-\sqrt{\frac{2\log\frac{1}{\delta}}{n\beta}}\right)n\beta}}.
\end{align*}
This implies  that with probability $1-\delta$ over the sample $(x_i,y_i,a_i)_{i=1}^n$ we have
\begin{align*}
    \sup_f |\DPviol(f;\Psymb)-\DPviol(f;\mathcal{D})| \leq M k^2 \sqrt{\left(d+\log\frac{2|\mathcal{A}|}{\delta}\right)\cdot\left[{\left(1-\sqrt{\frac{2\log\frac{2|\mathcal{A}|}{\delta}}{n\beta}}\right)n\beta}\right]^{-1}}.
\end{align*}
In particular, this provides a bound on $\DPviol(f;\Psymb)$ for the threshold model $f$ learned by our strategy.

\vspace{2mm}
\item \textbf{Bound on $\EOviol(f;\Psymb)$}

The bound on $\EOviol(f;\Psymb)$ can be derived in a similar way as the bound on $\DPviol(f;\Psymb)$, and we refrain from presenting the details here.

\vspace{1mm}
\end{itemize}

\subsection{Proof of Lemma~\ref{lemma_fairness_in_both_steps_necessary}}\label{appendix_proof_both_steps_necessary}

We choose $S$ to consist only of the two projections onto the first and second coordinate, respectively, that is $S=\{p_1:\R^2\rightarrow \R~\text{with}~p_1(u_1,u_2)=u_1, p_2:\R^2\rightarrow \R~\text{with}~p_2(u_1,u_2)=u_2\}$.
\begin{enumerate}[wide, labelwidth=0pt, labelindent=0pt]
\item ~

\begin{itemize}
    \item 
Aiming for pairwise DP:

Consider $n=2m+1$ many 
datapoints $(x_1,y_1,a_1)\ldots,(x_n,y_n,a_n)\in\R^2\times[4]\times\{0,1\}$ with 
\begin{align*}
x_i=
\begin{cases}
(1,3) &i=1,\\
(i,2)& 2\leq i\leq m+1\\
(i,4)&  m+2\leq i\leq 2m+1
\end{cases},  
\qquad 
y_i=
\begin{cases}
1 & i=1,\\
2 &  2\leq i \leq m+1,\\
4 & m+2\leq i\leq 2m+1
\end{cases},
\qquad 
a_i=
\begin{cases}
0 & i=1,\\
1 &  2\leq i \leq 2m+1
\end{cases}.
\end{align*}
The scoring function $p_1$ does not suffer from any label flips whereas the scoring function $p_2$ incurs label flips for $m$ many datapoint pairs. However, $p_2$ satisfies pairwise DP whereas $p_1$ does not. 
When choosing thresholds that minimize the $\MAE$ under the constraint that the resulting predictor~$f$ has to satisfy pairwise DP, when using $p_1$ as scoring function, we must choose thresholds such that $f(x_i)=2$ for all $i\in\N$, and  if we use $p_1$, we obtain $\MAE=1$. If we
use $p_2$ as scoring function, we can choose thresholds $\theta_1=0, \theta_2=2.5, \theta_3=3.5$ and obtain a predictor~$f$ with $\MAE=\frac{2}{2m+1}$.

\vspace{4mm}
\item Aiming for pairwise EO: 

Consider $n=2m+5$ many 
datapoints $(x_1,y_1,a_1)\ldots,(x_n,y_n,a_n)\in\R^2\times[4]\times\{0,1\}$ with 
\begin{align*}
&(x_1,y_1,a_1)=((1,1),1,0),\qquad ~    (x_2,y_2,a_2)=((2,3),1,1), \qquad ~  (x_3,y_3,a_3)=((3,2),2,0),\\
&(x_4,y_4,a_4)=((4,4),2,1),\qquad  ~   (x_5,y_5,a_5)=((5,n+1),4,1)\\
&(x_i,a_i,y_i)=((i,i),3,1), \quad 6\leq i \leq m+5,\qquad (x_i,a_i,y_i)=((i,i),4,1), \quad m+6\leq i \leq 2m+5.
\end{align*}
The scoring function~$p_1$ incurs label flips for $m$ many datapoint pairs whereas the scoring function~$p_2$ incurs label flips for only one datapoint pair. 
However, $p_1$ satisfies pairwise EO whereas $p_2$ does not. 
When choosing thresholds that minimize the $\MAE$ under the constraint that the resulting predictor~$f$ has to satisfy pairwise EO, when using $p_2$ as scoring function, we have to choose thresholds such that $f(x_i)=3$ for all $i\in\N$, and  if we use $p_2$, we obtain $\MAE=\frac{6+m}{2m+5}>\frac{1}{2}$. If we
use $p_1$ as scoring function, we can choose thresholds $\theta_1=2.5, \theta_2=4.5, \theta_3=m+5.5$ and obtain a predictor~$f$ with $\MAE=\frac{1}{n}$.

\end{itemize}

\vspace{6mm}
\item ~

\begin{itemize}
    \item Aiming for pairwise DP:

Consider six datapoints $(x_1,y_1,a_1)\ldots,(x_6,y_6,a_6)\in\R^2\times[4]\times\{0,1\}$ with 
\begin{align*}
    (x_1,y_1,a_1)=((1,0),1,0),\qquad ~    (x_2,y_2,a_2)=((2,0),1,0), \qquad ~  (x_3,y_3,a_3)=((3,1),1,1),\\
    (x_4,y_4,a_4)=((4,1),1,1),\qquad  ~   (x_5,y_5,a_5)=((5,0),2,0), \qquad  ~  (x_6,y_6,a_6)=((6,0),2,0).
\end{align*}
The scoring function~$p_1$ satisfies pairwise DP whereas $p_2$ does not.
Using $p_1$, we can choose thresholds $\theta_1=4.5, \theta_2=\theta_3=7$ in order to obtain a predictor with $\MAE=0$. However, for this predictor we have  $\DPviol=0.5$.

\vspace{4mm}
\item Aiming for pairwise EO:

Consider six datapoints $(x_1,y_1,a_1)\ldots,(x_6,y_6,a_6)\in\R^2\times[4]\times\{0,1\}$ with 
\begin{align*}
    (x_1,y_1,a_1)=((1,0),1,0),\qquad ~    (x_2,y_2,a_2)=((2,1),2,1), \qquad ~  (x_3,y_3,a_3)=((3,1),1,1),\\
    (x_4,y_4,a_4)=((4,1),1,1),\qquad  ~   (x_5,y_5,a_5)=((5,0),2,0), \qquad  ~  (x_6,y_6,a_6)=((6,0),2,0).
\end{align*}
The scoring function~$p_1$ satisfies pairwise EO whereas $p_2$ does not.
Using $p_1$, if we choose thresholds that minimize the $\MAE$, we can choose $\theta_1=4.5, \theta_2=\theta_3=7$ and obtain a predictor with $\MAE=\frac{1}{6}$. However, for this predictor we have  $\EOviol=1$.

\end{itemize}

\end{enumerate}
\hfill$\square$

\subsection{Proof of Lemma~\ref{lemma_fairness_scoring_vs_threshold_model}}\label{appendix_proof_fairness_scoring_vs_threshold_model}

Let 
\begin{align*}
     &A=\Psymb[s(x_1)> s(x_2) | a_1=\tilde{a},a_2=\hat{a}],\\
     &B=\Psymb[s(x_1)< s(x_2) | a_1=\tilde{a},a_2=\hat{a}],\\
     &C=\Psymb[s(x_1)= s(x_2) | a_1=\tilde{a},a_2=\hat{a}],\\
     &D=\Psymb[f(x_1)> f(x_2) | a_1=\tilde{a},a_2=\hat{a}],\\
     &E=\Psymb[f(x_1)< f(x_2) | a_1=\tilde{a},a_2=\hat{a}].
\end{align*}

 It is $A+B+C=1$ 
and $|A-B|\leq \DPviol(s)$.  
It follows that 
\begin{align*}
A,B\in\left[\frac{1-C-\DPviol(s)}{2},\frac{1-C+\DPviol(s)}{2}\right].    
\end{align*}
Since $f(x_1)> f(x_2)$ implies that $s(x_1)> s(x_2)$ and $f(x_1)<f(x_2)$ implies that $s(x_1)< s(x_2)$, we have $D\leq A$ and $E\leq B$. Hence,
\begin{align*}
D,E\in\left[0,\frac{1-C+\DPviol(s)}{2}\right]
\end{align*}
and 
\begin{align*}
    |D-E|\leq \frac{1-C+\DPviol(s)}{2}\leq \frac{1}{2}+\frac{\DPviol(s)}{2}.
\end{align*}
\hfill$\square$

\subsection{Performing Local Search to Minimize \eqref{threshold_objective}}\label{appendix_local_search}

We first sort the datapoints such that $s(x_j)\leq s(x_{j+1})$, $j\in[n-1]$. Including the time it takes to evaluate $s$, this can be done in time~$\mathcal{O}(nd+n\log n)$. 
We can then perform a local search in order to find a local minimum of \eqref{threshold_objective} as follows: given 
thresholds $\theta_1\leq\ldots\leq\theta_{k-1}$, in each round, we move one threshold~$\theta_i$ to the left or to the right (within $[\theta_{i-1},\theta_{i+1}]$ and thus maintaining the order of the thresholds) such that we decrease the value of  \eqref{threshold_objective} as much as possible. Since the value of \eqref{threshold_objective} does not depend on the exact location of $\theta_i$ within $[s(x_j), s(x_{j+1}))$ , it is enough to consider moving $\theta_i$ to a position in $[\theta_{i-1},\theta_i)\cap\{s(x_j):j\in[n]\}$ or $(\theta_i,\theta_{i+1}]\cap\{s(x_j):j\in[n]\}$.

\vspace{2mm}
\underline{When aiming for pairwise DP:}

\vspace{1mm}
For thresholds~$\boldsymbol\theta=(\theta_1,\ldots,\theta_{k-1})$, set $\theta_0=-\infty$ and $\theta_k=+\infty$, and let for $i\in [k]$ and  $\tilde{a},\hat{a}\in \mathcal{A}$ with $\tilde{a}\neq \hat{a}$  
\begin{align*}
T(i,\tilde{a};\boldsymbol\theta)&=|\{j\in[n]: a_j=\tilde{a}, s(x_j)\in(\theta_{i-1},\theta_i]\}|,\\
P(\tilde{a},\hat{a};\boldsymbol\theta)&=|\{(j,l)\in [n]^2: a_j=\tilde{a}, a_l=\hat{a},f(x_j;s,\boldsymbol\theta)>f(x_l;s,\boldsymbol\theta)\}|,\\
Q(\tilde{a},\hat{a};\boldsymbol\theta)&=|\{(j,l)\in [n]^2: a_j=\tilde{a}, a_l=\hat{a},f(x_j;s,\boldsymbol\theta)<f(x_l;s,\boldsymbol\theta)\}|,
\end{align*}
where $f(\cdot\,;s,\boldsymbol\theta)$ denotes the threshold model-predictor with scoring function~$s$ and thresholds~$\boldsymbol\theta$. 
The value of the objective function~\eqref{threshold_objective} using thresholds~$\boldsymbol\theta$ is
\begin{align*}
   \Obj(\boldsymbol\theta)&=
   \Cost(\boldsymbol\theta)+
\lambda\cdot \DPviol(f(\cdot\,;s,\boldsymbol\theta);\mathcal{D})\\
&=\frac{1}{n} \sum_{i=1}^n C_{y_i,f(x_i;s,\boldsymbol\theta)}+
\lambda\cdot \max_{\tilde{a}\neq\hat{a}}\frac{|P(\tilde{a},\hat{a};\boldsymbol\theta)-Q(\tilde{a},\hat{a};\boldsymbol\theta)|}{|\{j\in[n]:a_j=\tilde{a}\}|\cdot |\{j\in[n]:a_j=\hat{a}\}|}.
\end{align*}
Given some initial thresholds~$\boldsymbol\theta=(\theta_1,\ldots,\theta_{k-1})$, we can compute 
$T(i,\tilde{a};\boldsymbol\theta)$, 
$P(\tilde{a},\hat{a};\boldsymbol\theta)$, 
$Q(\tilde{a},\hat{a};\boldsymbol\theta)$ and $\Obj(\boldsymbol\theta)$ in time~$\mathcal{O}(n+k|\mathcal{A}|^2)$ (we assume $k\leq n$ and $|\mathcal{A}|\leq n$). We can then compute the best possible right-move in time $\mathcal{O}(n|\mathcal{A}|^2)$ by going through the sorted array of scores $(s(x_j))_{j=1}^n$ once and using the following: if $\theta_{i_0}$ (for some $i_0\in[k-1])$ is located at $s(x_j)$ and we move it to $s(x_{j+1})$, thus yielding new thresholds~$\boldsymbol\theta'$, we have %
\begin{align*}
T(i_0,a_{j+1};\boldsymbol\theta')&=T(i_0,a_{j+1};\boldsymbol\theta)+1,\\
T(i_0+1,a_{j+1};\boldsymbol\theta')&=T(i_0+1,a_{j+1};\boldsymbol\theta)-1,\\
T(i,\tilde{a};\boldsymbol\theta')&=T(i,\tilde{a};\boldsymbol\theta)~~~\text{for}~~~(i,\tilde{a})\notin\{(i_0,a_{j+1}),(i_0+1,a_{j+1})\},\\[6pt]
P(\tilde{a},\hat{a};\boldsymbol\theta')&=\begin{cases}
P(\tilde{a},\hat{a};\boldsymbol\theta)-T(i_0,\hat{a};\boldsymbol\theta)&\text{if}~a_{j+1}=\tilde{a}\\
P(\tilde{a},\hat{a};\boldsymbol\theta)+T(i_0+1,\tilde{a};\boldsymbol\theta)&\text{if}~a_{j+1}=\hat{a}\\
P(\tilde{a},\hat{a};\boldsymbol\theta)&\text{else}
\end{cases},\\[6pt]
Q(\tilde{a},\hat{a};\boldsymbol\theta')&=\begin{cases}
Q(\tilde{a},\hat{a};\boldsymbol\theta)+T(i_0+1,\hat{a};\boldsymbol\theta)&\text{if}~a_{j+1}=\tilde{a}\\
Q(\tilde{a},\hat{a};\boldsymbol\theta)-T(i_0,\tilde{a};\boldsymbol\theta)&\text{if}~a_{j+1}=\hat{a}\\
Q(\tilde{a},\hat{a};\boldsymbol\theta)&\text{else}
\end{cases},\\[6pt]
\Obj(\boldsymbol\theta')&=\Cost(\boldsymbol\theta)-\frac{1}{n}\cdot C_{y_{j+1},i+1}+ \frac{1}{n}\cdot C_{y_{j+1},i}+
\lambda\cdot \max_{\tilde{a}\neq\hat{a}}\frac{|P(\tilde{a},\hat{a};\boldsymbol\theta')-Q(\tilde{a},\hat{a};\boldsymbol\theta')|}{|\{j\in[n]:a_j=\tilde{a}\}|\cdot |\{j\in[n]:a_j=\hat{a}\}|}.
\end{align*}

Similarly, we can compute the best possible left-move in time $\mathcal{O}(n|\mathcal{A}|^2)$. Some care has to be taken when there are datapoints $x_j,x_{j+1}$ with $s(x_j)=s(x_{j+1})$ since we cannot move a threshold to $s(x_j)$  then.
Clearly, we have $\mathcal{O}(n+k|\mathcal{A}|^2)\subseteq \mathcal{O}(n|\mathcal{A}|^2)$.

\vspace{5mm}
\underline{When aiming for pairwise EO:}

We can proceed similarly as in case of pairwise DP. For  thresholds~$\boldsymbol\theta=(\theta_1,\ldots,\theta_{k-1})$, set $\theta_0=-\infty$ and $\theta_k=+\infty$, and let for $i,i'\in[k]$ and $\tilde{a},\hat{a}\in \mathcal{A}$ with $\tilde{a}\neq \hat{a}$ 
\begin{align*}
T(i,i',\tilde{a};\boldsymbol\theta)&=|\{j\in[n]: a_j=\tilde{a}, y_j=i',  s(x_j)\in(\theta_{i-1},\theta_i]\}|,\\
P(\tilde{a},\hat{a};\boldsymbol\theta)&=|\{(j,l)\in [n]^2: a_j=\tilde{a}, a_l=\hat{a},y_j>y_l,f(x_j;s,\boldsymbol\theta)>f(x_l;s,\boldsymbol\theta)\}|,\\
Q(\tilde{a},\hat{a};\boldsymbol\theta)&=|\{(j,l)\in [n]^2: a_j=\tilde{a}, a_l=\hat{a},y_j<y_l,f(x_j;s,\boldsymbol\theta)<f(x_l;s,\boldsymbol\theta)\}|.
\end{align*}
It is
\begin{align*}
   \Obj(\boldsymbol\theta)&=
   \Cost(\boldsymbol\theta)+
\lambda\cdot \EOviol(f(\cdot\,;s,\boldsymbol\theta);\mathcal{D})\\
&=\frac{1}{n} \sum_{i=1}^n C_{y_i,f(x_i;s,\boldsymbol\theta)}+\\
&~~~~~~~~~~~~~~
\lambda\cdot \max_{\tilde{a}\neq\hat{a}}\left|\frac{P(\tilde{a},\hat{a};\boldsymbol\theta)}{|\{(j,l)\in [n]^2: a_j=\tilde{a}, a_l=\hat{a},y_j>y_l\}|}-\frac{Q(\tilde{a},\hat{a};\boldsymbol\theta)}{|\{(j,l)\in [n]^2: a_j=\tilde{a}, a_l=\hat{a},y_j<y_l\}|}\right|.
\end{align*}
Given some initial thresholds~$\boldsymbol\theta=(\theta_1,\ldots,\theta_{k-1})$, we can compute $T(i,i',\tilde{a};\boldsymbol\theta)$, 
$P(\tilde{a},\hat{a};\boldsymbol\theta)$, 
$Q(\tilde{a},\hat{a};\boldsymbol\theta)$ and $\Obj(\boldsymbol\theta)$ in time~$\mathcal{O}(n+k^2|\mathcal{A}|^2)$. 
We can then compute the best possible right-move in time $\mathcal{O}(nk|\mathcal{A}|+n|\mathcal{A}|^2)$ by going through the sorted array of scores $(s(x_j))_{j=1}^n$ once and using the following: 
if $\theta_{i_0}$ (for some $i_0\in[k-1])$ is located at $s(x_j)$ and we move it to $s(x_{j+1})$, thus yielding new thresholds~$\boldsymbol\theta'$, we have
\begin{align*}
T(i_0,y_{j+1},a_{j+1};\boldsymbol\theta')&=T(i_0,y_{j+1},a_{j+1};\boldsymbol\theta)+1,\\
T(i_0+1,y_{j+1},a_{j+1};\boldsymbol\theta')&=T(i_0+1,y_{j+1},a_{j+1};\boldsymbol\theta)-1,\\
T(i,i',\tilde{a};\boldsymbol\theta')&=T(i,i',\tilde{a};\boldsymbol\theta)~~~\text{for}~~~(i,i',\tilde{a})\notin\{(i_0,y_{j+1},a_{j+1}),(i_0+1,y_{j+1},a_{j+1})\}
\end{align*}
and
\begin{align*}
P(\tilde{a},\hat{a};\boldsymbol\theta')&=\begin{cases}
P(\tilde{a},\hat{a};\boldsymbol\theta)-\sum_{i'=1}^{y_{j+1}-1}T(i_0,i',\hat{a};\boldsymbol\theta)&\text{if}~a_{j+1}=\tilde{a}\\
P(\tilde{a},\hat{a};\boldsymbol\theta)+\sum_{i'=y_{j+1}+1}^k T(i_0+1,i',\tilde{a};\boldsymbol\theta)&\text{if}~a_{j+1}=\hat{a}\\
P(\tilde{a},\hat{a};\boldsymbol\theta)&\text{else}
\end{cases},\\[7pt]
Q(\tilde{a},\hat{a};\boldsymbol\theta')&=\begin{cases}
Q(\tilde{a},\hat{a};\boldsymbol\theta)+\sum_{i'=y_{j+1}+1}^k T(i_0+1,i',\hat{a};\boldsymbol\theta)&\text{if}~a_{j+1}=\tilde{a}\\
Q(\tilde{a},\hat{a};\boldsymbol\theta)-\sum_{i'=1}^{y_{j+1}-1}T(i_0,i',\tilde{a};\boldsymbol\theta)&\text{if}~a_{j+1}=\hat{a}\\
Q(\tilde{a},\hat{a};\boldsymbol\theta)&\text{else}
\end{cases},\\[7pt]
\Obj(\boldsymbol\theta')&=\Cost(\boldsymbol\theta)-\frac{1}{n}\cdot C_{y_{j+1},i+1}+\frac{1}{n}\cdot C_{y_{j+1},i}+\\
&~~~~~~~~~~~~
\lambda\cdot \max_{\tilde{a}\neq\hat{a}}\left|\frac{P(\tilde{a},\hat{a};\boldsymbol\theta')}{|\{(j,l)\in [n]^2: a_j=\tilde{a}, a_l=\hat{a},y_j>y_l\}|}-\frac{Q(\tilde{a},\hat{a};\boldsymbol\theta')}{|\{(j,l)\in [n]^2: a_j=\tilde{a}, a_l=\hat{a},y_j<y_l\}|}\right|.
\end{align*}

Similarly, we can compute the best possible left-move in time 
$\mathcal{O}(nk|\mathcal{A}|+n|\mathcal{A}|^2)$. Clearly, we have $\mathcal{O}(n+k^2|\mathcal{A}|^2),\mathcal{O}(nk|\mathcal{A}|+n|\mathcal{A}|^2)\subseteq \mathcal{O}(nk|\mathcal{A}|^2)$.

\vspace{5mm}
\underline{Initializing the local search:}

To initialize the local search, we can 
 choose $\theta_1,\ldots,\theta_{k-1}$ at random or, 
 for small values of $\lambda$, 
 we 
 can 
 choose them as a solution to \eqref{threshold_objective} with $\lambda=0$, which we 
 can 
 compute in time~$\mathcal{O}(n^2k^3)$ (see~Appendix~\ref{appendix_choose_thresh_lambda_eq_0}).

\subsection{Solving \eqref{threshold_objective} When $\lambda=0$}\label{appendix_choose_thresh_lambda_eq_0}

We write $s_i=s(x_i)$, $i\in[n]$, for the values of the scoring function~$s$ on the training data~$\mathcal{D}=((x_i,y_i,a_i))_{i=1}^n$ and assume the values to be sorted, that is $s_i\leq s_{i+1}$, $i\in[n-1]$, and given, that is we do not take the time for evaluating $s$ into account. 

\vspace{2mm}
 We build a table $T\in (\R_{\geq 0}\cup\{\infty\})^{(n+1)\times k\times k}$ with
\begin{align*}
T(i,f,l)=\min_{\hat{y}\in \mathcal{H}_{i,f,l}}\sum_{j=1}^i C_{y_j,\hat{y}_j}
\end{align*}
for $i\in\{0\}\cup[n]$ and $f,l\in[k]$, where
\begin{align*}
\mathcal{H}_{i,f,l}&=\{\hat{y}=(\hat{y}_1,\ldots,\hat{y}_i)\in[k]^i: \hat{y}~\text{are predictions for %
$x_1,\ldots,x_i$
that are sorted, that is $y_r\leq y_{r+1}$ for $r\in [i-1]$,}\\
&~~~~~~~~~~~~\text{with $\hat{y}_i=l$, take at most $f$ different values and satisfy $\hat{y}_r=\hat{y}_{r'}$ for $s_r=s_{r'}$}\}.
\end{align*}
The optimal value of \eqref{threshold_objective} with $\lambda=0$ is 
given by $(1/n)\cdot\min_{l\in[k]}T(n,k,l)$.

\vspace{4mm}
It is
\begin{align*}
T(0,f,l)&=0,\quad f,l\in[k],\\
T(1,f,l)&=C_{y_1,l} \quad f,l\in[k],\\
T(i,1,l)&=\sum_{j=1}^i C_{y_j,l} \quad i\in[n],l\in[k],\\
T(i,f,1)&=\sum_{j=1}^i C_{y_j,1} \quad i\in[n],f\in[k],
\end{align*}
and 
\begin{align*}
T(i,f,l)&=\min_{i'<i,l'<l}\left[T(i',f-1,l')+\infty\cdot\charfct\{i'>0 \wedge (s_{i'}=s_{i'+1})\}+\sum_{r=i'+1}^i C_{y_r,l}\right].
\end{align*}
We can build $T$ in time $\mathcal{O}(n^2k^3)$. If we store the minimizing $(i',l')$ with $T(i,f,l)$, we can easily construct an optimal solution from $T$.

\subsection{Simulations Illustrating the Observation Made in Section~\ref{subsection_both_steps_necessary} }\label{appendix_simulations_fairness_Score_vs_threshold}

\begin{figure}[ht]
    \centering
    \includegraphics[height=5cm]{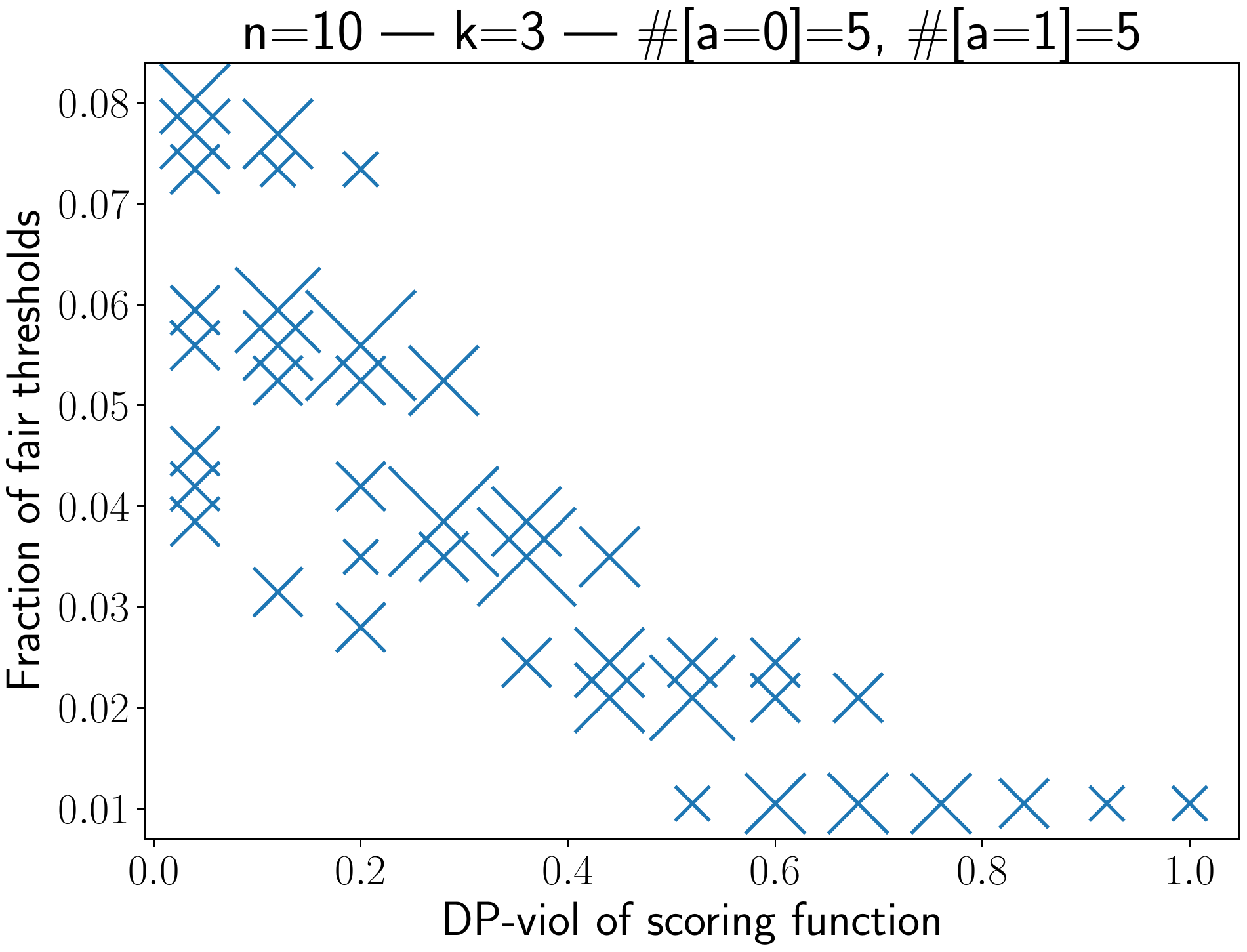}
    \hspace{12mm}
    \includegraphics[height=5cm]{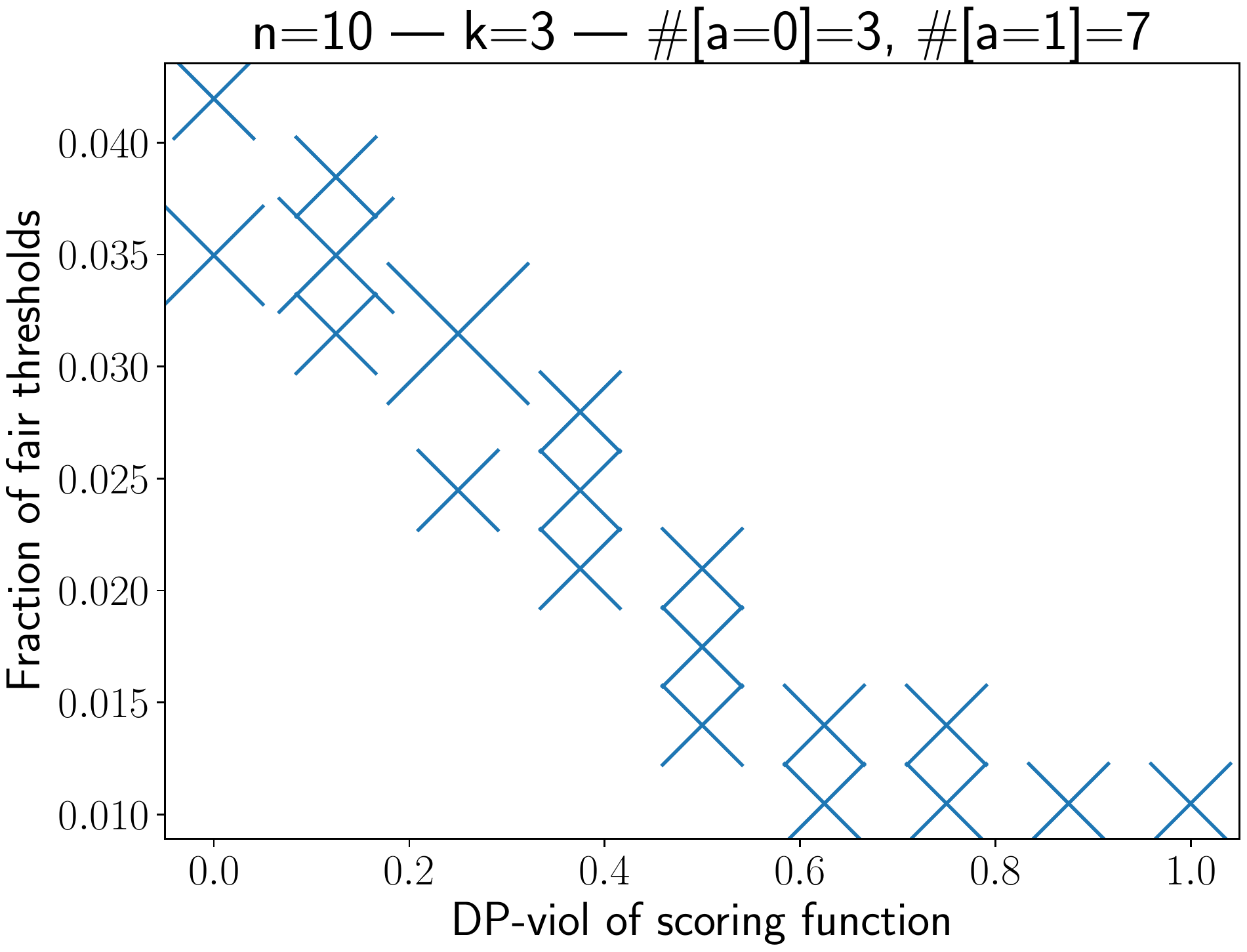}
    
    \vspace{6mm}
    \includegraphics[height=5cm]{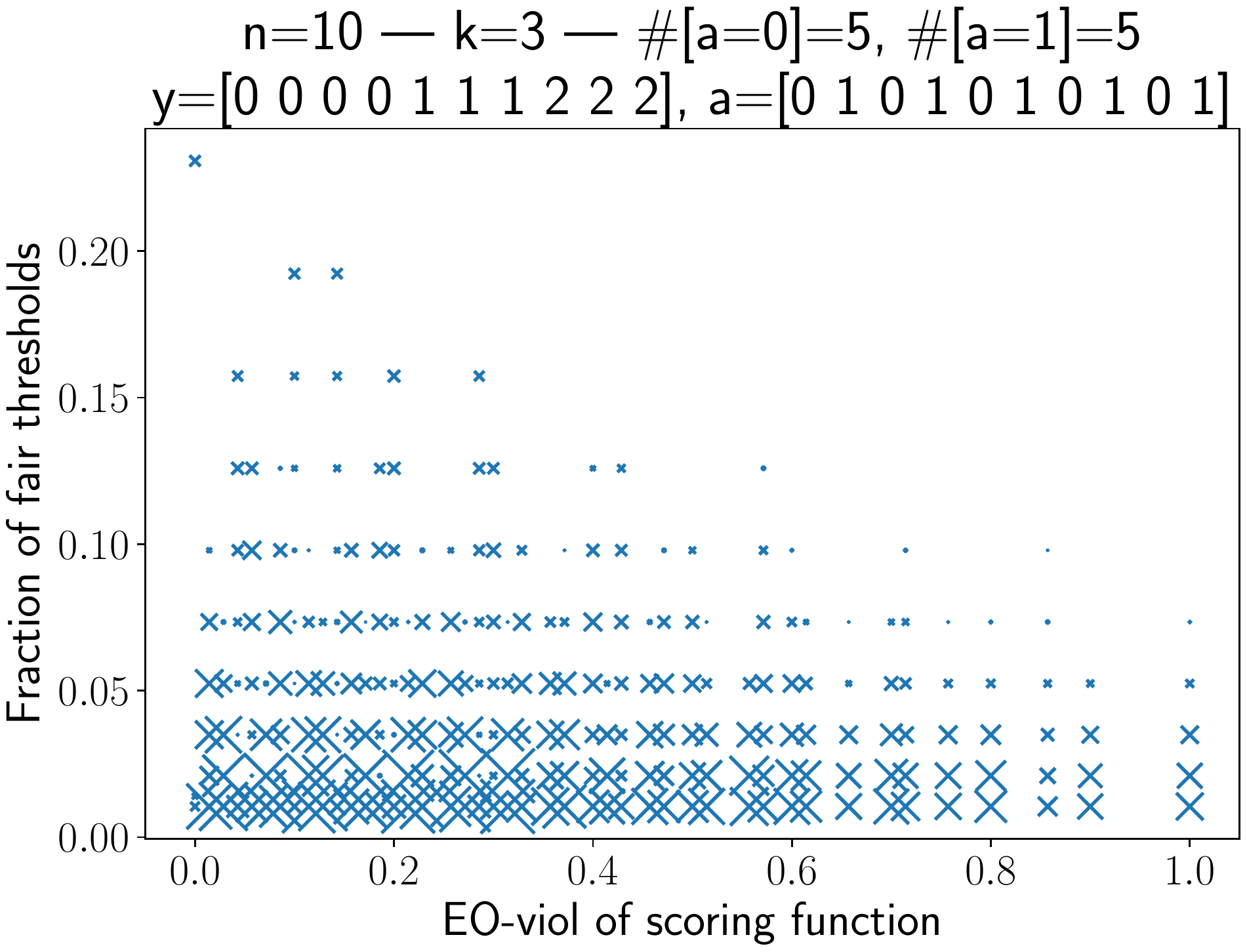}
    \hspace{12mm}
    \includegraphics[height=5cm]{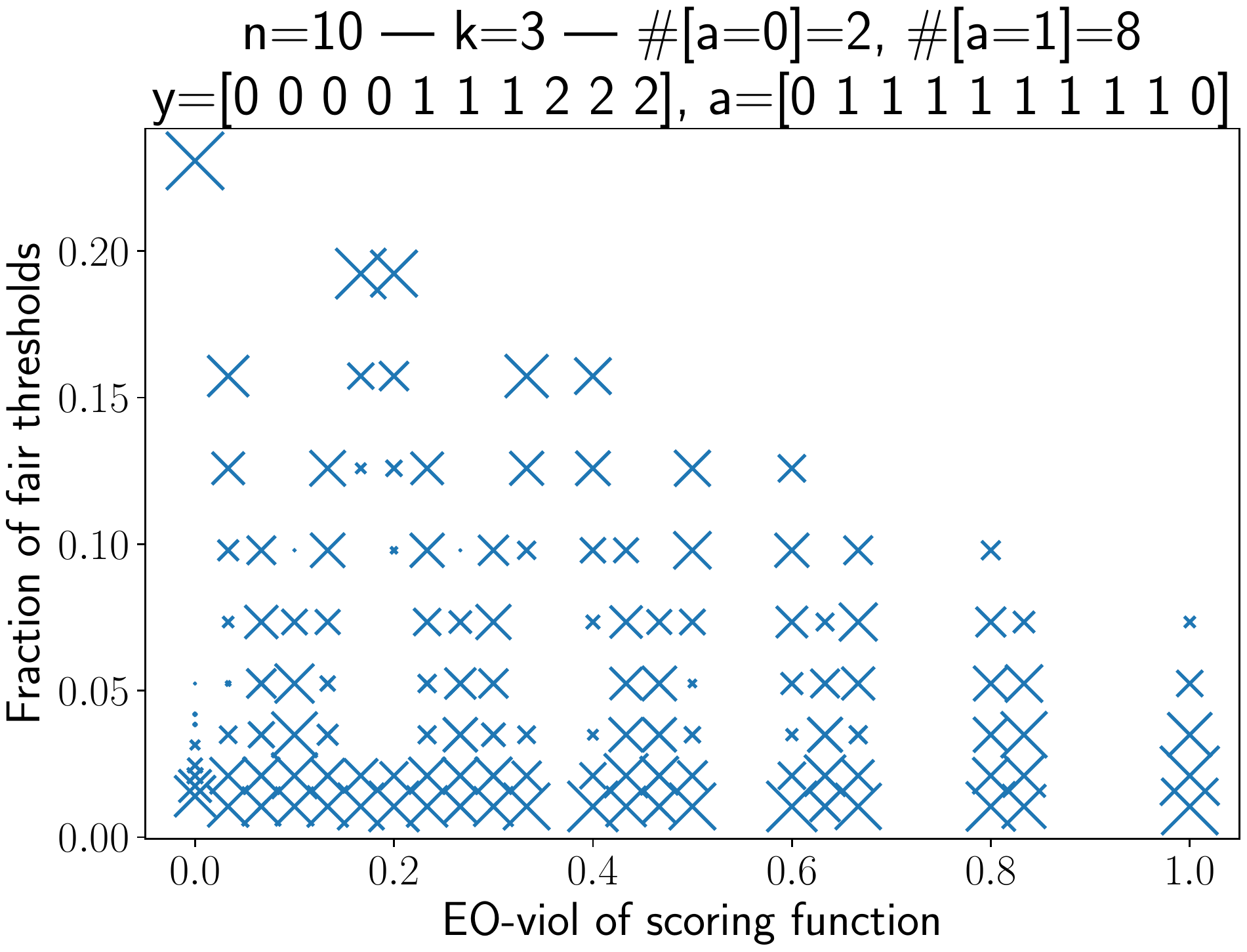}

    \caption{Simulations illustrating the observation made in Section~\ref{subsection_both_steps_necessary}: the more fair the scoring function, the more choices of thresholds 
    yield a fair predictor.
    The size of a marker is proportional to how many times we observed a particular outcome.}\label{fig:appendix_simulations}
\end{figure}

Figure~\ref{fig:appendix_simulations} shows some simulations 
illustrating 
the observation that we made in  Section~\ref{subsection_both_steps_necessary}: if the scoring function is fair to some extent, then there exist more choices of thresholds that yield a fair predictor. For $n=10$ 
many 
datapoints and $k=3$ we considered all possible injective scoring functions 
by considering all permutations of $1,\ldots,10$. We then computed the fairness violation~$\Fairviol$ of the scoring function and considering all possible choices of thresholds that yield a different predictor, we computed the fraction of choices of thresholds for which the resulting predictor is fair. The plots 
of Figure~\ref{fig:appendix_simulations}
show that fraction on the $y$-axis and   the fairness violation~$\Fairviol$ of the scoring function on the $x$-axis. In the plots in the top row we 
studied 
pairwise DP and there $\Fairviol=\DPviol$;  in the plots in the bottom row we 
studied 
pairwise EO and there $\Fairviol=\EOviol$. The size of a marker in these plots is proportional to how many times we observed a particular outcome (since there can be several different scoring functions with the same value of $\Fairviol$ and the same %
fraction of fair thresholds). 
We can see that there is a 
strong 
correlation between the fairness of the scoring function and the number of choices of thresholds that yield a fair predictor. Note that when considering pairwise DP (top row), the results do not depend on which ground-truth labels we assign to the datapoints, but only on the number of datapoints with $a=0$ and $a=1$, respectively.  When considering pairwise EO (bottom row), the results depend on the ground-truth labels and the protected attribute~$a$ of each datapoint. Each row shows two plots in different settings---the relevant information can be read from the titles of the plots.

\section{SOME DETAILS AND FURTHER EXPERIMENTS}\label{appendix_partB}

\subsection{Details About Implementation}\label{appendix_detail_hyperparameters}

We chose to build on the reduction  approach of \citet{agarwal_reductions_approach} for solving the fair binary classification problem described in Section~\ref{subsection_learning_scoring_function} since 
that 
approach 
is theoretically well-established. 
Furthermore, 
its implementation  is available as part of the official Python package \textsc{Fairlearn}\footnote{\url{https://fairlearn.github.io/} (MIT License)}.

We used the \textsc{GridSearch}-method, which corresponds to Section~3.4 of \citeauthor{agarwal_reductions_approach} and returns a deterministic rather than a randomized classifier as the \textsc{ExponentiatedGradient}-method does. As  \citeauthor{agarwal_reductions_approach} discuss, the \textsc{GridSearch}-method is only feasible when $|\mathcal{A}|\leq 3$. In fact, in the current version~v0.6.1 of \textsc{Fairlearn} the \textsc{GridSearch}-method allows 
for 
$|\mathcal{A}|= 3$ only when aiming for demographic parity; it requires $|\mathcal{A}|=2$ when aiming for equal opportunity. If we wanted to extend our implementation to arbitrary $|\mathcal{A}|\geq 2$ for both DP and EO, we would have to use the \textsc{ExponentiatedGradient}-method instead of the \textsc{GridSearch}-method. Since we aim to learn a deterministic classifier, we 
then would 
have to use the predictor with the highest weight in the mixture returned by  the \textsc{ExponentiatedGradient}-method.
Alternatively, we could try to use the other methods that are available for fair binary classification such as the various relaxation based approaches \citep[e.g.,][]{donini2018,wu2019,zafar2019}. However, the  relaxation based approaches  have recently been criticized for several reasons \citep{lohaus2020}.

In the \textsc{GridSearch}-method we set the parameters \texttt{grid\_size} and \texttt{grid\_limit} to 100 and 3, respectively. 
As base classification method we used logistic regression in the  implementation available in Scikit-learn. 
Its main parameter is the 
parameter~\texttt{C},  
which is the inverse of a regularization parameter. We set it to $1/(2\cdot \text{size of training data}\cdot \gamma)$ for some $\gamma\in\{10^{-i}:i\in[5]\}$ that we chose by means of 10-fold cross validation on the classification problem described in Section~\ref{subsection_learning_scoring_function} without any fairness constraint, aiming for small 01-loss.
We set all other parameters to their default values, except for \texttt{max\_iter} and \texttt{fit\_intercept}, which we set to 2500 and \texttt{False}, respectively.

Finally, 
in case the dataset~$\mathcal{D}'$ 
comprised more than $6\cdot10^5$ datapoints, 
we subsampled 
$6\cdot10^5$ 
datapoints 
(cf. Section~\ref{subsection_limitations}, third item).

For fitting a POM model, 
we used the implementation 
that comes with 
\textsc{Matlab}.

\subsection{Details About Datasets}\label{appendix_detail_datasets}

In Section~\ref{subsection_experiment_motivation} we used the Drug Consumption dataset \citep{drug_consumption_data} and the Communities and Crime dataset \citep{comm_and_crime_data,department_of_commerce1,department_of_commerce2,department_of_justice1,department_of_justice2}, which are both publicly available in the UCI repository \citep{uci_repository}.
 
The benchmark datasets used in Section~\ref{subsection_experiment_comparison} together with their splits into training and test sets  are available on the website accompanying the survey paper of \citet{gutierrez2016ordinal}. 

They extracted the real ordinal regression datasets from 
 the UCI repository \citep{uci_repository} and 
 OpenML \citep{openml}, and they generated one synthetic dataset as proposed by \citet{PINTODACOSTA2008}. 
We provide links to the data sources and references as requested by the data creator / donor if applicable:
\begin{itemize}
    \item automobile: 
    \url{https://archive.ics.uci.edu/ml/datasets/automobile}
    \item balance-scale:
    \url{https://archive.ics.uci.edu/ml/datasets/balance+scale}
    \item car: 
    \url{https://archive.ics.uci.edu/ml/datasets/car+evaluation}
    \item ERA: 
    \url{https://www.openml.org/d/1030}
    \item ESL: 
    \url{https://www.openml.org/d/1027}
    \item eucalyptus: 
    \url{https://www.openml.org/d/188}
    \item LEV: 
    \url{https://www.openml.org/d/1029}
    \item newthyroid: 
    \url{https://archive.ics.uci.edu/ml/datasets/thyroid+disease}
    \item SWD: 
    \url{https://www.openml.org/d/1028}
    \item toy: synthetic dataset generated as proposed by \citet{PINTODACOSTA2008}
    \item winequality-red (\citealp{winequality_data_set}):
    \url{https://archive.ics.uci.edu/ml/datasets/wine+quality}
\end{itemize}

\citet{gutierrez2016ordinal} downloaded the discretized regression datasets from the website accompanying the paper of \citet{chu2005}. \citeauthor{chu2005}, in turn, obtained them from the website of Luis Torgo\footnote{\url{https://www.dcc.fc.up.pt/~ltorgo/Regression/DataSets.html}}, who obtained them from the UCI repository \citep{uci_repository}, the Delve project\footnote{\url{http://www.cs.toronto.edu/~delve/}}, the  StatLib datasets archive\footnote{\url{http://lib.stat.cmu.edu/datasets/}},  %
and 
the Bilkent University Function Approximation Repository \citep{guvenir2000}.
We provide links to the data sources and references as requested by the data creator / donor if applicable:
\begin{itemize}
    \item abalone: 
    \url{http://archive.ics.uci.edu/ml/datasets/Abalone}
    \item bank: 
    \url{http://www.cs.toronto.edu/~delve/data/bank/desc.html}
    \item calhousing: 
    \url{http://lib.stat.cmu.edu/datasets/} as \texttt{houses.zip}; submitted by Kelley Pace
    \item census: 
    \url{http://www.cs.toronto.edu/~delve/data/census-house/desc.html}
    \item computer: 
    \url{http://www.cs.toronto.edu/~delve/data/comp-activ/desc.html}
    \item housing:
    \url{http://lib.stat.cmu.edu/datasets/boston}
    \item machine: 
    \url{http://archive.ics.uci.edu/ml/datasets/Computer+Hardware}
    \item stock: 
    \url{http://pcaltay.cs.bilkent.edu.tr/DataSets/}
\end{itemize}

 We normalized every dataset such that each feature has zero mean and unit variance on the training set. 
 Table~\ref{table_drug_consumption_data} and Table~\ref{table_comm_and_crime_data} provide  
 some additional statistics of the Drug Consumption dataset and the Communities and Crime dataset, respectively,  which we used in Section~\ref{subsection_experiment_motivation}.
 Table~\ref{table_statistics_data_sets2} and Table~\ref{table_statistics_data_sets1} provide  
 some statistics of the real ordinal 
 datasets 
 and discretized regression datasets
 used in Section~\ref{subsection_experiment_comparison}. Further statistics, such as the distribution per class, that is $\Psymb[y=i]$, $i\in[k]$, can be found in Section~4 of \citet{gutierrez2016ordinal}.

 \begin{table}[ht]
  \caption{Statistics of the Drug Consumption dataset  used in Section~\ref{subsection_experiment_motivation}.}\label{table_drug_consumption_data}
  \centering
\renewcommand{\arraystretch}{1.5}
\begin{tabular}{c|ccccc}
\toprule
$\Psymb[a=\text{f}]$ & $\Psymb[y=1]$ & $\Psymb[y=2]$ & $\Psymb[y=3]$ & $\Psymb[y=4]$ & $\Psymb[y=5]$ \\
\midrule
0.50 & 0.33 & 0.14 & 0.11 & 0.17 & 0.24\\
\bottomrule
\end{tabular}
\end{table}

~ 
 
\begin{table}[h]
  \caption{Statistics of the Communities and Crime dataset  used in Section~\ref{subsection_experiment_motivation}.}\label{table_comm_and_crime_data}
  \centering
\renewcommand{\arraystretch}{1.5}
\begin{tabular}{c|cccccccc}
\toprule
$\Psymb[a=\text{white}]$ & $\Psymb[y=1]$ & $\Psymb[y=2]$ & $\Psymb[y=3]$ & $\Psymb[y=4]$ & $\Psymb[y=5]$ & $\Psymb[y=6]$ & $\Psymb[y=7]$ & $\Psymb[y=8]$ \\
\midrule
0.57 & 0.20 & 0.18 & 0.22 & 0.13 & 0.08 & 0.05 & 0.05 & 0.09\\
\bottomrule
\end{tabular}
\end{table}

   \begin{sidewaystable}
  \caption{Statistics of the real ordinal regression  datasets used in Section~\ref{subsection_experiment_comparison}\label{table_statistics_data_sets2}.}
  \centering
\renewcommand{\arraystretch}{1.5}
    \begin{small}
\begin{tabular}{ccccccccc}
\toprule
& \multirow{2}{*}{$\#$ train} & \multirow{2}{*}{$\#$ test} & \multirow{2}{*}{$\#$ features} &  \multirow{2}{*}{$\#$ classes} & \multirow{2}{*}{$\Psymb[a=0]$} & \multirow{2}{*}{$\Psymb[y_1>y_2 | a_1=0,a_2=1]$} & \multirow{2}{*}{$\Psymb[y_1< y_2 | a_1=0,a_2=1]$} & $|\Psymb[y_1> y_2 | a_1=0,a_2=1]-$\\
& & & & & & & &  $~~~~~~~\Psymb[y_1< y_2 | a_1=0,a_2=1]|$ \\
\midrule
automobile & 153 & 52 & 70 & 6 & 0.53 & 0.62 & 0.20 & 0.42\\ 
balance-scale & 468 & 157 & 3 & 3 & 0.41 & 0.51 & 0.13 & 0.38\\
car & 1296 & 432 & 20 & 4 & 0.67 & 0.45 & 0.0 & 0.45\\
ERA & 750 & 250 & 3 & 9 & 0.46 & 0.25 & 0.63 &  0.38\\
ESL & 366 & 122 & 3 & 9 & 0.40 & 0.06 & 0.81 & 0.75\\
eucalyptus & 552 & 184 & 90 & 5 & 0.59 & 0.38 & 0.41 & 0.06\\
LEV & 750 & 250 & 3 & 5 & 0.44 & 0.17 & 0.57 & 0.41\\
newthyroid & 161 & 54 & 4 & 3 & 0.49 & 0.42 & 0.07 &  0.35\\
SWD & 750 & 250 & 9 & 4 & 0.55 & 0.23 & 0.46 &  0.23 \\
toy & 225 & 75 & 1 & 5 & 0.49 & 0.29 & 0.49 & 0.20 \\
winequality-red & 1199 & 400 & 10 & 6 & 0.48 & 0.27 & 0.38 & 0.11\\ 
\bottomrule
\end{tabular}
\end{small}
\end{sidewaystable}

\begin{sidewaystable}    
  \caption{Statistics of the discretized regression datasets used in Section~\ref{subsection_experiment_comparison}\label{table_statistics_data_sets1}.}
  \centering
\renewcommand{\arraystretch}{1.5}
\begin{small}
\begin{tabular}{ccccccccc}
\toprule
& \multirow{2}{*}{$\#$ train} & \multirow{2}{*}{$\#$ test} & \multirow{2}{*}{$\#$ features} &  \multirow{2}{*}{$\#$ classes} & \multirow{2}{*}{$\Psymb[a=0]$}& \multirow{2}{*}{$\Psymb[y_1> y_2 | a_1=0,a_2=1]$} & \multirow{2}{*}{$\Psymb[y_1< y_2 | a_1=0,a_2=1]$} & $|\Psymb[y_1> y_2 | a_1=0,a_2=1]-$\\
& & & && & & & $~~~~~~~\Psymb[y_1< y_2 | a_1=0,a_2=1]|$ \\
\midrule
 abalone-5   & 1000 & 3177 & 9 & 5 & 0.63 & 0.28 & 0.53 & 0.25\\
 bank1-5  & 50 & 8142 & 7 & 5 &0.50 & 0.38 & 0.42 & 0.05\\
bank2-5  & 75 & 8117 & 31 & 5 &0.50 & 0.41 & 0.39 & 0.02\\
 calhousing-5  & 150 & 20490 & 7 & 5 &0.50 & 0.35 & 0.46 & 0.11\\
 census1-5  & 175 & 22609 & 7 & 5 &0.50 & 0.14 & 0.72 & 0.58\\
 census2-5  & 200 & 22584 & 15 & 5 & 0.50 & 0.14 & 0.72 & 0.58\\
 computer1-5  & 100 & 8092 & 11 & 5 &0.50 & 0.70 & 0.15 & 0.55\\
 computer2-5 & 125 & 8067 & 20 & 5 &0.50 & 0.70 & 0.15 & 0.55\\
housing-5  & 300 & 206 & 12 & 5 &0.50 & 0.65 & 0.19 & 0.46\\
 machine-5  & 150 & 59 & 5 & 5 &0.49 & 0.75 & 0.12 & 0.63\\
 stock-5 & 600 & 350 & 8 & 5 & 0.50 & 0.18 & 0.69 & 0.51\\
 \midrule
 abalone-10   & 1000 & 3177 & 9 & 10 & 0.63 & 0.32 & 0.58 & 0.26\\
 bank1-10  & 50 & 8142 & 7 & 10 &0.50 & 0.43 & 0.47 & 0.05\\
bank2-10  & 75 & 8117 & 31 & 10 &0.50 & 0.46 & 0.44 & 0.01\\
 calhousing-10  & 150 & 20490 & 7 & 10 &0.50 & 0.40 & 0.51 & 0.11\\
 census1-10  & 175 & 22609 & 7 & 10 &0.50 & 0.17 & 0.76 & 0.59\\
 census2-10  & 200 & 22584 & 15 & 10 & 0.50 & 0.16 & 0.76 & 0.60\\
 computer1-10  & 100 & 8092 & 11 & 10 &0.49 & 0.75 & 0.18 & 0.57\\
 computer2-10 & 125 & 8067 & 20 & 10 &0.50 & 0.75 & 0.18 & 0.57\\
housing-10  & 300 & 206 & 12 & 10 &0.50 & 0.71 & 0.22 & 0.48\\
 machine-10  & 150 & 59 & 5 & 10 &0.48 & 0.80 & 0.14 & 0.67\\
 stock-10 & 600 & 350 & 8 & 10 & 0.50 & 0.20 & 0.75 & 0.54\\
\bottomrule
\end{tabular}
\end{small}
\end{sidewaystable}

\clearpage 
 
\subsection{Addendum to the Experiment of Section~\ref{subsection_experiment_motivation} on the Communities and Crime Dataset
}\label{appendix_comm_and_crime_finer_grid} 
 
\begin{figure}[ht]
    \begin{center}
    \includegraphics[width=0.92\linewidth]{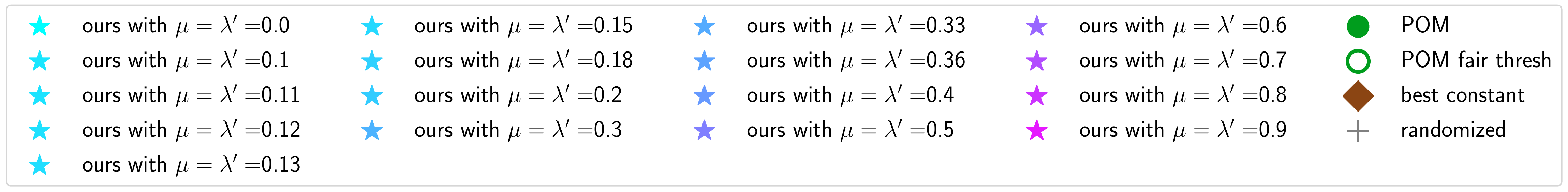}
    
    \vspace{4mm}
    \includegraphics[height=4.2cm]{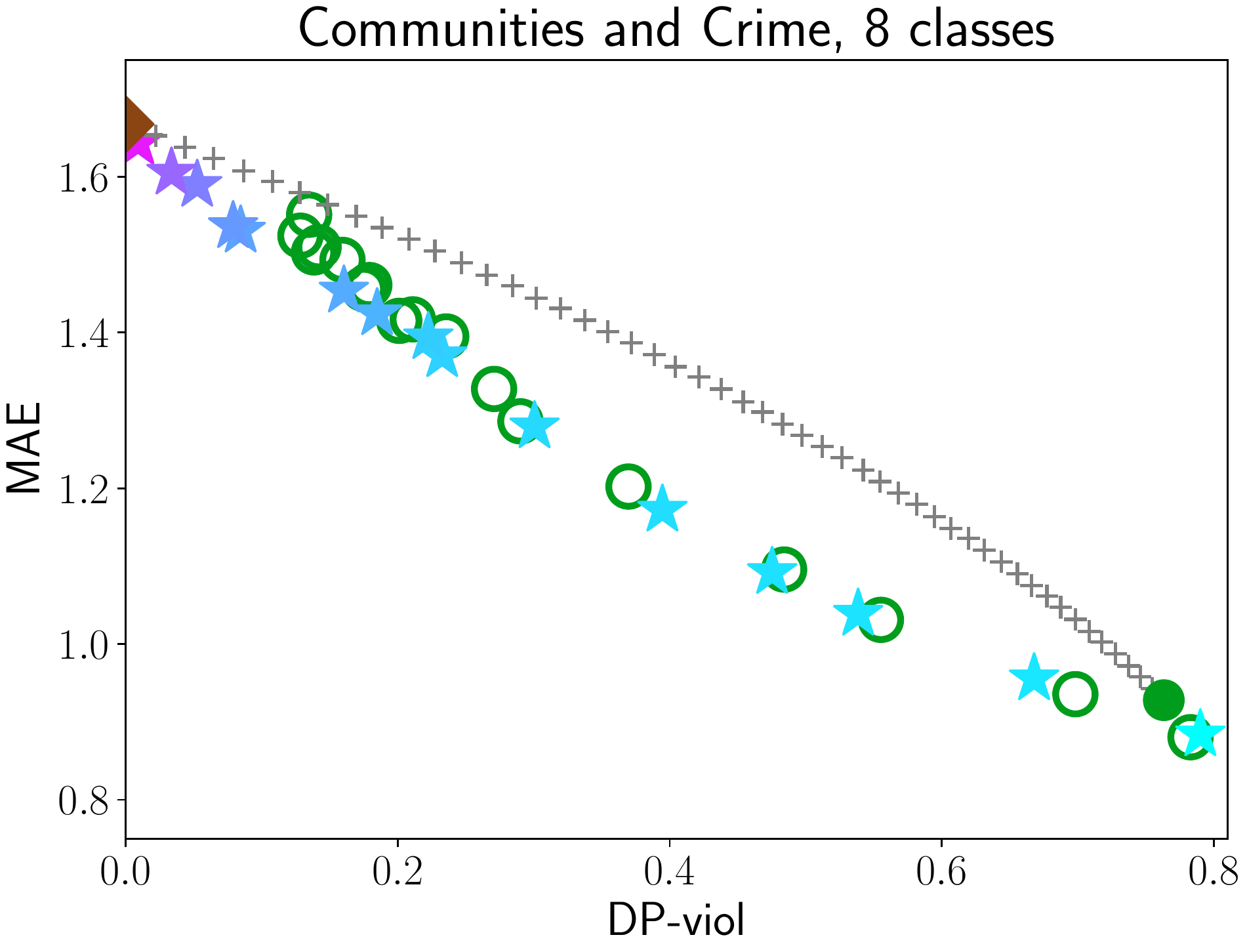}
    \hspace{0.1cm}
    \includegraphics[height=4.2cm]{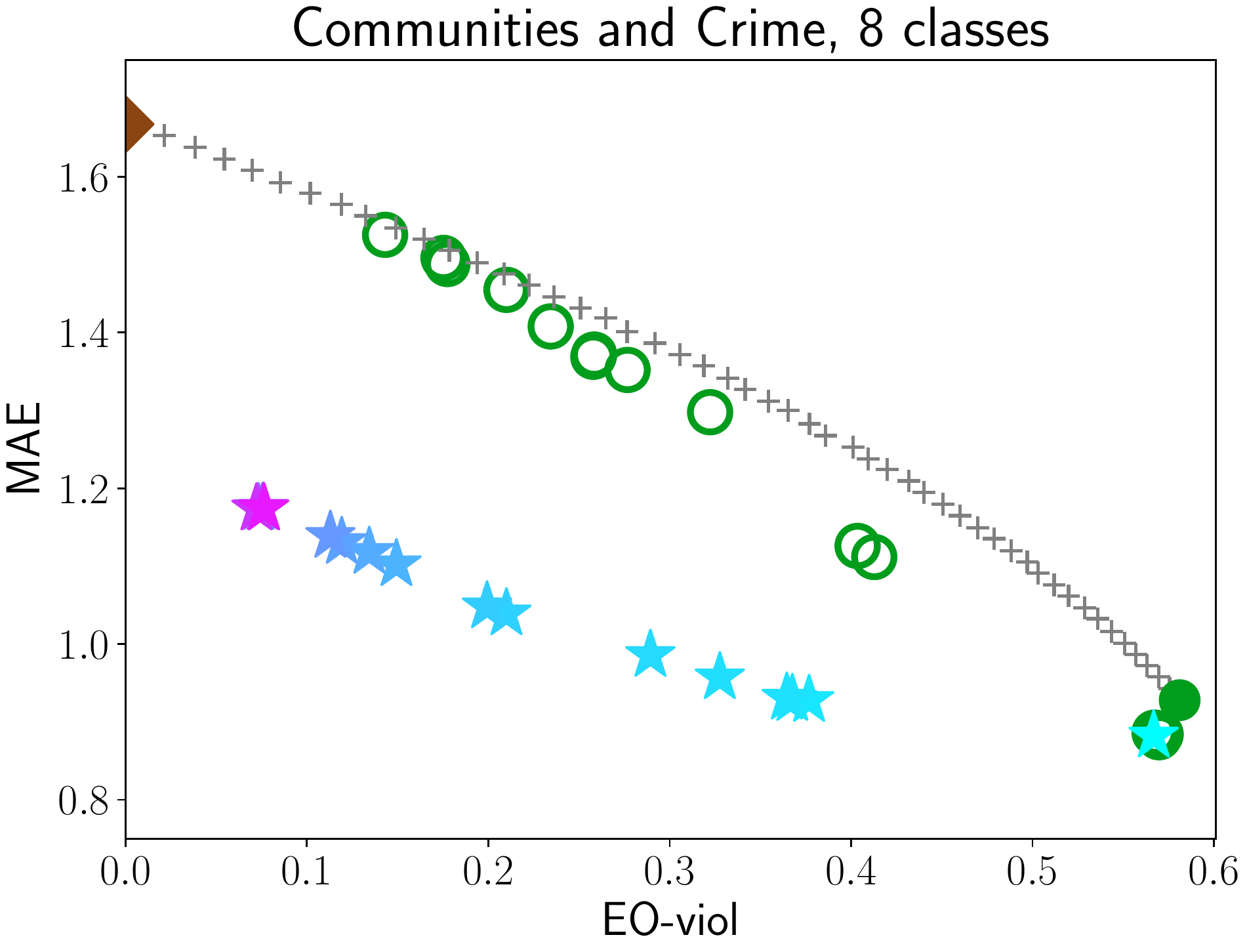}
    \hspace{0.1cm}
    \includegraphics[height=4.2cm]{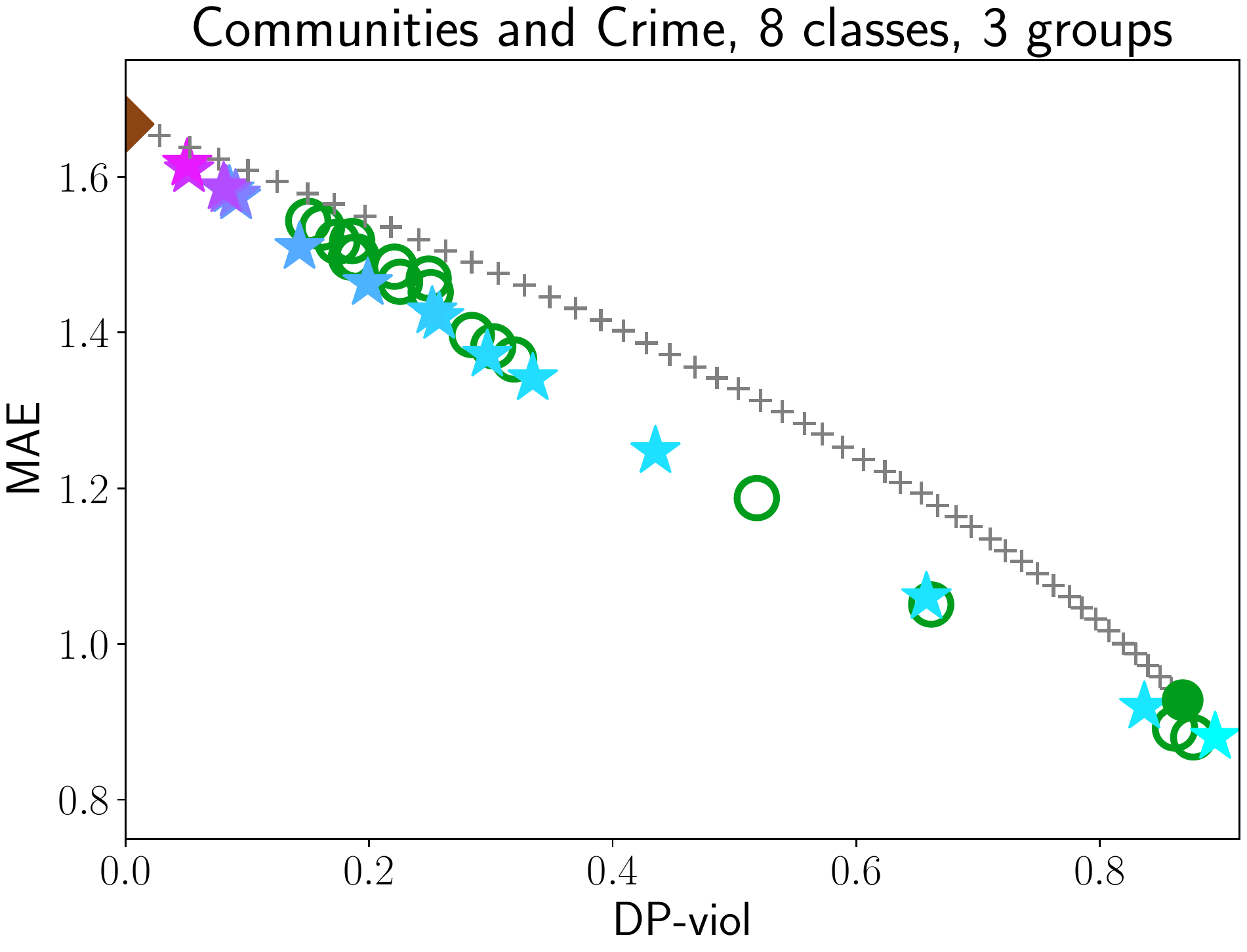}
    
    \end{center}
    \caption{\textbf{Left and center:} 
    Results 
    of the experiment of Section~\ref{subsection_experiment_motivation} on the Communities and Crime dataset when we run our strategy for additional values of $\mu=\lambda'$. \textbf{Right:} $\MAE$ vs $\DPviol$ for the various predictors on a version of the dataset with three protected groups. }\label{exp_comm_and_crime_additional}

\end{figure}

Figure~\ref{exp_comm_and_crime_additional} (left and center) show the results of the experiment on the Communities and Crime dataset presented in Section~\ref{subsection_experiment_motivation}, when we additionally run our strategy  with $\mu=\lambda'\in\{0.11,0.12,0.13,0.15,0.18\}$. This results in nicely exploring the $\MAE$-vs-$\DPviol$ trade-off over the whole range of $\DPviol\in[0,0.8]$. 

The right plot shows the results (when aiming for pairwise DP) for a version of the dataset with three protected groups: rather than $a\in\{\text{white},\text{diverse}\}$, we consider $a\in\{\text{white},\text{African-American},\text{Hispanic or Asian}\}$. We set $a=\text{African-American}$ if at least 25\% of a community's population are African-Americans, $a=\text{Hispanic or Asian}$ if at least 25\% are Hispanics or Asians and less than 25\% are African-Americans, and $a=\text{white}$ otherwise. It is $\Psymb[a=\text{white}]=0.46$, $\Psymb[a=\text{African-American}]=0.24$ and  $\Psymb[a=\text{Hispanic or Asian}]=0.3$. We can see that also in this case the predictors produced by our approach nicely explore the accuracy-vs-fairness trade-off. Note that our implementation does not allow us to aim for pairwise EO in the case of three groups due to the limitations of \textsc{Fairlearn}'s \textsc{GridSearch}-method (cf. Appendix~\ref{appendix_detail_hyperparameters}).

 \subsection{Experiment with Asymmetric Cost Matrix}\label{appendix_asymmetric_cost}

 \begin{figure}[h]
     \centering
    \includegraphics[height=4cm]{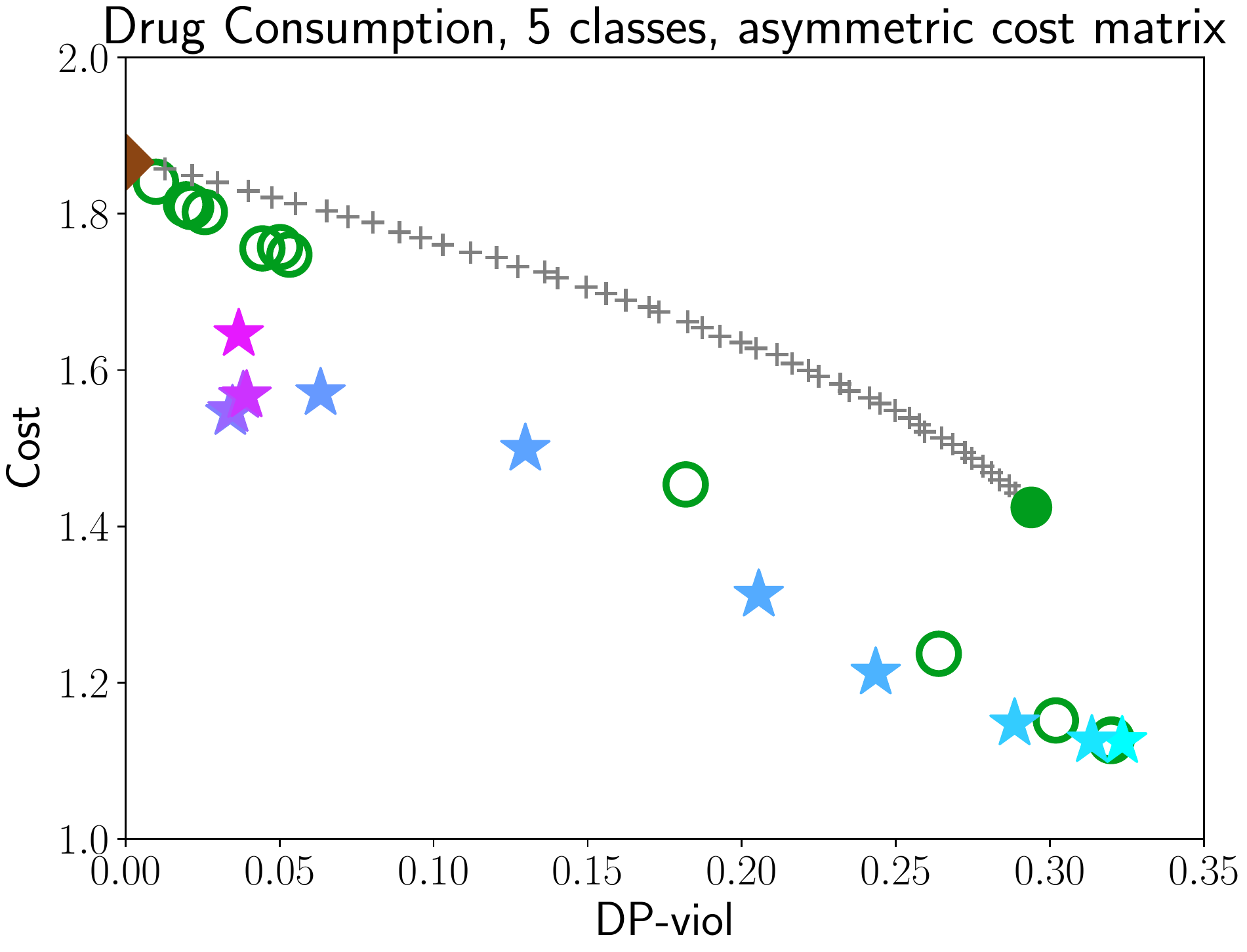}
    \includegraphics[height=4cm]{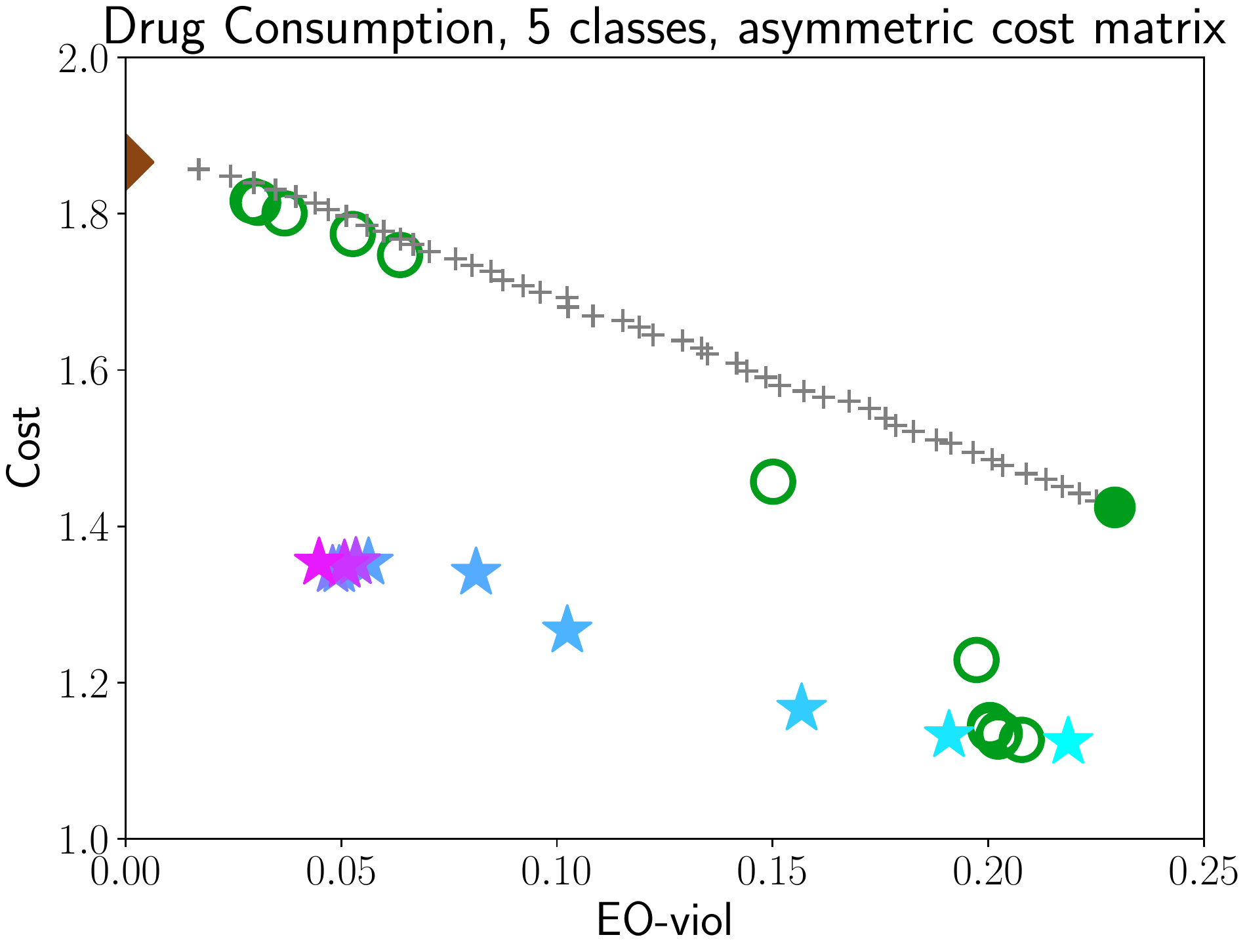}
    \begin{minipage}[c]{5.3cm}
    
    \vspace{-4cm}
     \includegraphics[height=2.5cm]{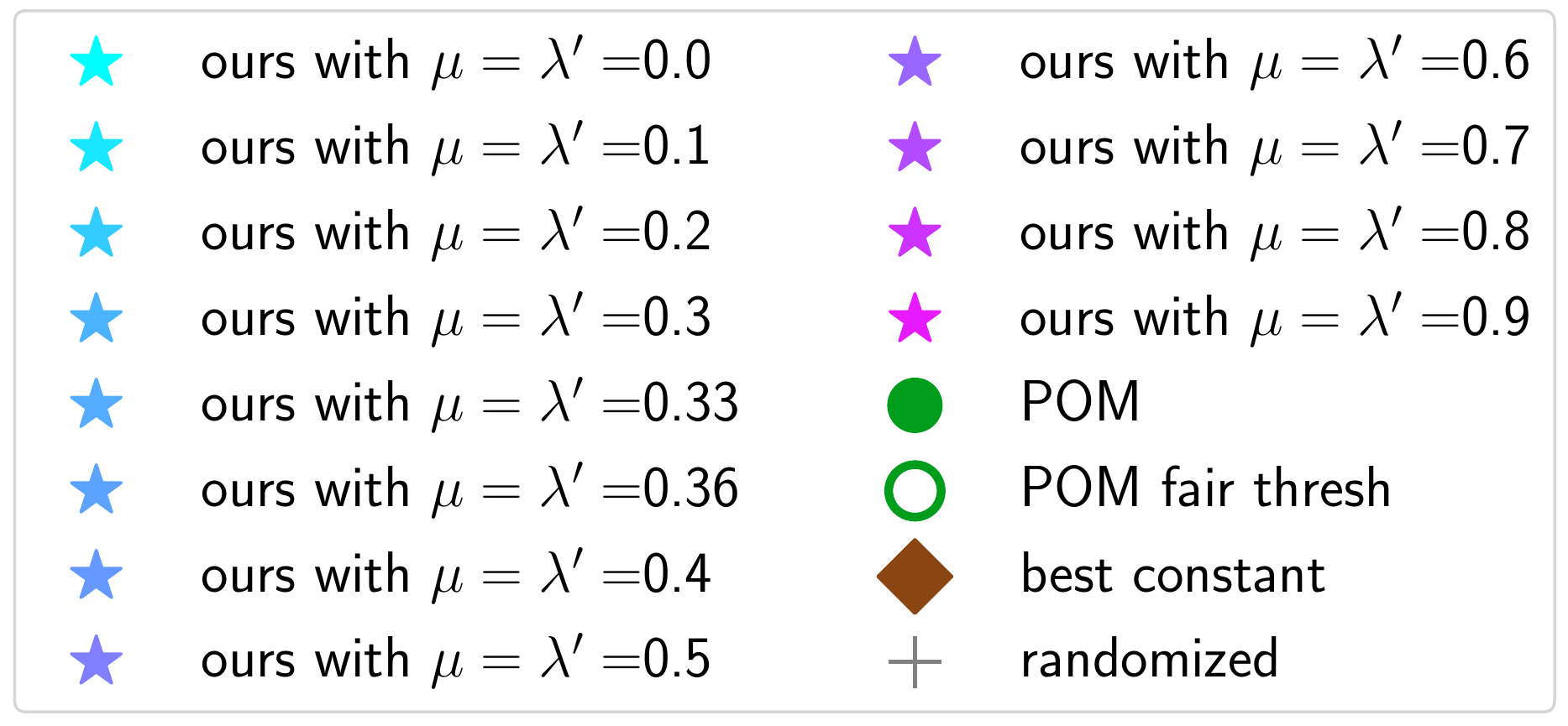}
     \end{minipage}
     
     \caption{Same experiment on the 
     DC 
     dataset as shown in Figure~\ref{fig_drug_consumption}, but with an asymmetric cost~matrix.}\label{figure_appendix_asymmetric}
 \end{figure}

 We performed the same experiment on the Drug Consumption dataset as described in Section~\ref{subsection_experiment_motivation}, but with the asymmetric cost matrix $C_{i,j}=|i-j|+|i-j|\cdot\charfct\{j>i\}$ instead of the absolute cost matrix $C_{i,j}=|i-j|$. 
 All 
 observations 
 made in Section~\ref{subsection_experiment_motivation} for the symmetric case hold similarly in the asymmetric case.

\subsection{
Experiments 
of 
Section~\ref{subsection_experiment_comparison}}\label{appendix_further_experiments}

Figures~\ref{fig:exp_comparison_APPENDIX_real_ord_reg_DP} to \ref{fig:exp_comparison_APPENDIX_DISC_10classes_EO}
show the results of the experiments described in Section~\ref{subsection_experiment_comparison}. Our observations discussed in Section~\ref{section_experiments} generally hold across all datasets, with a few exceptions: on the eucalyptus dataset, all methods are quite fair with $\DPviol\leq 0.08$ and $\EOviol\leq 0.07$, and our method does not provide any significant improvement; note that our predictors are competitive with all state-of-the-art predictors in terms of 
the
$\MAE$. On the newthyroid dataset, we observe two interesting phenomena when aiming for pairwise EO. First, it is striking that the randomized predictors are significantly 
less fair 
than the predictor with the smallest $\MAE$ for values of $p$ around $0.5$ (``the grey curve strings a big bow''). Second, the $\EOviol$ of our predictors increases as $\mu=\lambda'$ increases, which is in stark contrast to what we would expect. We do not have an explanation for these two phenomena, but suspect that they have a common cause. On the toy dataset, the best constant predictor outperforms all other predictors. Since the toy dataset has only a single feature that is provided as input to a predictor, these results are not too meaningful. On the bank1-5 and bank1-10 datasets (when aiming for pairwise DP) and on the bank2-5 and bank2-10 datasets (when aiming for pairwise EO) we observe the same two phenomena as on the newthyroid dataset, which we do not have an explanation for. However, note that 
in these cases 
\emph{all} predictors 
satisfy
$\Fairviol\leq 0.012$ (!) and 
are almost perfectly fair.

Figures~\ref{fig:exp_comparison_APPENDIX_real_ord_reg_DP_with_STD} to \ref{fig:exp_comparison_APPENDIX_DISC_10classes_EO_with_STD} 
show the same results, but additionally show the standard deviation of the $\MAE$ and $\Fairviol$ over the various splits.
We can see that the standard deviations for 
our predictors 
are mostly comparable with the standard deviations for the predictors produced by the \textsc{Orca} algorithms, both in case of the $\MAE$ and $\Fairviol$.

\renewcommand{\scaleparameterA}{0.18}
\renewcommand{\abstA}{6pt}
\renewcommand{\scaleparameterA}{0.2}
\renewcommand{\abstA}{2pt}

\begin{figure*}
    \centering
    \includegraphics[width=\linewidth]{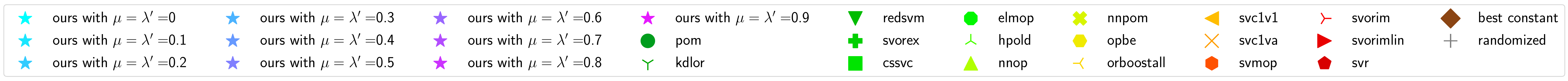}

    \includegraphics[scale=\scaleparameterA]{experiment_real_ord/DP/automobile.pdf}
    \hspace{\abstA}
    \includegraphics[scale=\scaleparameterA]{experiment_real_ord/DP/balance-scale.pdf}
    \hspace{\abstA}
    \includegraphics[scale=\scaleparameterA]{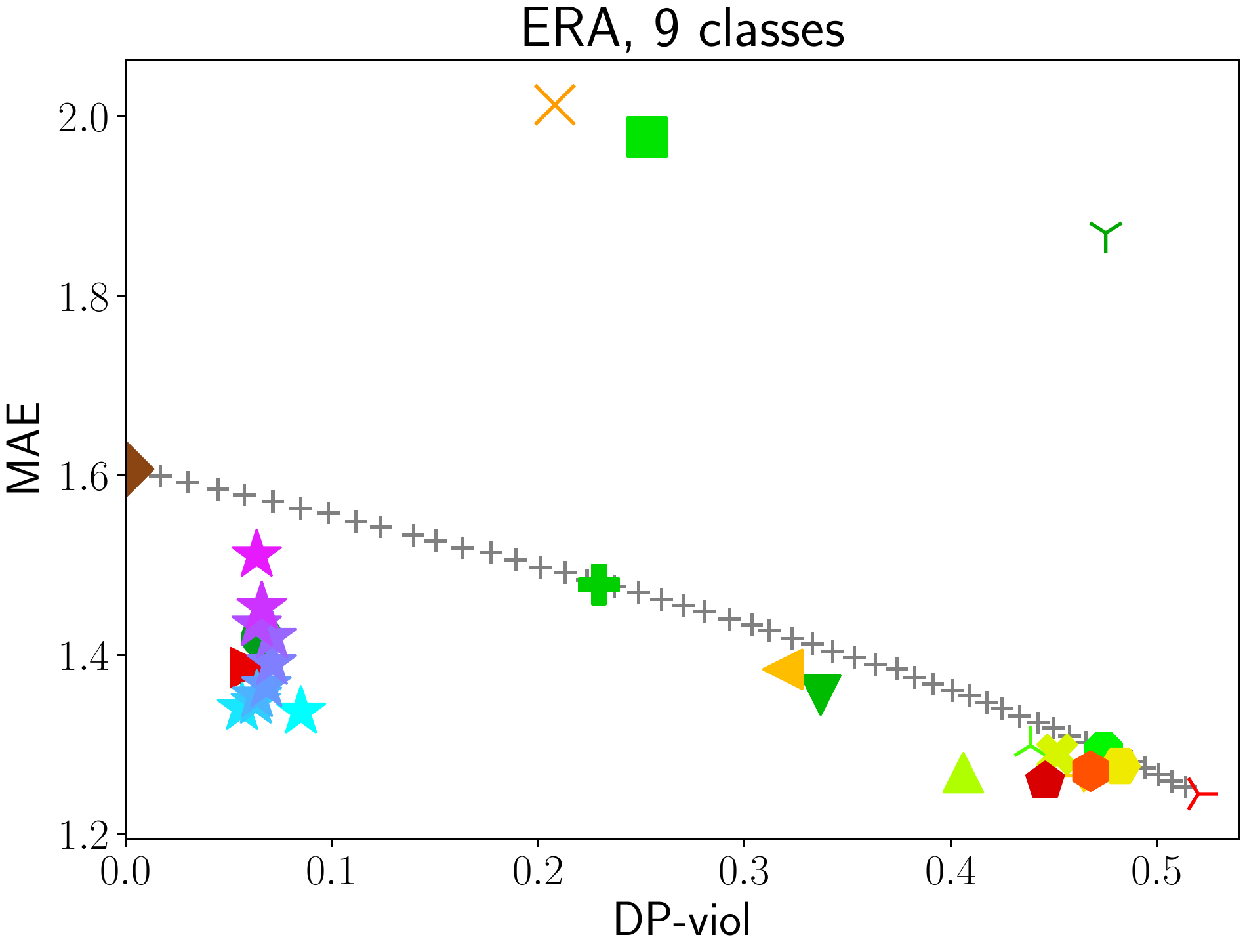}
    \hspace{\abstA}
    \includegraphics[scale=\scaleparameterA]{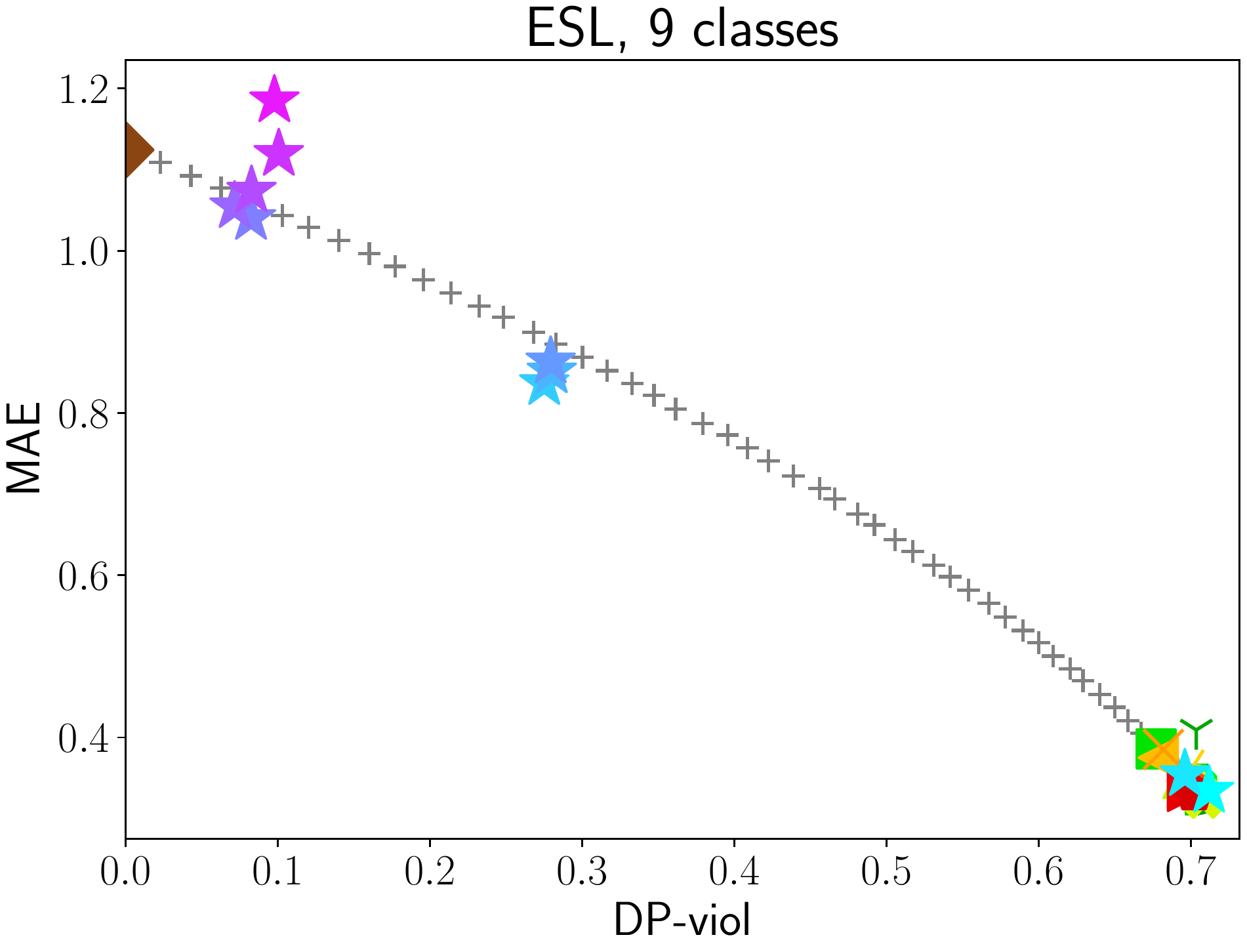}

    \includegraphics[scale=\scaleparameterA]{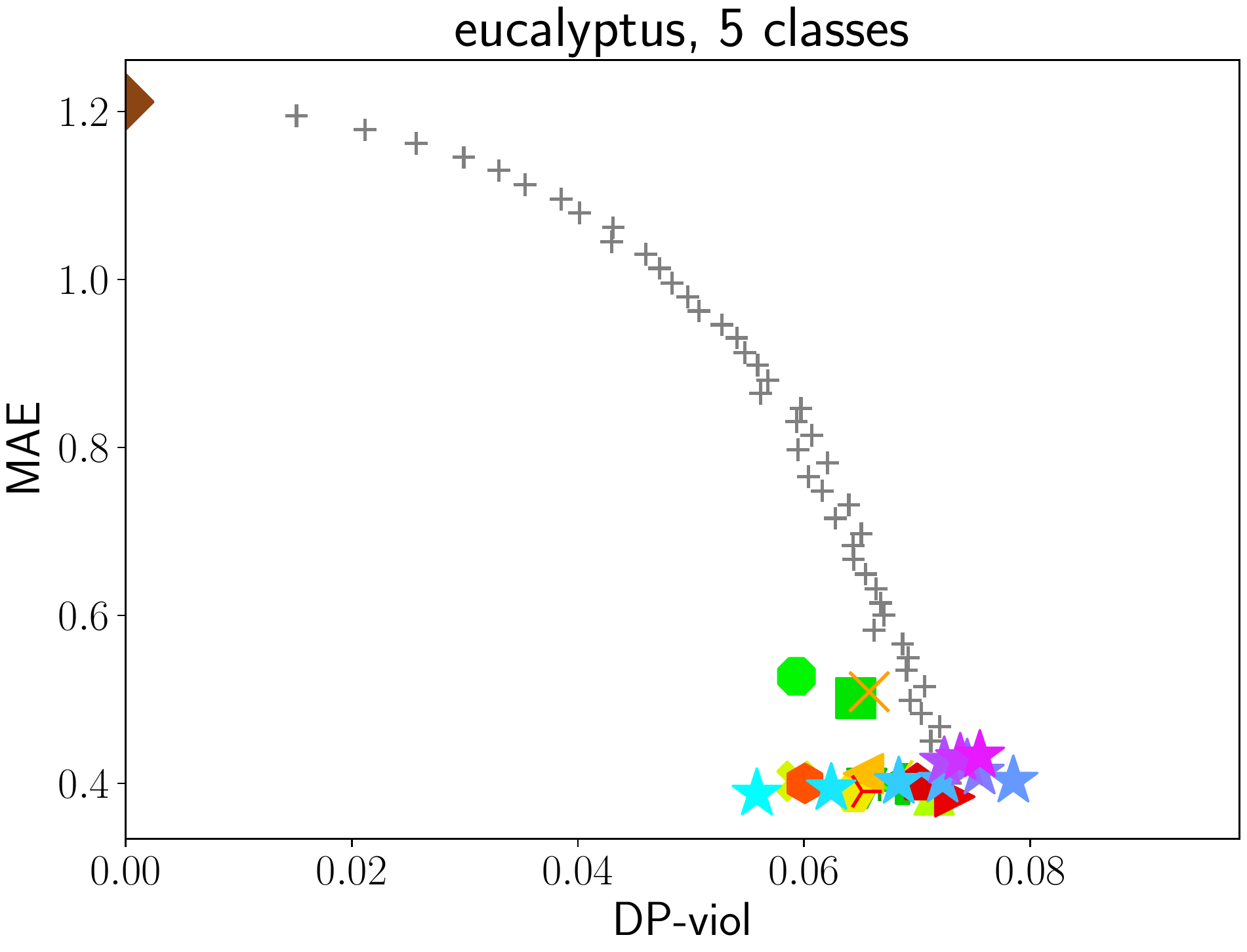}
    \hspace{\abstA}
    \includegraphics[scale=\scaleparameterA]{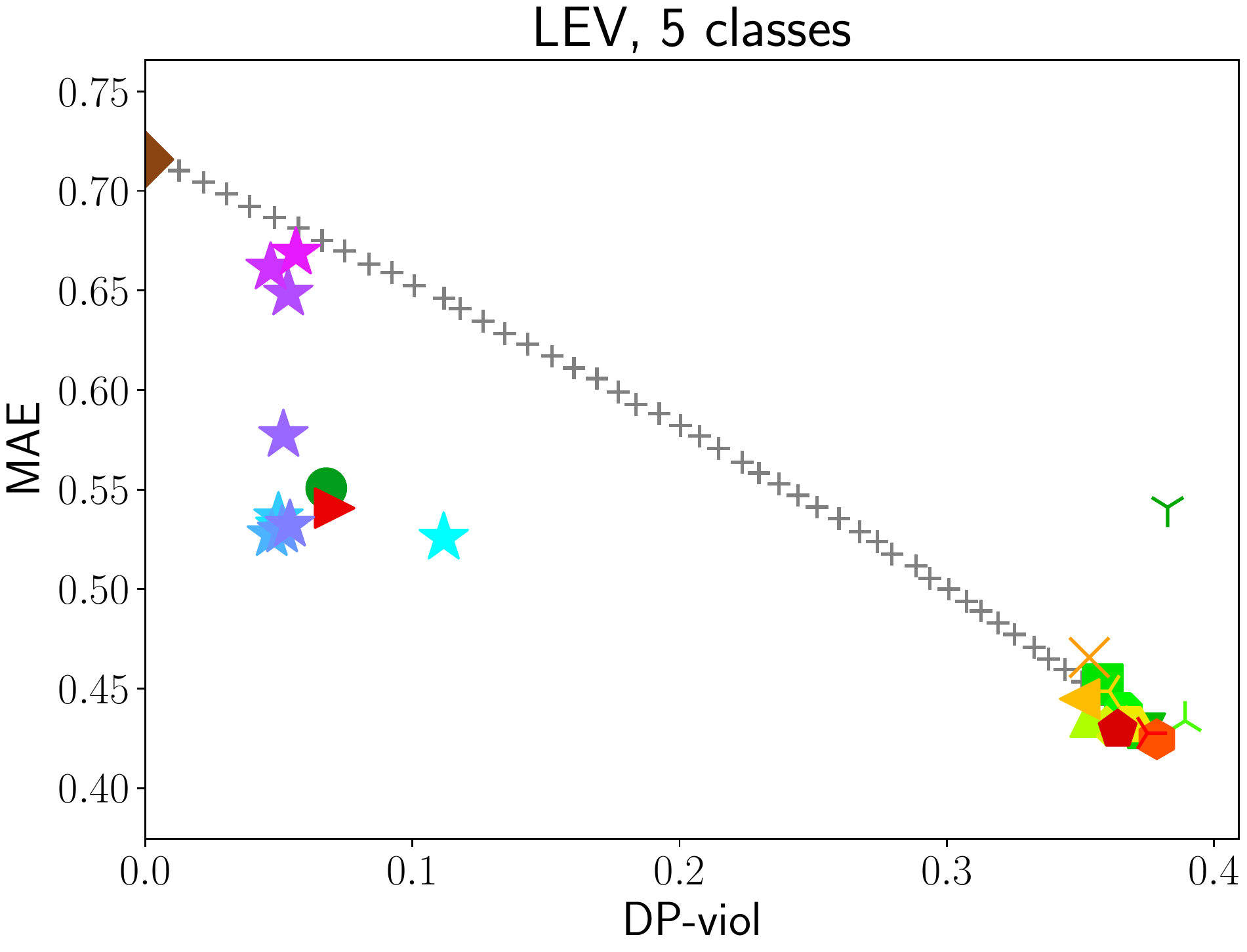}
    \hspace{\abstA}
    \includegraphics[scale=\scaleparameterA]{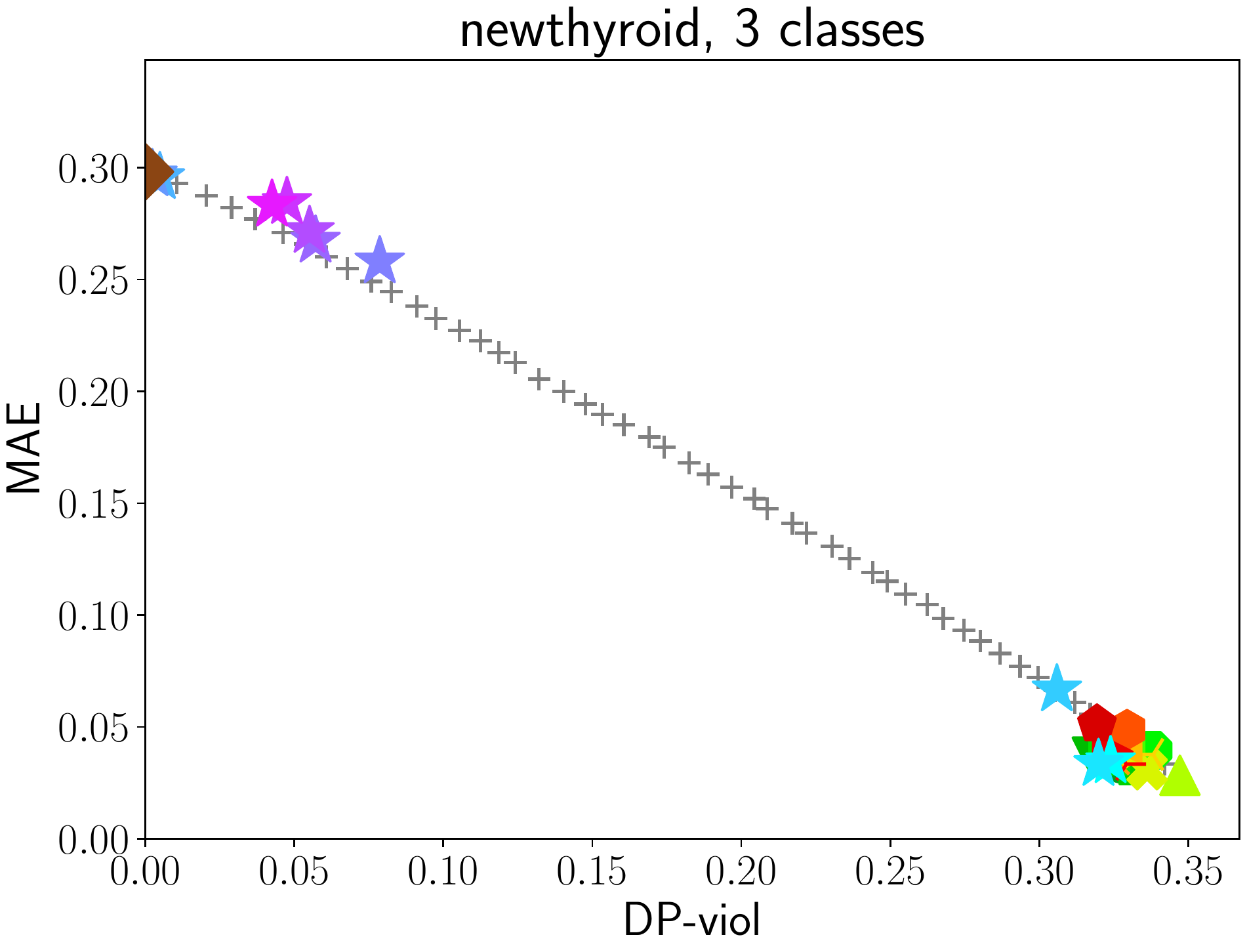}
    \hspace{\abstA}
    \includegraphics[scale=\scaleparameterA]{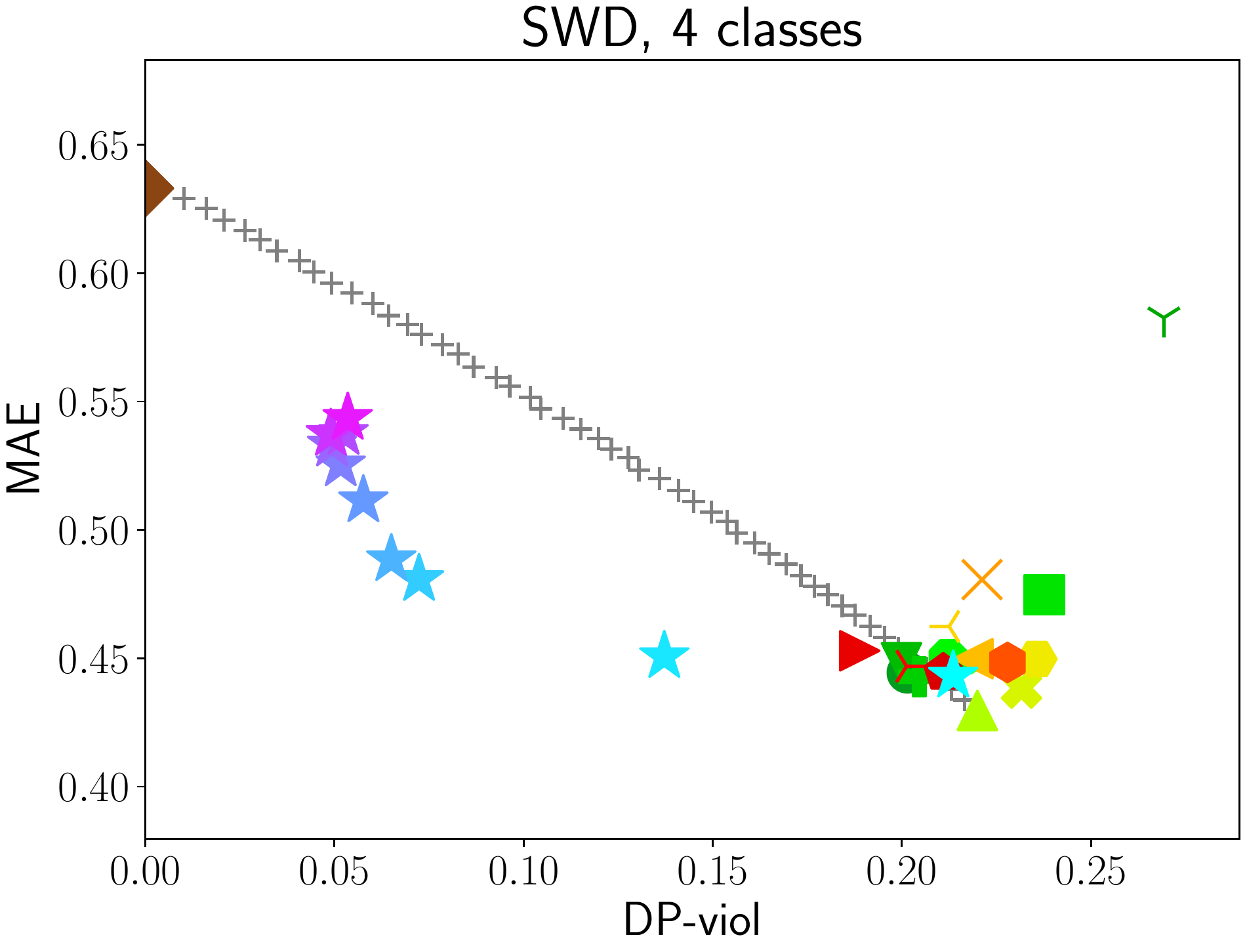}
    
    \includegraphics[scale=\scaleparameterA]{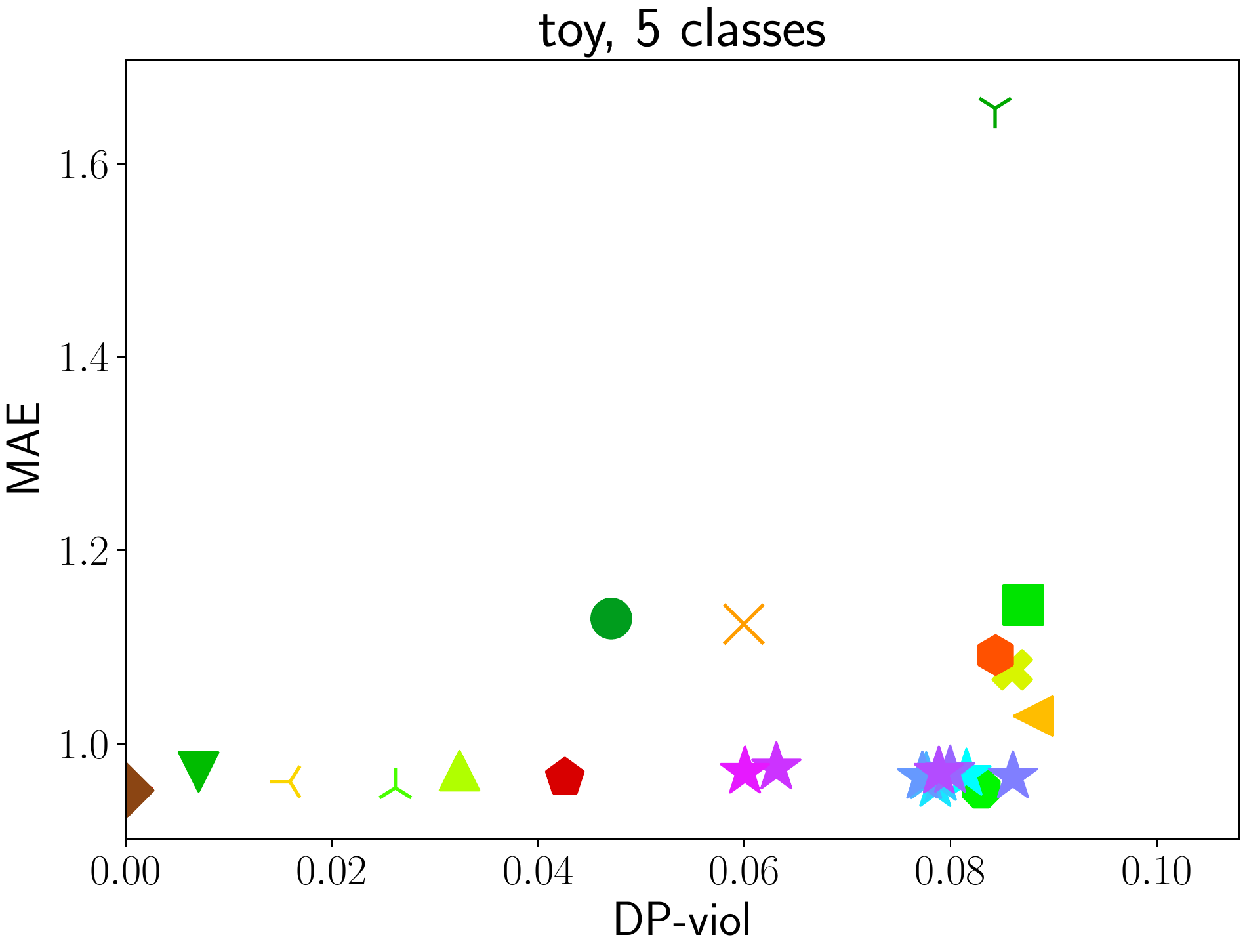}
    \hspace{\abstA}
    \includegraphics[scale=\scaleparameterA]{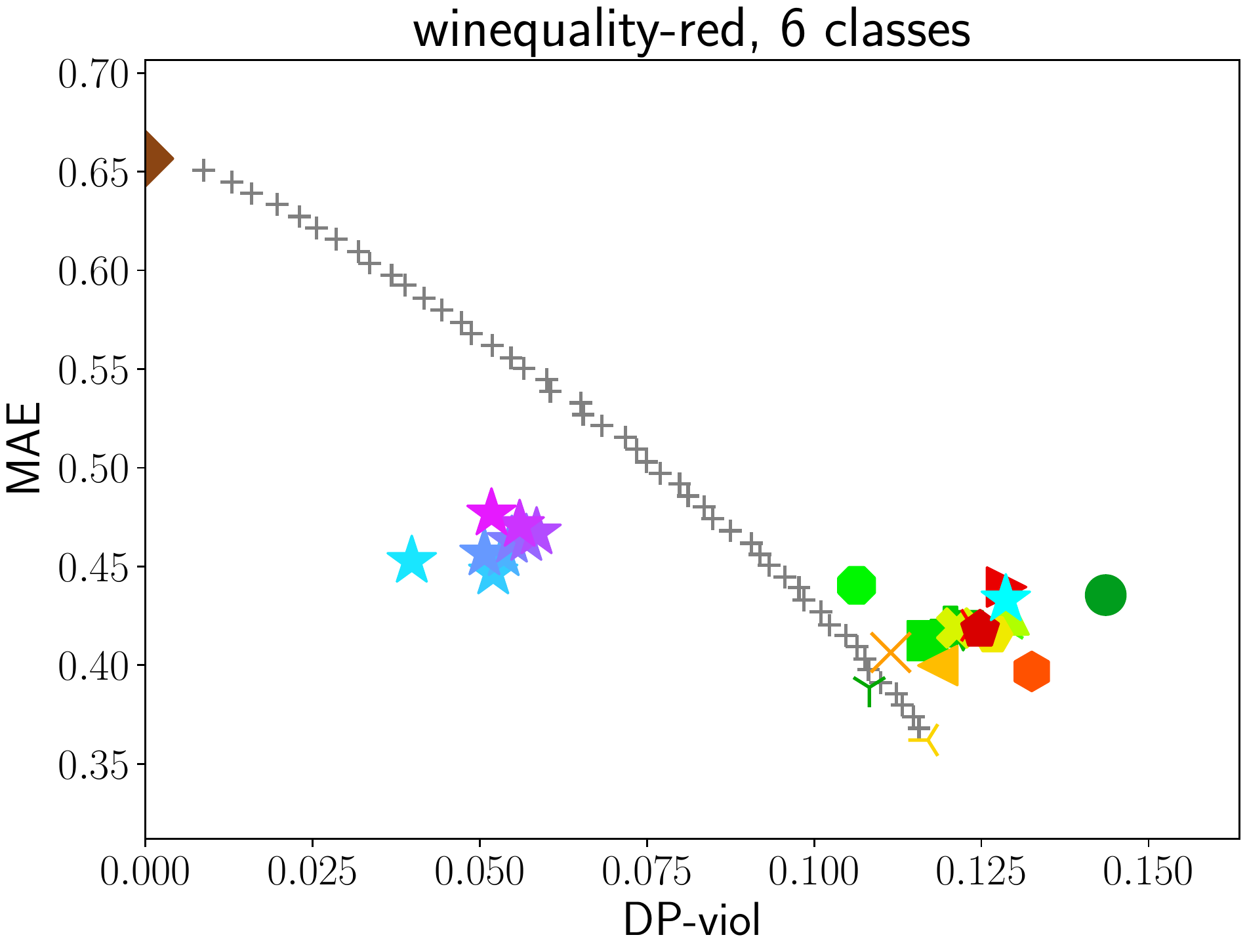}
    \hspace{4.1cm}
    \includegraphics[scale=\scaleparameterA]{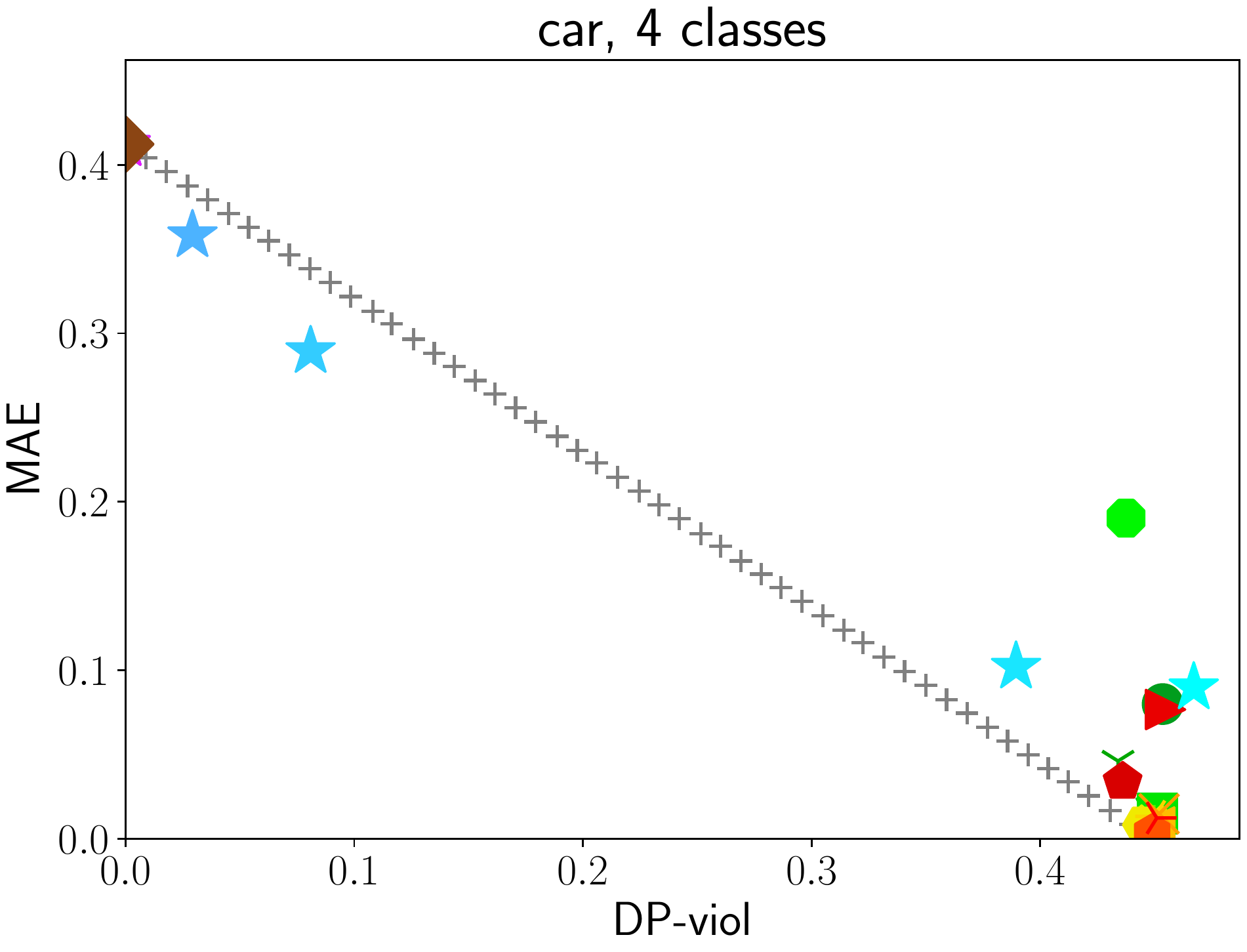}

    \caption{Experiments of Section~\ref{subsection_experiment_comparison} on the \textbf{real ordinal regression datasets} when aiming for \textbf{pairwise DP}. Note that the toy dataset has only a single feature that is provided as input to a predictor and that the best method on the toy dataset (svorex) coincides with the best constant predictor;  we do not see any grey crosses corresponding to randomly mixing the best predictor with the best constant one.}
    \label{fig:exp_comparison_APPENDIX_real_ord_reg_DP}
\end{figure*}

\begin{figure*}
    
    \centering
    \includegraphics[width=\linewidth]{experiment_real_ord/legend_big.pdf}

    \includegraphics[scale=\scaleparameterA]{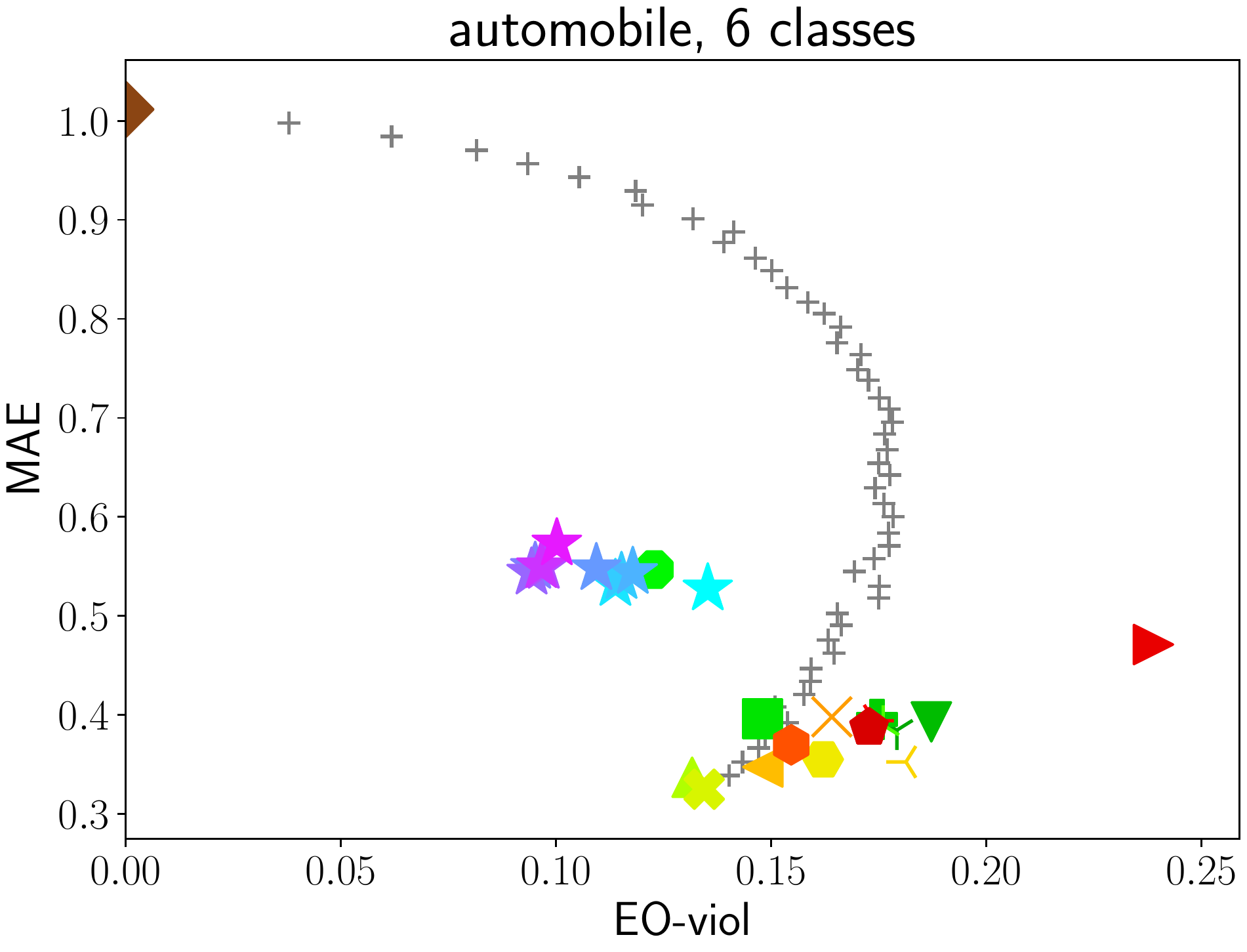}
    \hspace{\abstA}
    \includegraphics[scale=\scaleparameterA]{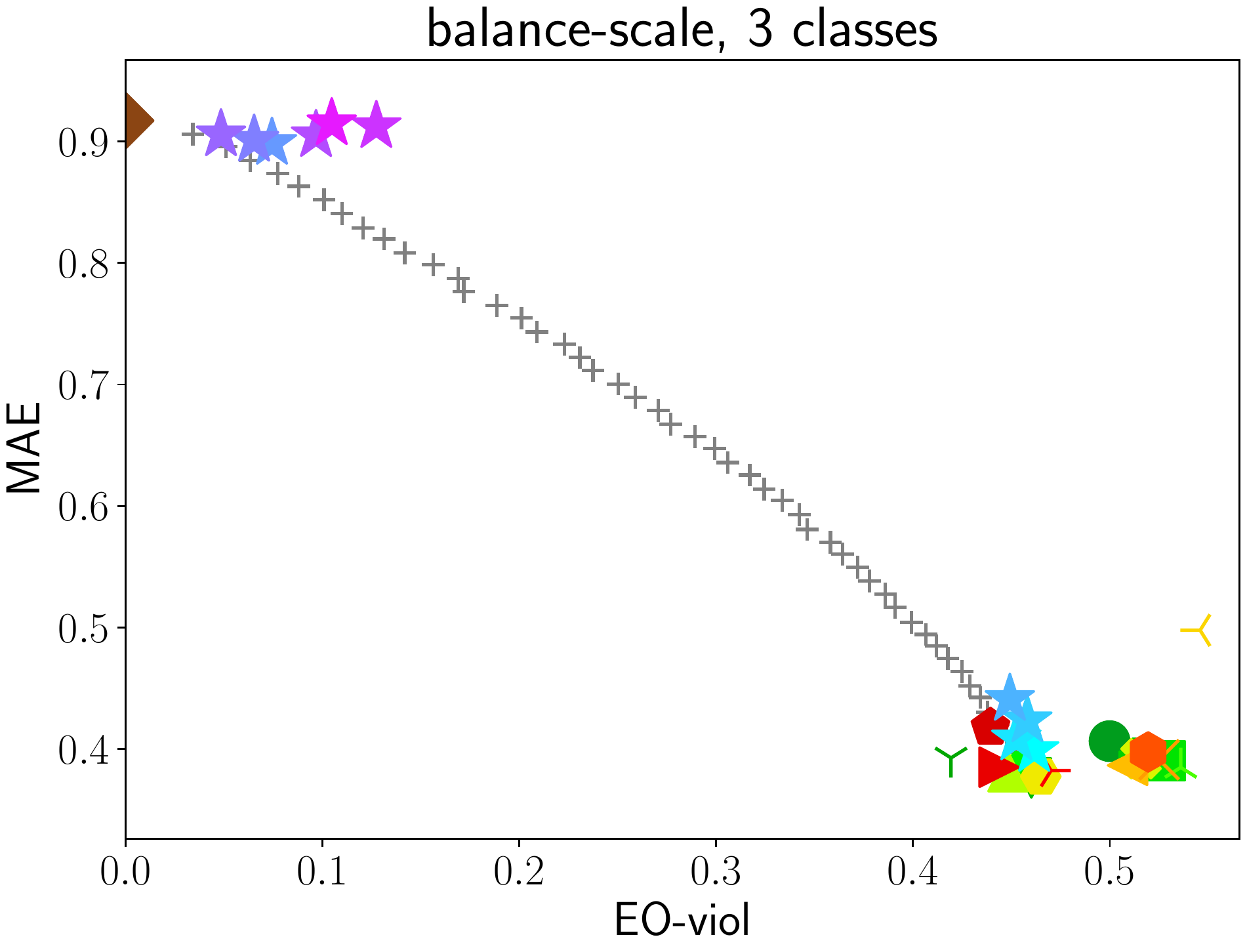}
    \hspace{\abstA}
    \includegraphics[scale=\scaleparameterA]{experiment_real_ord/EO/ERA.pdf}
    \hspace{\abstA}
    \includegraphics[scale=\scaleparameterA]{experiment_real_ord/EO/ESL.pdf}
    
    \includegraphics[scale=\scaleparameterA]{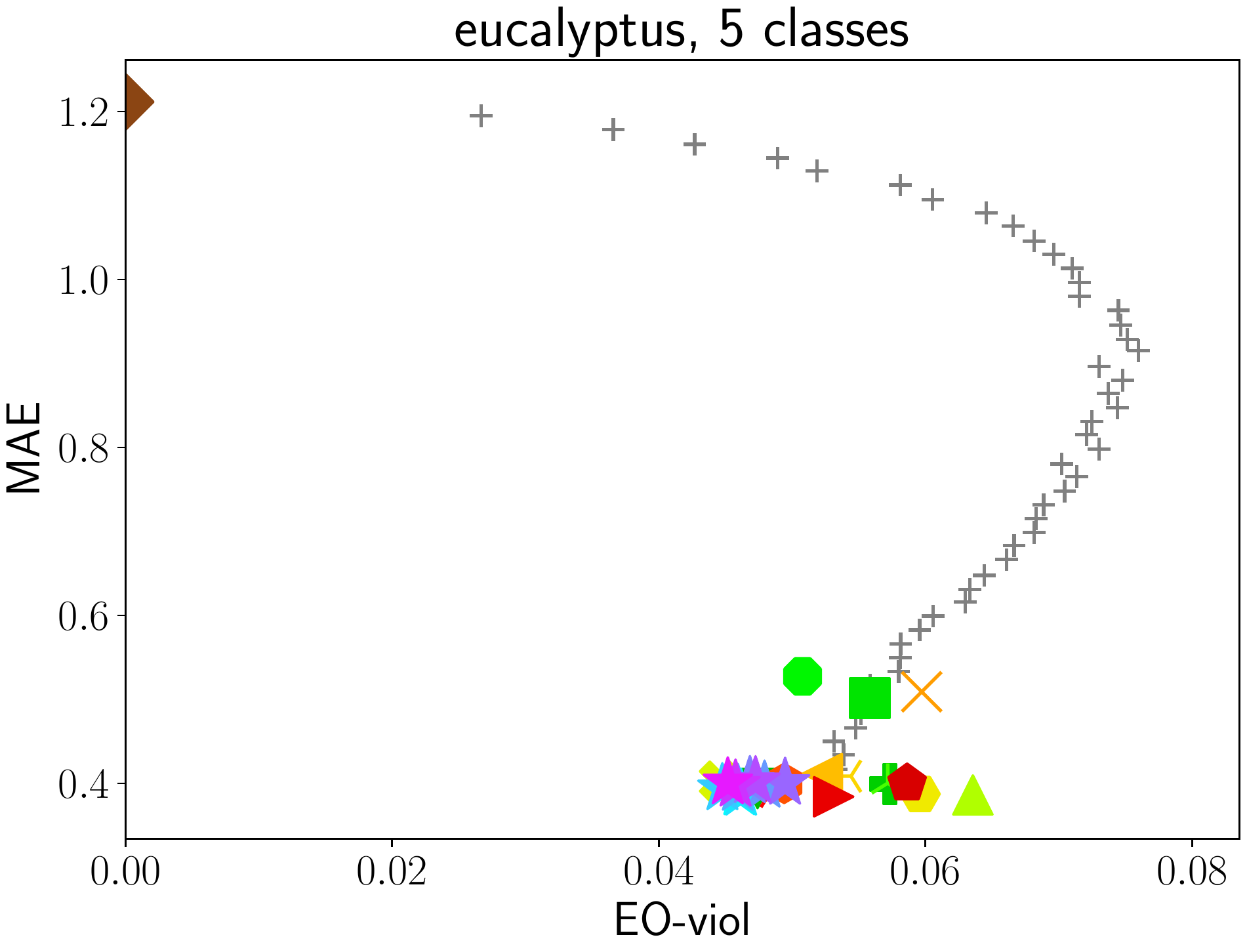}
    \hspace{\abstA}
    \includegraphics[scale=\scaleparameterA]{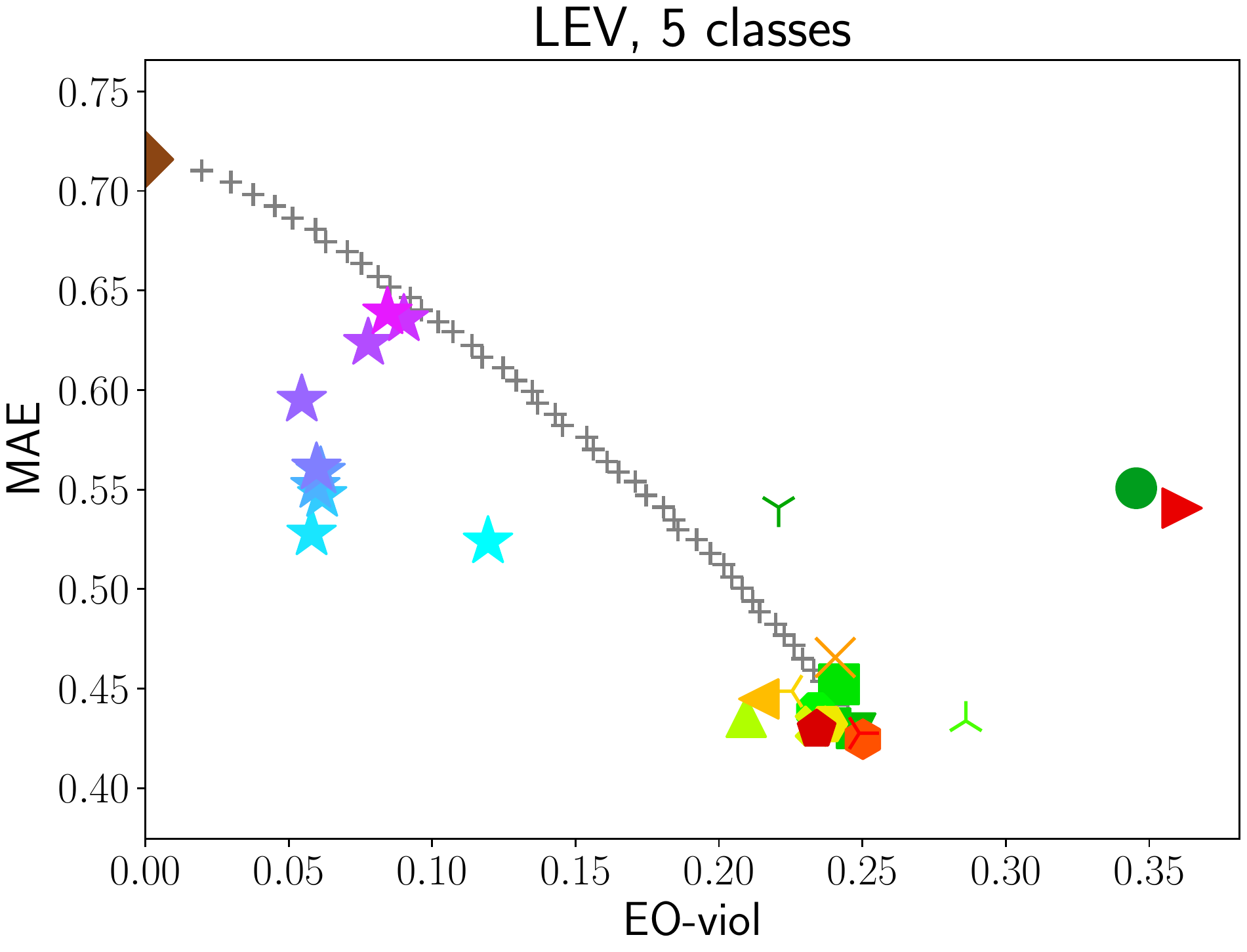}
    \hspace{\abstA}
    \includegraphics[scale=\scaleparameterA]{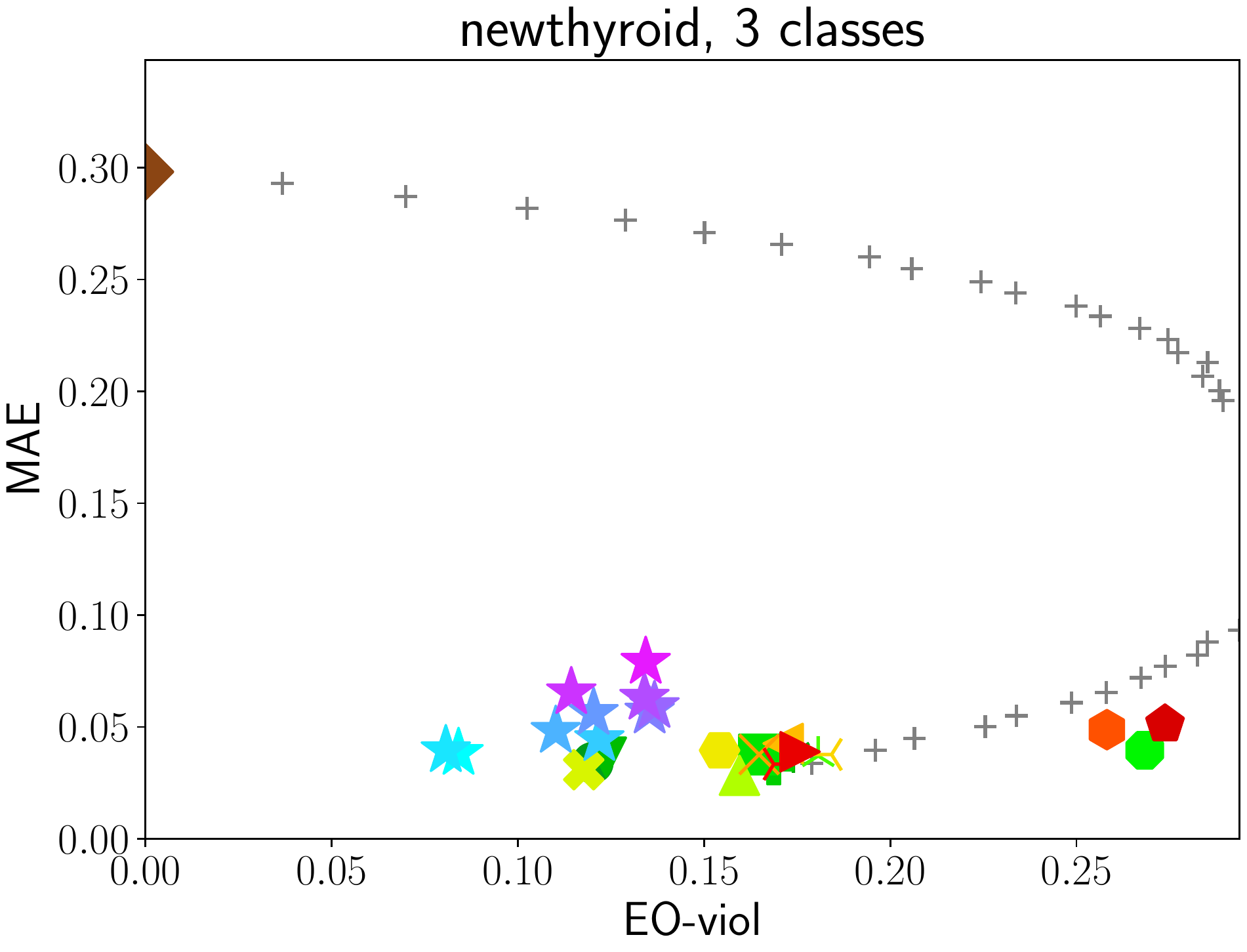}
    \hspace{\abstA}
    \includegraphics[scale=\scaleparameterA]{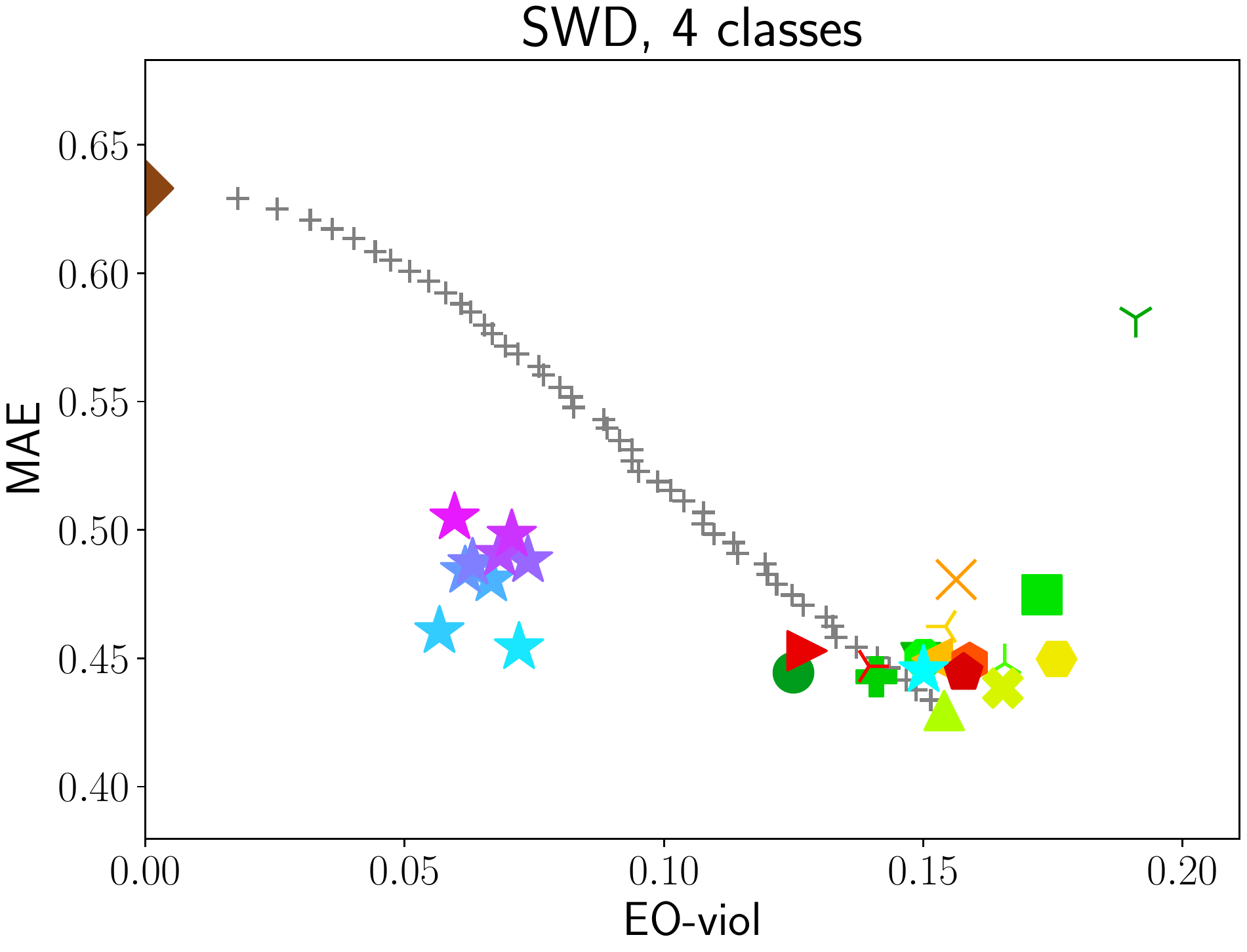}
    
    \includegraphics[scale=\scaleparameterA]{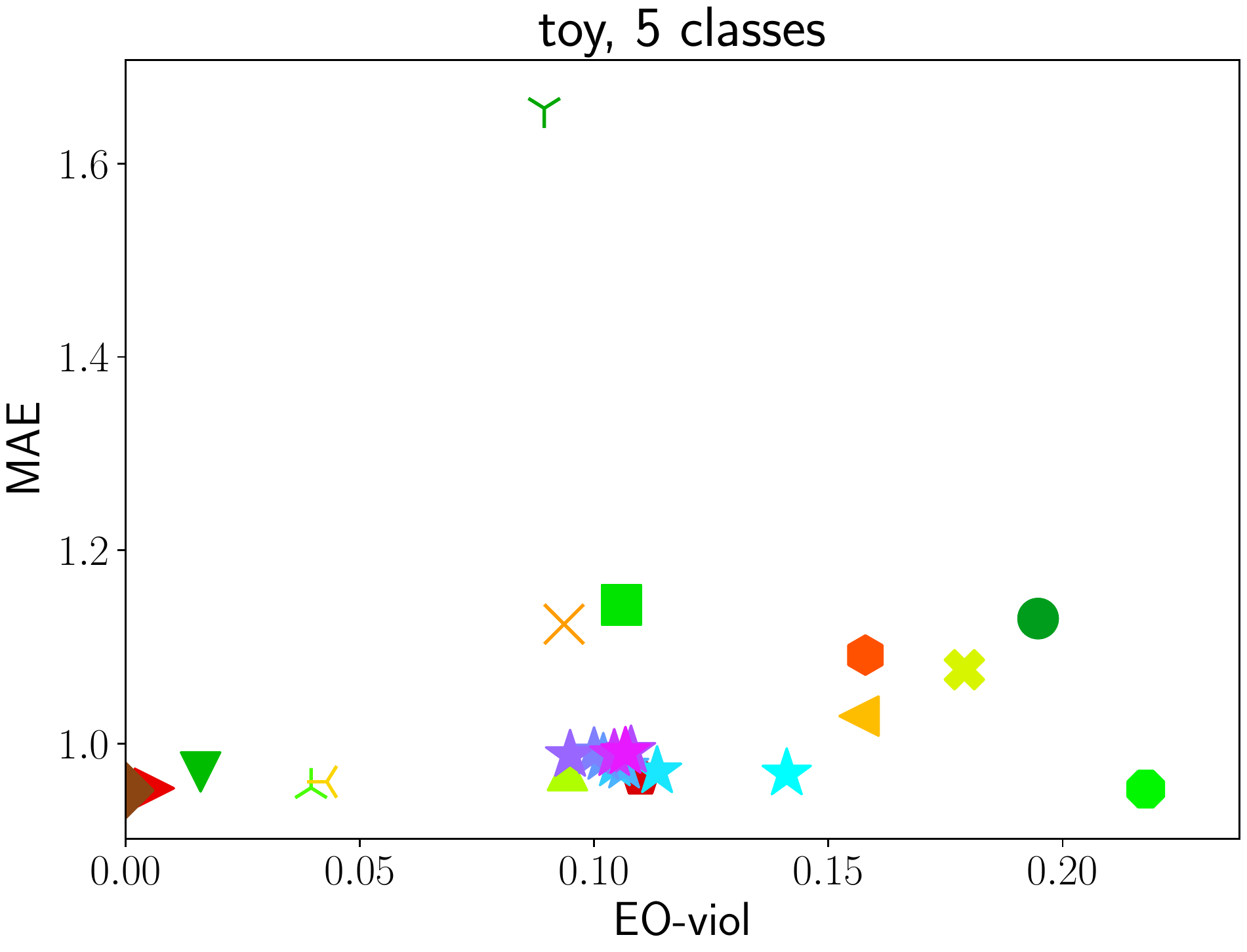}
    \hspace{\abstA}
    \includegraphics[scale=\scaleparameterA]{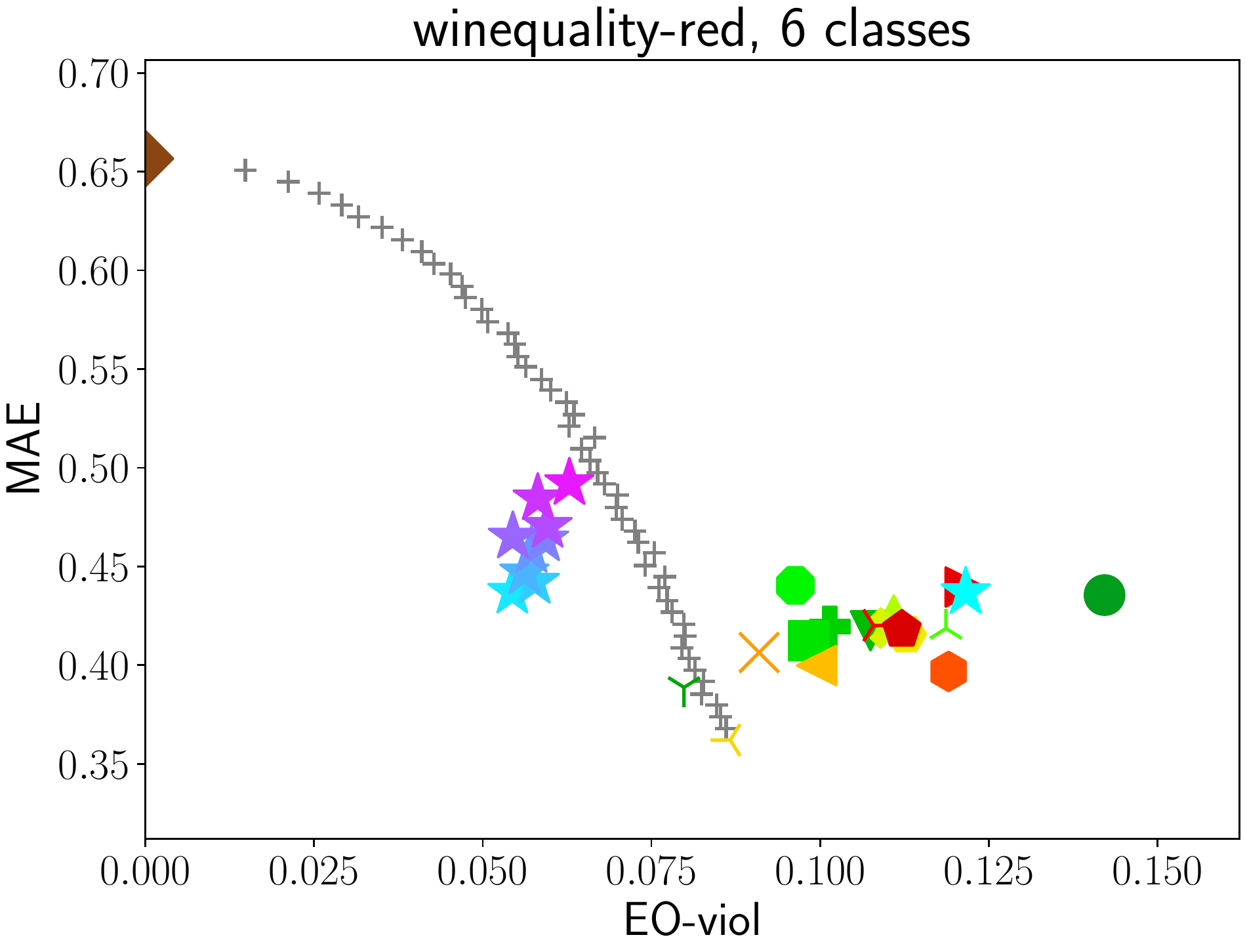}
    \hspace{7.8cm}
    
\caption{Experiments of Section~\ref{subsection_experiment_comparison} on the \textbf{real ordinal regression datasets} when aiming for \textbf{pairwise EO}. Note that the toy dataset has only a single feature that is provided as input to a predictor and that the best method on the toy dataset (svorex) coincides with the best constant predictor;  we do not see any grey crosses corresponding to randomly mixing the best predictor with the best constant one. Also note that we do not provide a plot for the car dataset; since $\Psymb[y_1<y_2, a_1=0,a_2=1]=0$ for the car dataset (cf. Table~\ref{table_statistics_data_sets2}), the notion of pairwise EO is not well-defined for this dataset.}
    \label{fig:exp_comparison_APPENDIX_real_ord_reg_EO}
\end{figure*}

\renewcommand{\scaleparameterA}{0.2}
\renewcommand{\abstA}{2pt}

\begin{figure*}
    
    \includegraphics[width=\linewidth]{experiment_real_ord/legend_big.pdf}

    \includegraphics[scale=\scaleparameterA]{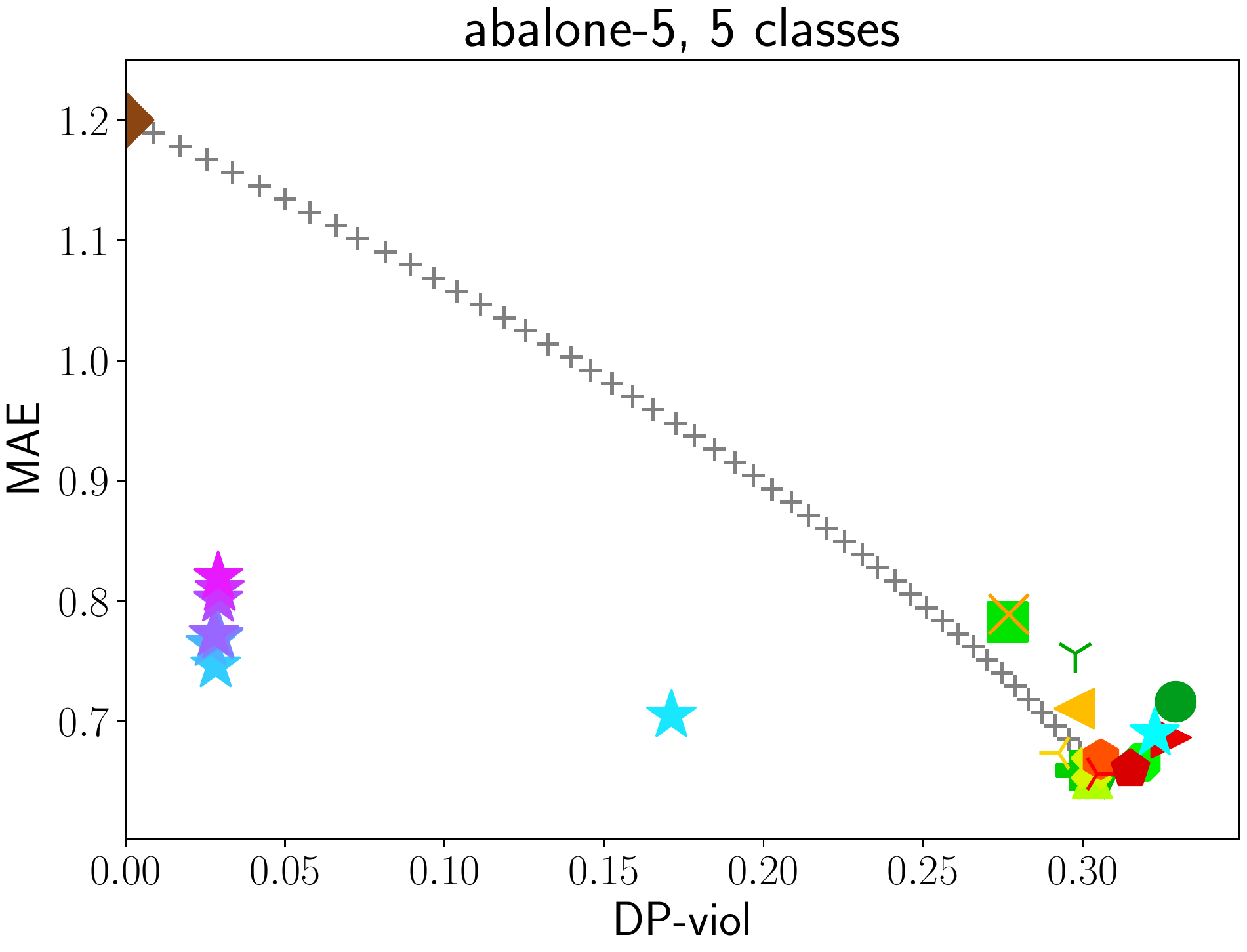}
    \hspace{\abstA}
    \includegraphics[scale=\scaleparameterA]{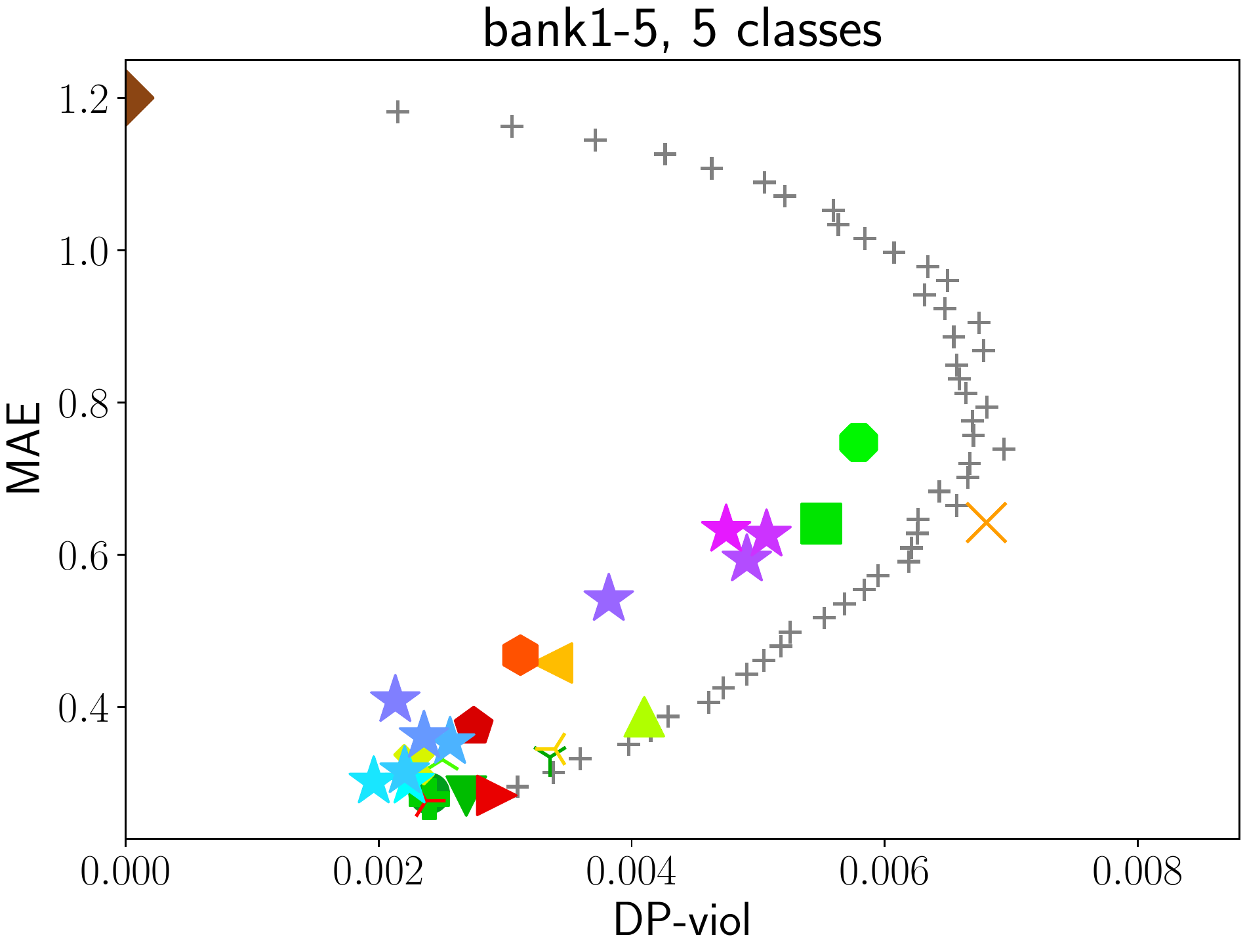}
    \hspace{\abstA}
    \includegraphics[scale=\scaleparameterA]{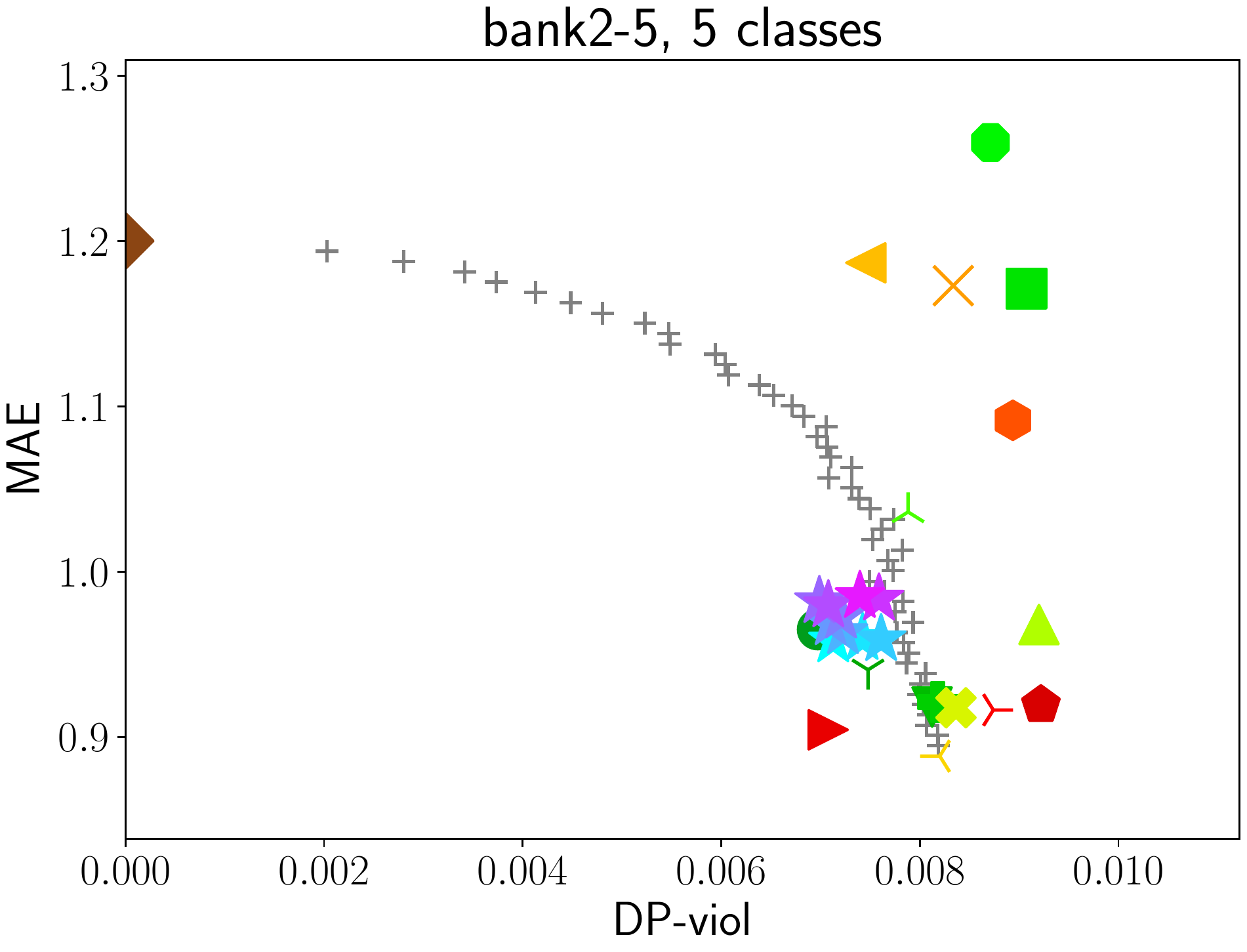}
    \hspace{\abstA}
    \includegraphics[scale=\scaleparameterA]{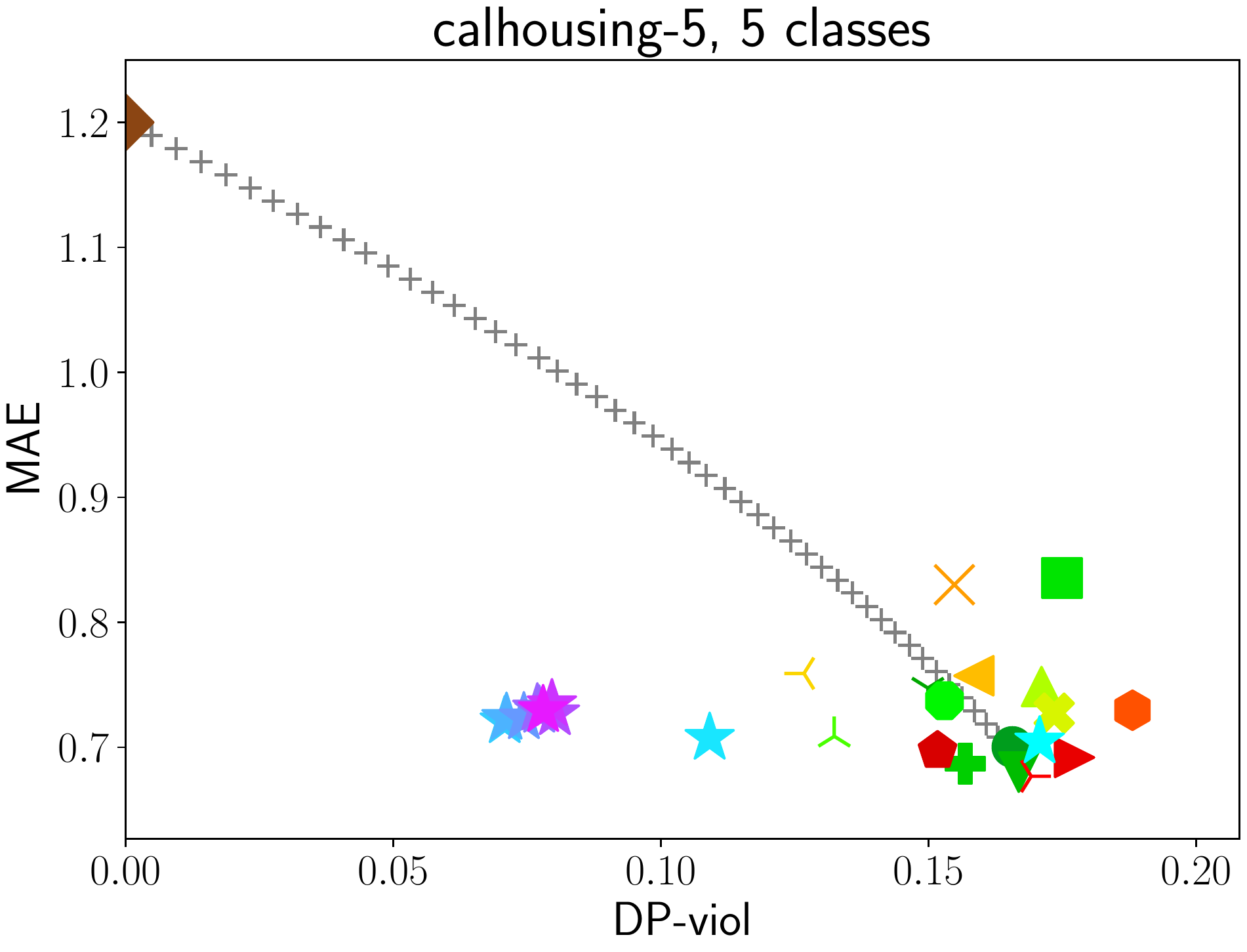}
    
    \includegraphics[scale=\scaleparameterA]{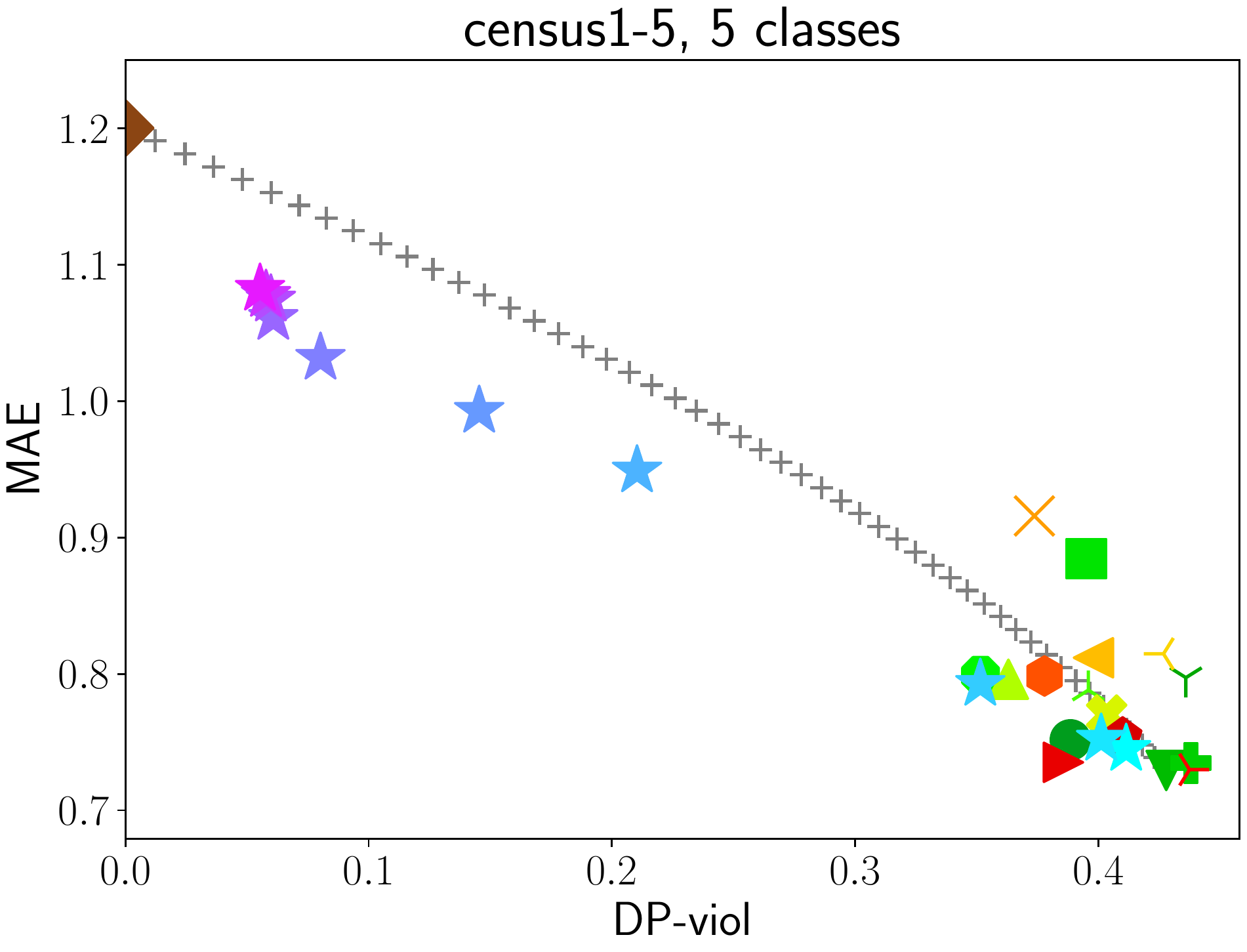}
    \hspace{\abstA}
    \includegraphics[scale=\scaleparameterA]{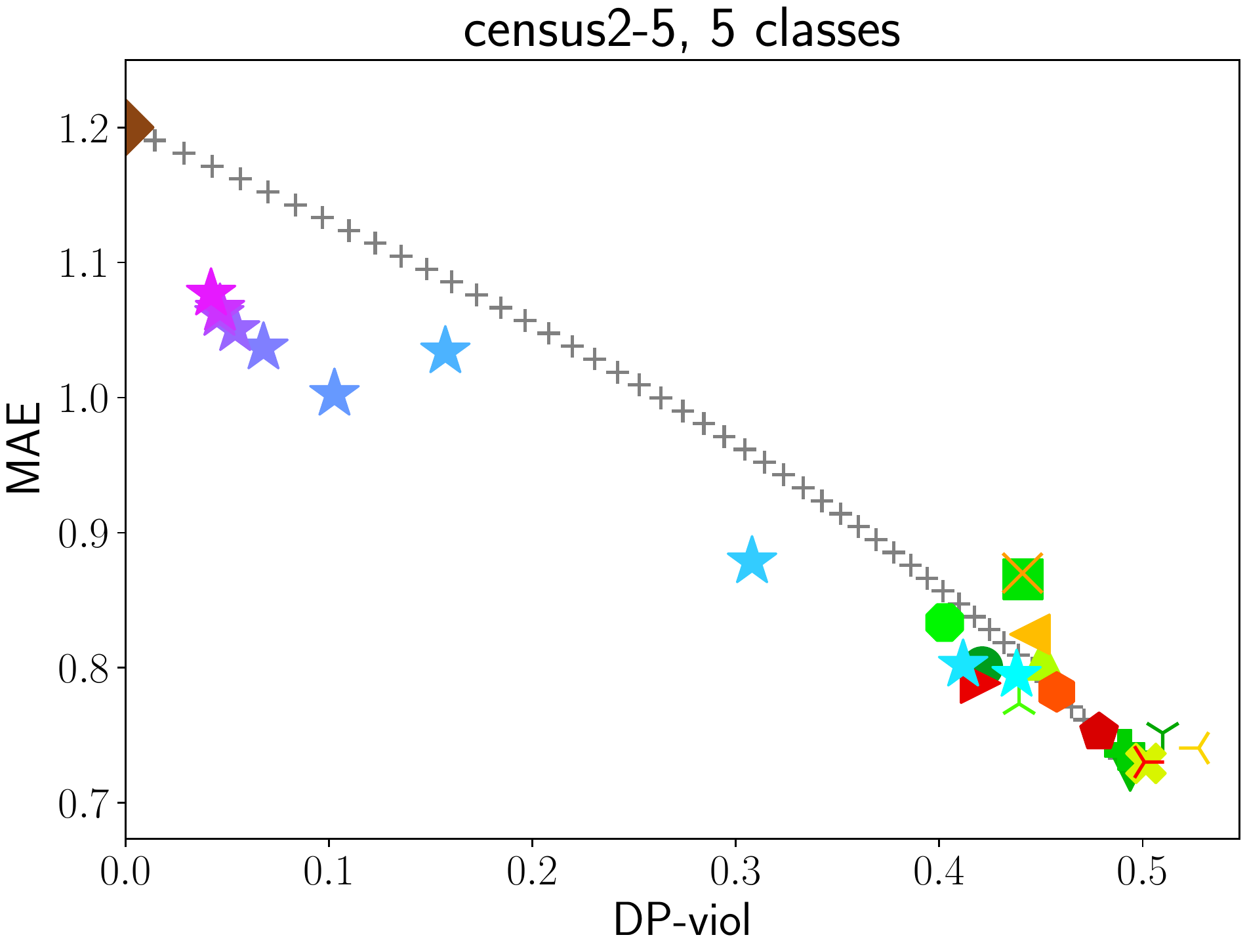}
    \hspace{\abstA}
    \includegraphics[scale=\scaleparameterA]{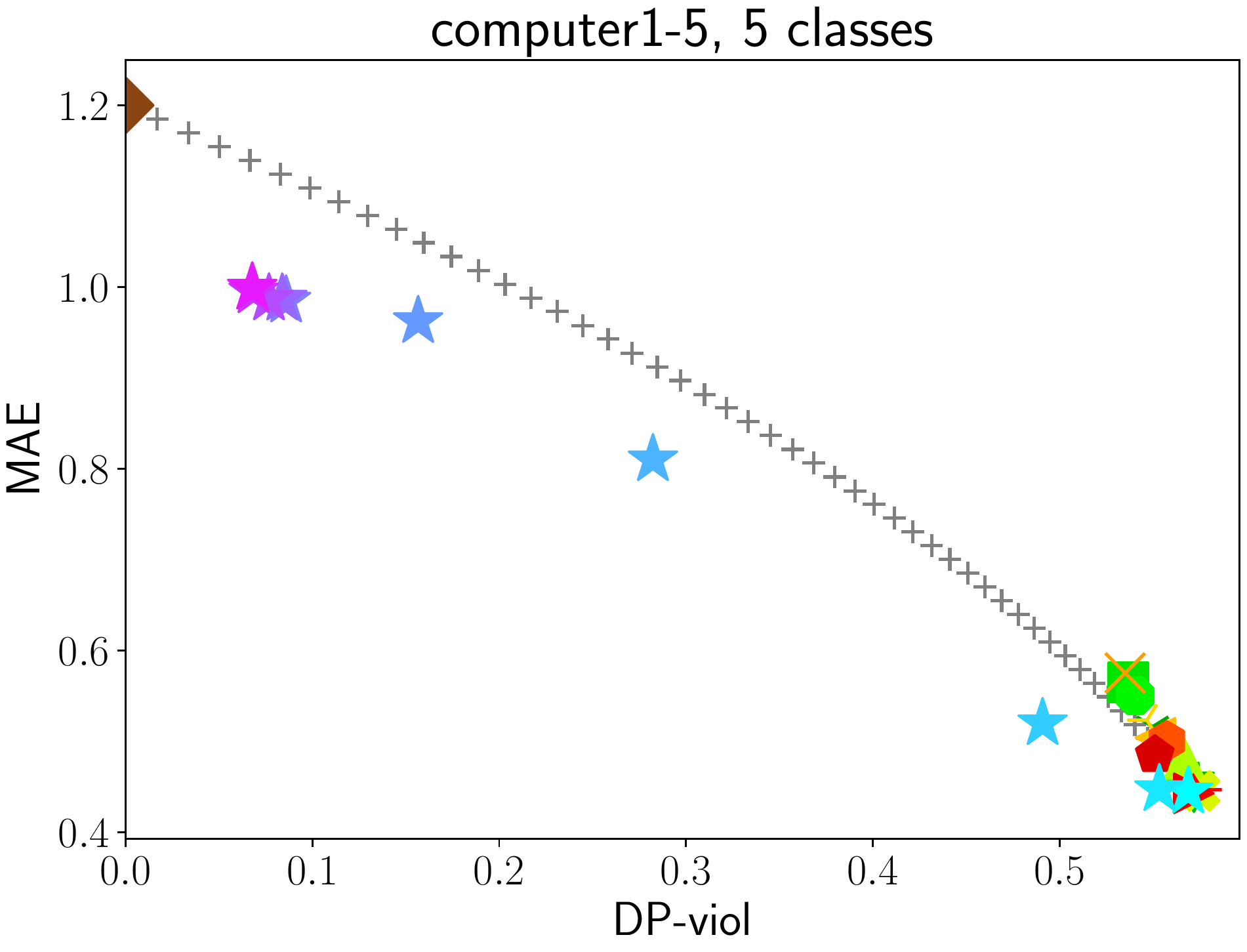}
    \hspace{\abstA}
    \includegraphics[scale=\scaleparameterA]{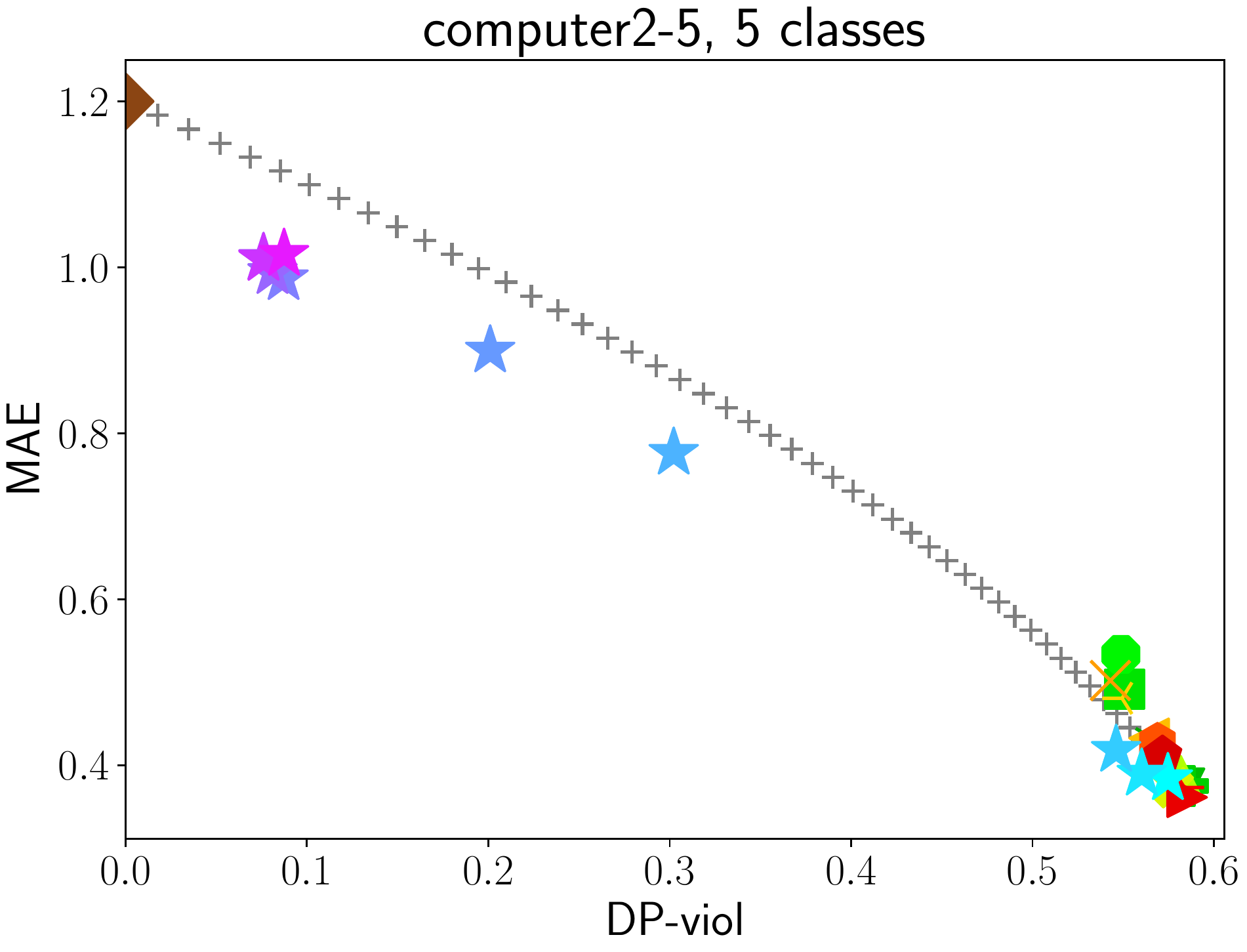}
    
    \includegraphics[scale=\scaleparameterA]{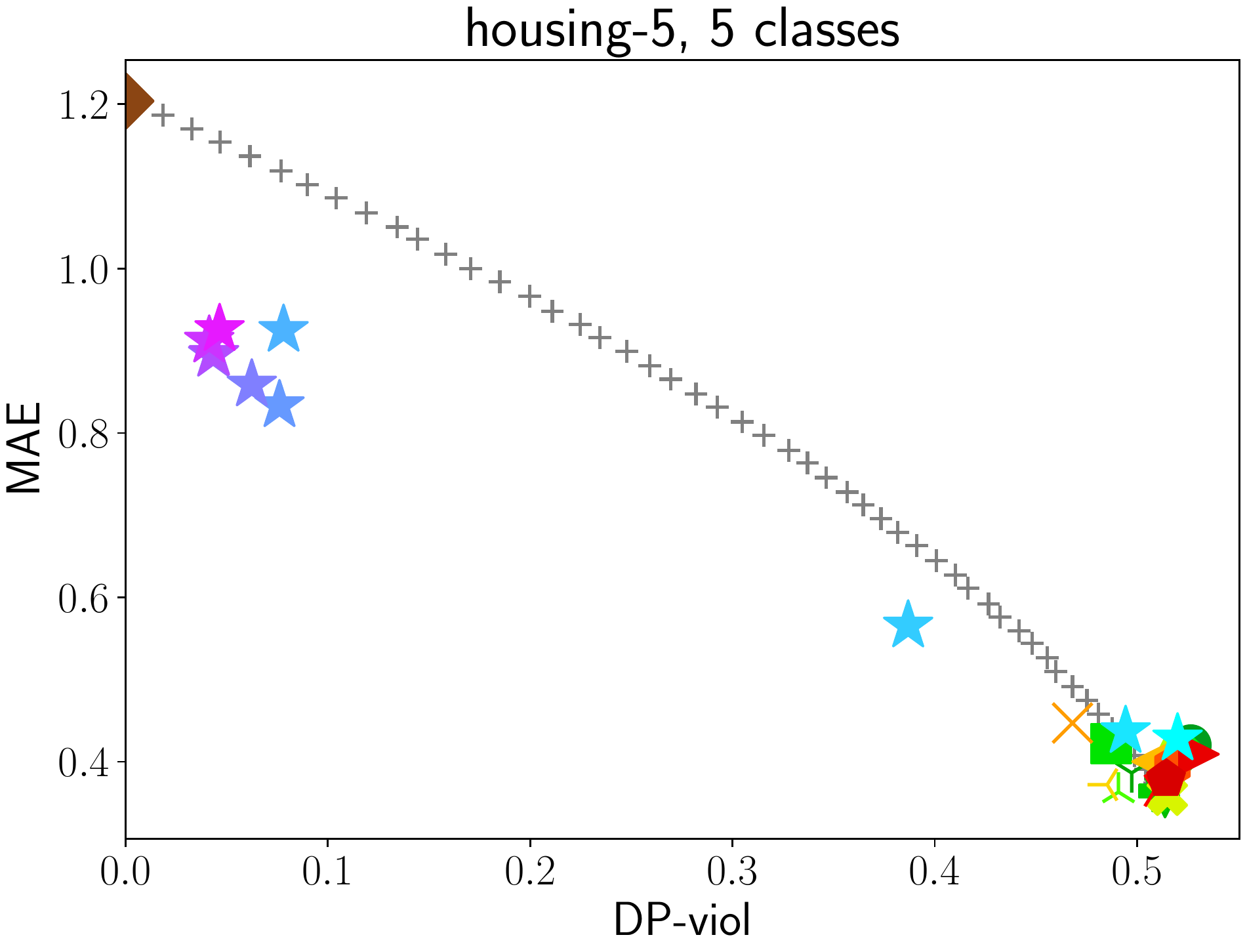}
    \hspace{\abstA}
    \includegraphics[scale=\scaleparameterA]{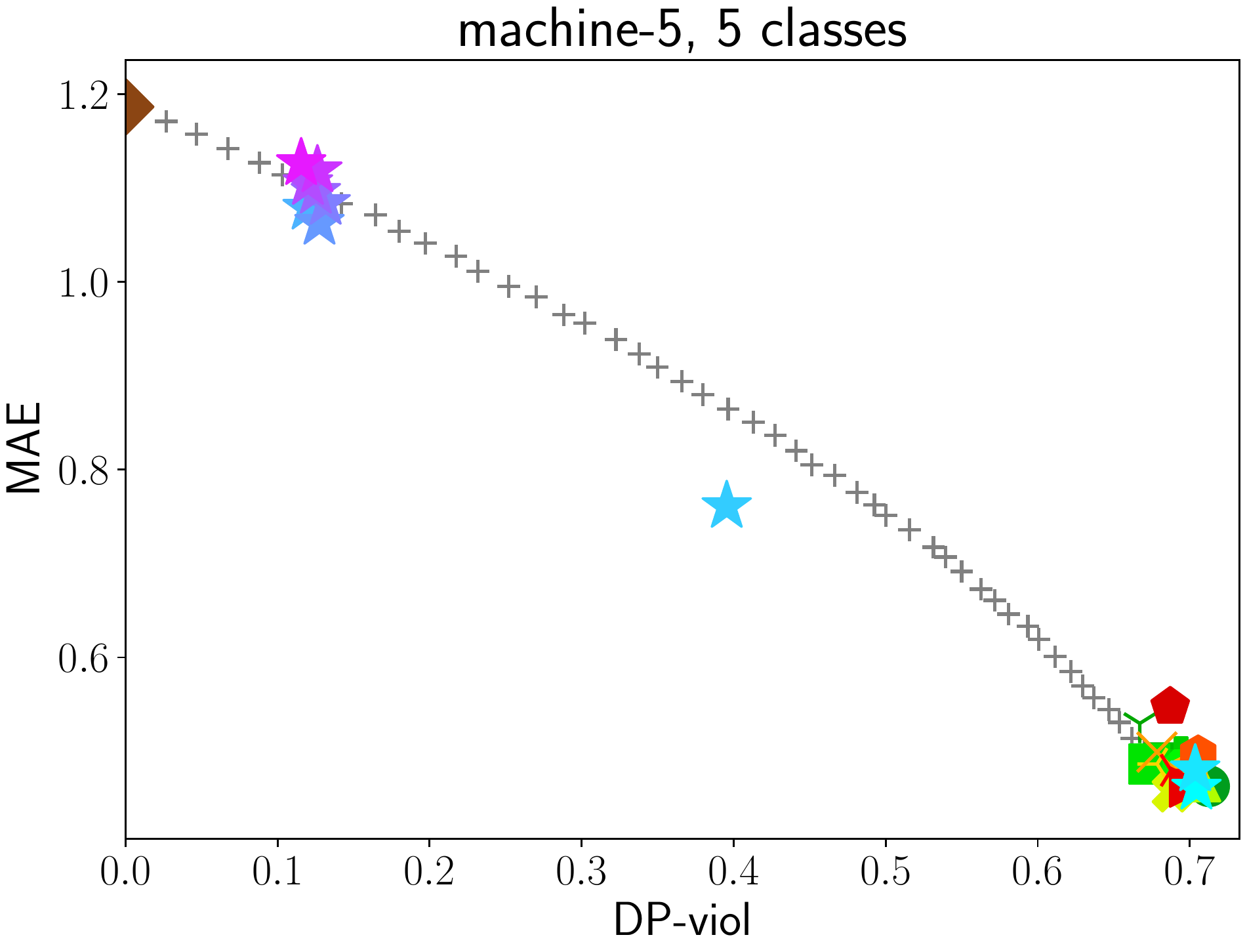}
    \hspace{\abstA}
    \includegraphics[scale=\scaleparameterA]{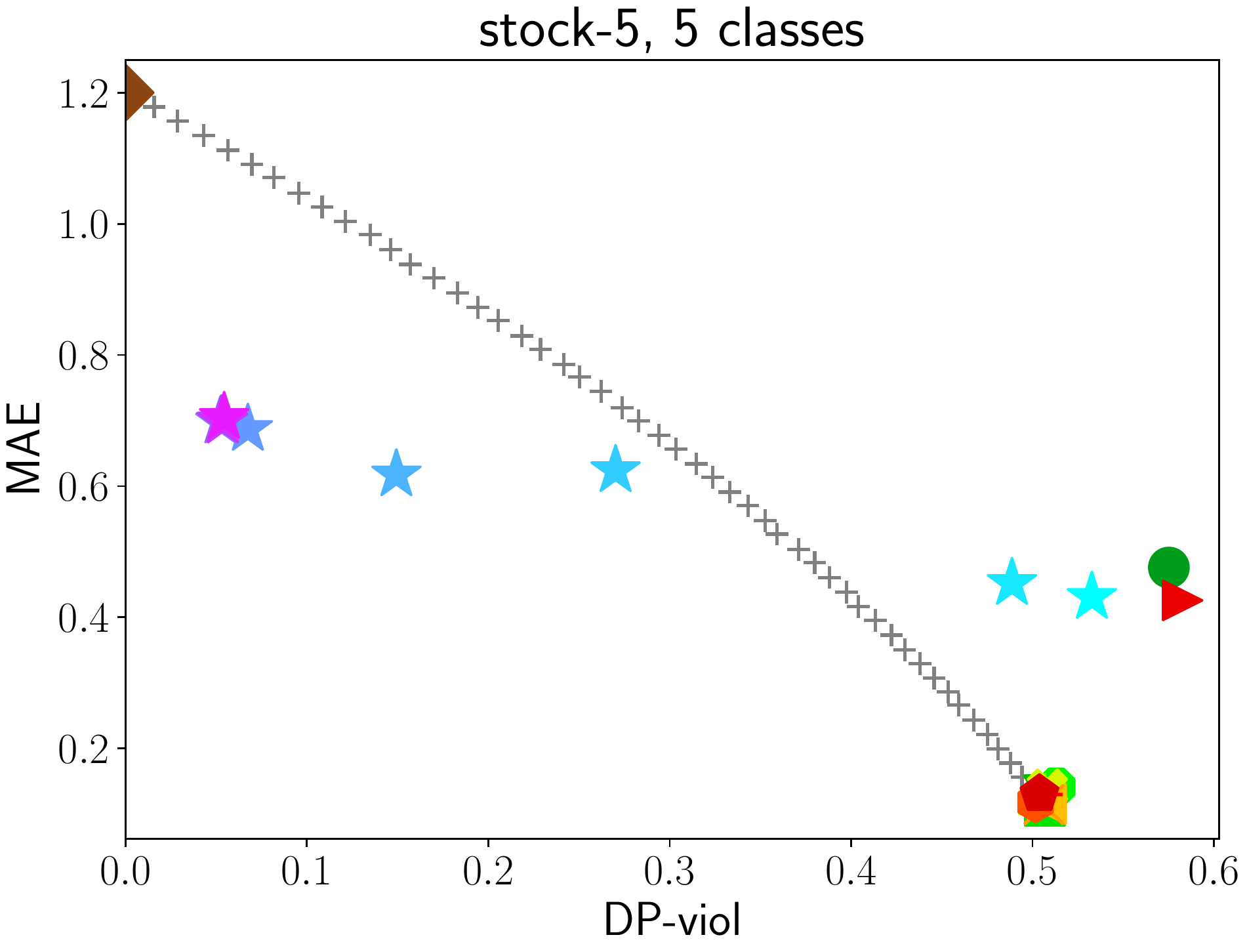}

    \caption{Experiments of Section~\ref{subsection_experiment_comparison} on the \textbf{discretized  regression datasets with 5 classes} when aiming for \textbf{pairwise DP}.}
    \label{fig:exp_comparison_APPENDIX_DISC_5classes_DP}
\end{figure*}

\begin{figure*}
    
    \includegraphics[width=\linewidth]{experiment_real_ord/legend_big.pdf}

    \includegraphics[scale=\scaleparameterA]{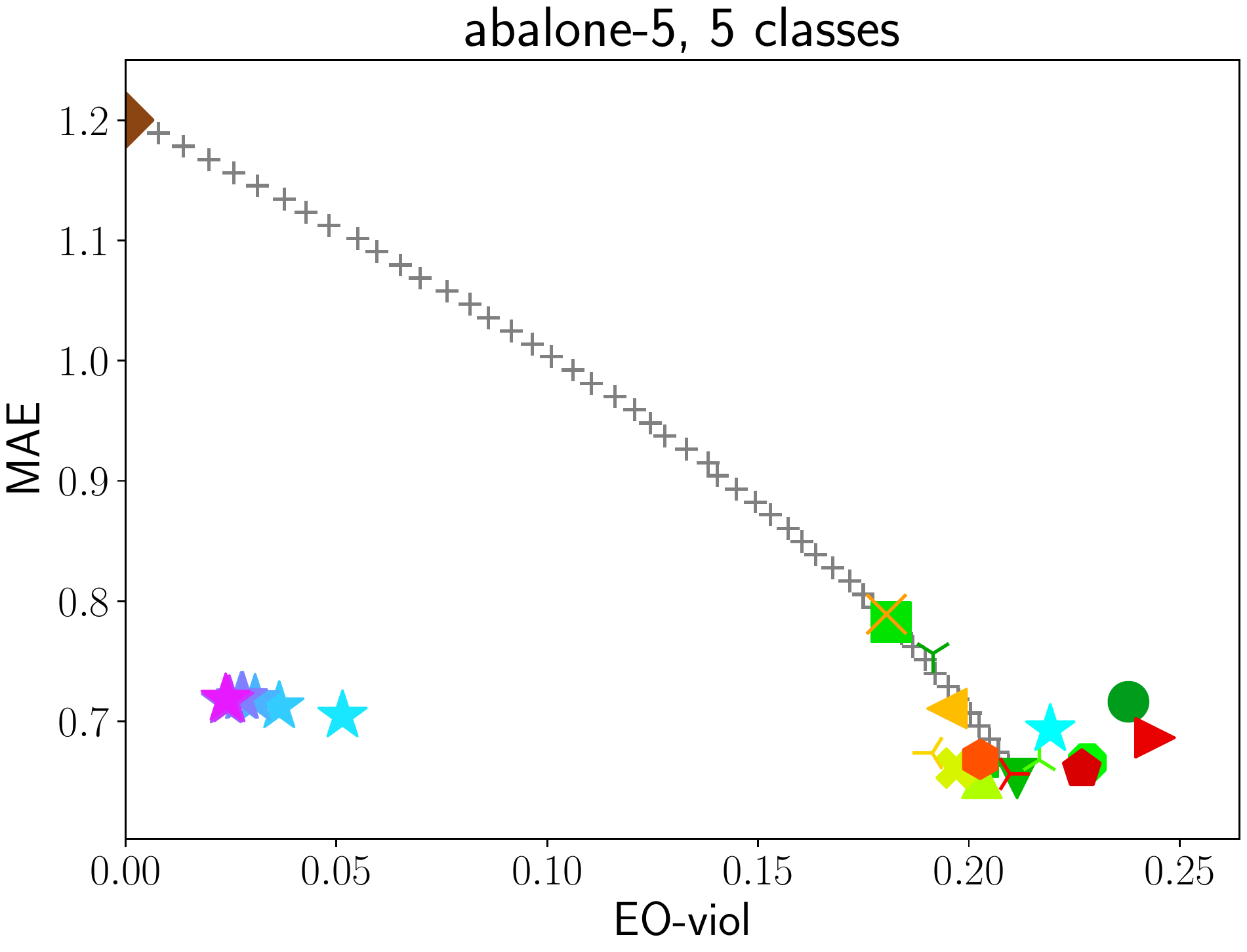}
    \hspace{\abstA}
    \includegraphics[scale=\scaleparameterA]{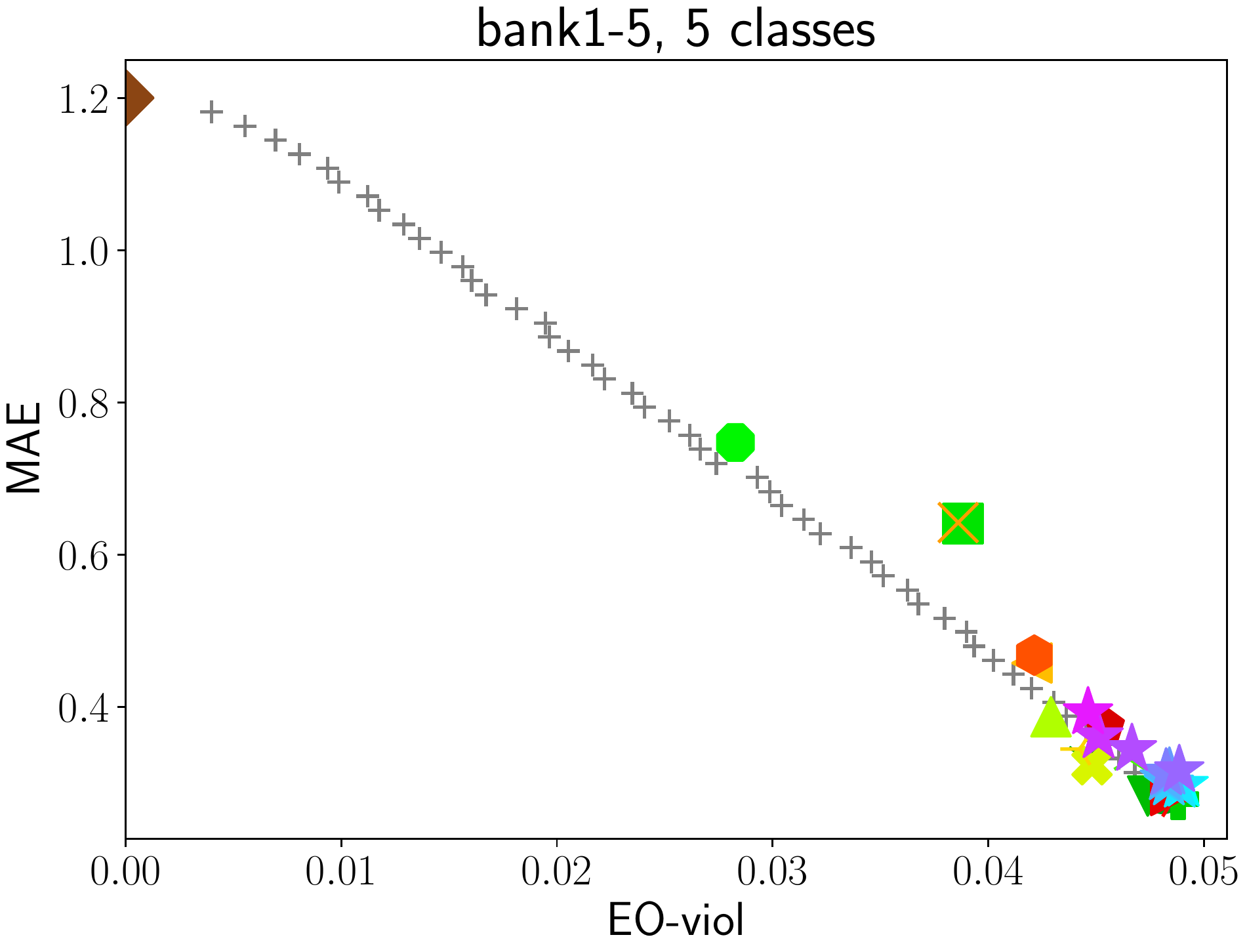}
    \hspace{\abstA}
    \includegraphics[scale=\scaleparameterA]{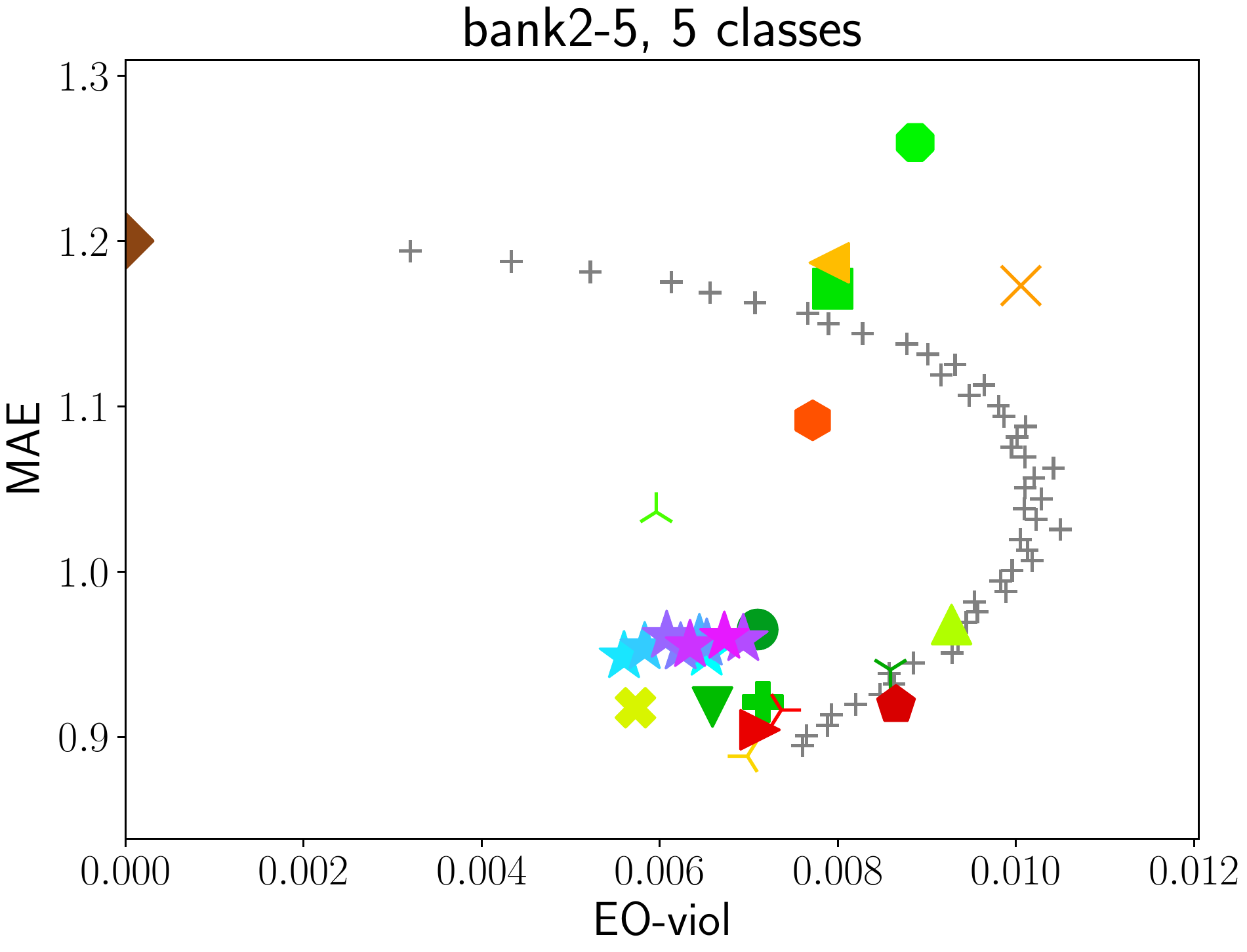}
    \hspace{\abstA}
    \includegraphics[scale=\scaleparameterA]{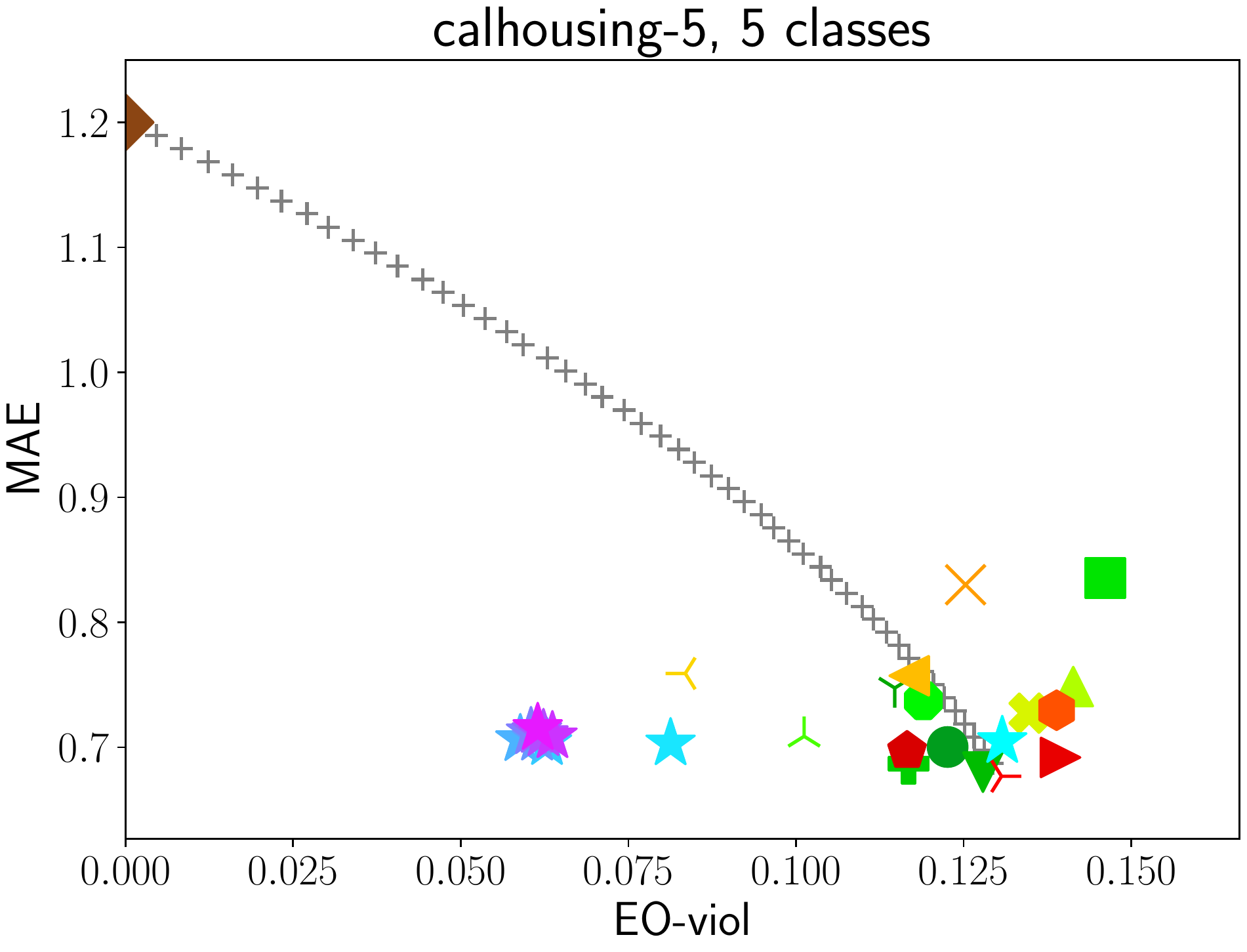}
    
    \includegraphics[scale=\scaleparameterA]{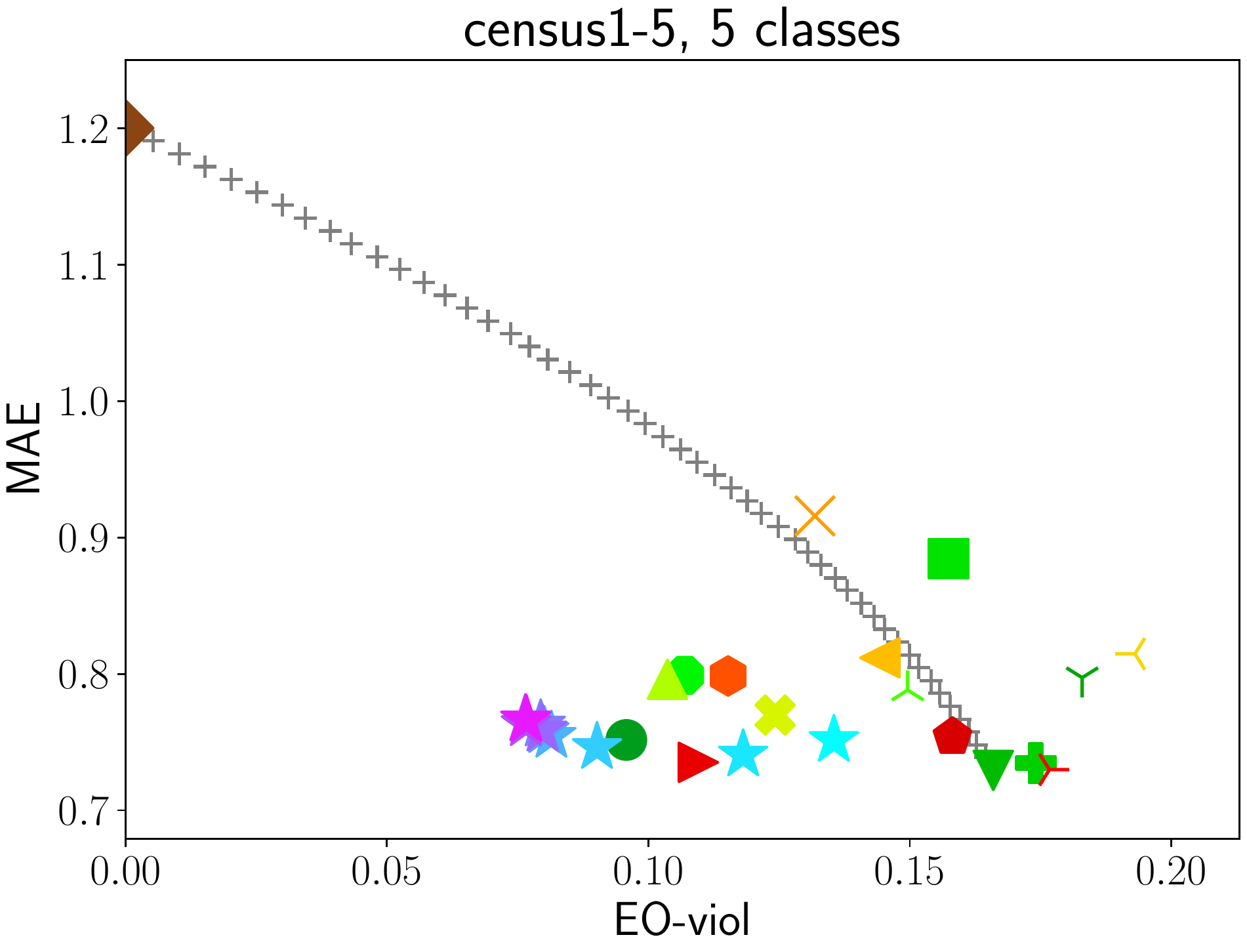}
    \hspace{\abstA}
    \includegraphics[scale=\scaleparameterA]{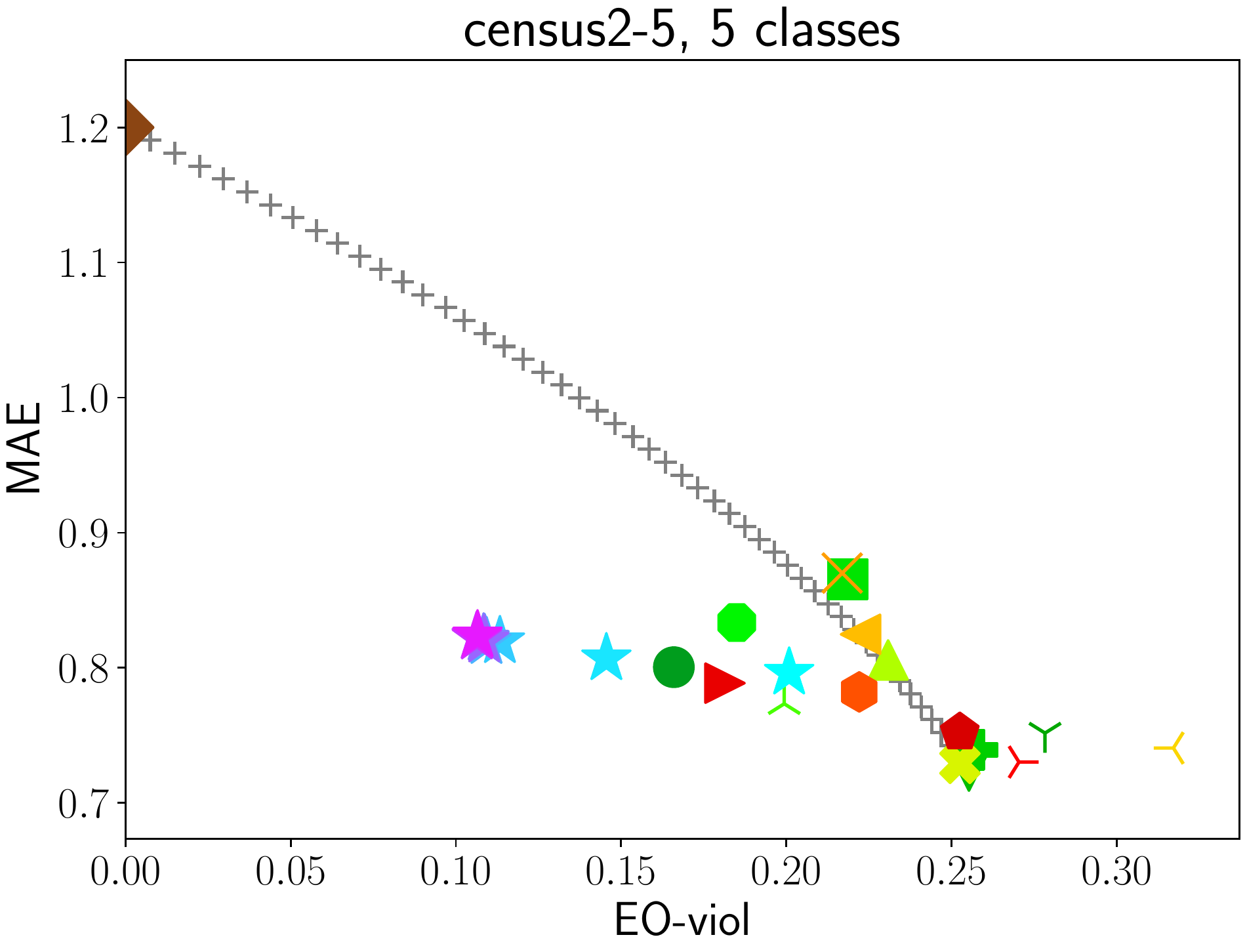}
    \hspace{\abstA}
    \includegraphics[scale=\scaleparameterA]{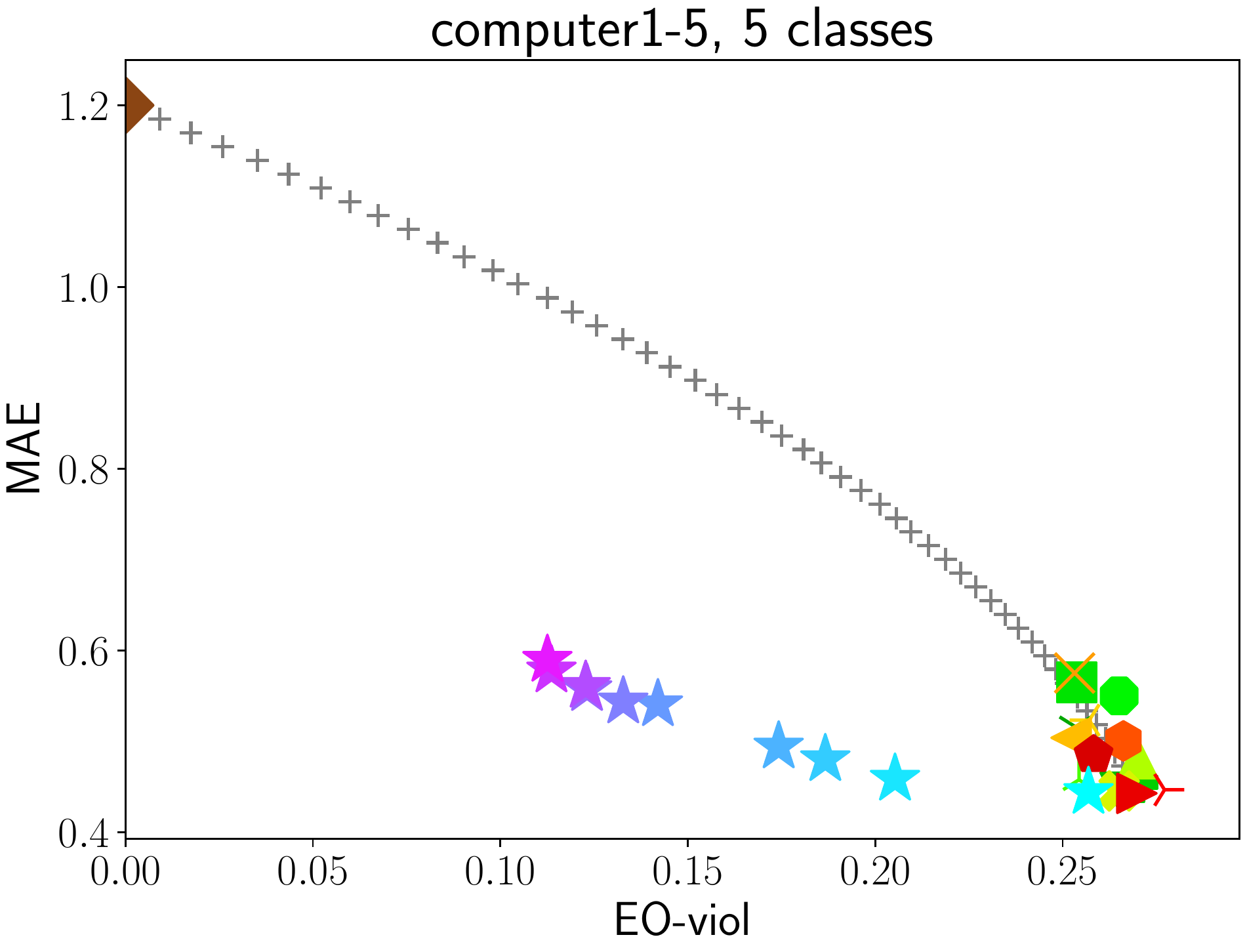}
    \hspace{\abstA}
    \includegraphics[scale=\scaleparameterA]{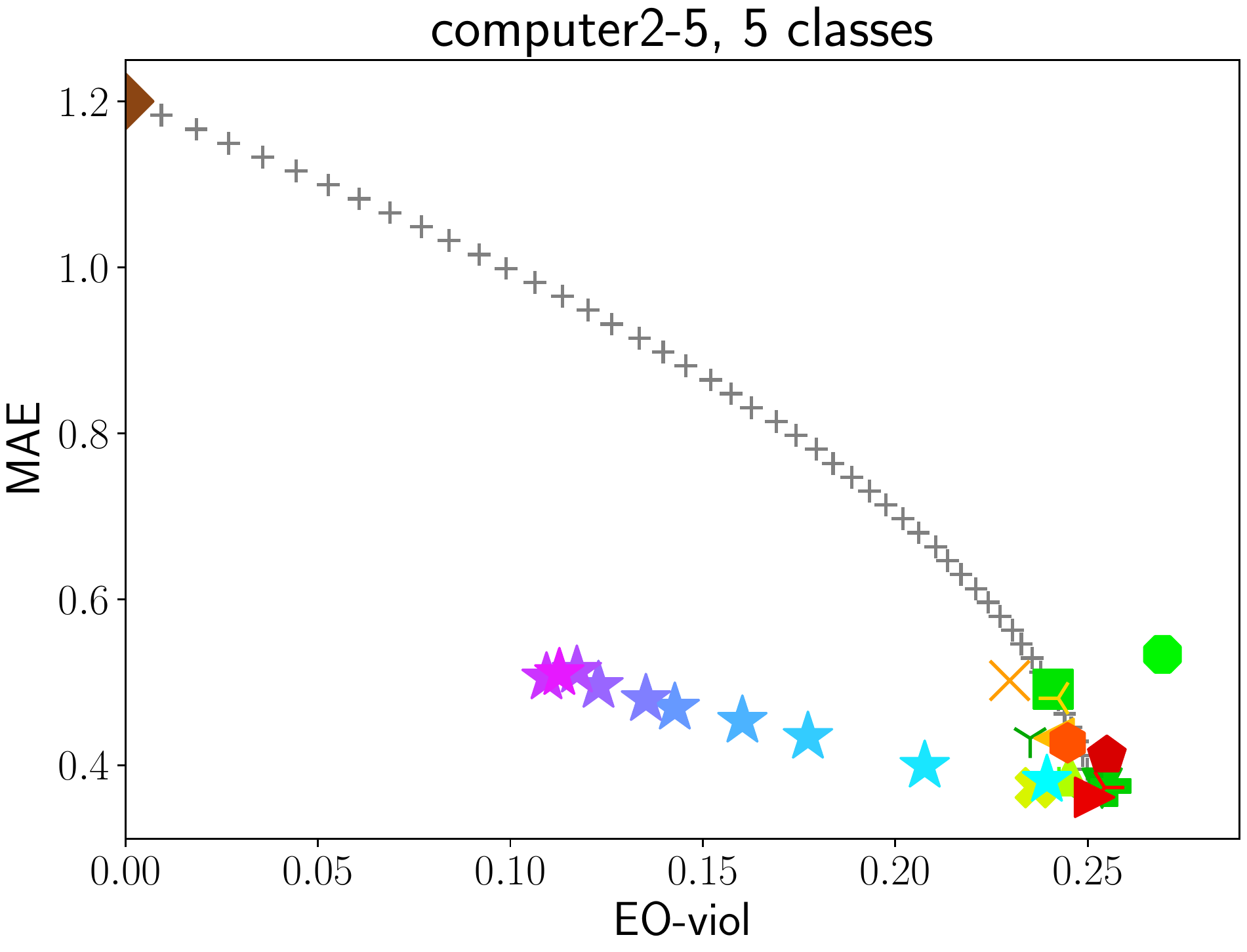}
    
    \includegraphics[scale=\scaleparameterA]{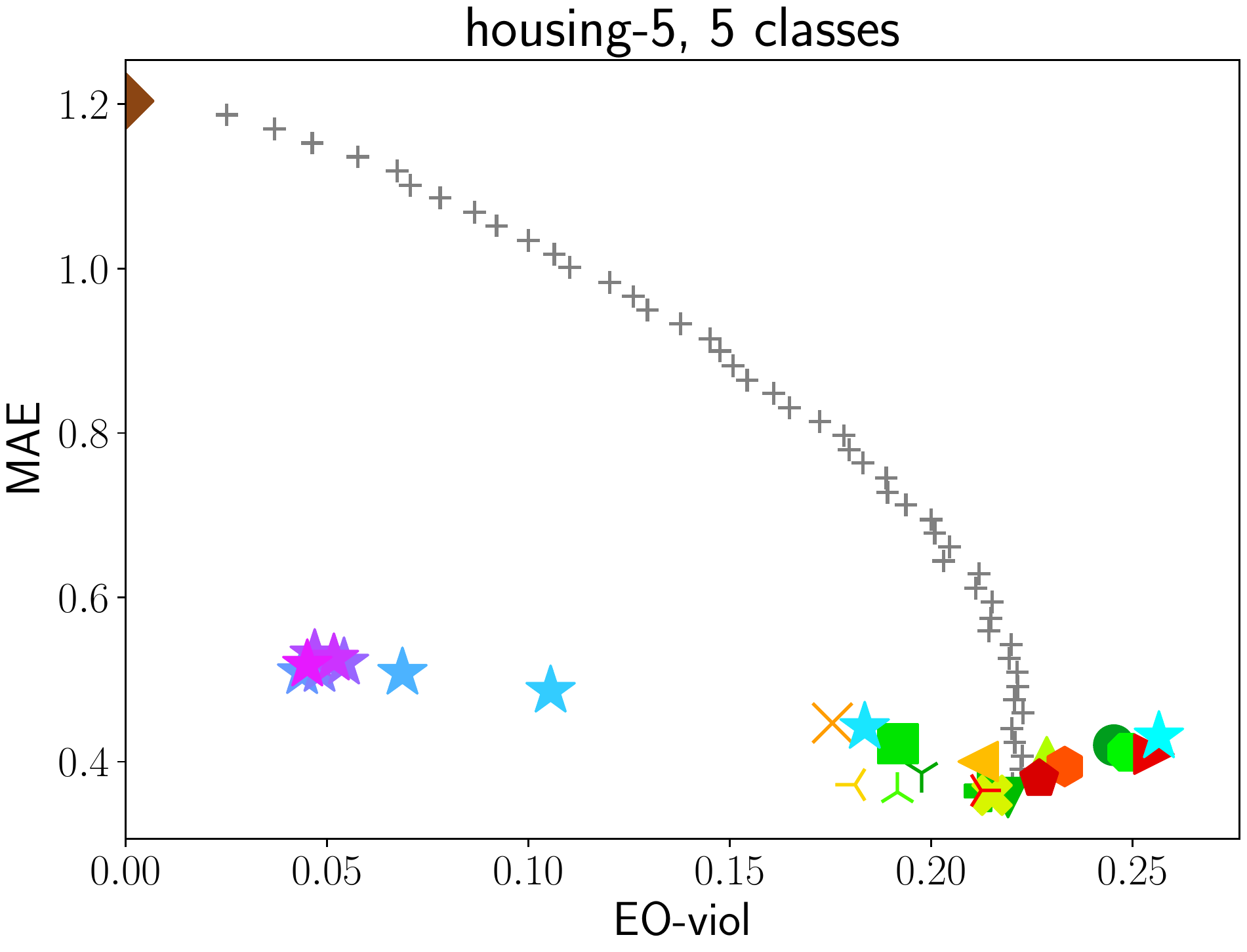}
    \hspace{\abstA}
    \includegraphics[scale=\scaleparameterA]{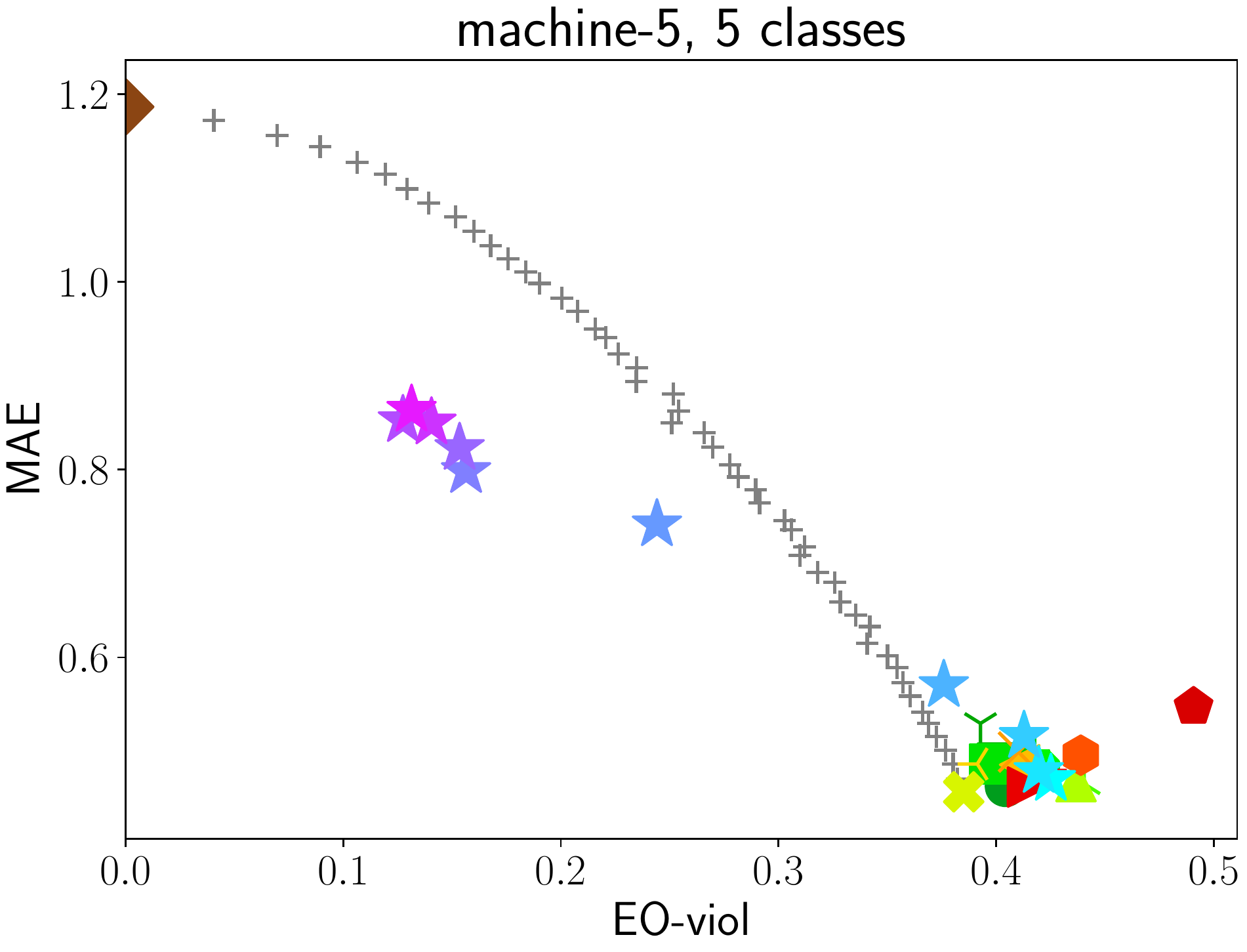}
    \hspace{\abstA}
    \includegraphics[scale=\scaleparameterA]{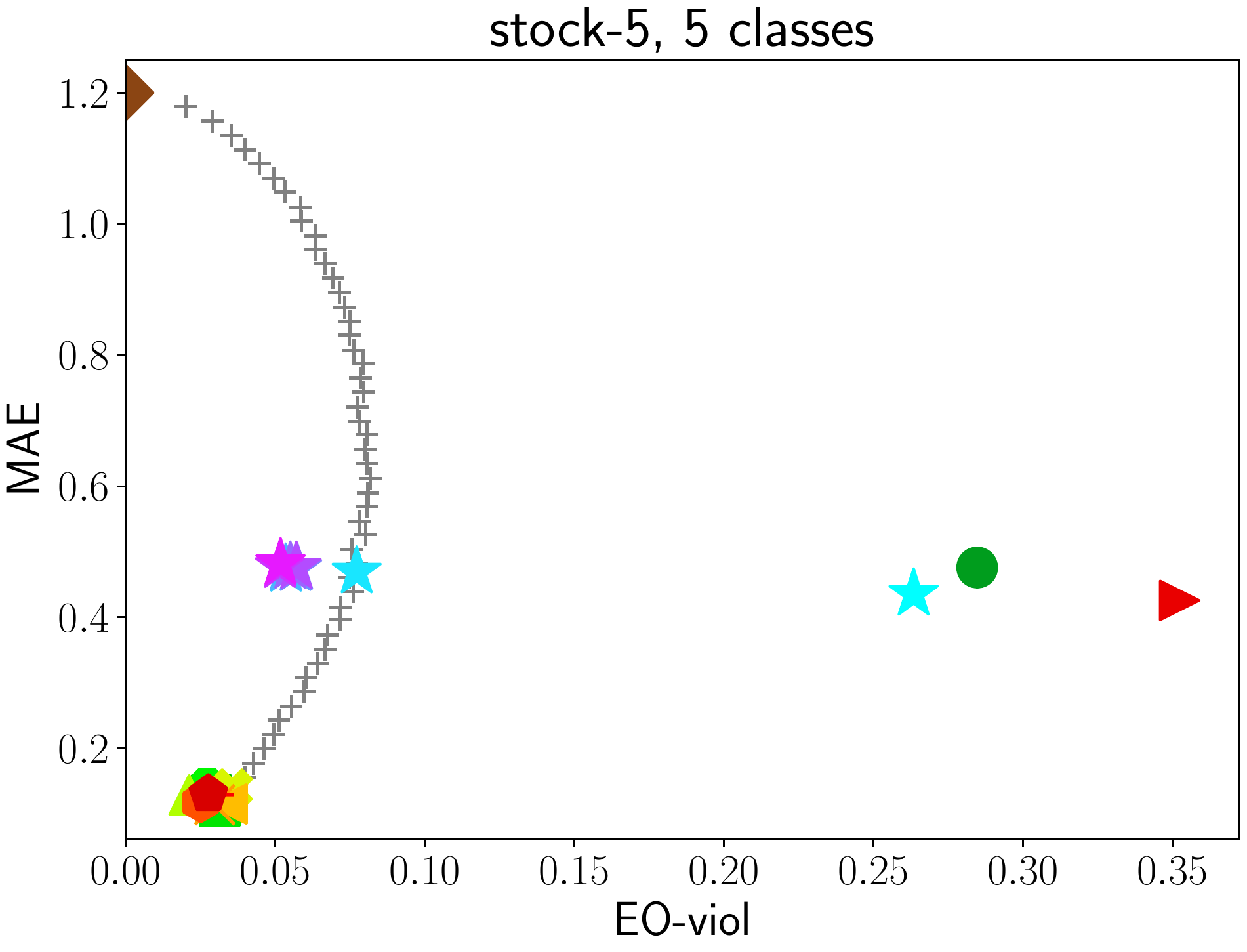}

    \caption{Experiments of Section~\ref{subsection_experiment_comparison} on the \textbf{discretized  regression datasets with 5 classes} when aiming for \textbf{pairwise EO}.}
    \label{fig:exp_comparison_APPENDIX_DISC_5classes_EO}
\end{figure*}

\begin{figure*}
    
    \includegraphics[width=\linewidth]{experiment_real_ord/legend_big.pdf}

    \includegraphics[scale=\scaleparameterA]{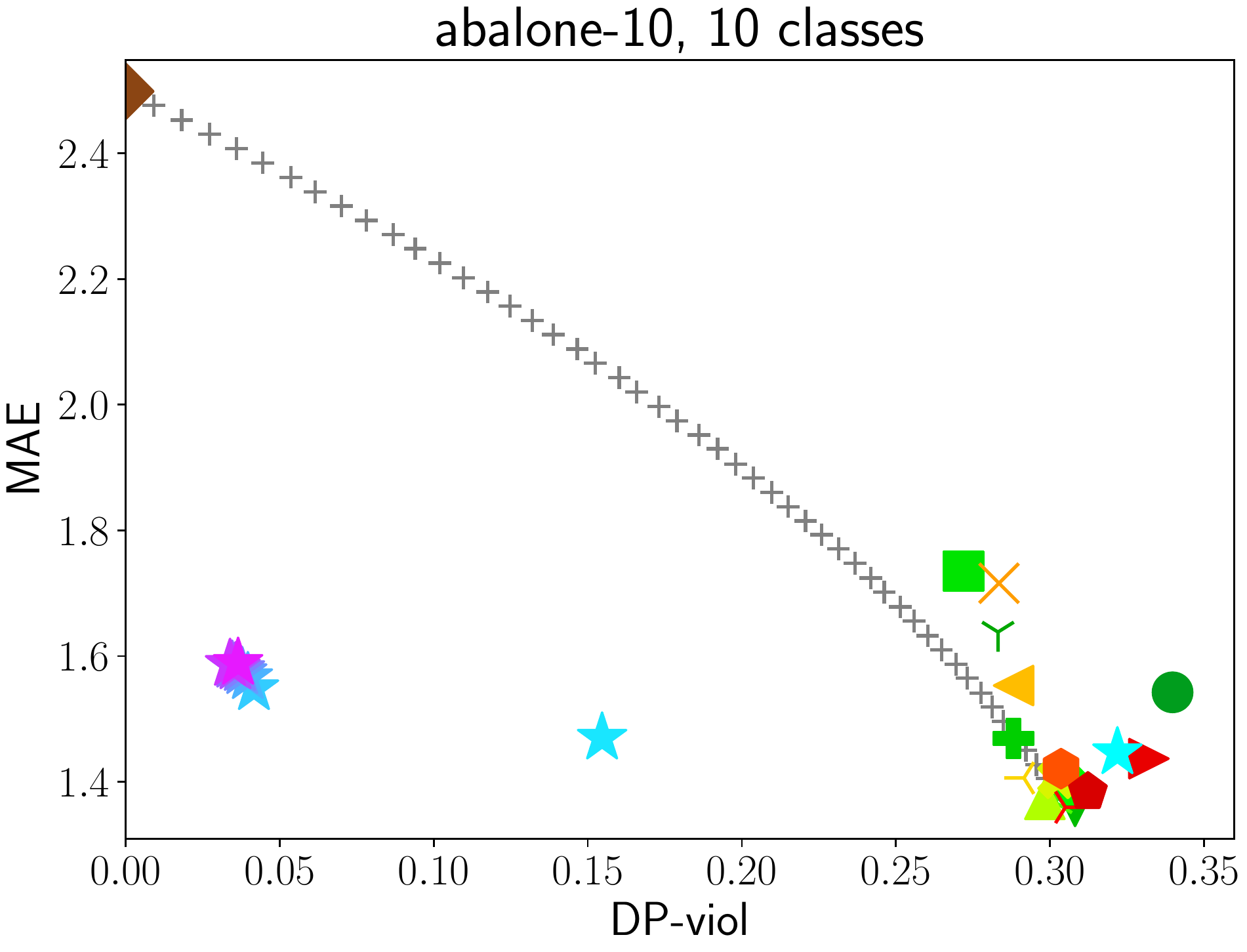}
    \hspace{\abstA}
    \includegraphics[scale=\scaleparameterA]{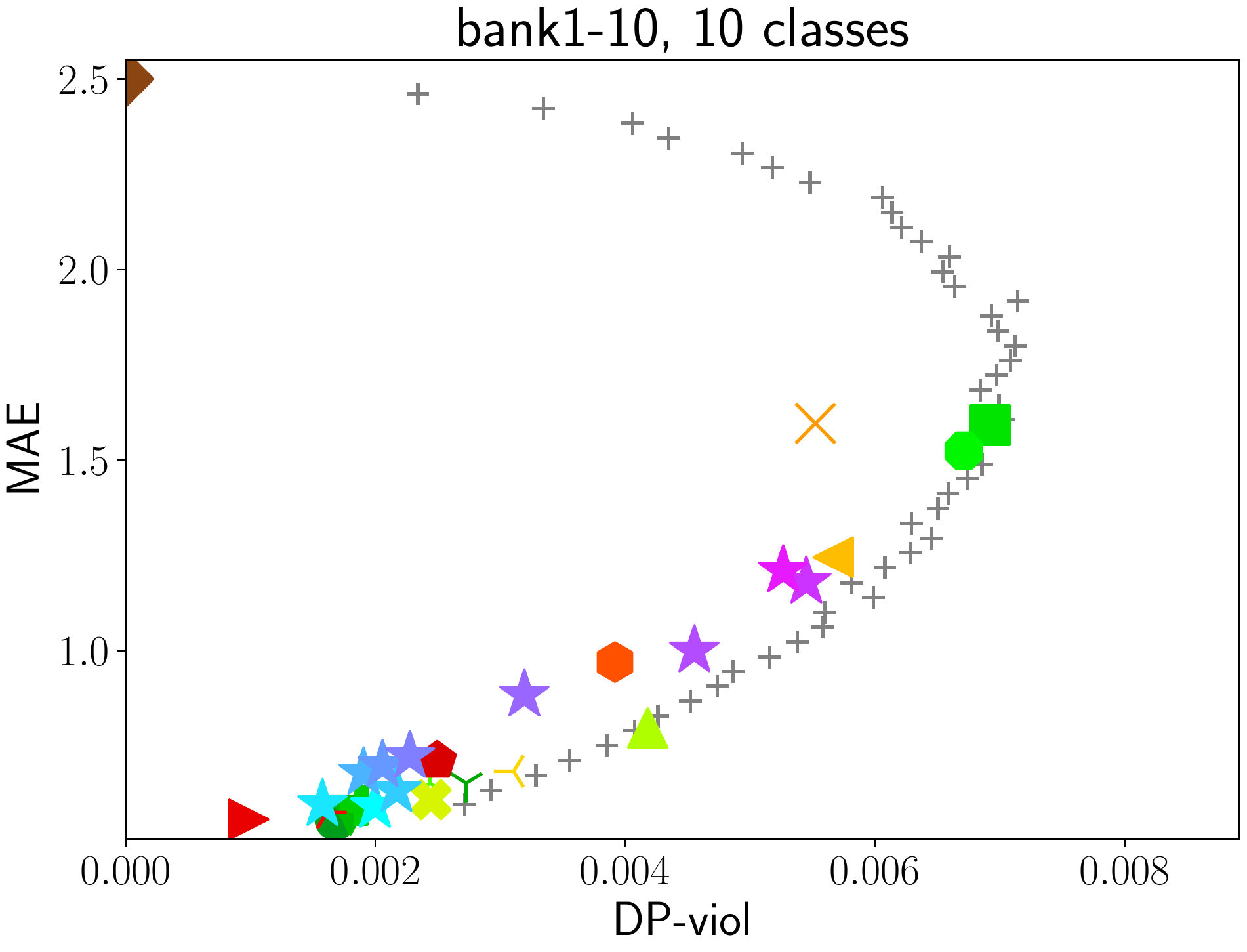}
    \hspace{\abstA}
    \includegraphics[scale=\scaleparameterA]{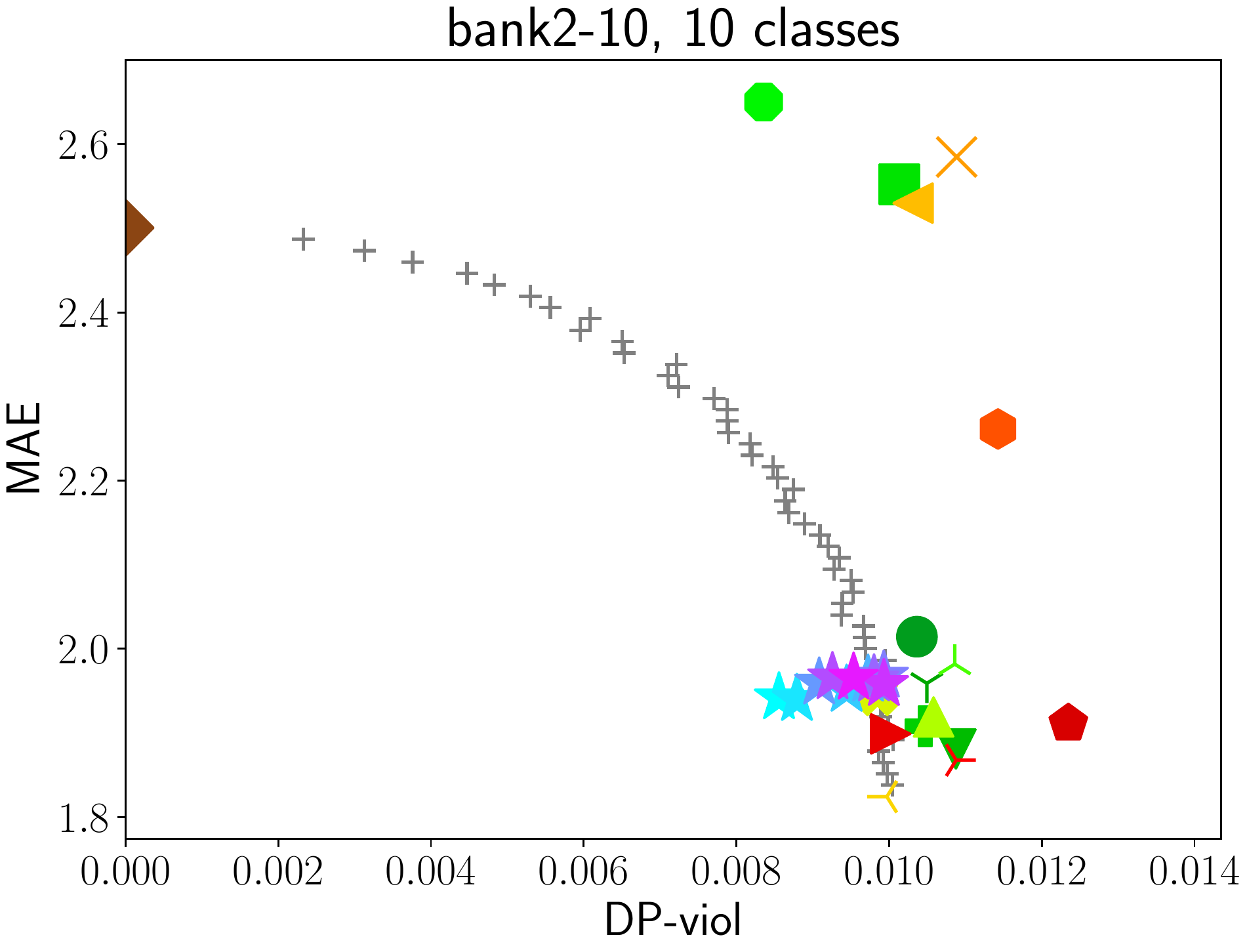}
    \hspace{\abstA}
    \includegraphics[scale=\scaleparameterA]{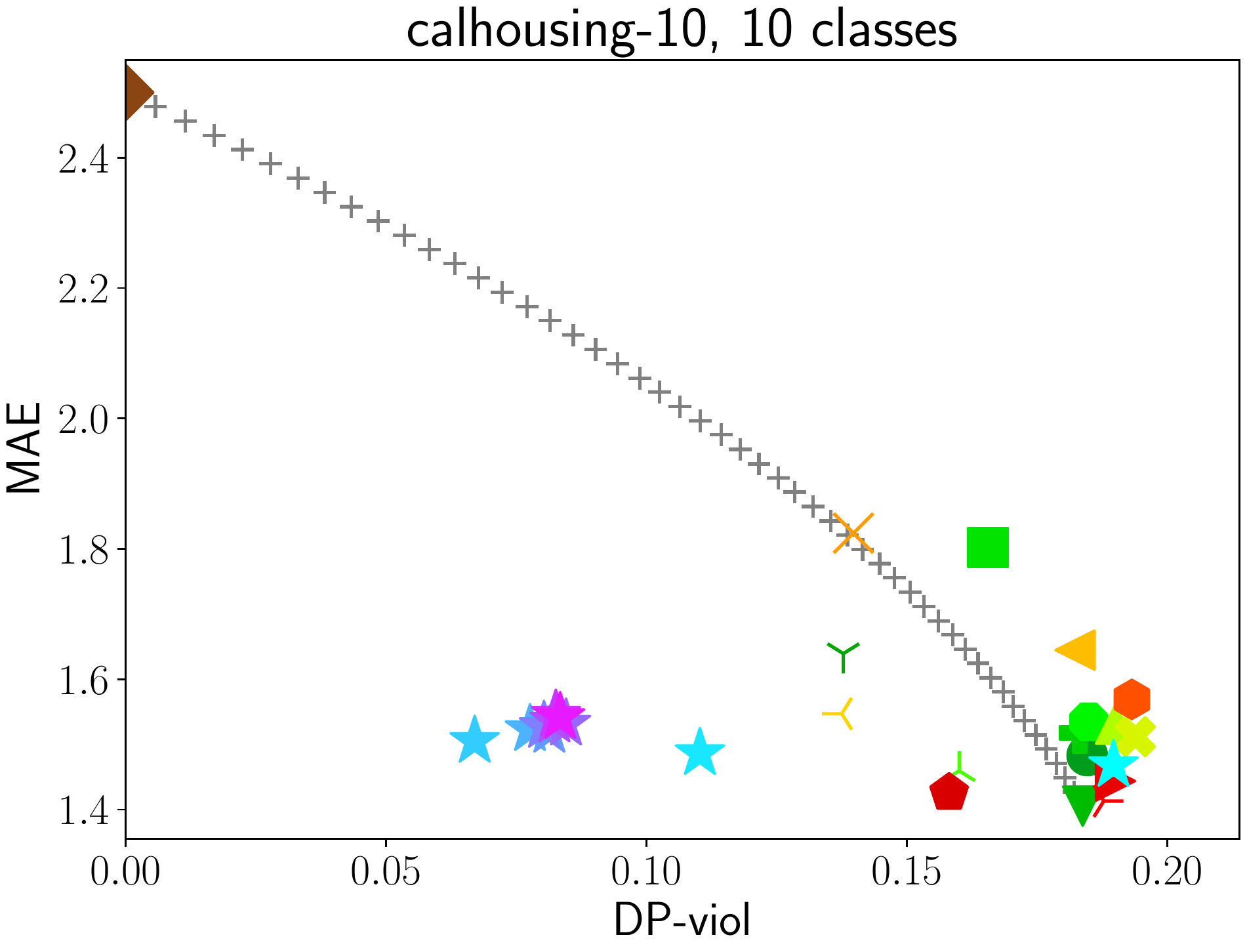}
    
    \includegraphics[scale=\scaleparameterA]{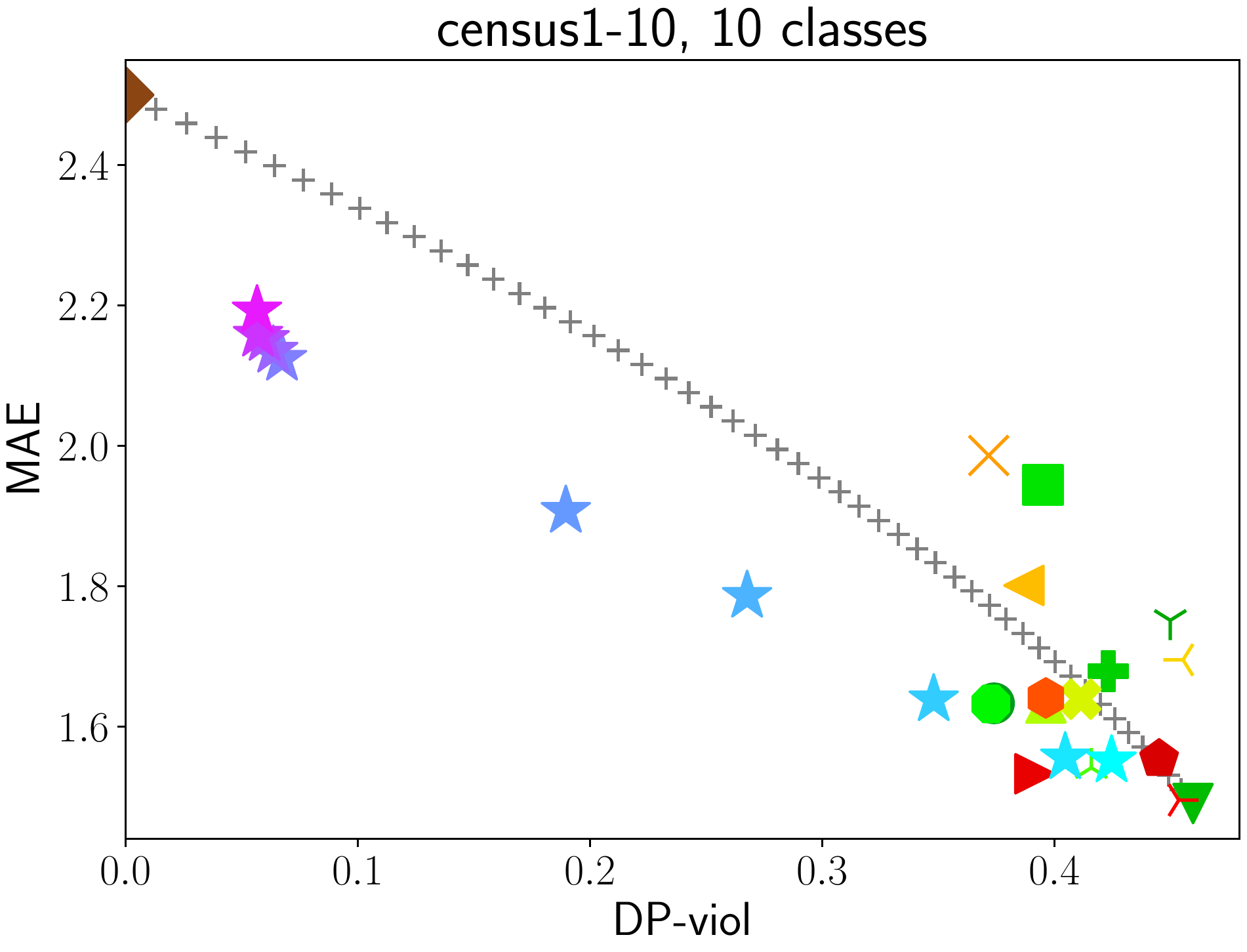}
    \hspace{\abstA}
    \includegraphics[scale=\scaleparameterA]{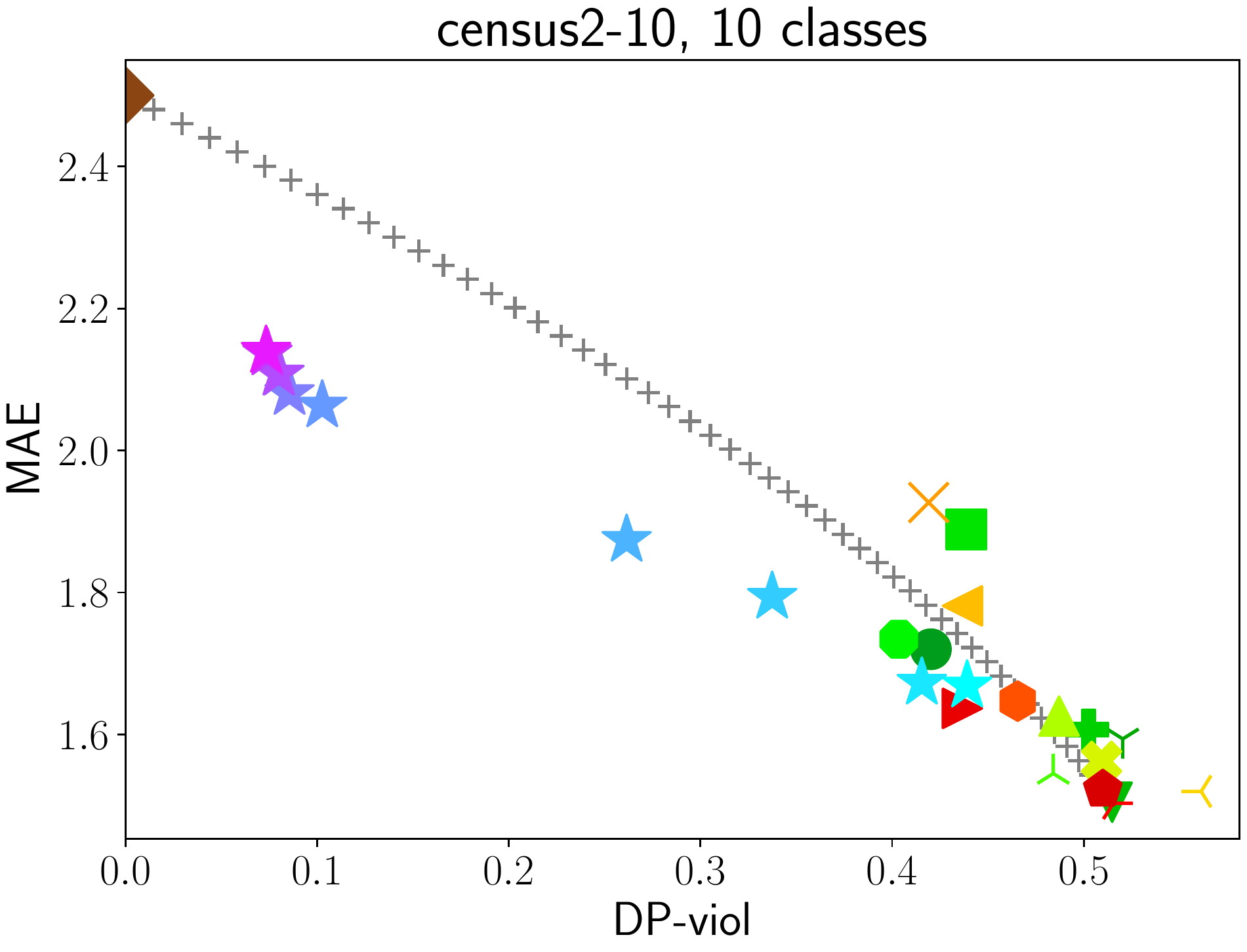}
    \hspace{\abstA}
    \includegraphics[scale=\scaleparameterA]{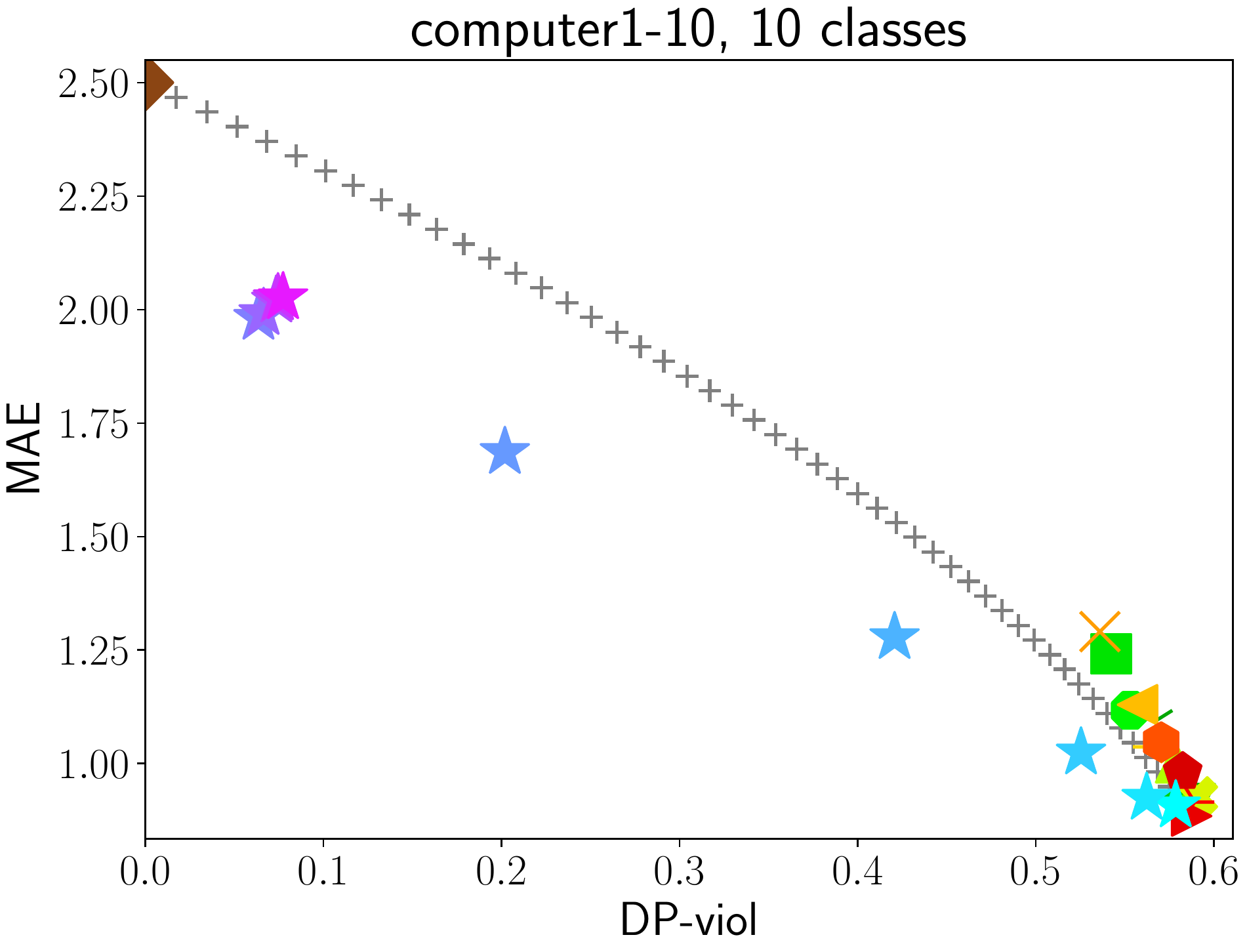}
    \hspace{\abstA}
    \includegraphics[scale=\scaleparameterA]{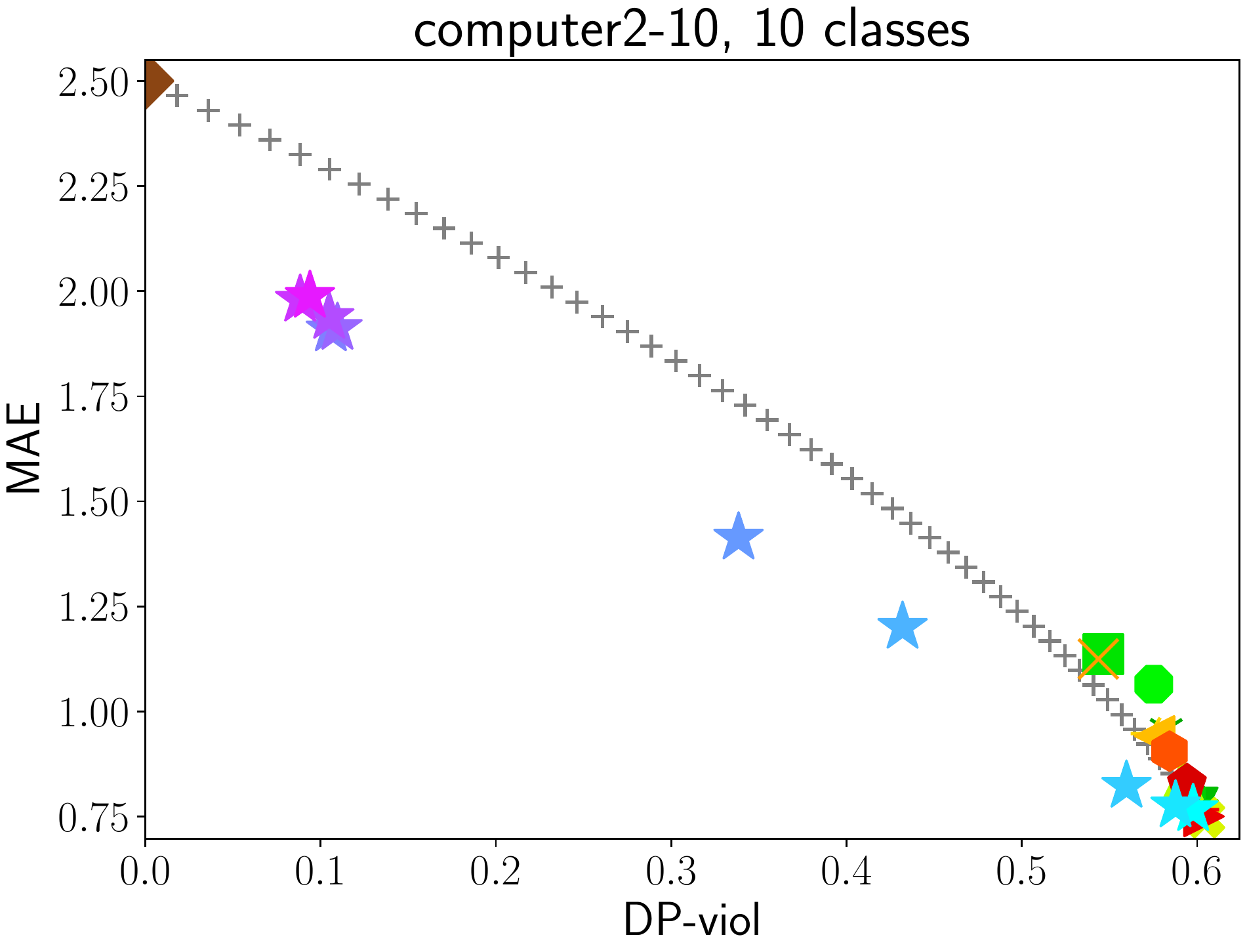}
    
    \includegraphics[scale=\scaleparameterA]{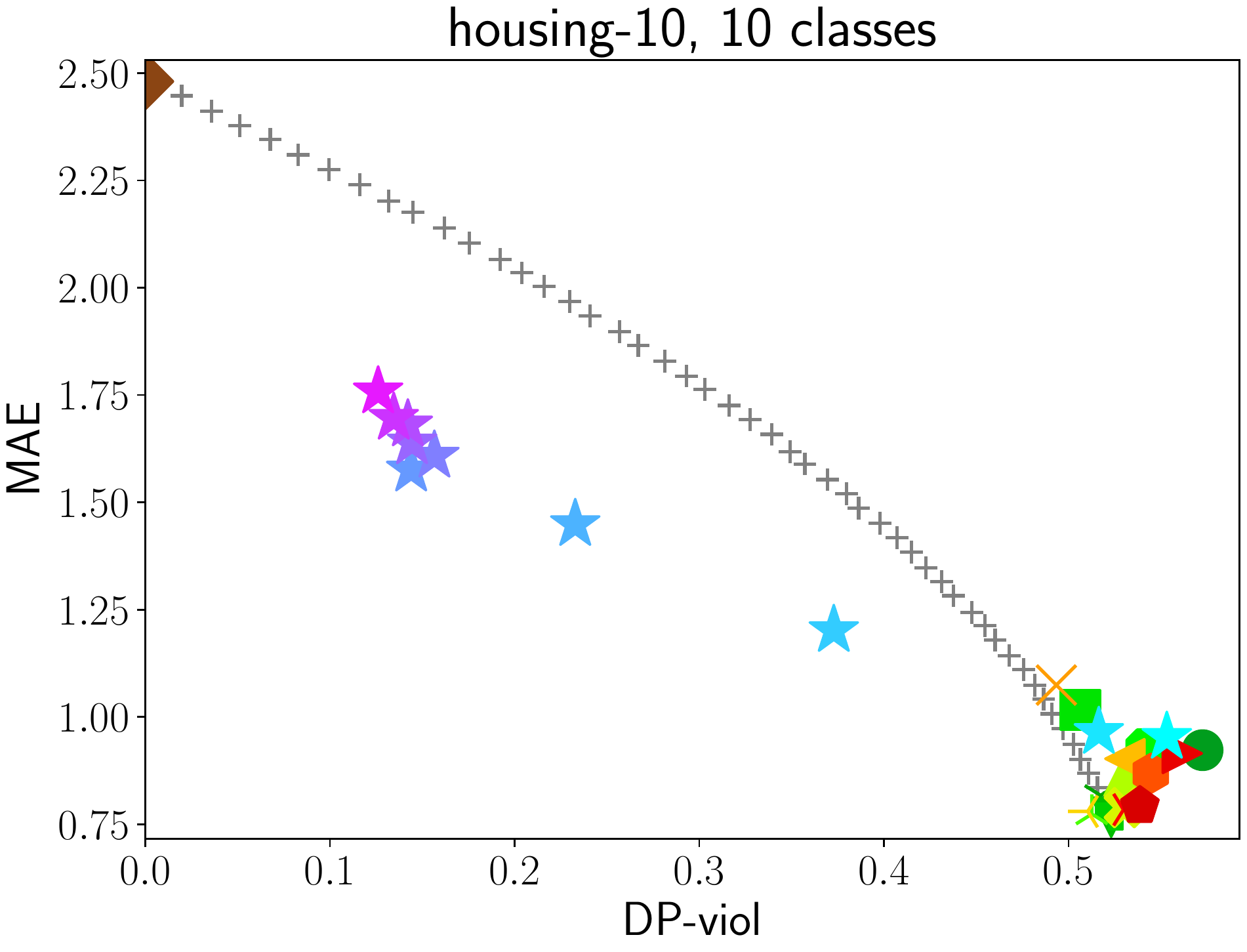}
    \hspace{\abstA}
    \includegraphics[scale=\scaleparameterA]{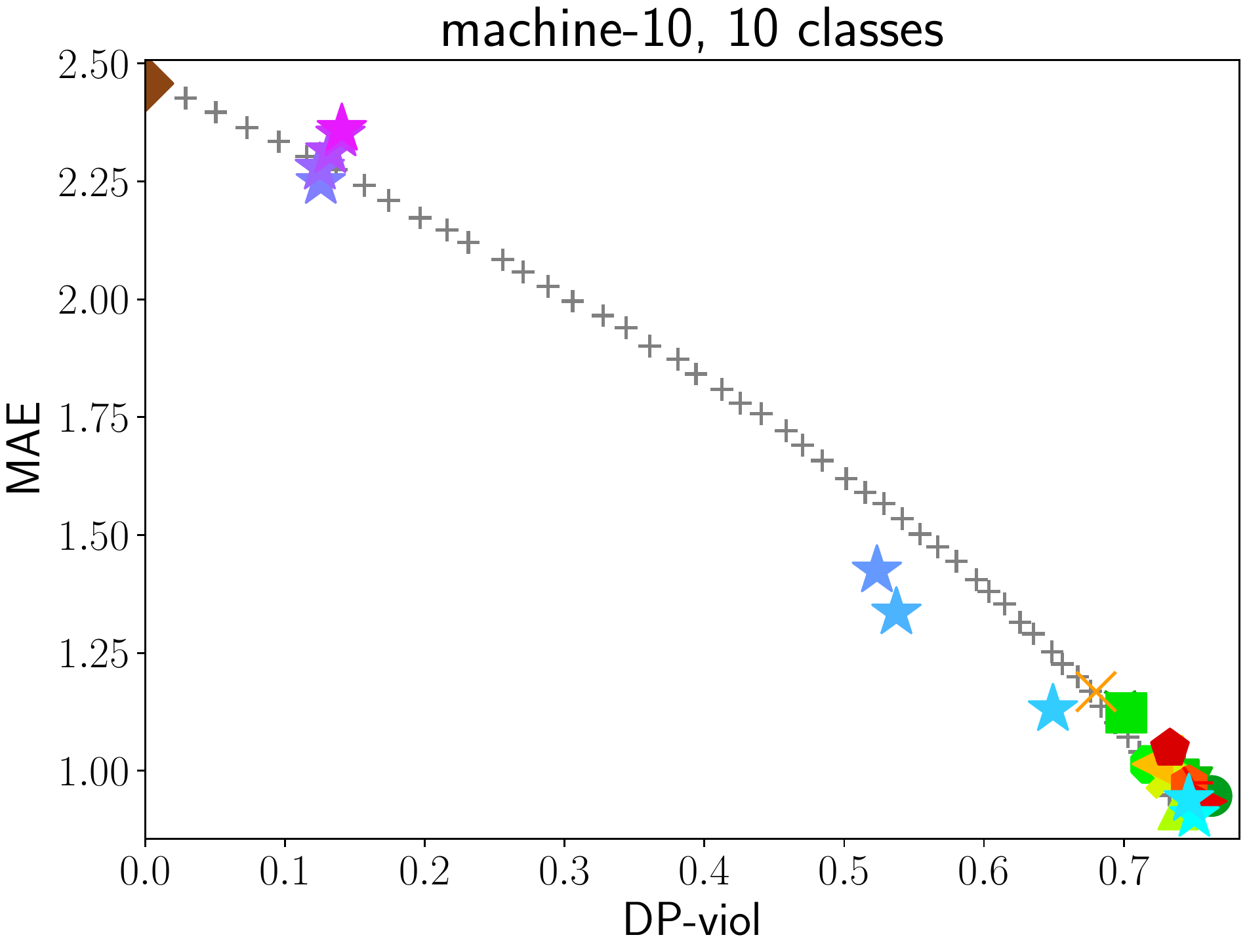}
    \hspace{\abstA}
    \includegraphics[scale=\scaleparameterA]{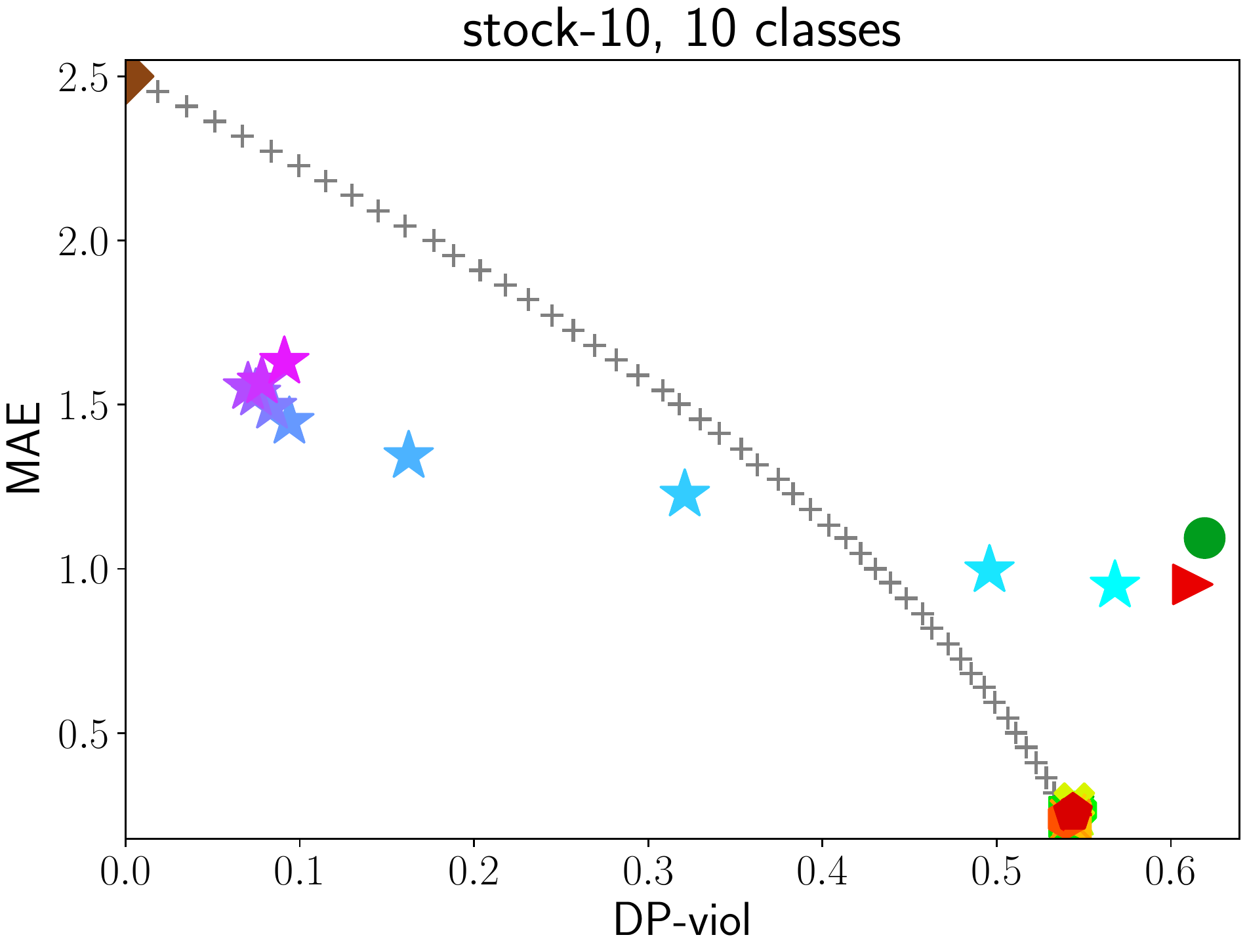}

    \caption{Experiments of Section~\ref{subsection_experiment_comparison} on the \textbf{discretized  regression datasets with 10 classes} when aiming for \textbf{pairwise~DP}.}
    \label{fig:exp_comparison_APPENDIX_DISC_10classes_DP}
\end{figure*}

\begin{figure*}
    
    \includegraphics[width=\linewidth]{experiment_real_ord/legend_big.pdf}

    \includegraphics[scale=\scaleparameterA]{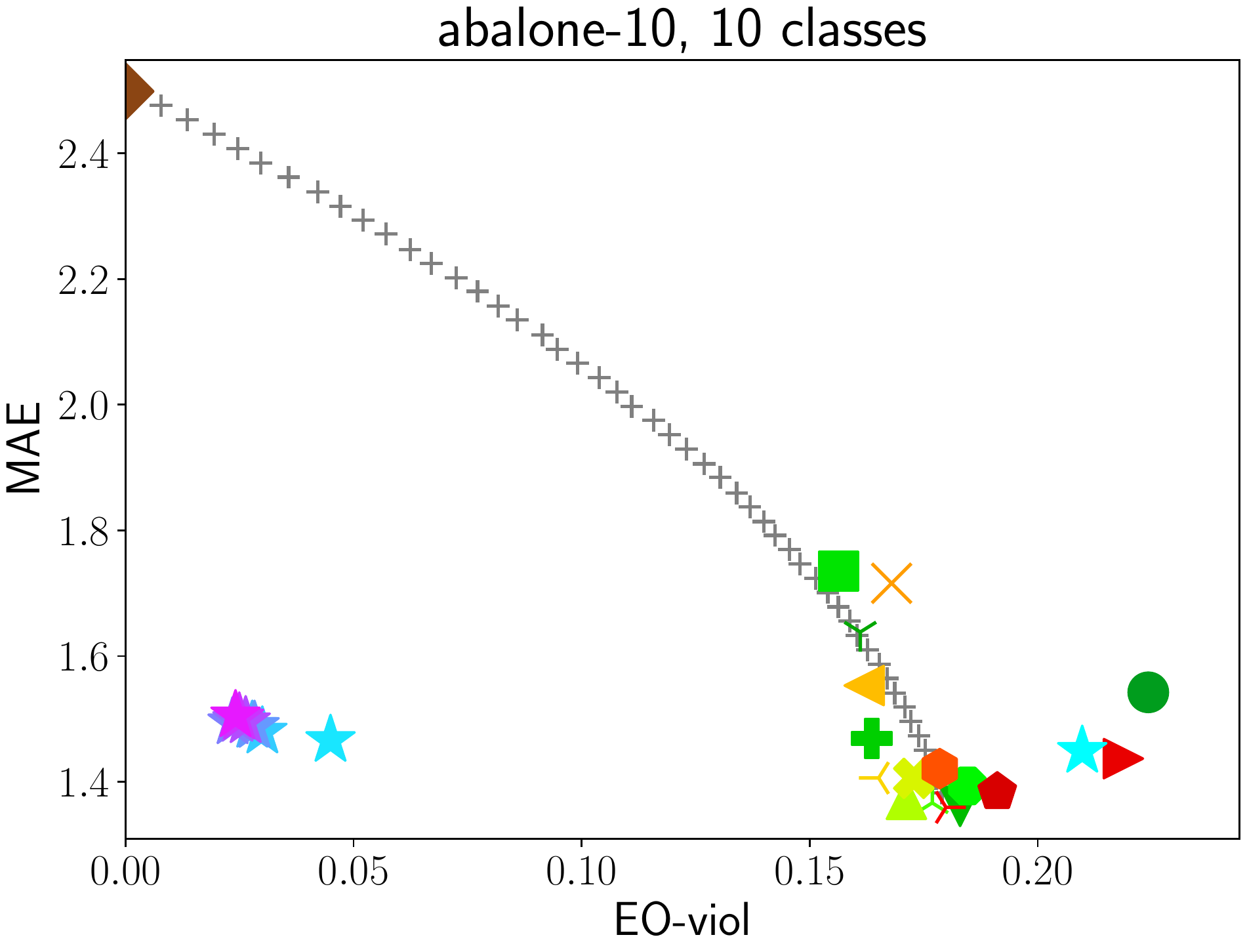}
    \hspace{\abstA}
    \includegraphics[scale=\scaleparameterA]{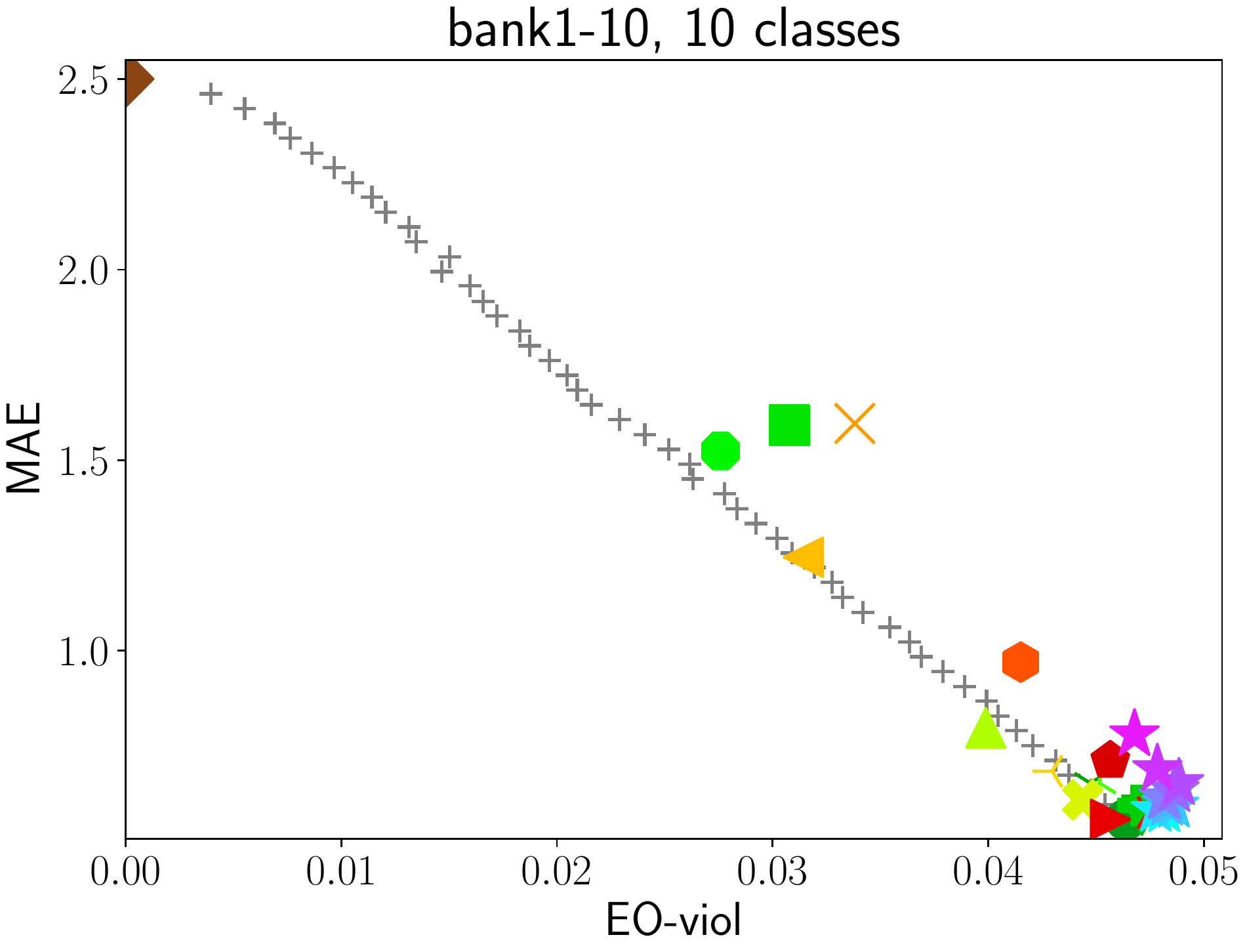}
    \hspace{\abstA}
    \includegraphics[scale=\scaleparameterA]{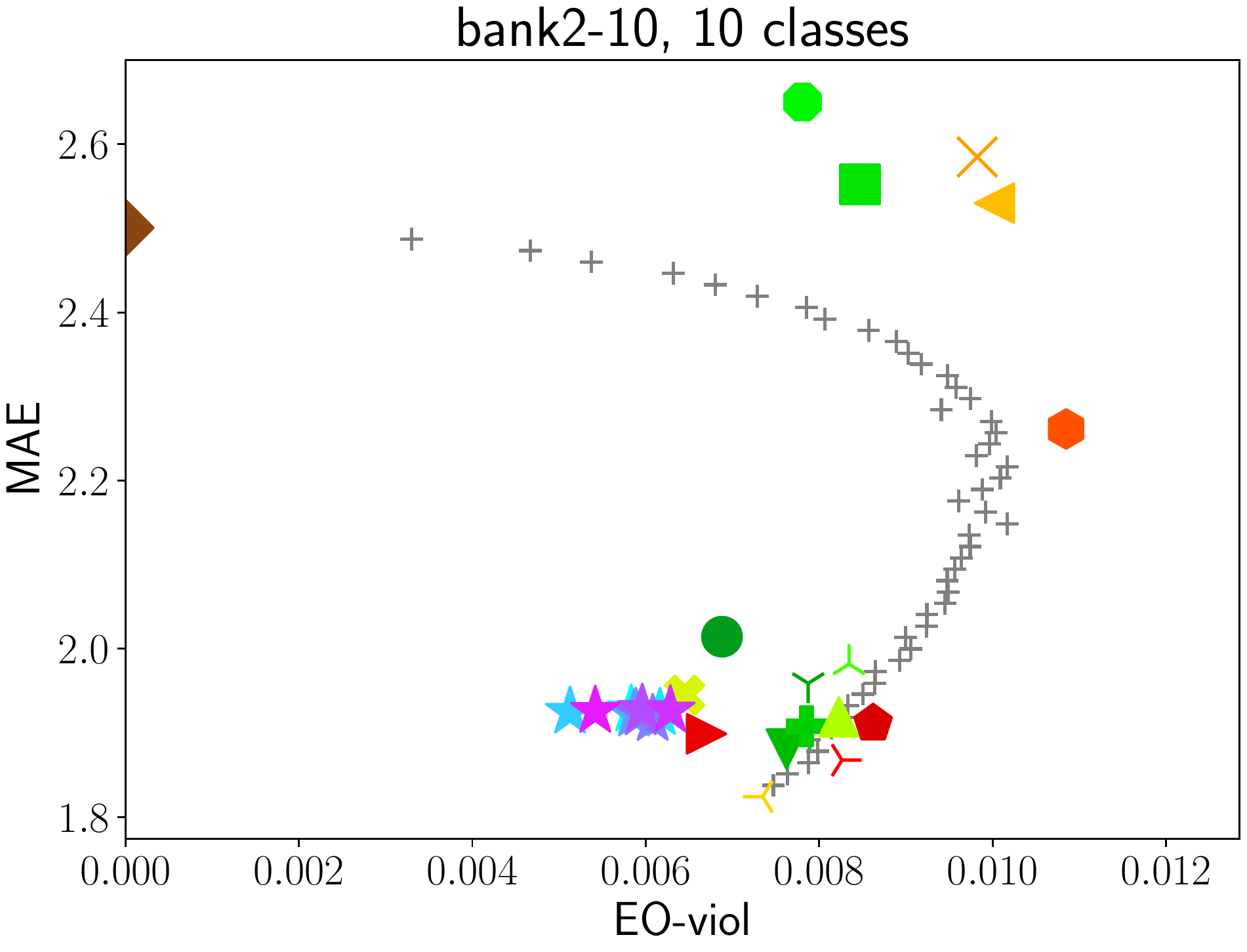}
    \hspace{\abstA}
    \includegraphics[scale=\scaleparameterA]{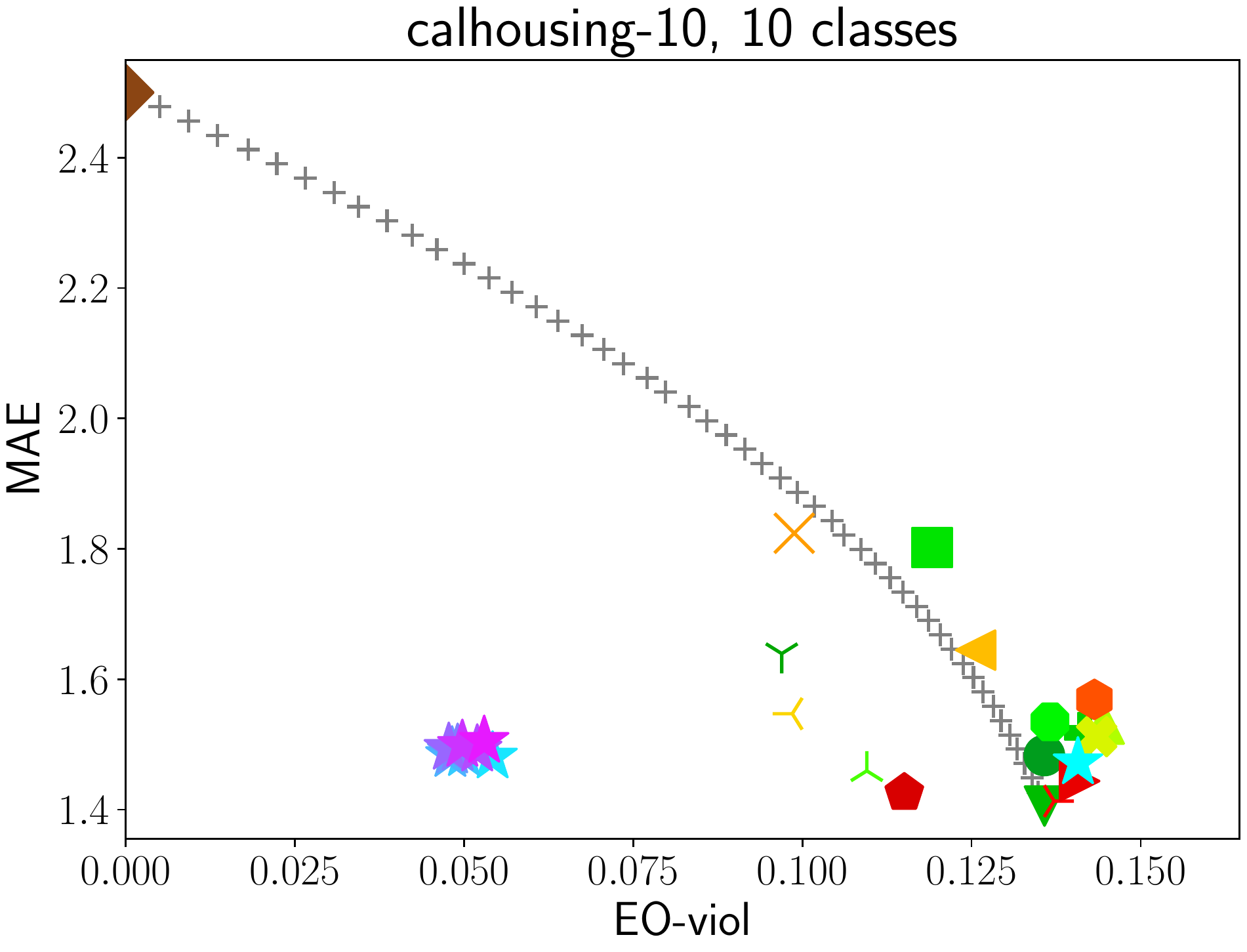}
    
    \includegraphics[scale=\scaleparameterA]{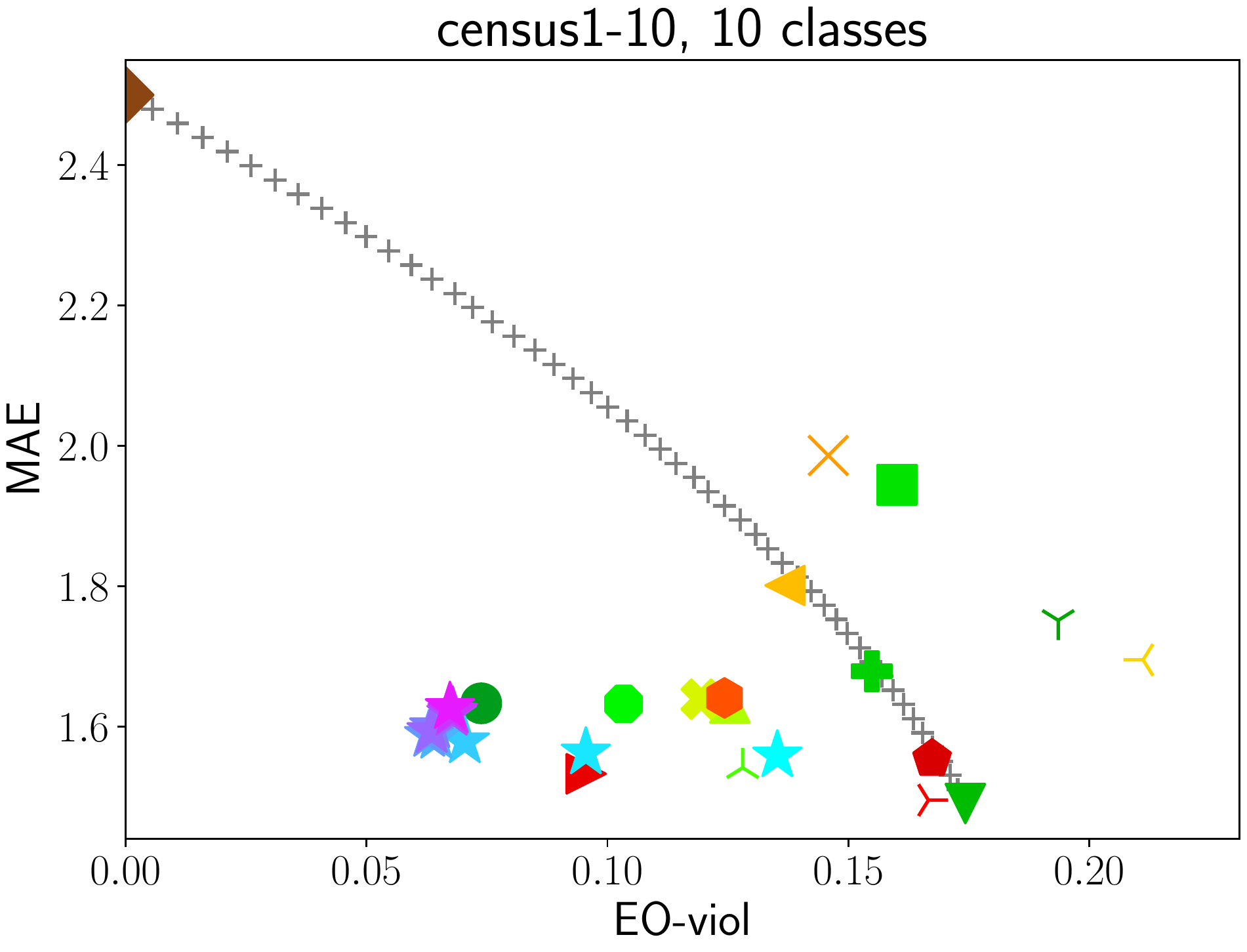}
    \hspace{\abstA}
    \includegraphics[scale=\scaleparameterA]{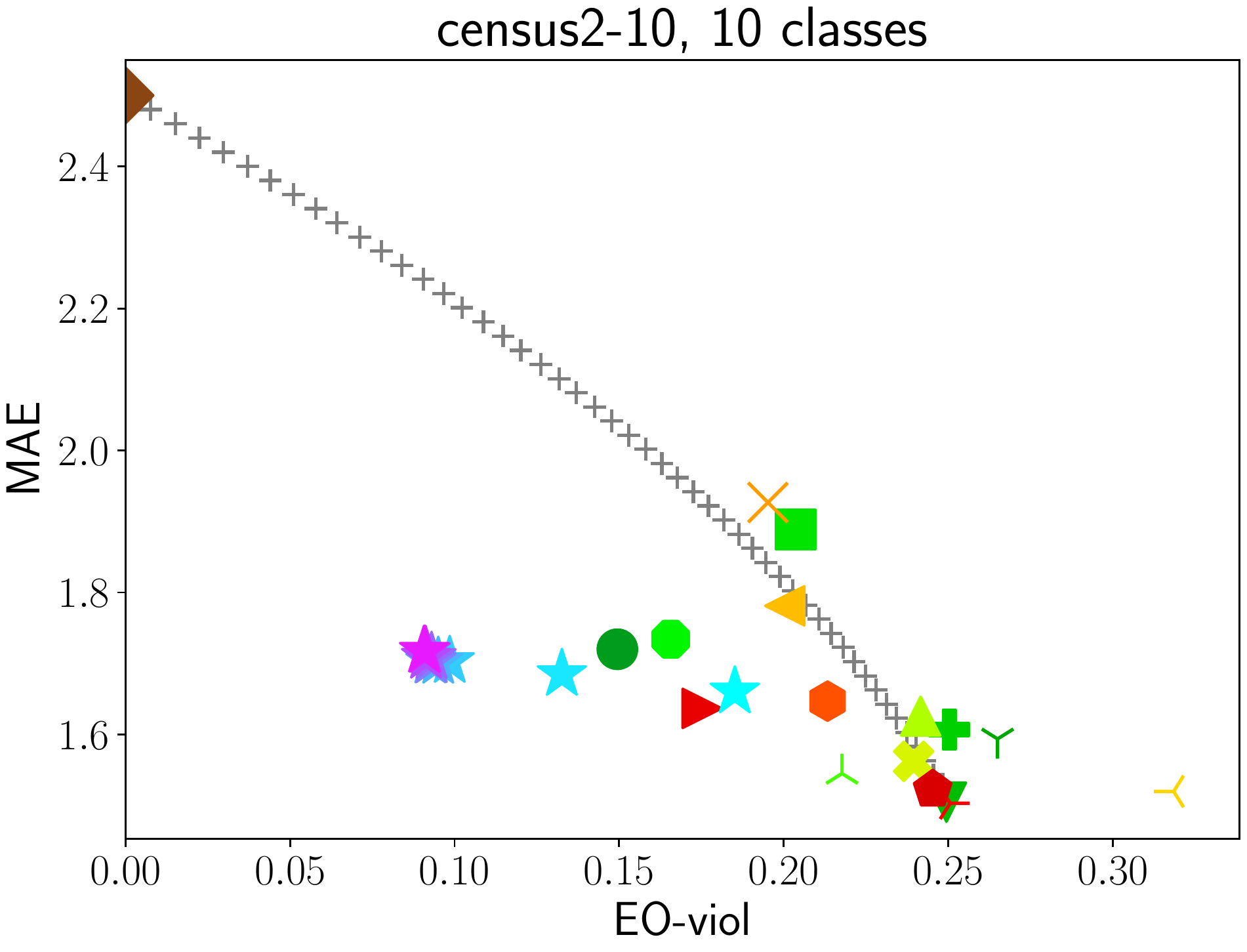}
    \hspace{\abstA}
    \includegraphics[scale=\scaleparameterA]{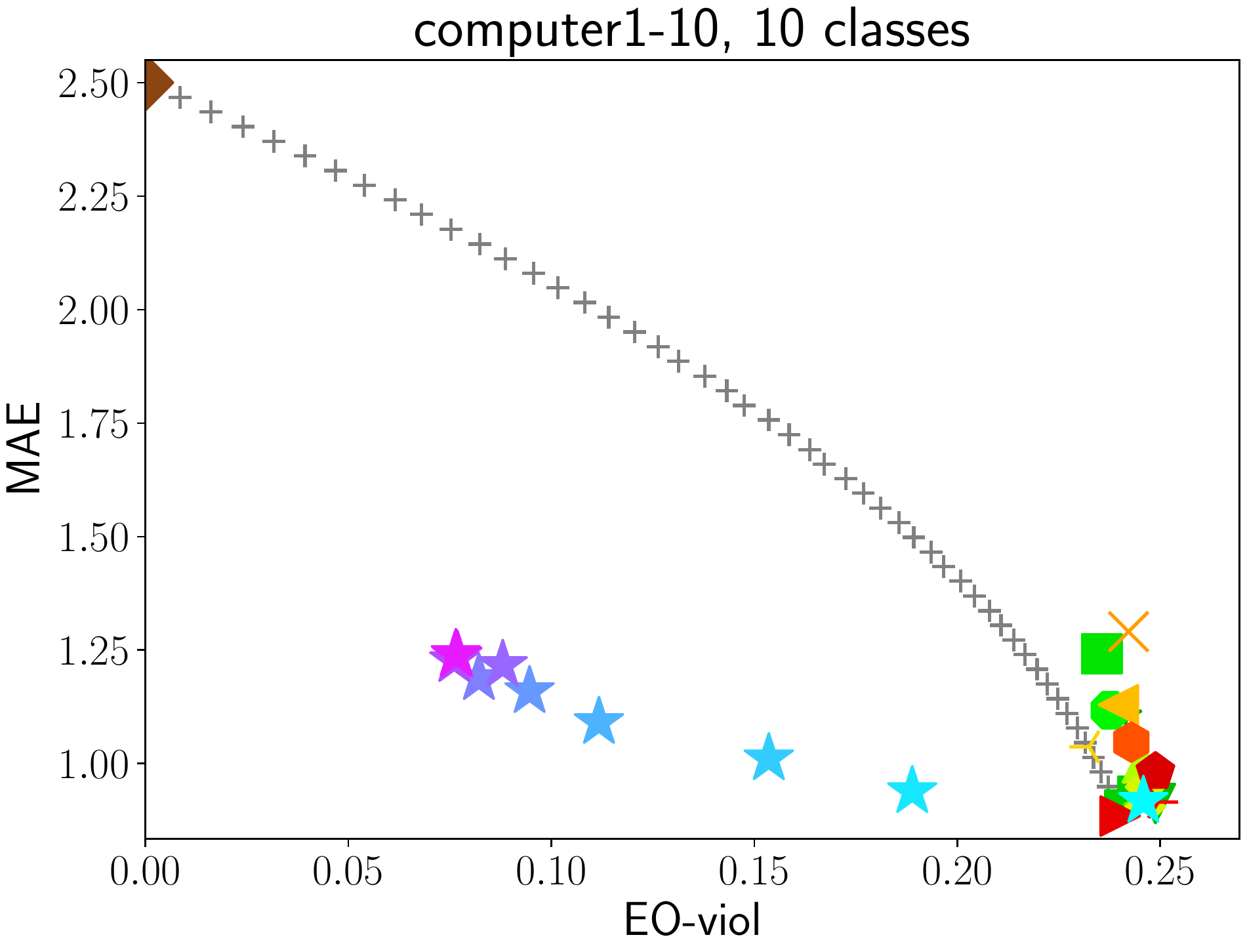}
    \hspace{\abstA}
    \includegraphics[scale=\scaleparameterA]{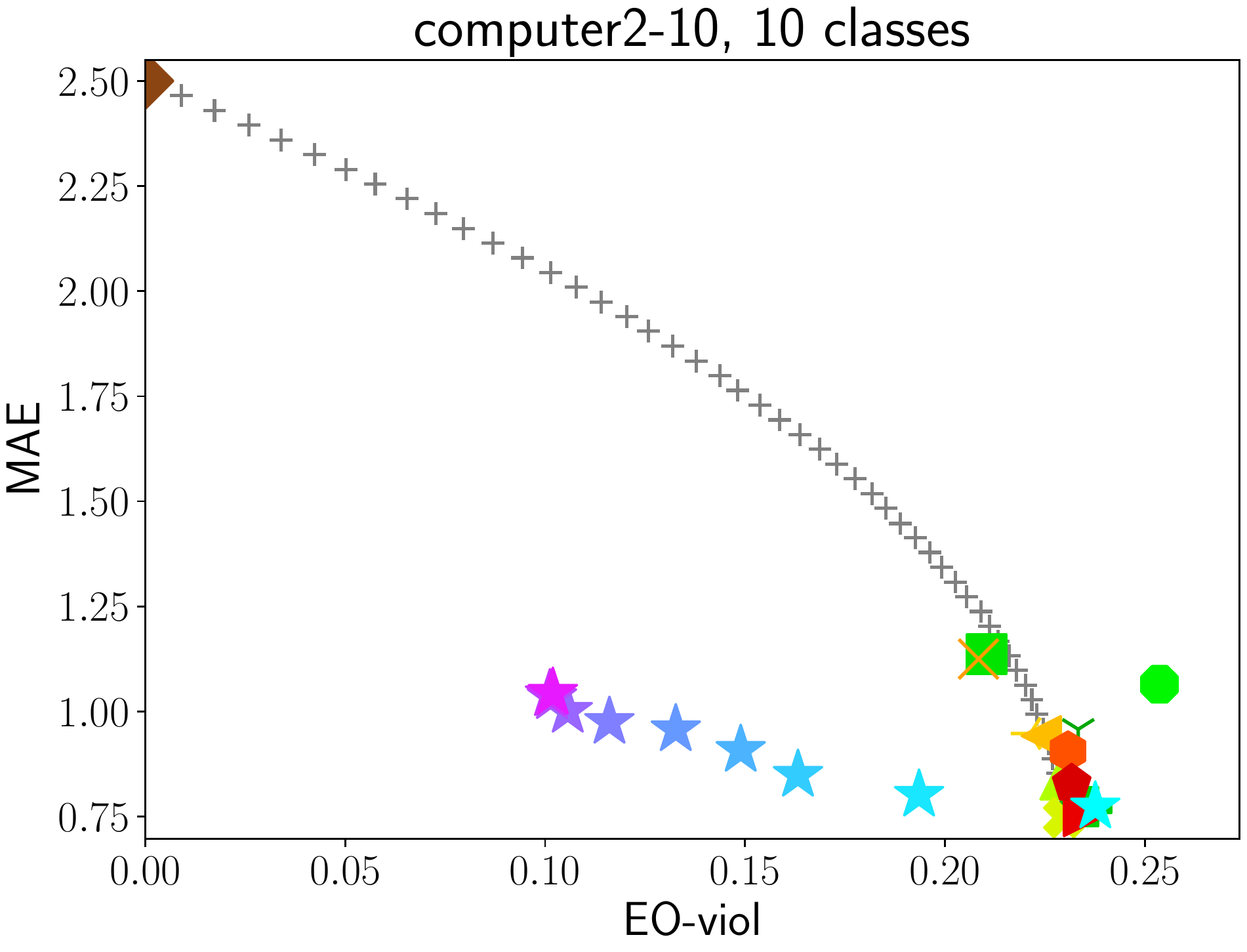}
    
    \includegraphics[scale=\scaleparameterA]{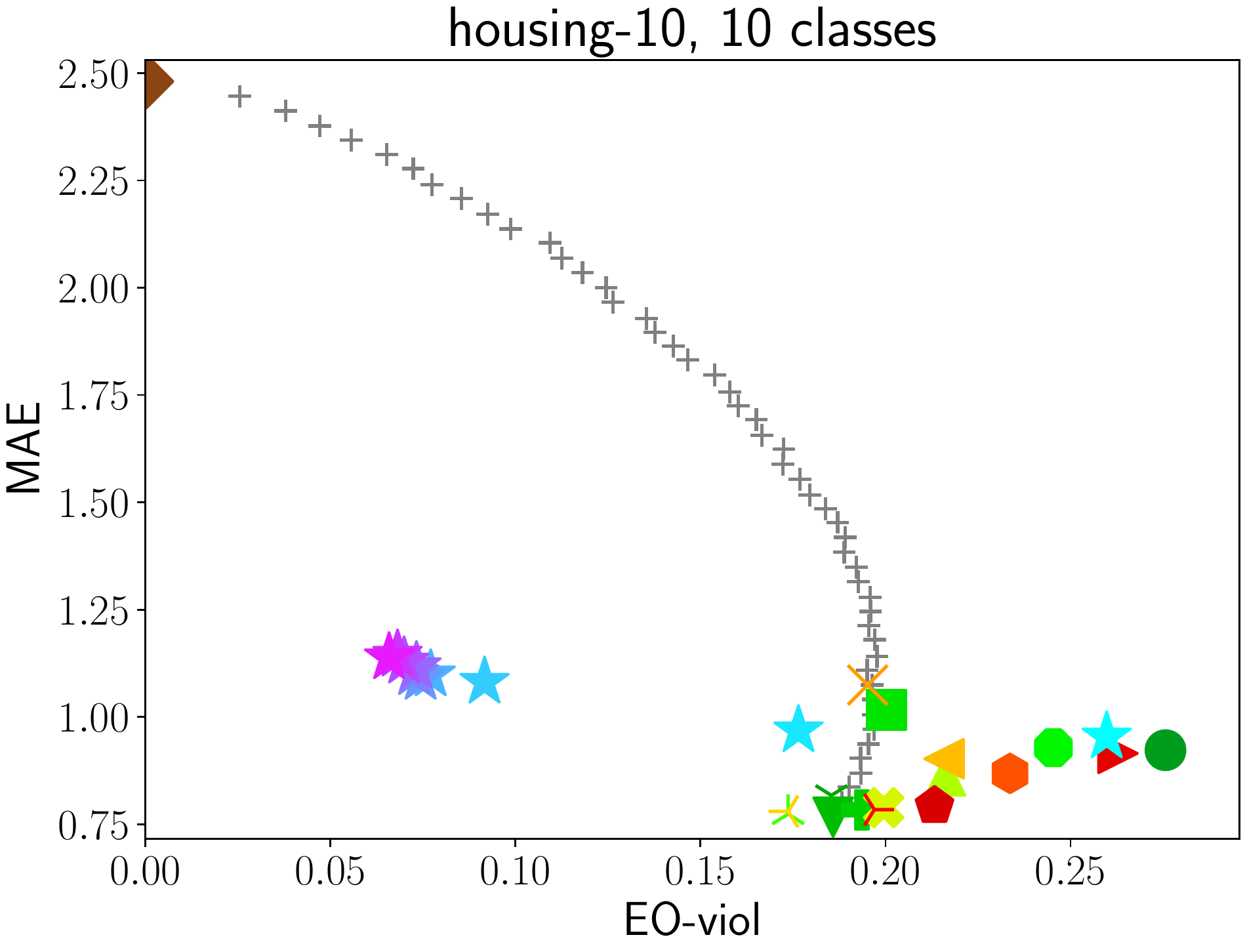}
    \hspace{\abstA}
    \includegraphics[scale=\scaleparameterA]{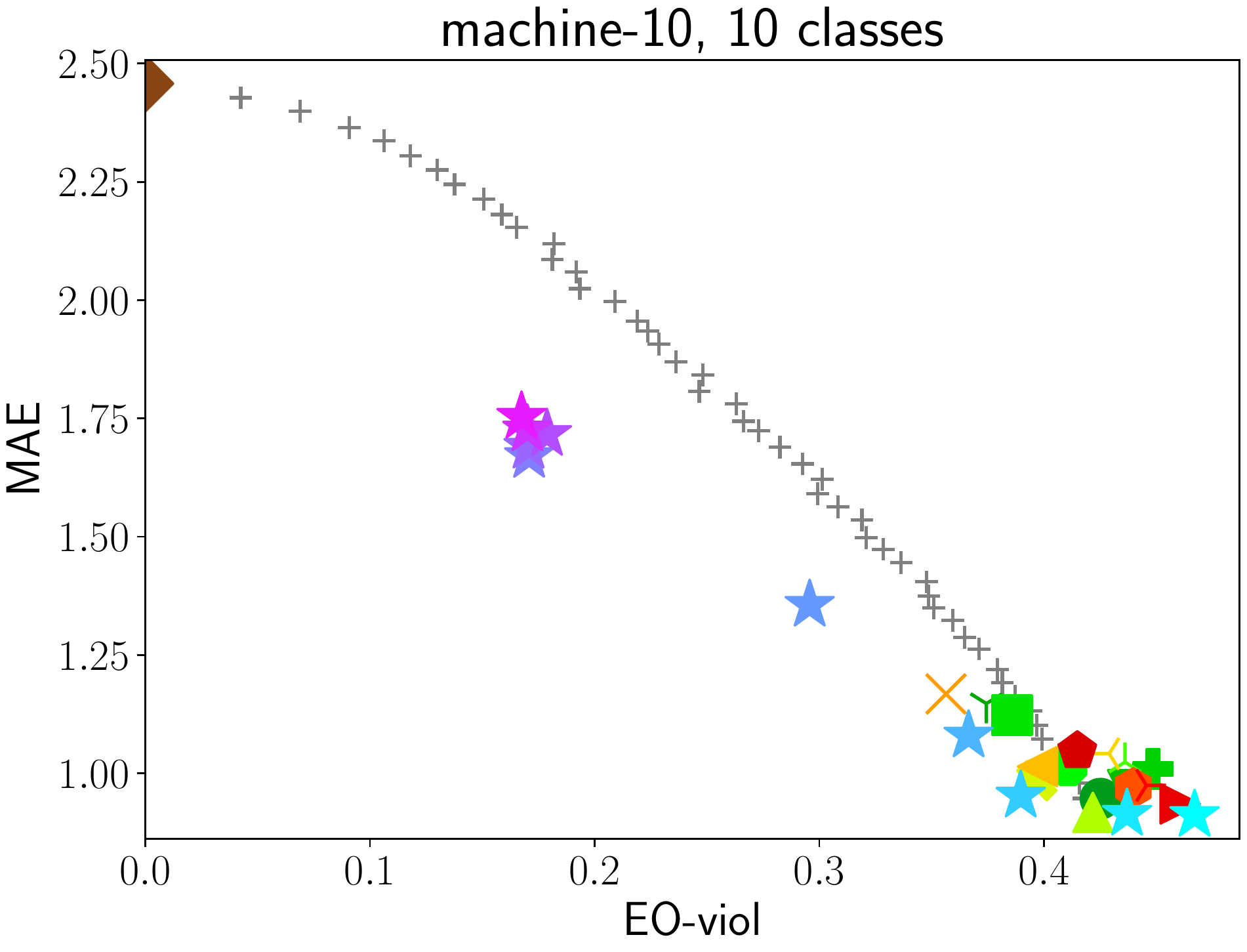}
    \hspace{\abstA}
    \includegraphics[scale=\scaleparameterA]{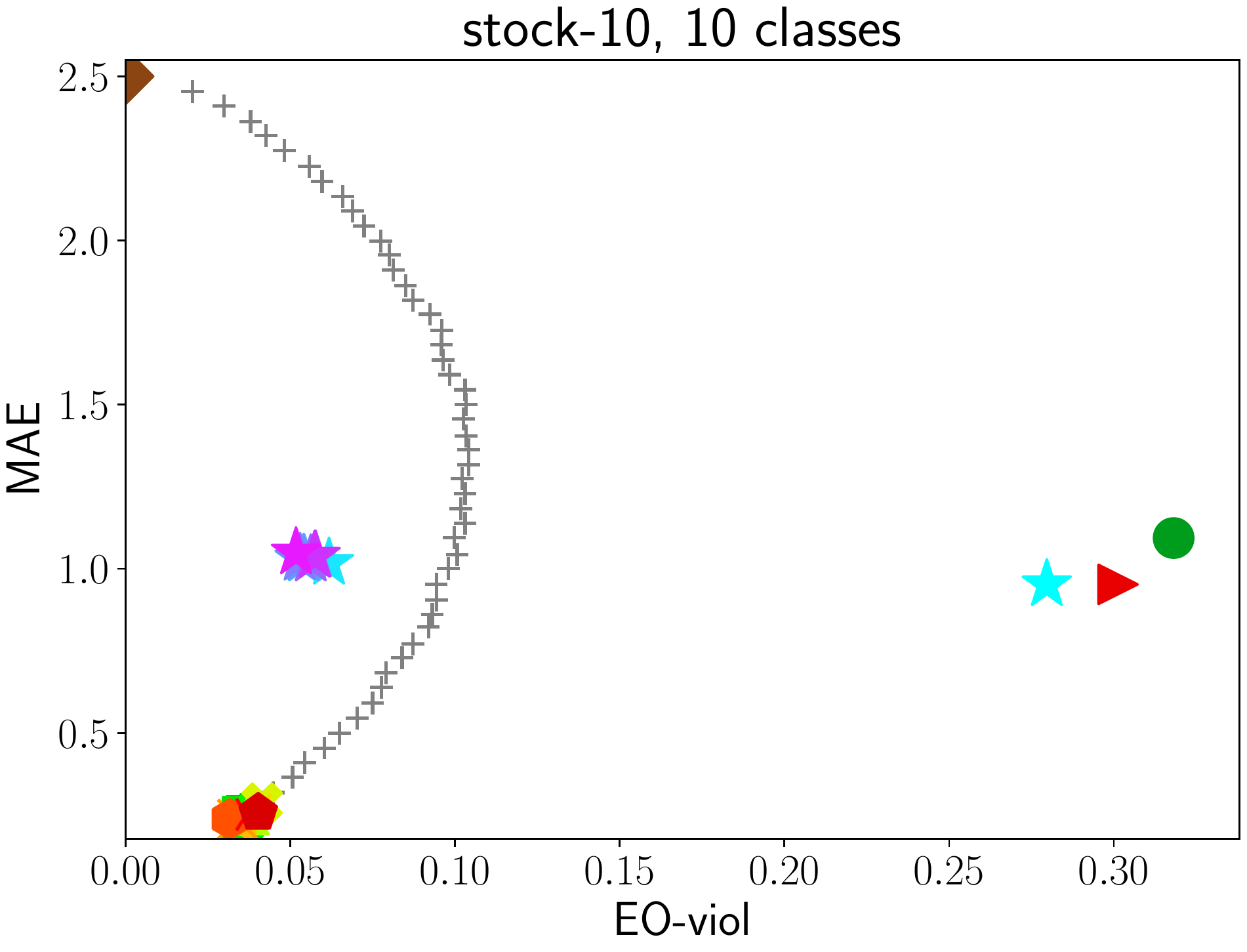}

    \caption{Experiments of Sec.~\ref{subsection_experiment_comparison} on the \textbf{discretized  regression datasets with 10 classes} when aiming for \textbf{pairwise~EO}.}
    \label{fig:exp_comparison_APPENDIX_DISC_10classes_EO}
\end{figure*}

\renewcommand{\scaleparameterA}{0.18}
\renewcommand{\abstA}{6pt}

\renewcommand{\scaleparameterA}{0.17}
\renewcommand{\abstA}{11pt}

\begin{figure*}
    \centering
    \includegraphics[width=\linewidth]{experiment_real_ord/legend_big.pdf}

    \includegraphics[scale=\scaleparameterA]{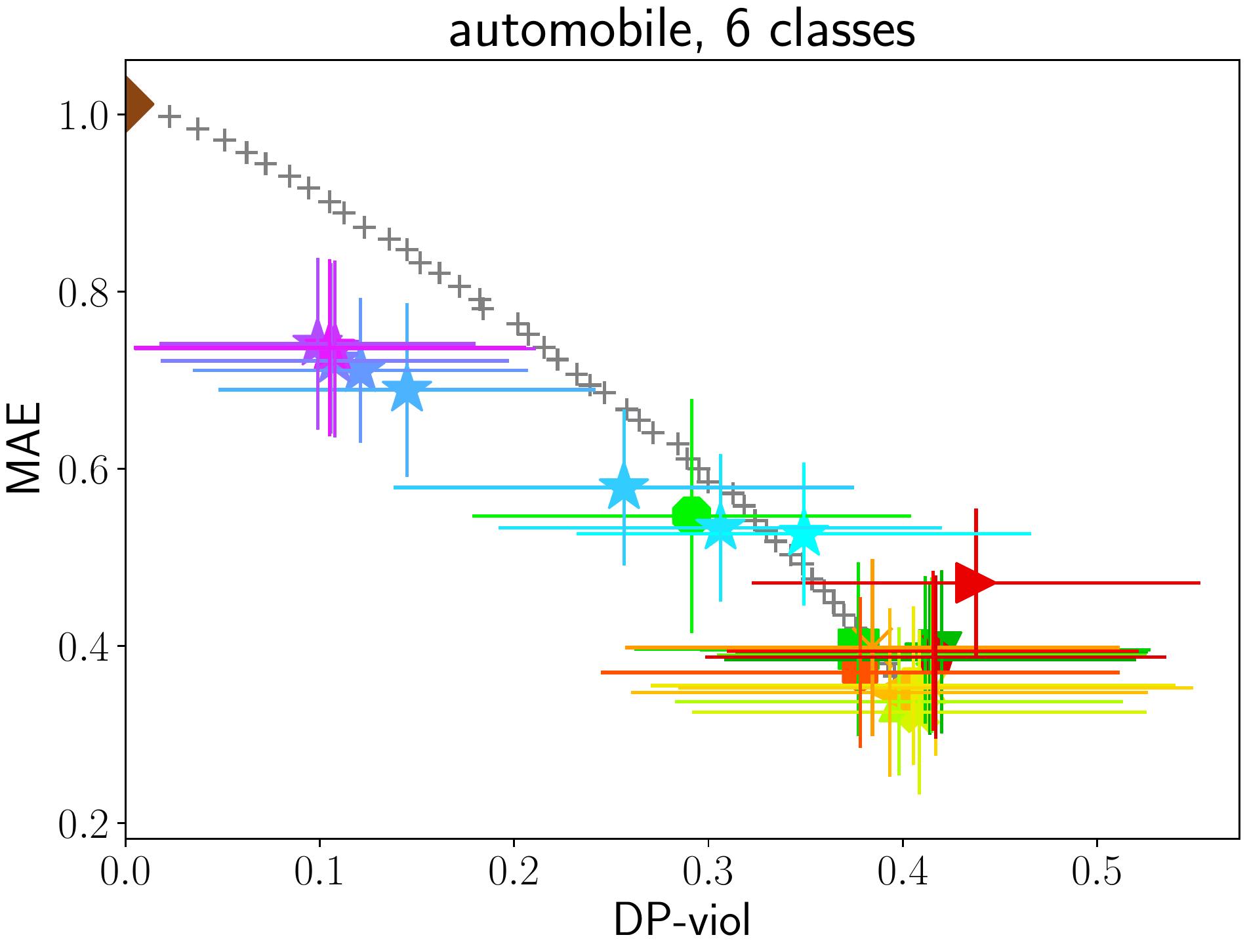}
    \hspace{\abstA}
    \includegraphics[scale=\scaleparameterA]{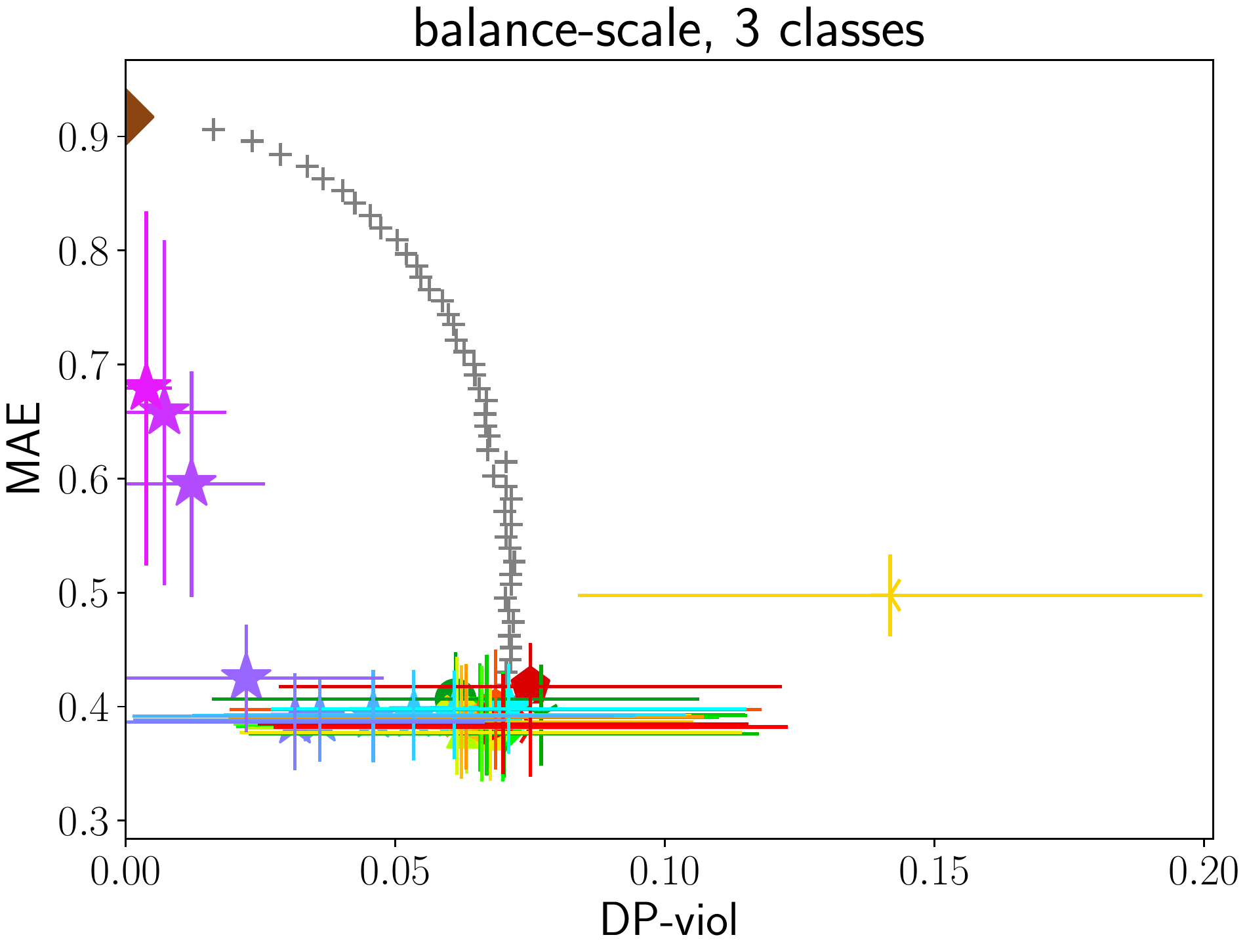}
    \hspace{\abstA}
    \includegraphics[scale=\scaleparameterA]{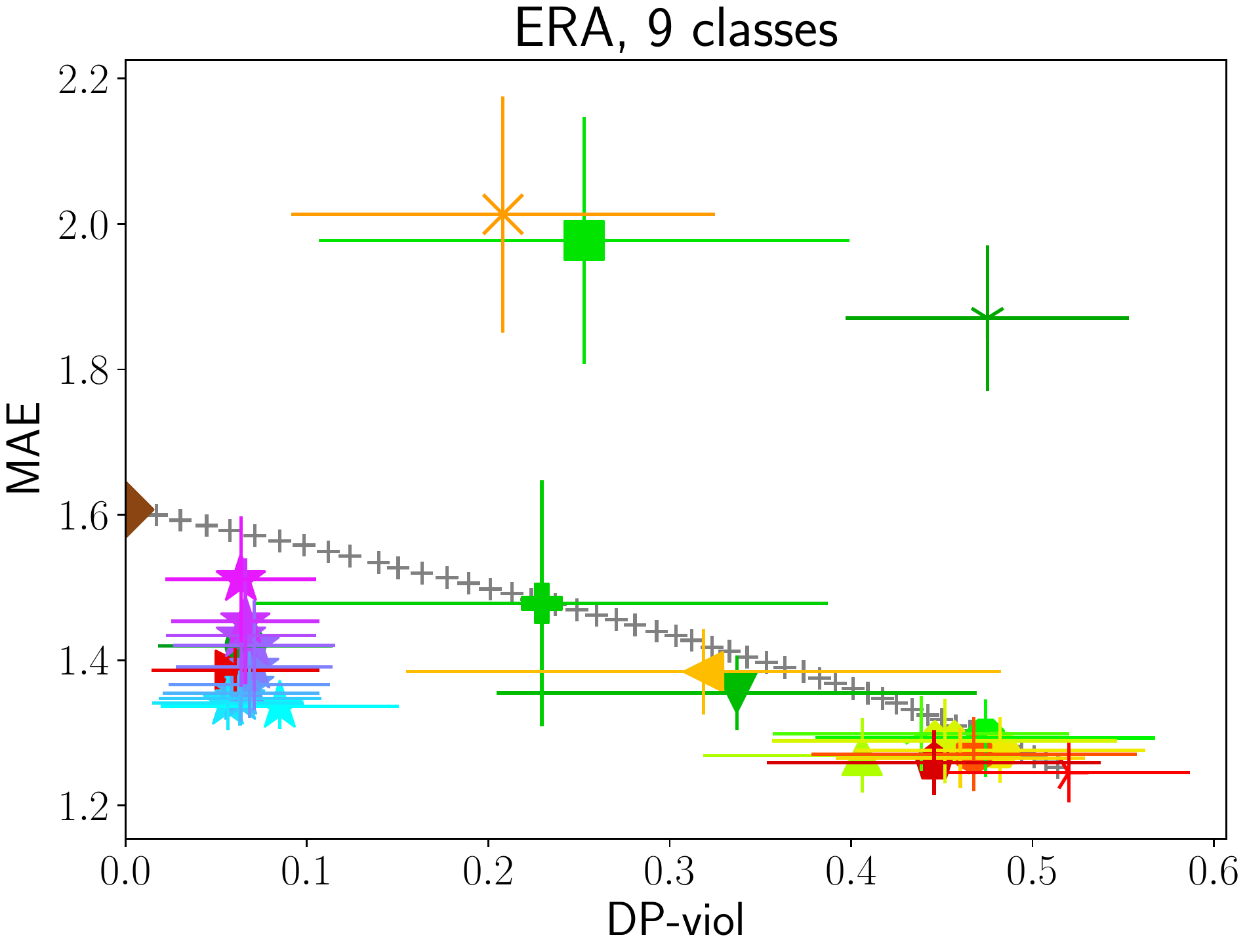}
    \hspace{\abstA}
    \includegraphics[scale=\scaleparameterA]{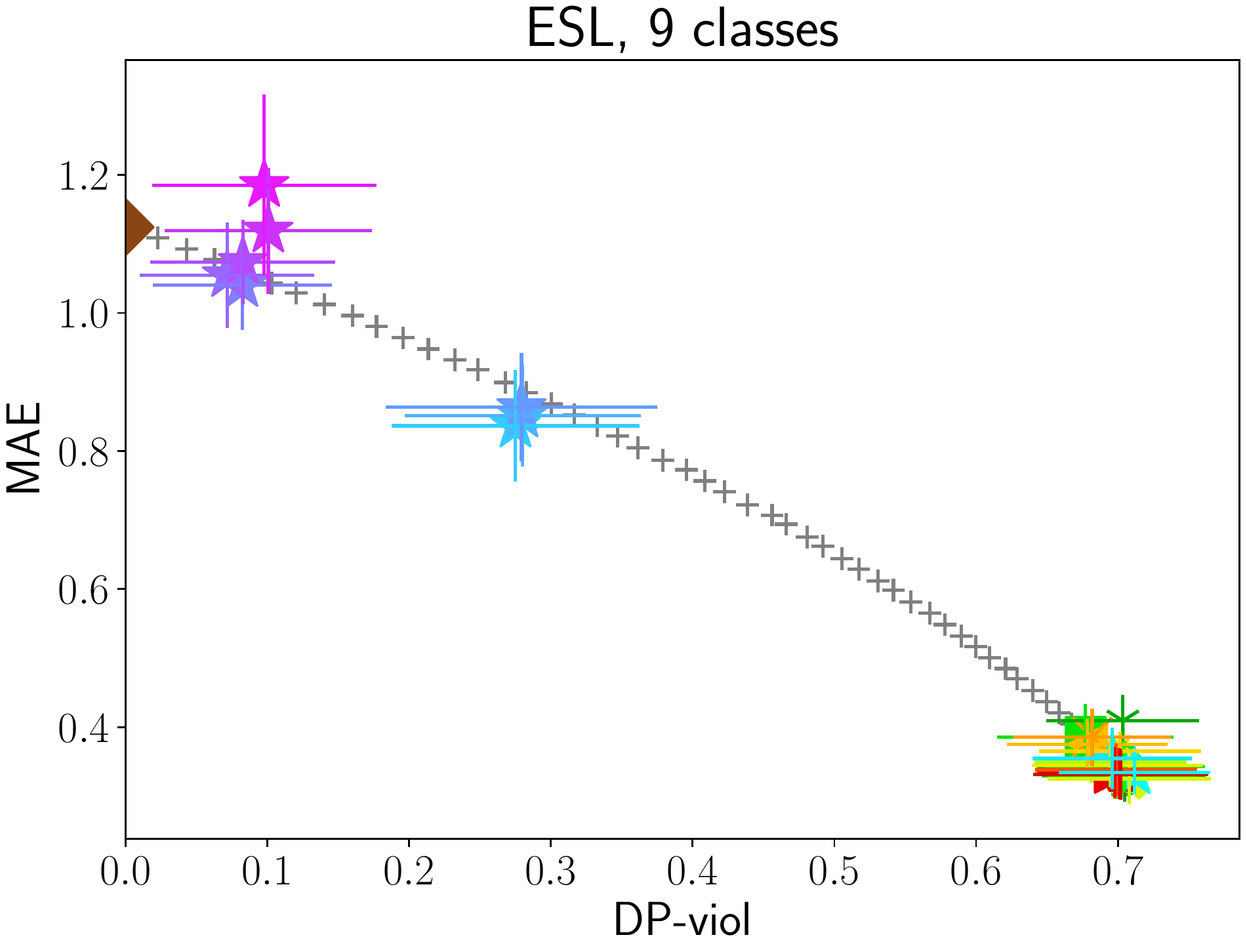}

    \includegraphics[scale=\scaleparameterA]{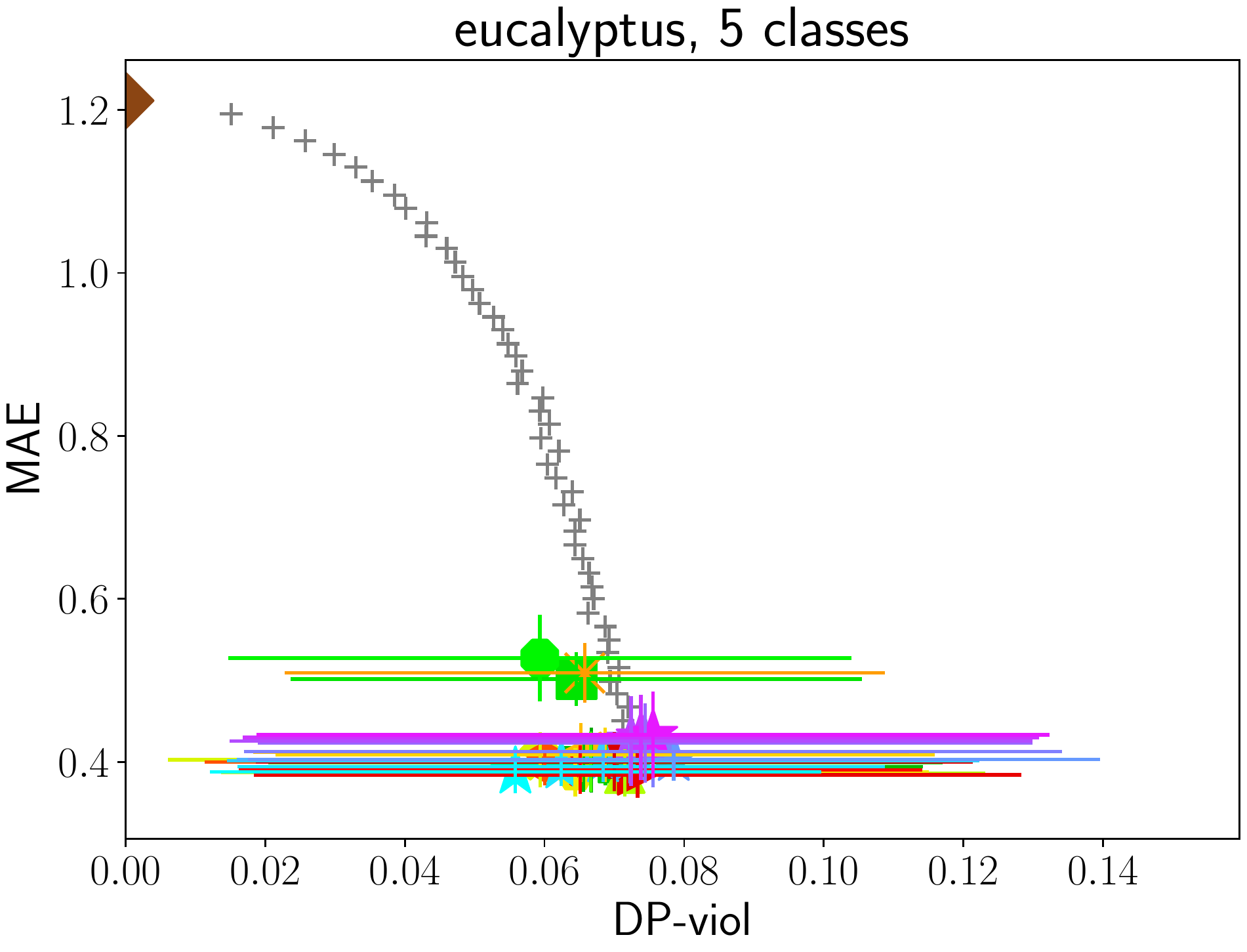}
    \hspace{\abstA}
    \includegraphics[scale=\scaleparameterA]{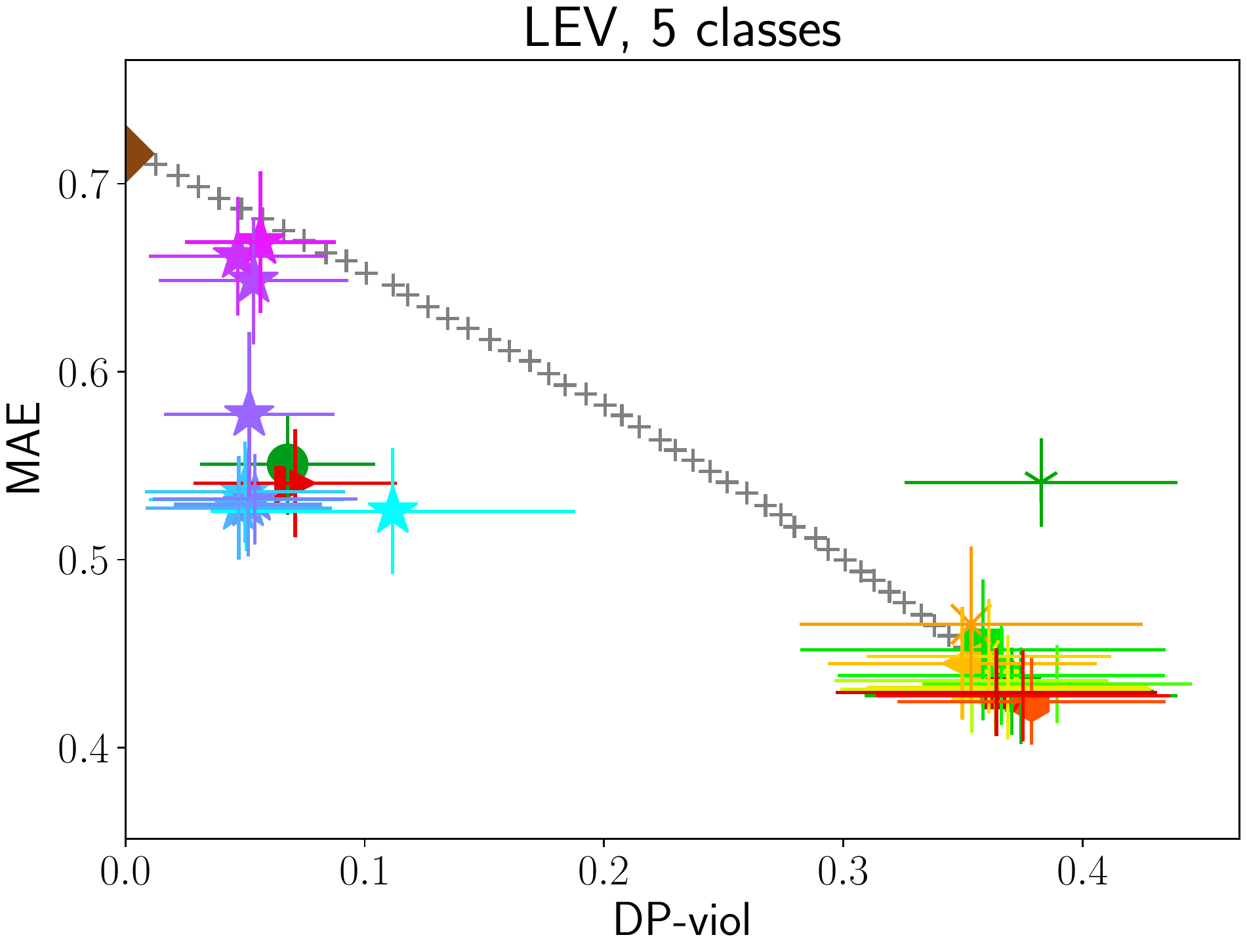}
    \hspace{\abstA}
    \includegraphics[scale=\scaleparameterA]{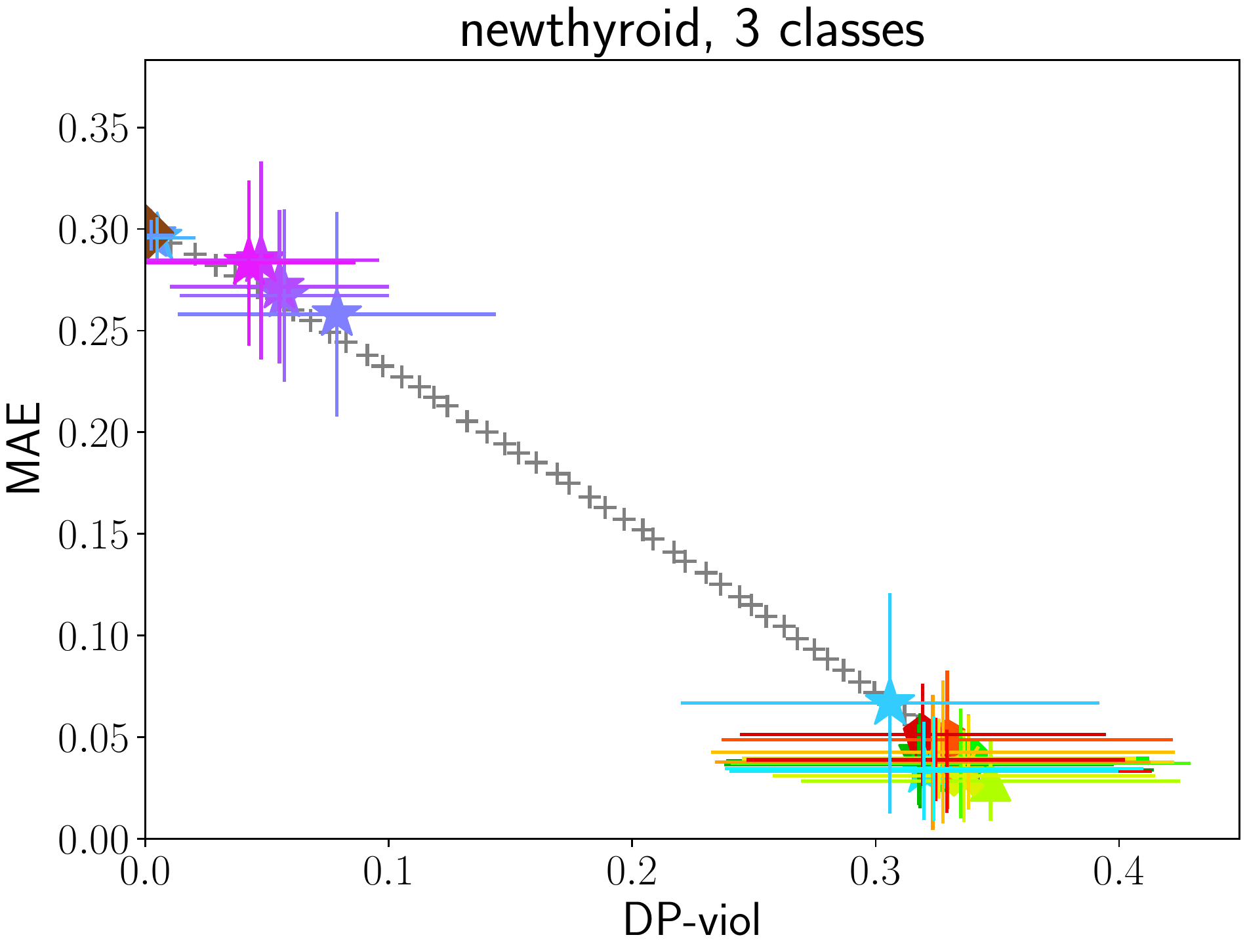}
    \hspace{\abstA}
    \includegraphics[scale=\scaleparameterA]{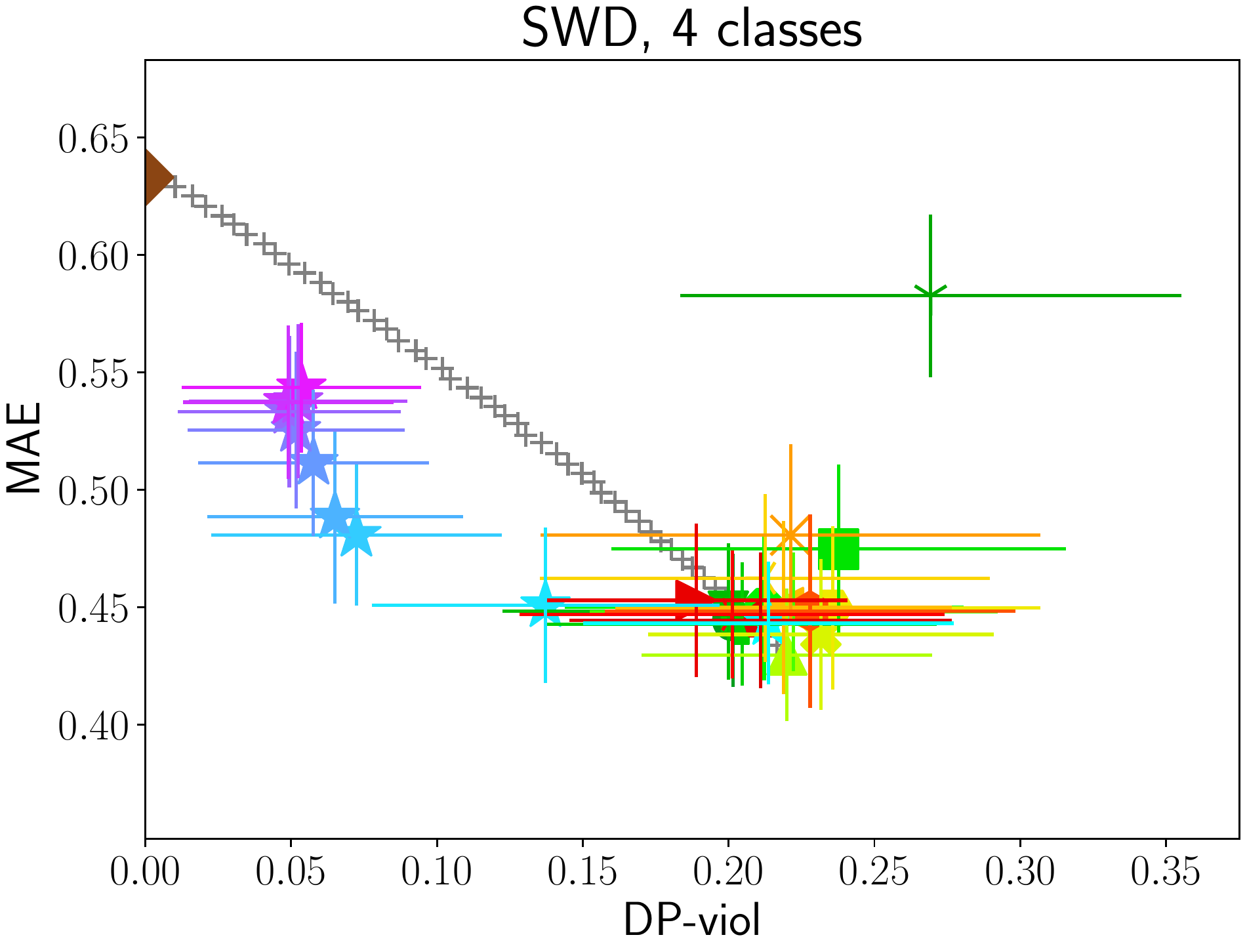}
    
    \includegraphics[scale=\scaleparameterA]{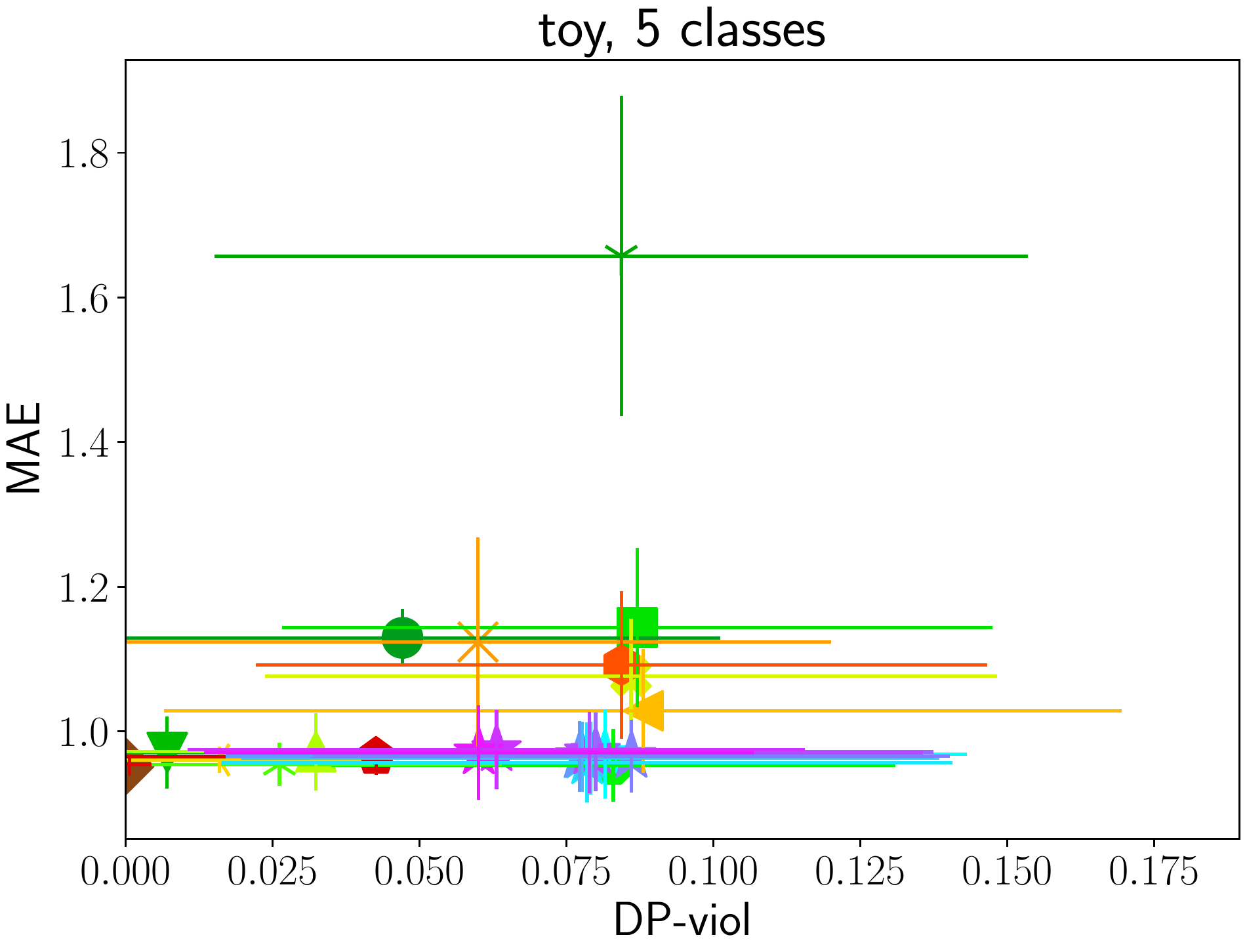}
    \hspace{\abstA}
    \includegraphics[scale=\scaleparameterA]{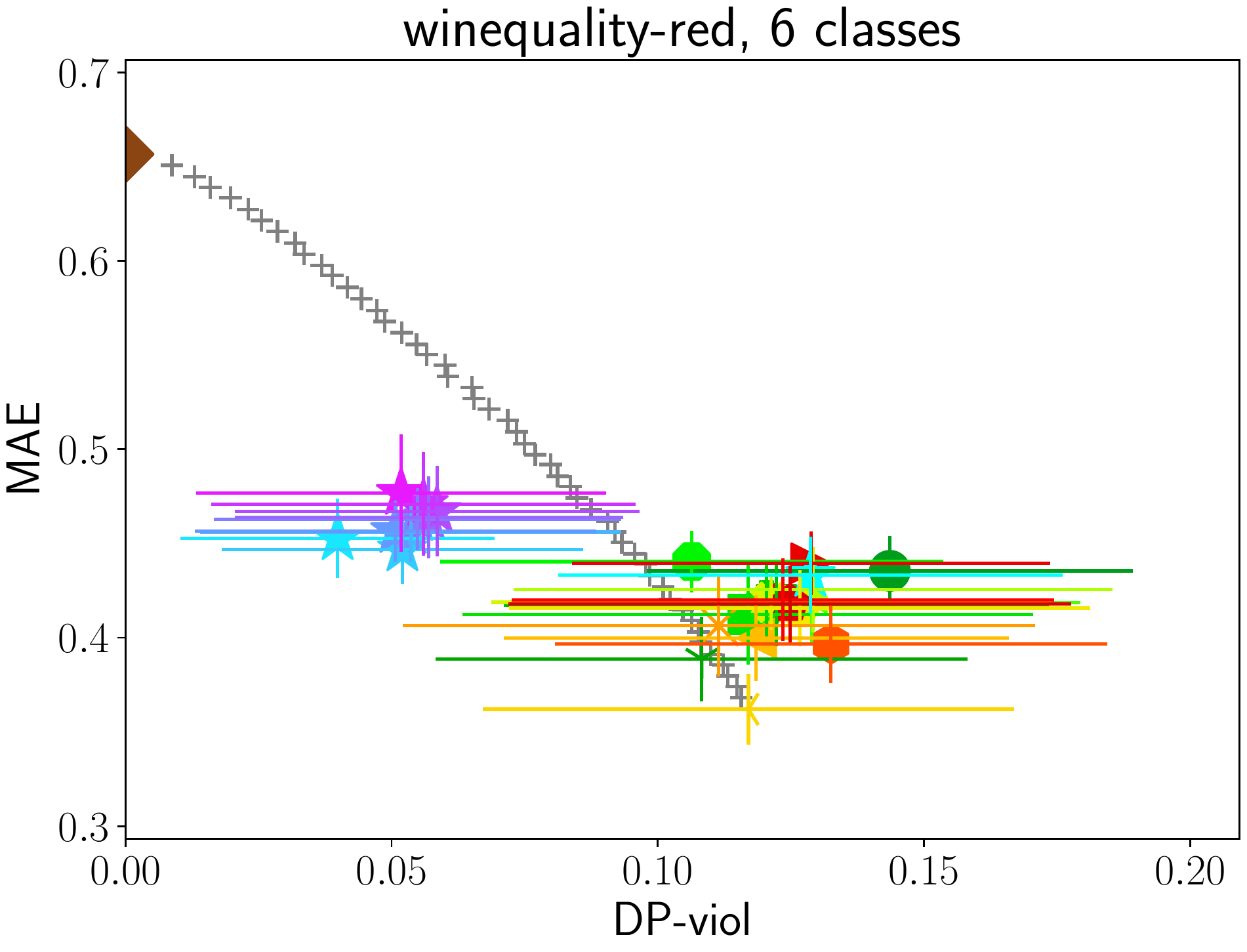}
    \hspace{4.3cm}
    \includegraphics[scale=\scaleparameterA]{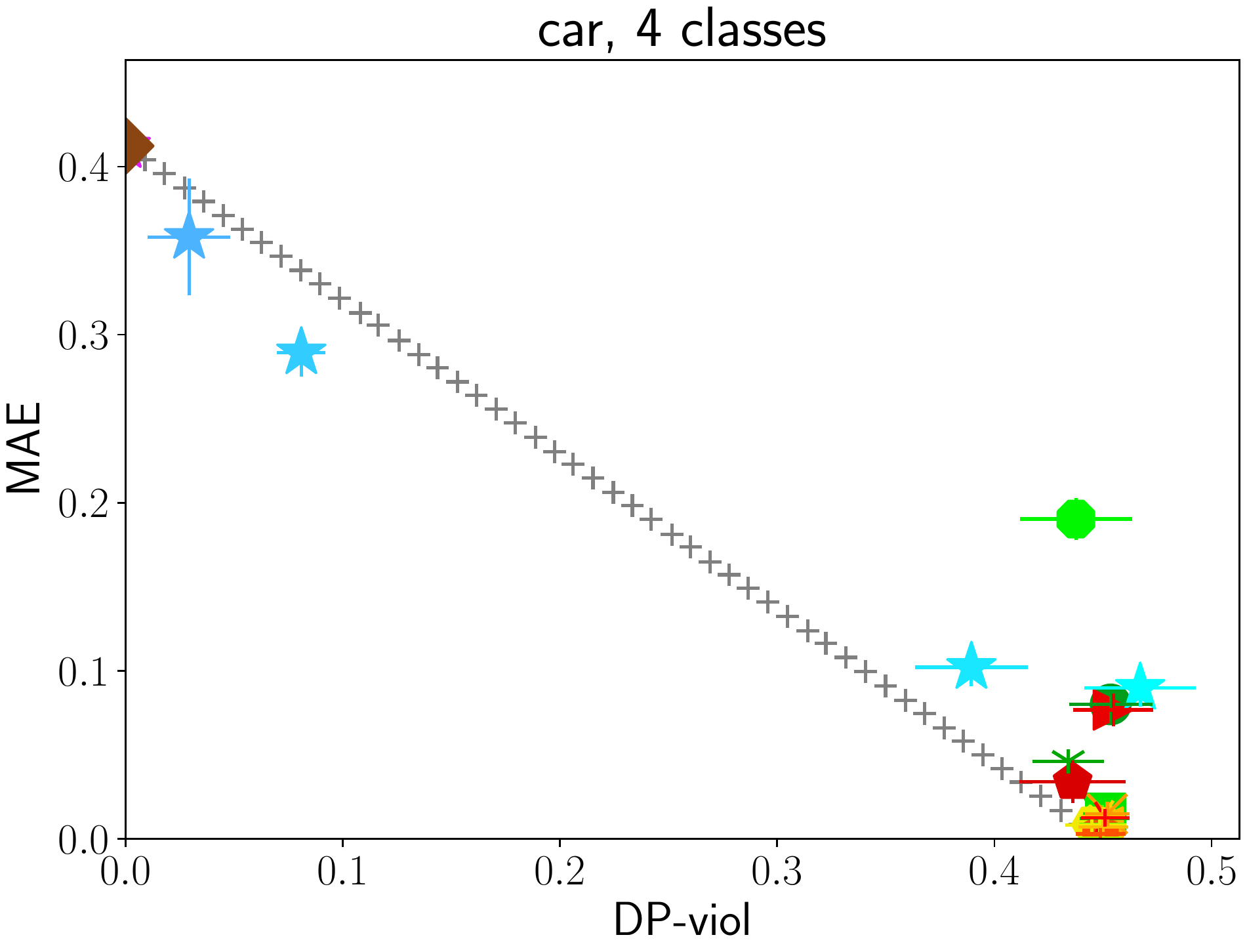}

    \caption{Experiments of Sec.~\ref{subsection_experiment_comparison} on the \textbf{real ordinal regression datasets} when aiming for \textbf{pairwise DP}. Note that the toy dataset has only a single feature that is provided as input to a predictor and that the best method on the toy dataset (svorex) coincides with the best constant predictor;  we do not see any grey crosses corresponding to randomly mixing the best predictor with the best constant one. The errorbars show the standard deviation over the 30 splits into training~and~test~sets.}
    \label{fig:exp_comparison_APPENDIX_real_ord_reg_DP_with_STD}
\end{figure*}

\begin{figure*}
    
    \centering
    \includegraphics[width=\linewidth]{experiment_real_ord/legend_big.pdf}

    \includegraphics[scale=\scaleparameterA]{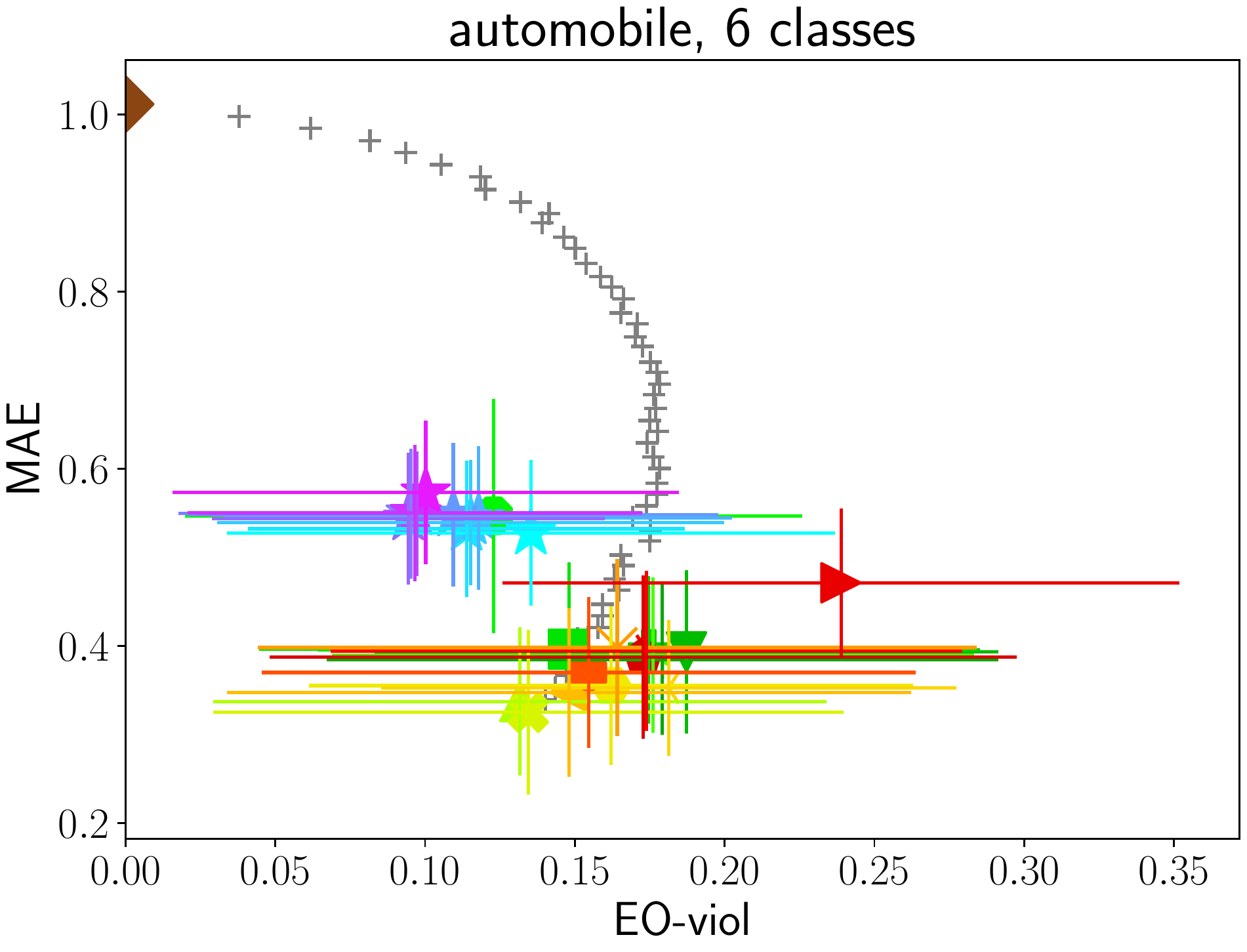}
    \hspace{\abstA}
    \includegraphics[scale=\scaleparameterA]{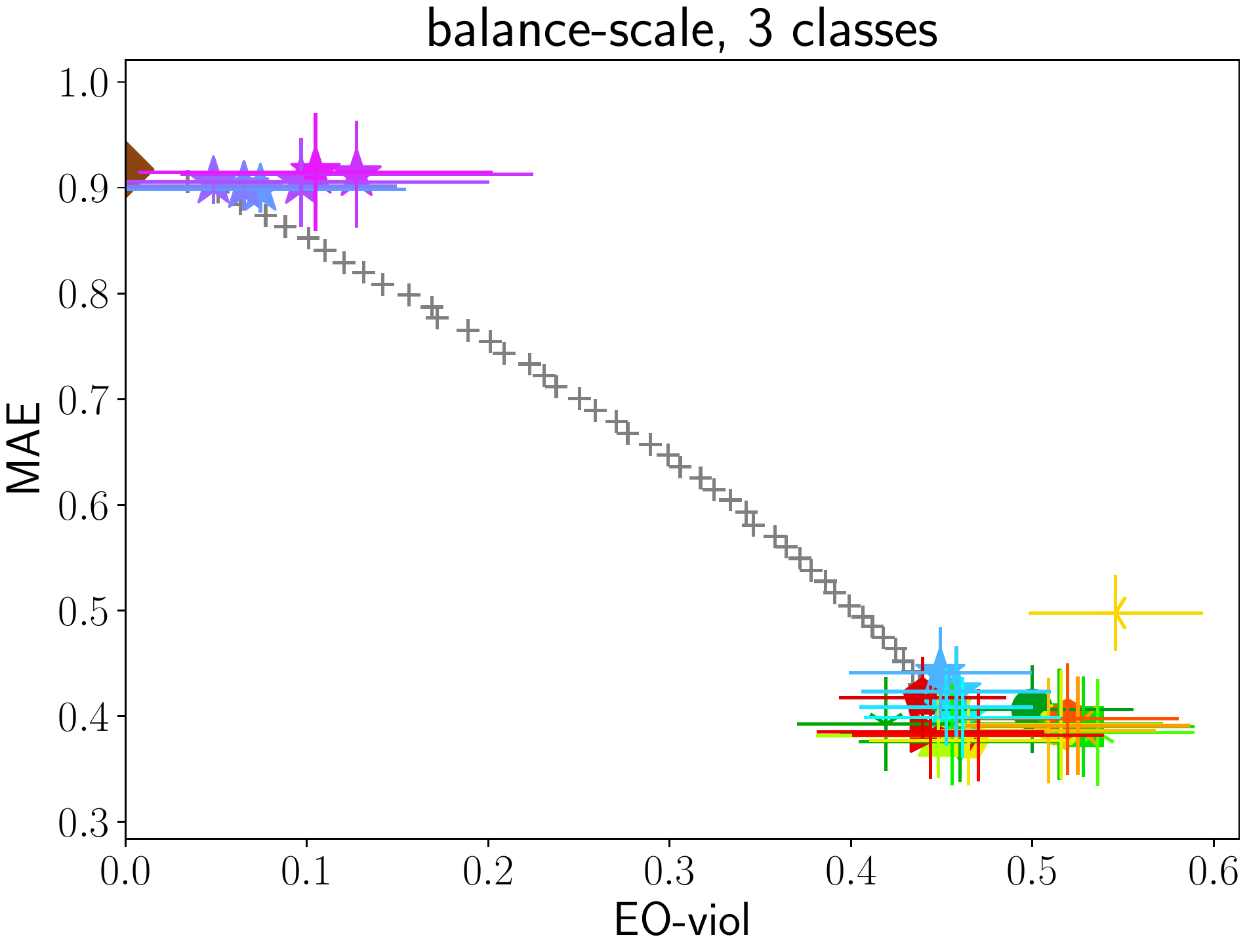}
    \hspace{\abstA}
    \includegraphics[scale=\scaleparameterA]{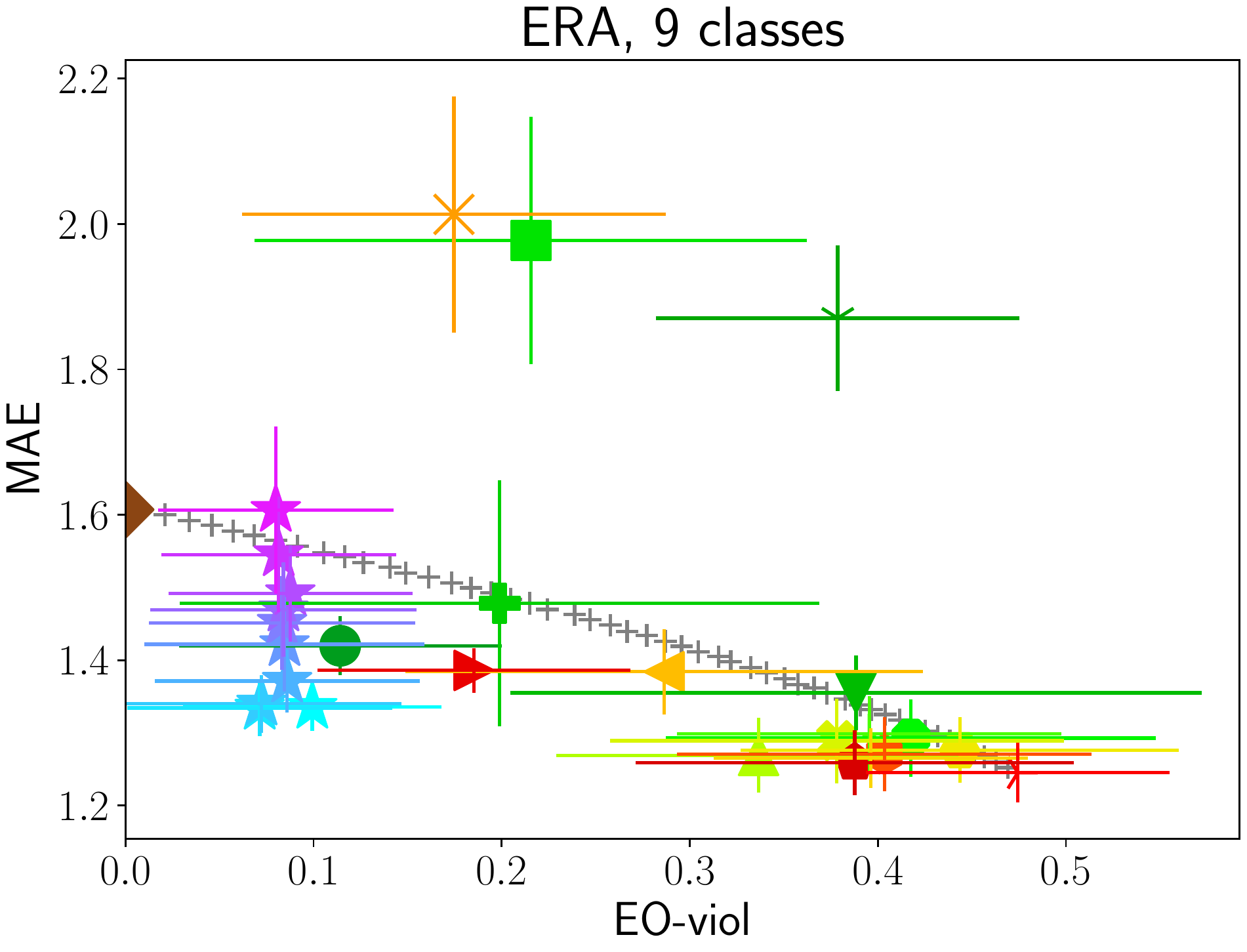}
    \hspace{\abstA}
    \includegraphics[scale=\scaleparameterA]{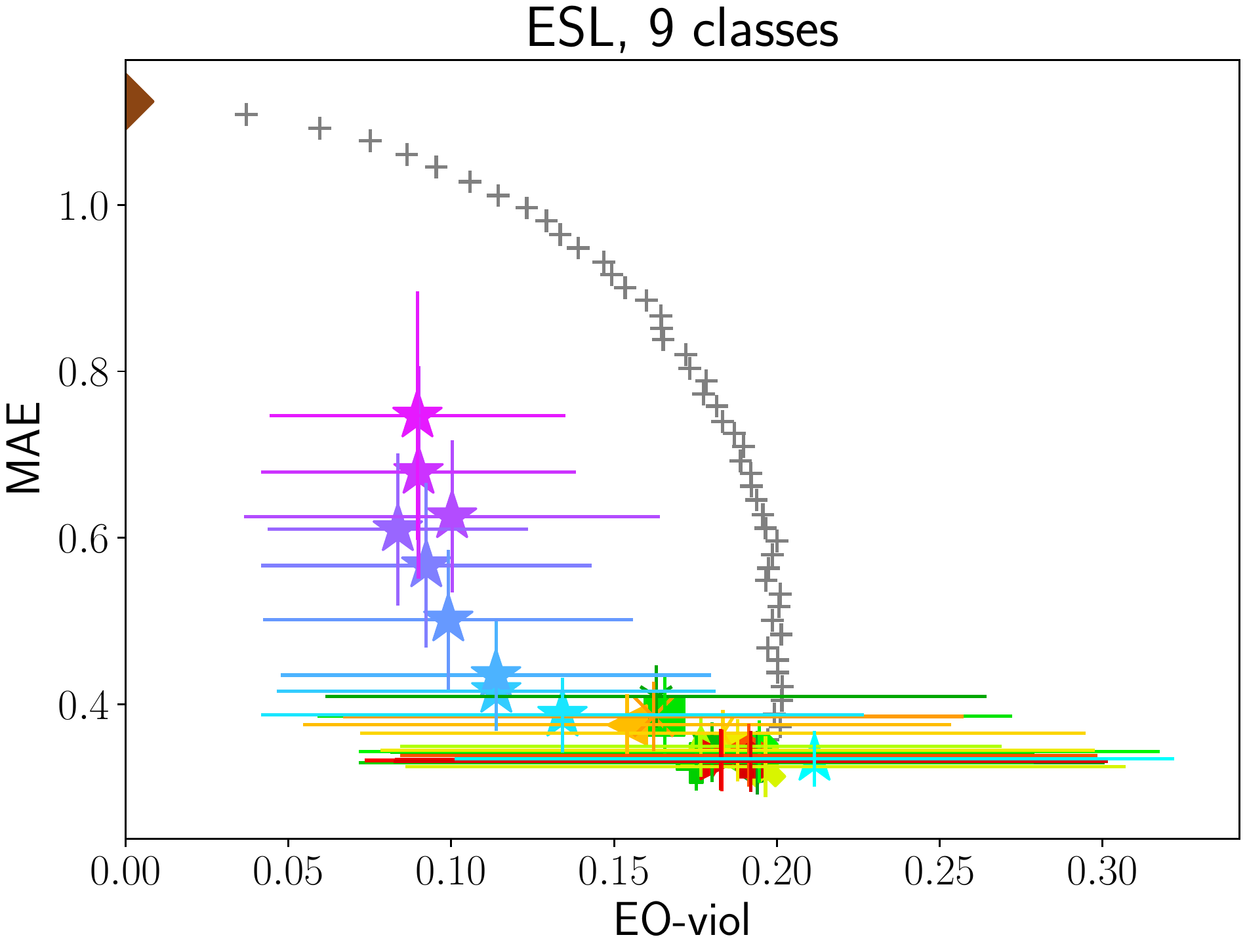}
    
    \includegraphics[scale=\scaleparameterA]{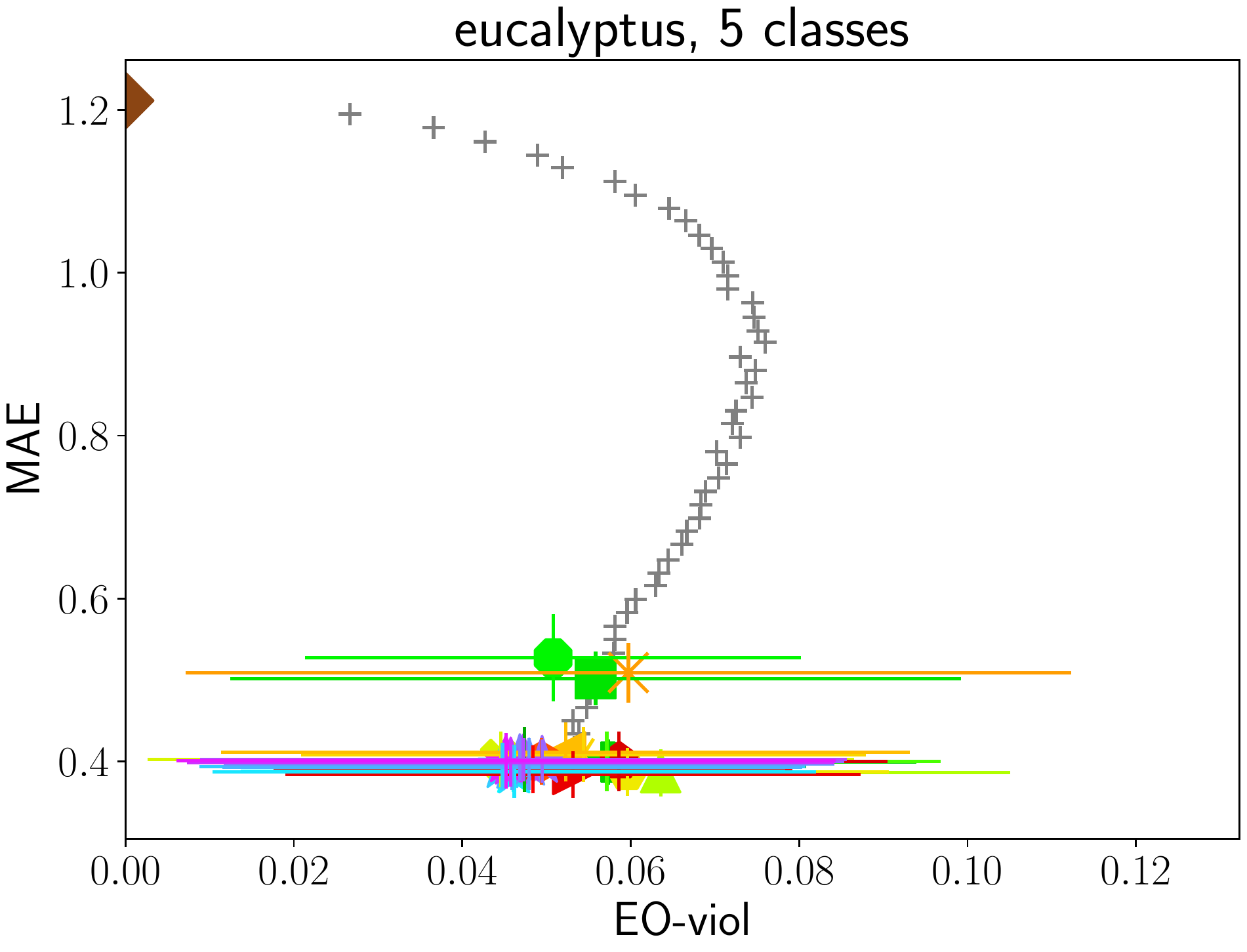}
    \hspace{\abstA}
    \includegraphics[scale=\scaleparameterA]{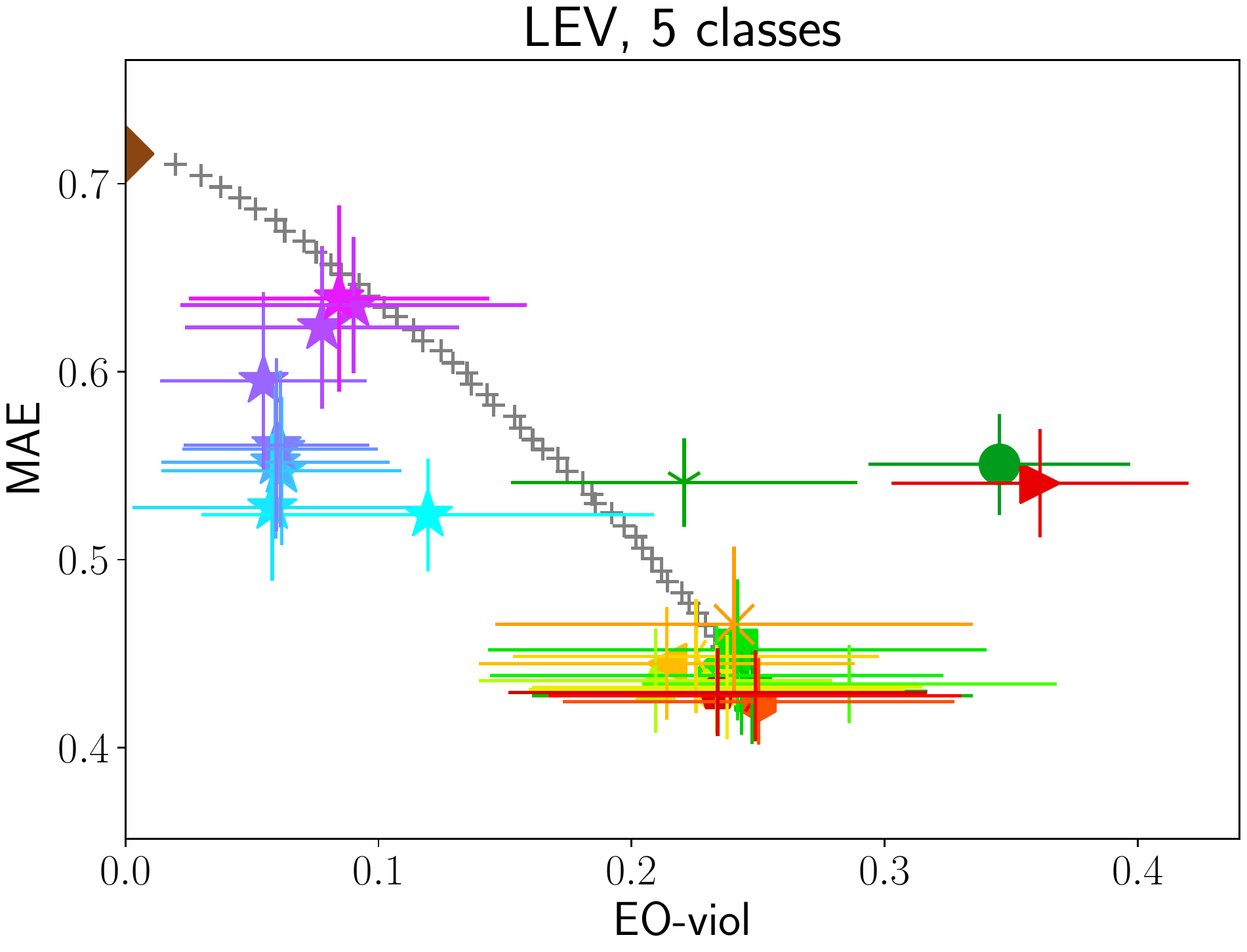}
    \hspace{\abstA}
    \includegraphics[scale=\scaleparameterA]{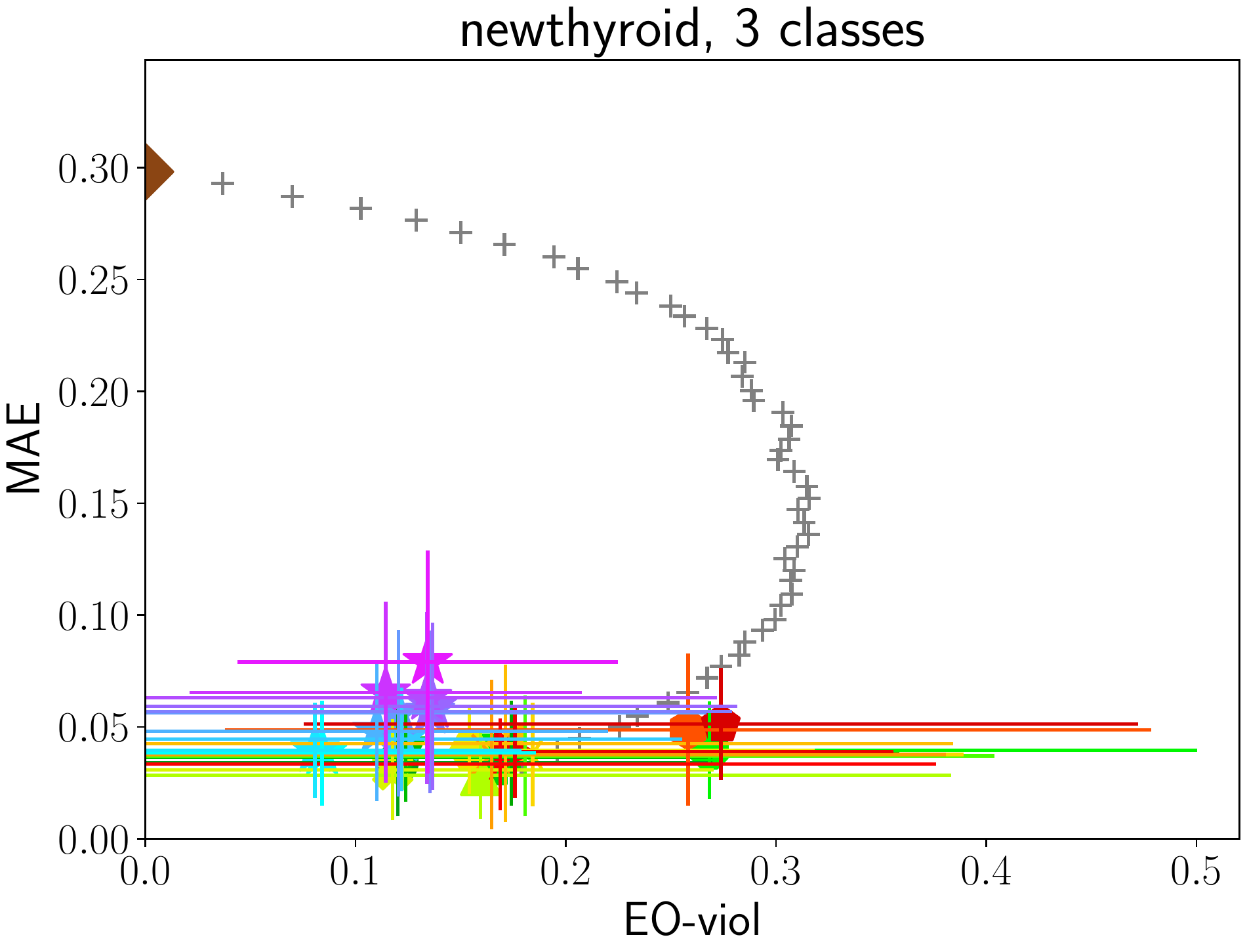}
    \hspace{\abstA}
    \includegraphics[scale=\scaleparameterA]{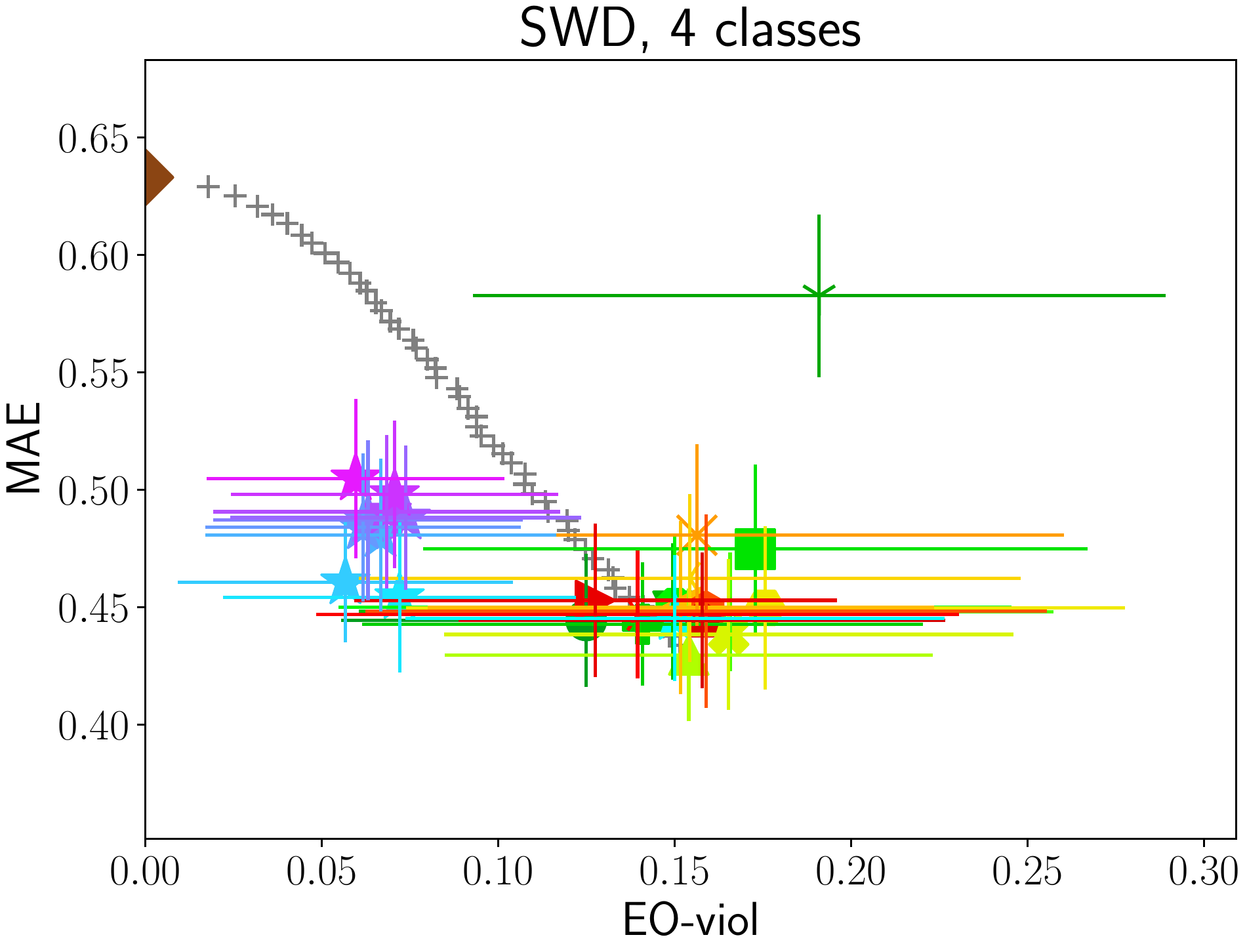}
    
    \includegraphics[scale=\scaleparameterA]{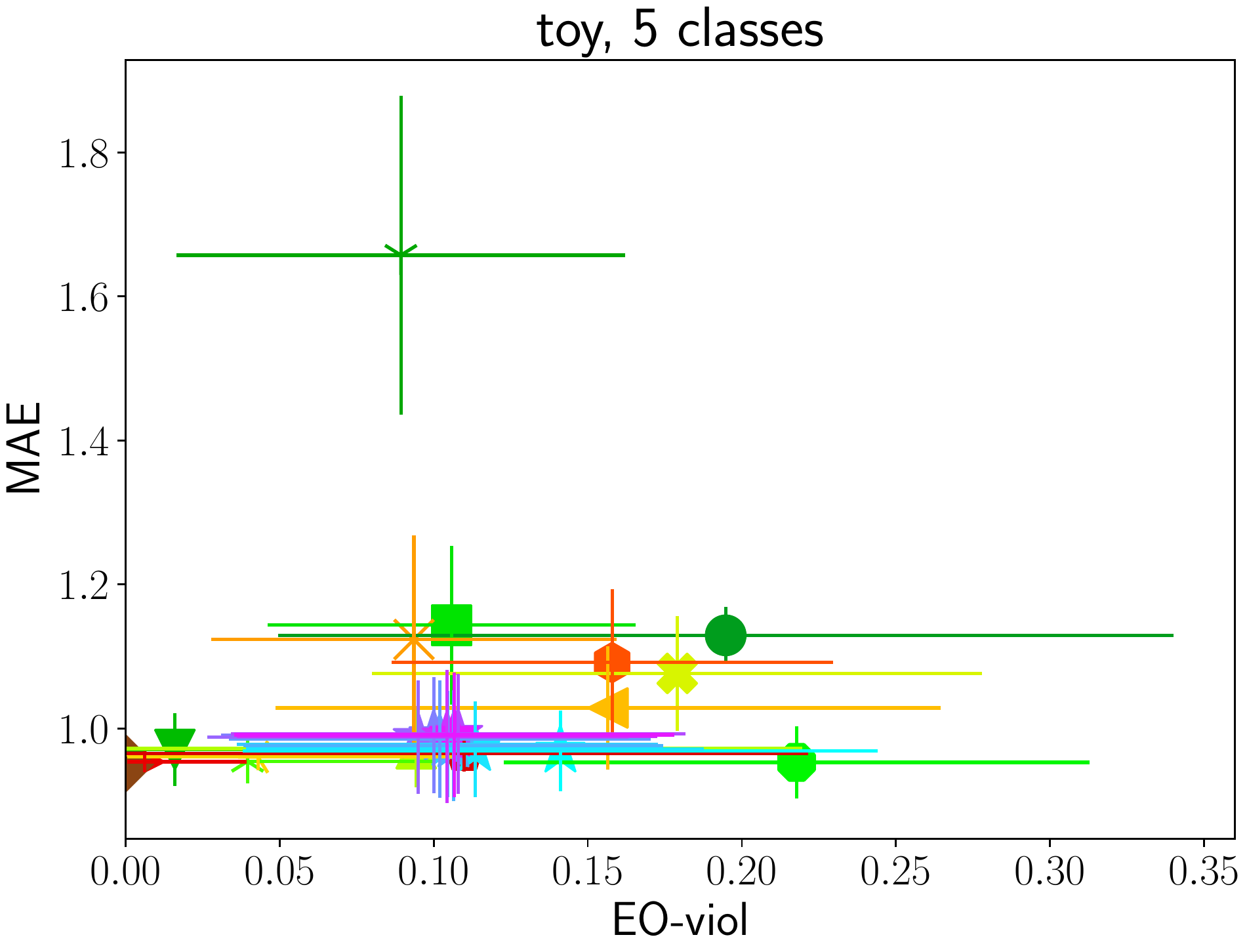}
    \hspace{\abstA}
    \includegraphics[scale=\scaleparameterA]{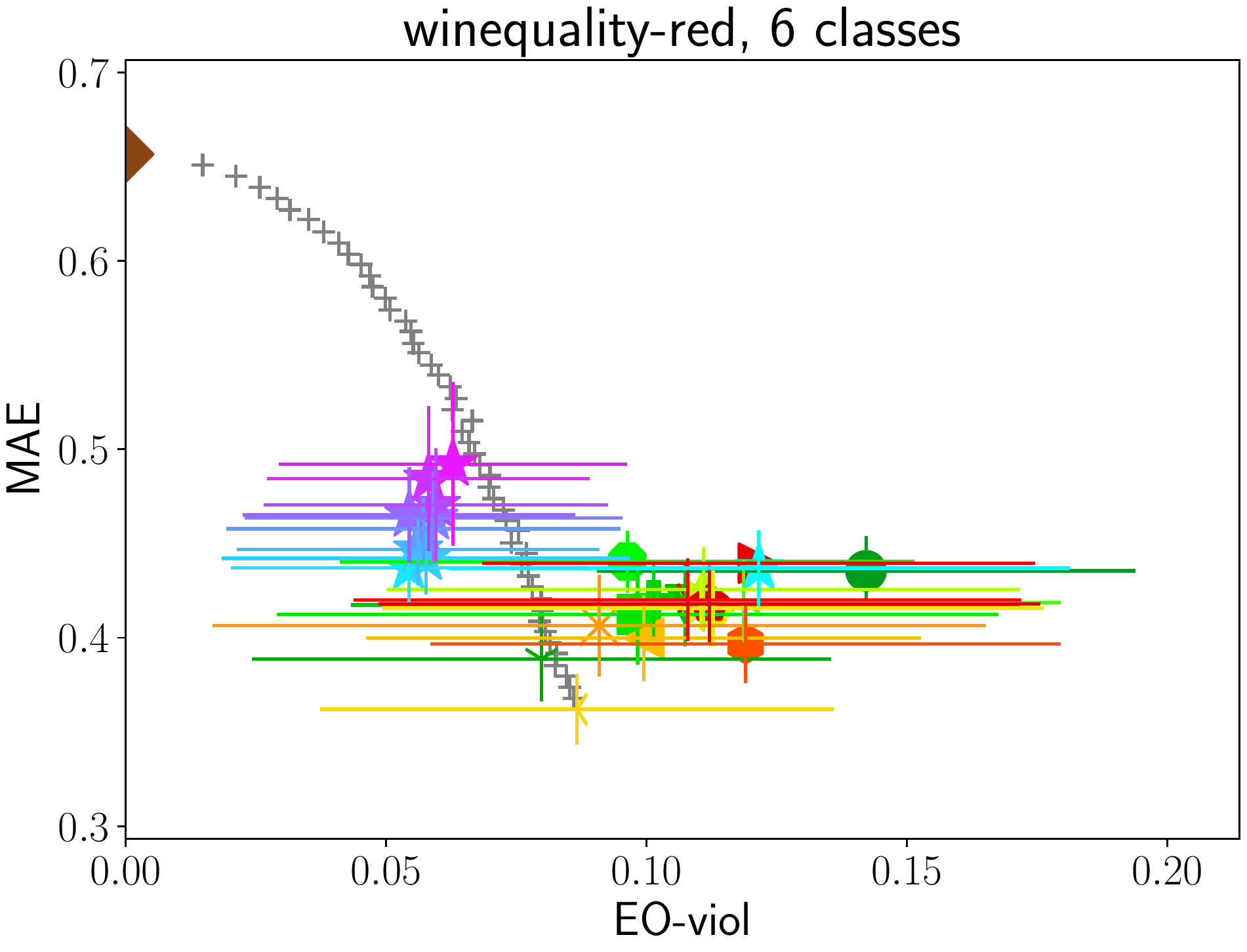}
    \hspace{7.7cm}
    
\caption{Experiments of Section~\ref{subsection_experiment_comparison} on the \textbf{real ordinal regression datasets} when aiming for \textbf{pairwise EO}. Note that the toy dataset has only a single feature that is provided as input to a predictor and that the best method on the toy dataset (svorex) coincides with the best constant predictor;  we do not see any grey crosses corresponding to randomly mixing the best predictor with the best constant one. Also note that we do not provide a plot for the car dataset; since $\Psymb[y_1<y_2, a_1=0,a_2=1]=0$ for the car dataset (cf. Table~\ref{table_statistics_data_sets2}), the notion of pairwise EO is not well-defined for this dataset. The errorbars show the standard deviation over the 30 splits into training and test~sets.}
    \label{fig:exp_comparison_APPENDIX_real_ord_reg_EO_with_STD}
\end{figure*}

\renewcommand{\scaleparameterA}{0.204}
\renewcommand{\abstA}{2pt}

\begin{figure*}
    
    \includegraphics[width=\linewidth]{experiment_real_ord/legend_big.pdf}

    \includegraphics[scale=\scaleparameterA]{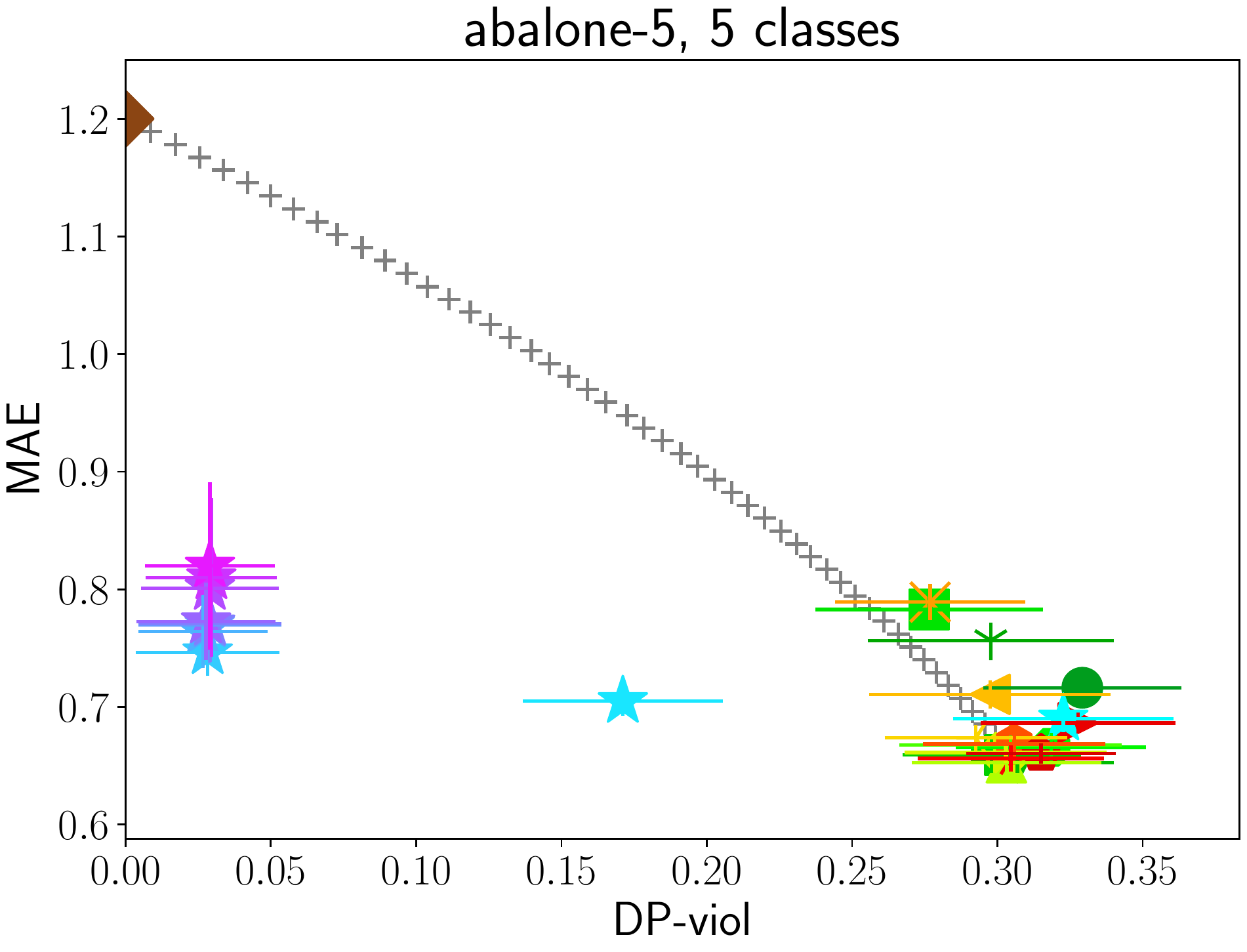}
    \hspace{\abstA}
    \includegraphics[scale=\scaleparameterA]{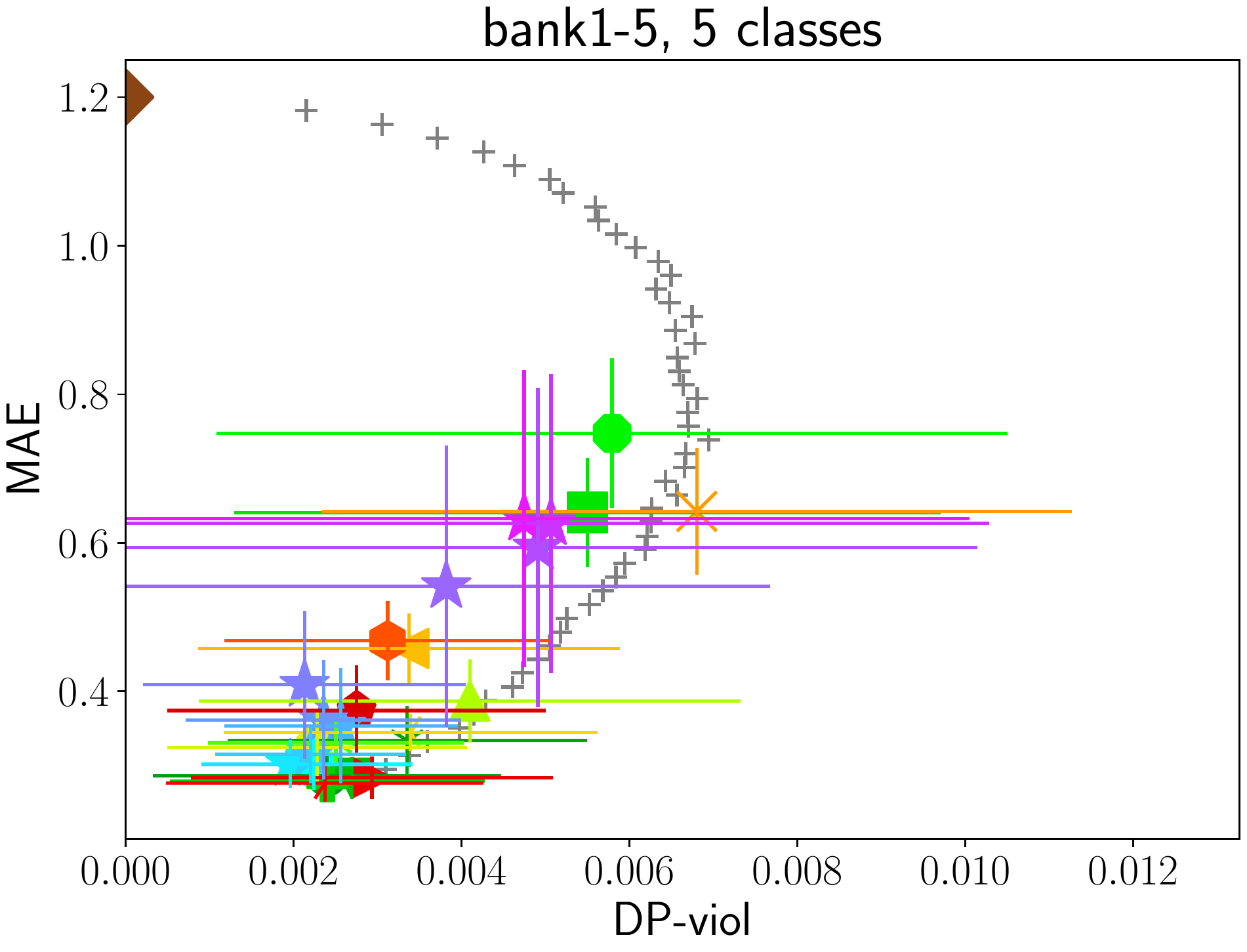}
    \hspace{\abstA}
    \includegraphics[scale=\scaleparameterA]{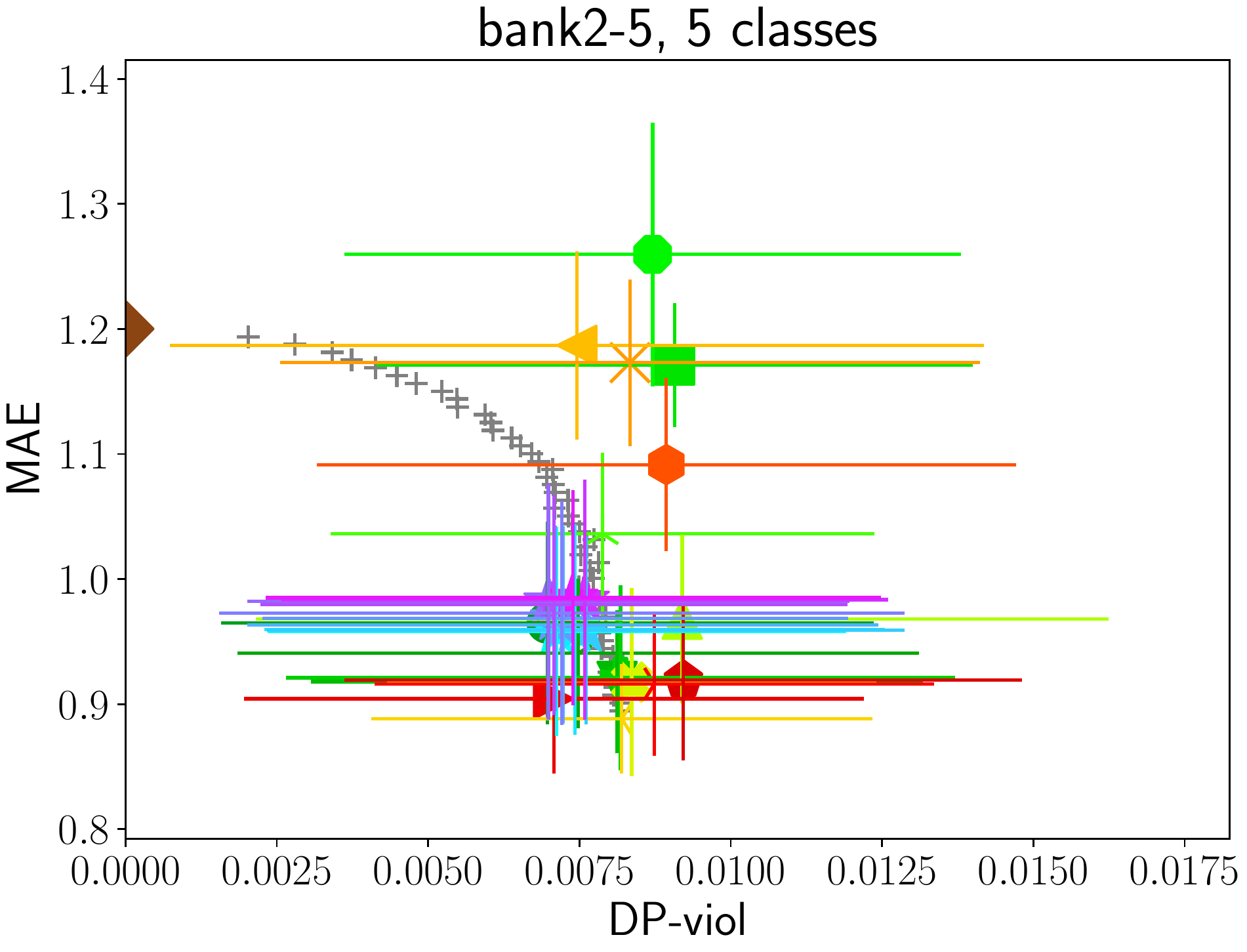}
    \hspace{\abstA}
    \includegraphics[scale=\scaleparameterA]{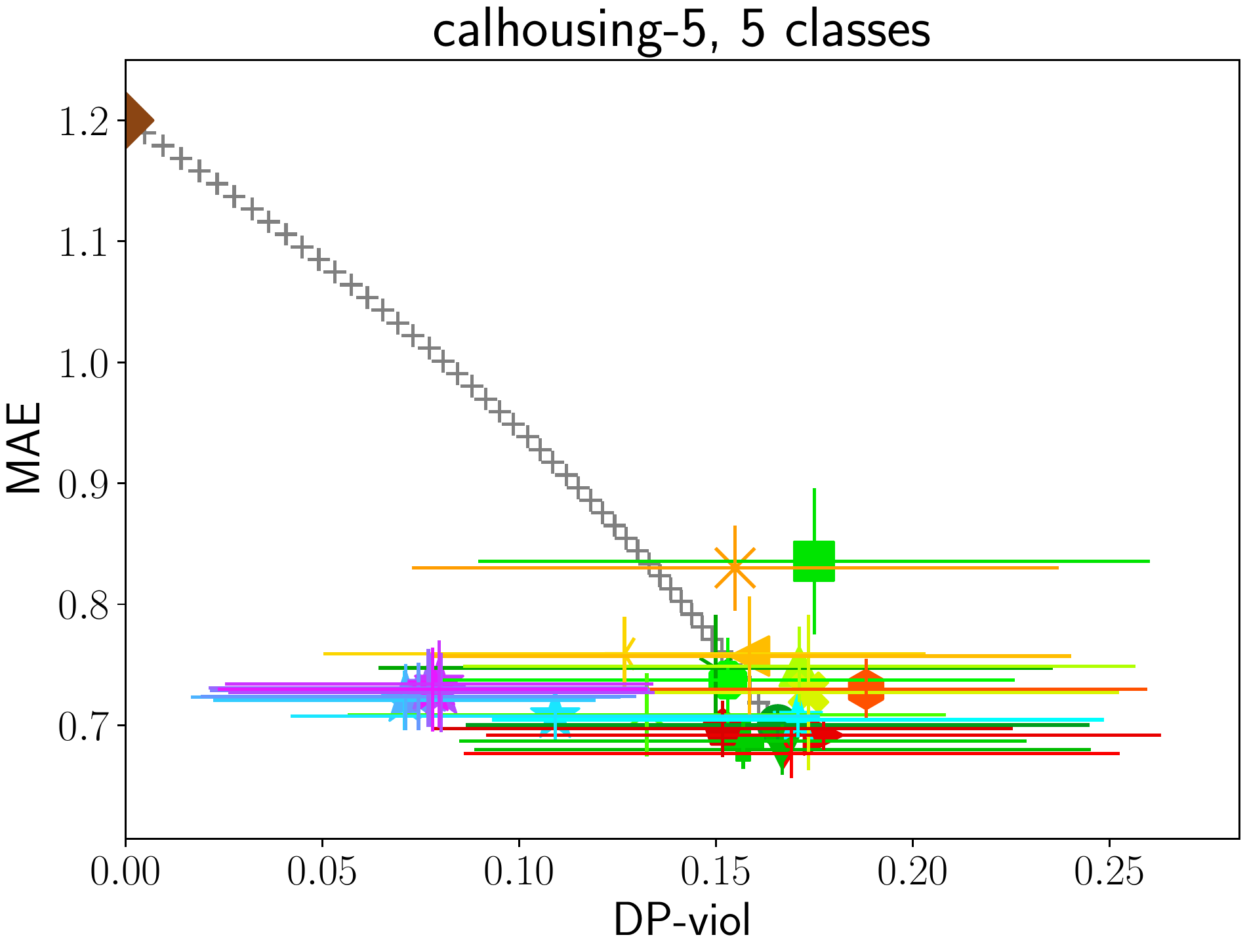}
    
    \includegraphics[scale=\scaleparameterA]{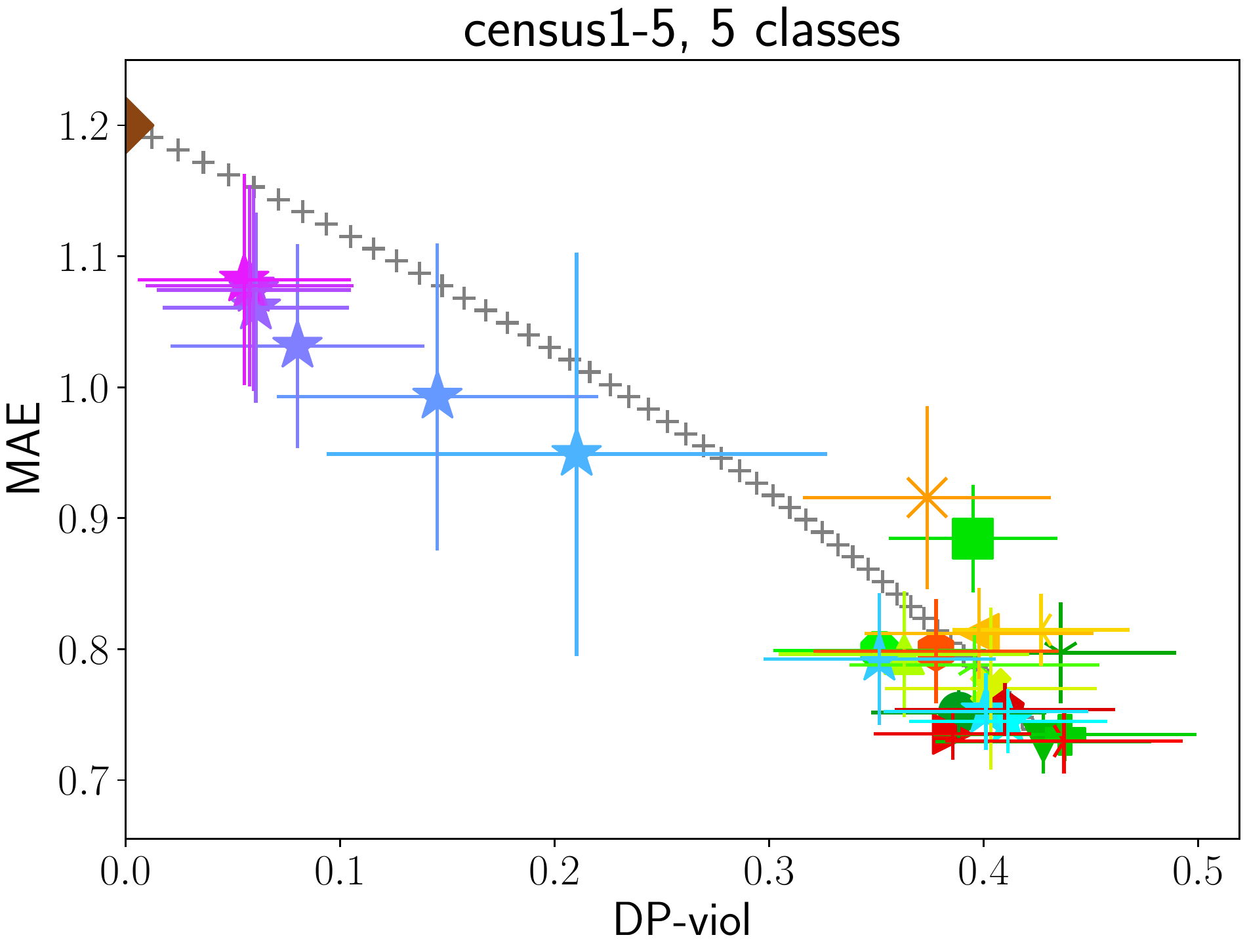}
    \hspace{\abstA}
    \includegraphics[scale=\scaleparameterA]{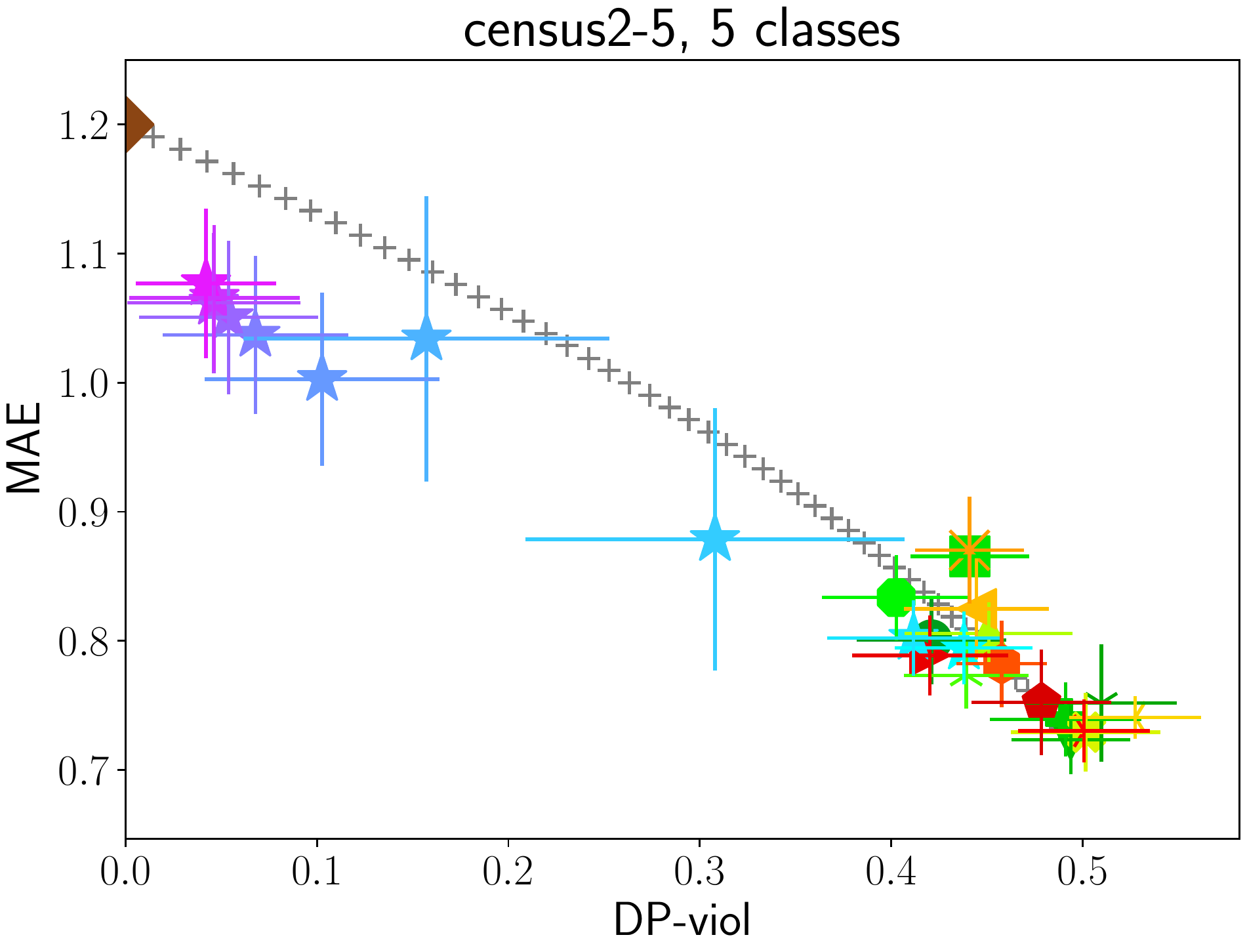}
    \hspace{\abstA}
    \includegraphics[scale=\scaleparameterA]{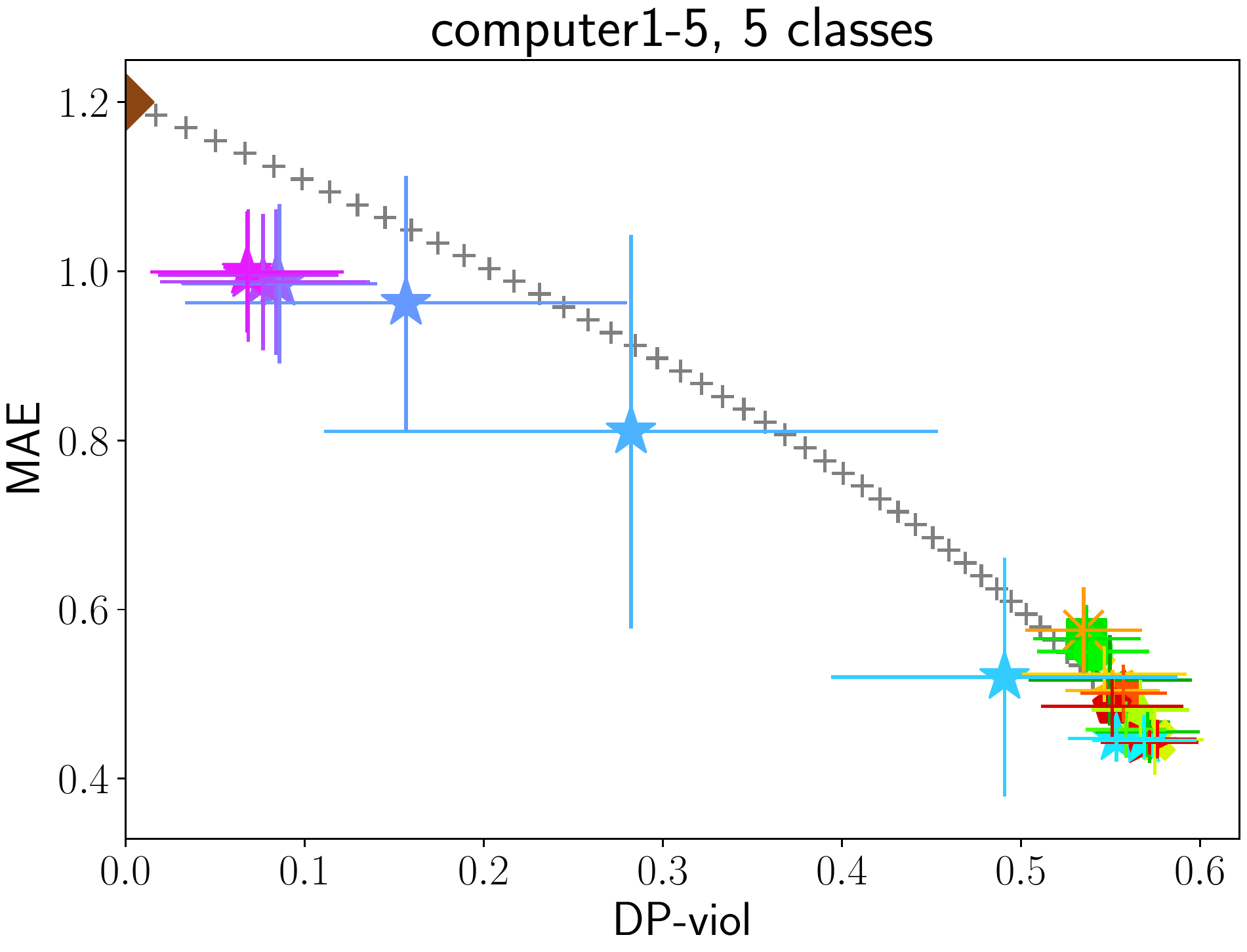}
    \hspace{\abstA}
    \includegraphics[scale=\scaleparameterA]{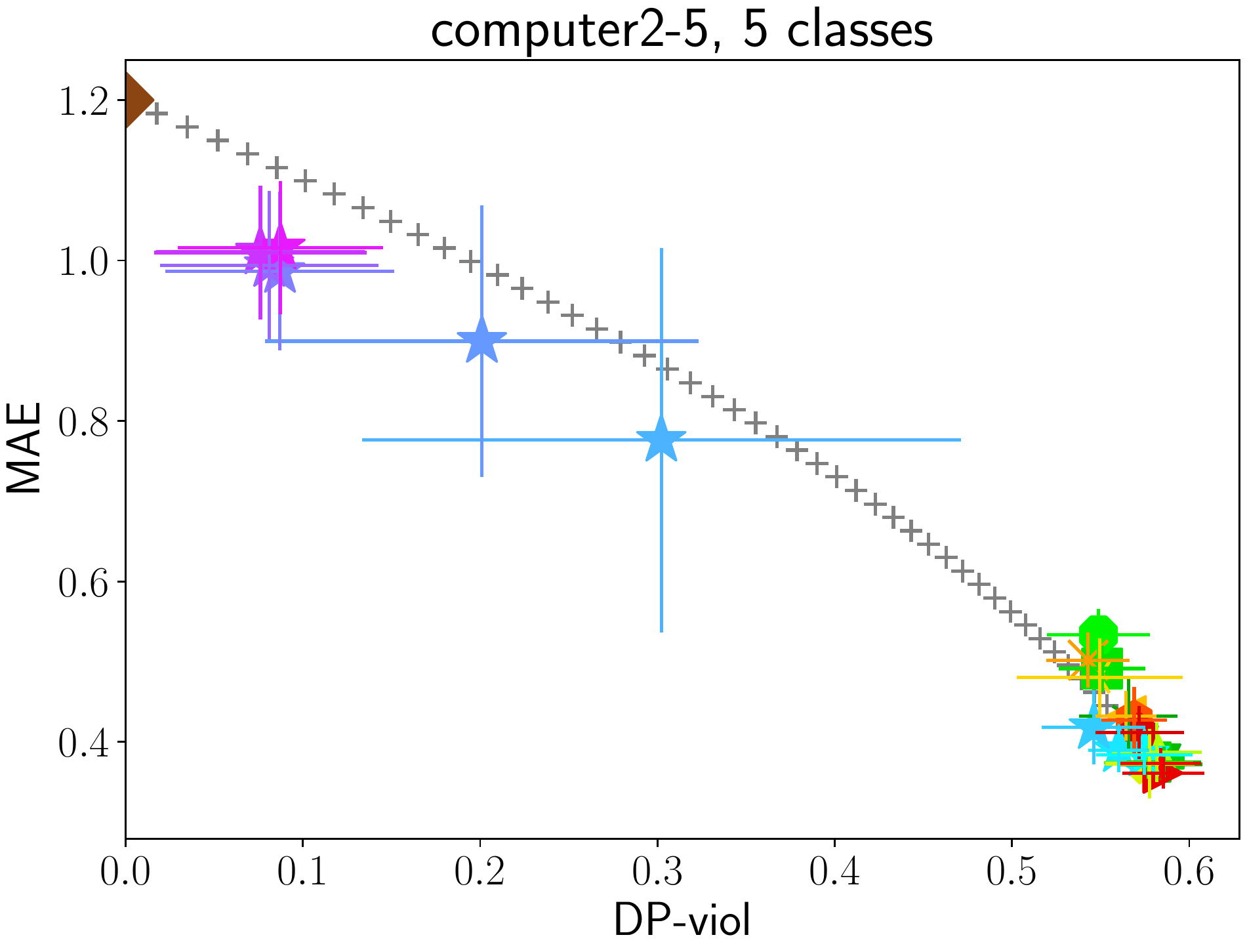}
    
    \includegraphics[scale=\scaleparameterA]{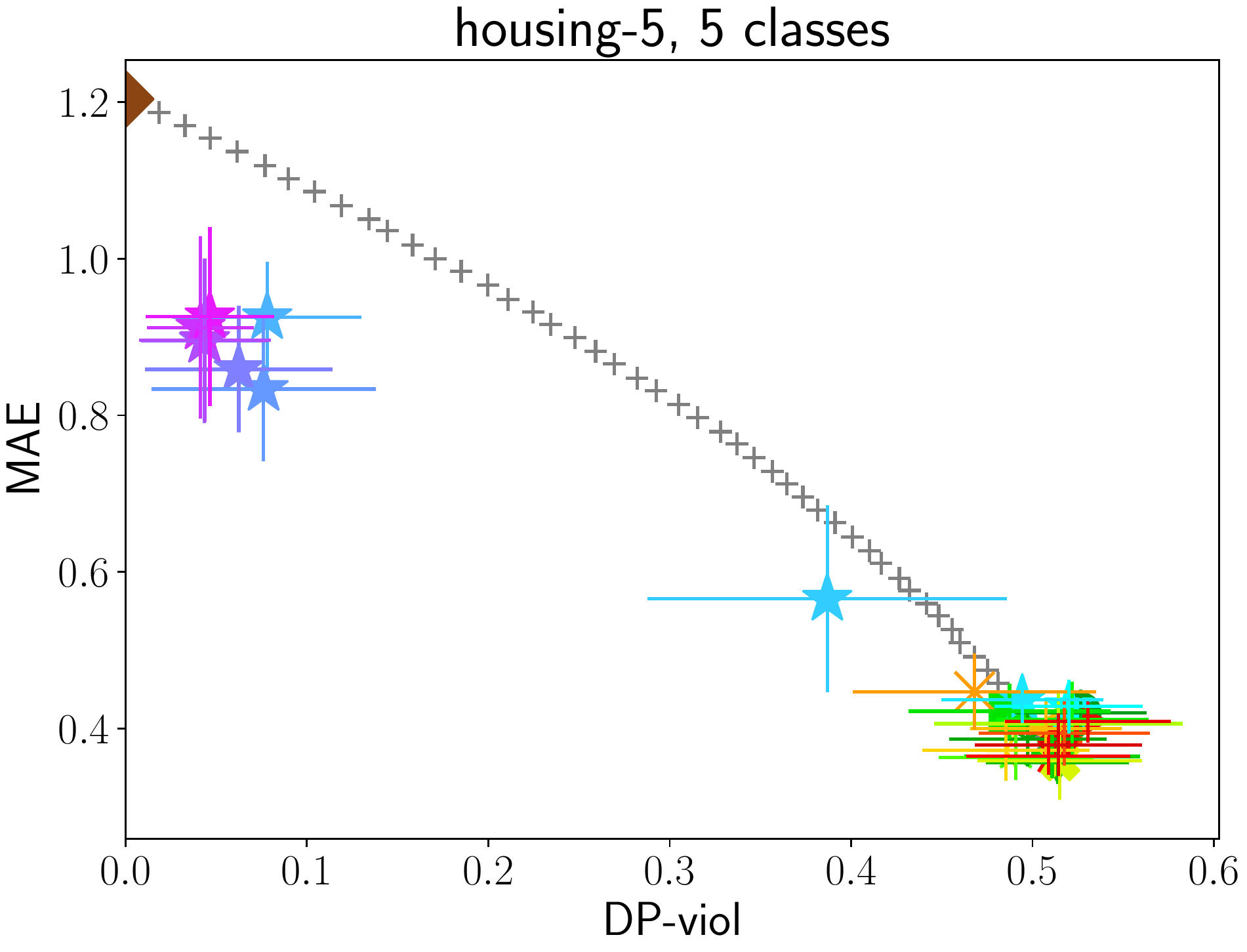}
    \hspace{\abstA}
    \includegraphics[scale=\scaleparameterA]{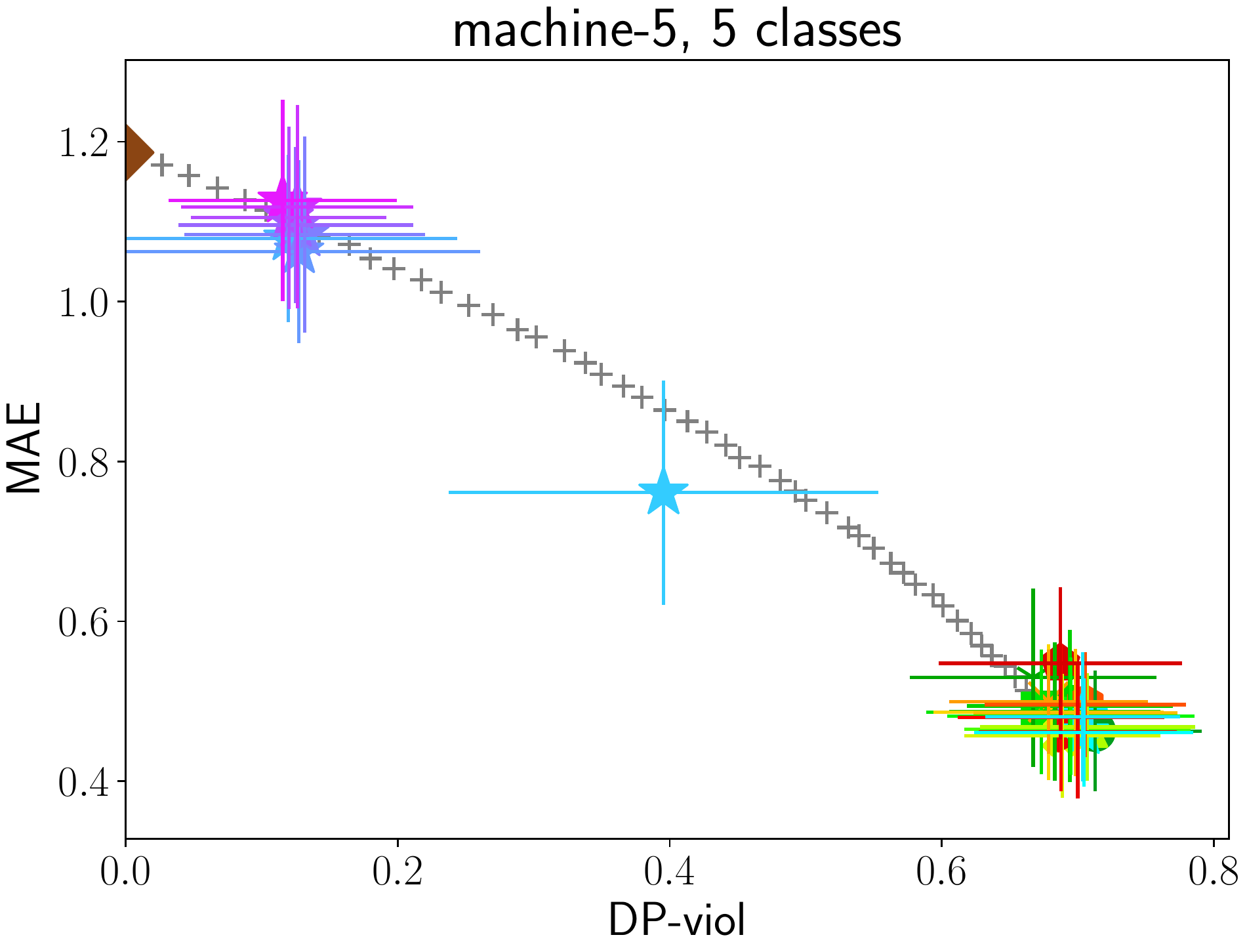}
    \hspace{\abstA}
    \includegraphics[scale=\scaleparameterA]{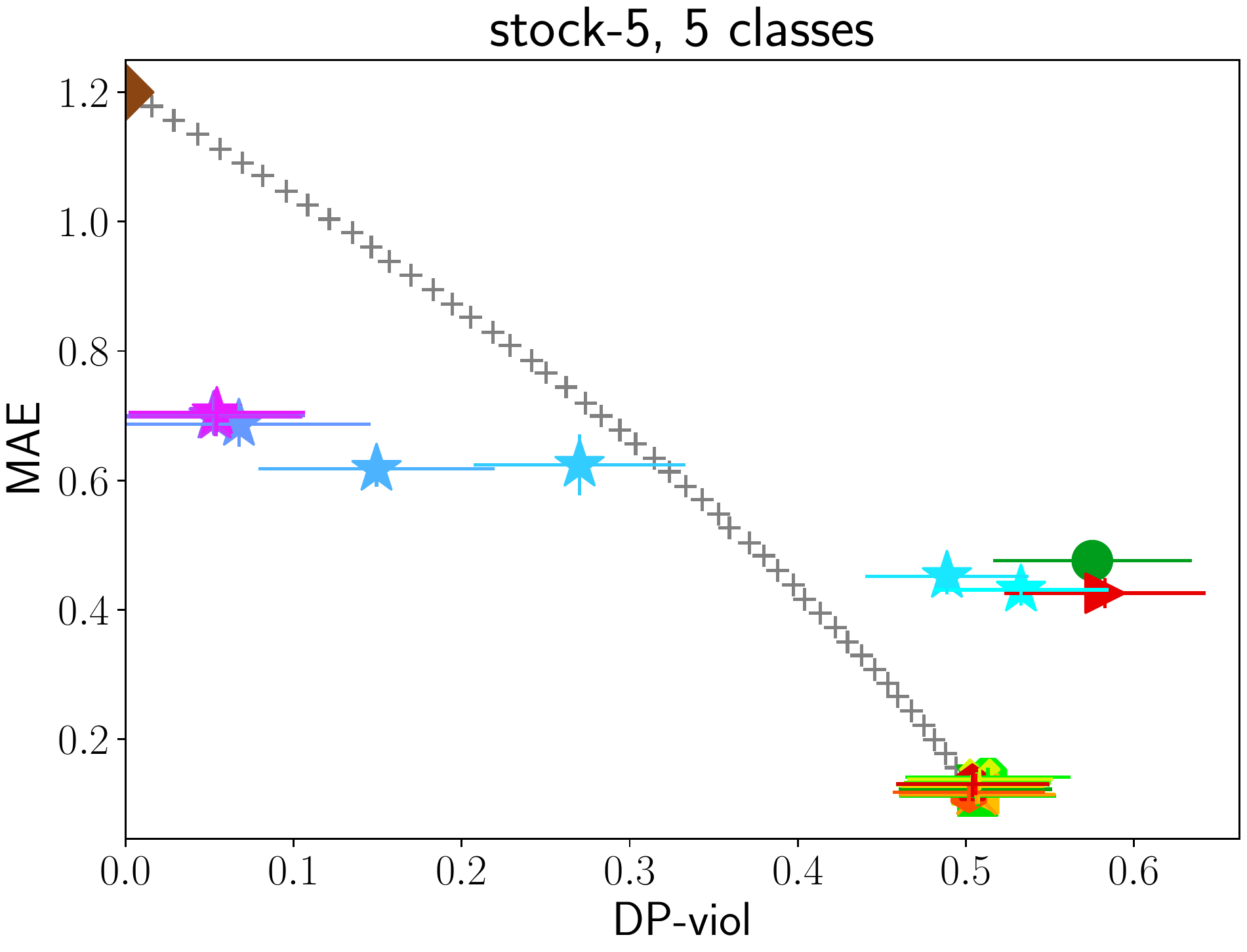}

    \caption{Experiments of Section~\ref{subsection_experiment_comparison} on the \textbf{discretized  regression datasets with 5 classes} when aiming for \textbf{pairwise DP}. The errorbars show the standard deviation over the 20 splits into training and test sets.}
    \label{fig:exp_comparison_APPENDIX_DISC_5classes_DP_with_STD}
\end{figure*}

\begin{figure*}
    
    \includegraphics[width=\linewidth]{experiment_real_ord/legend_big.pdf}

    \includegraphics[scale=\scaleparameterA]{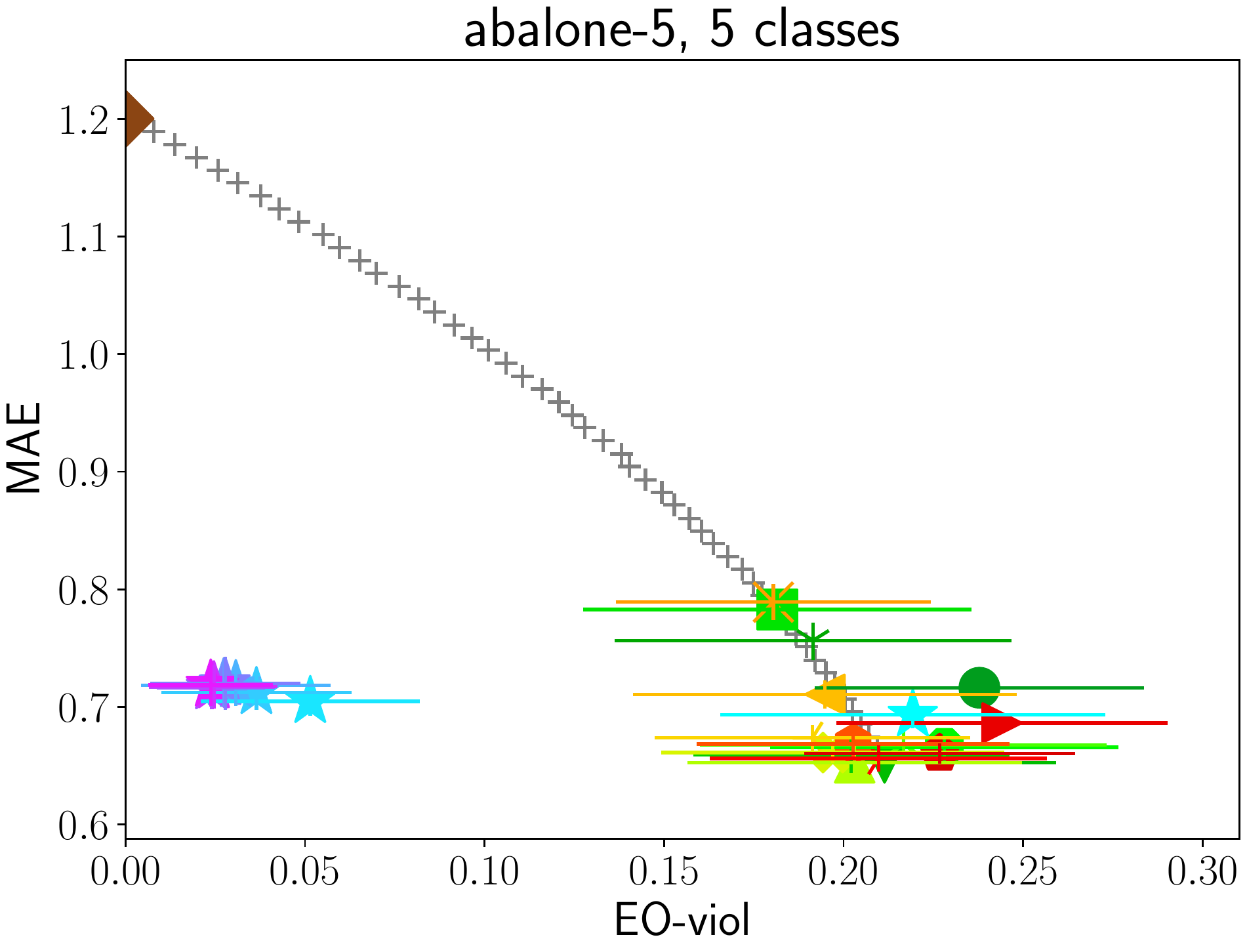}
    \hspace{\abstA}
    \includegraphics[scale=\scaleparameterA]{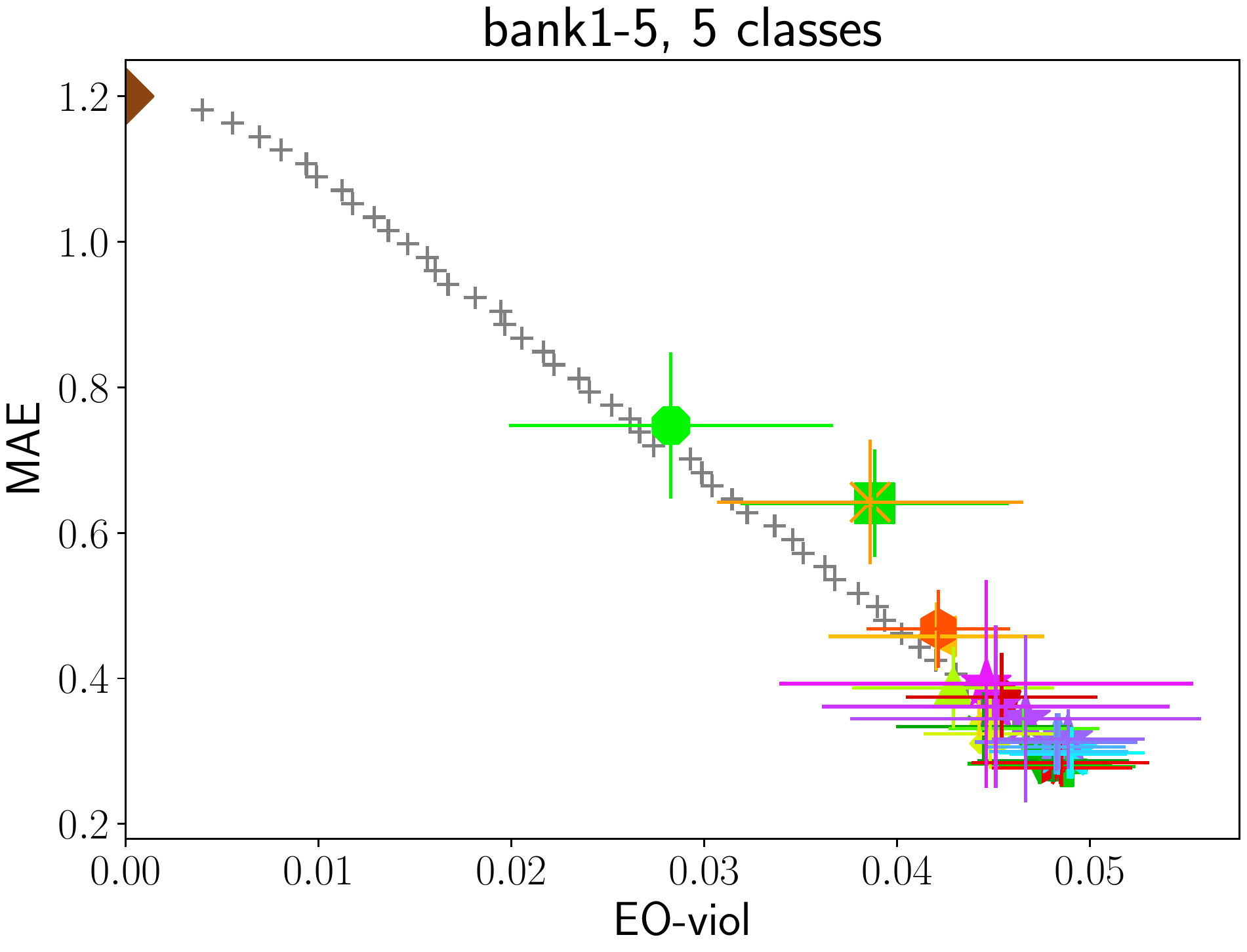}
    \hspace{\abstA}
    \includegraphics[scale=\scaleparameterA]{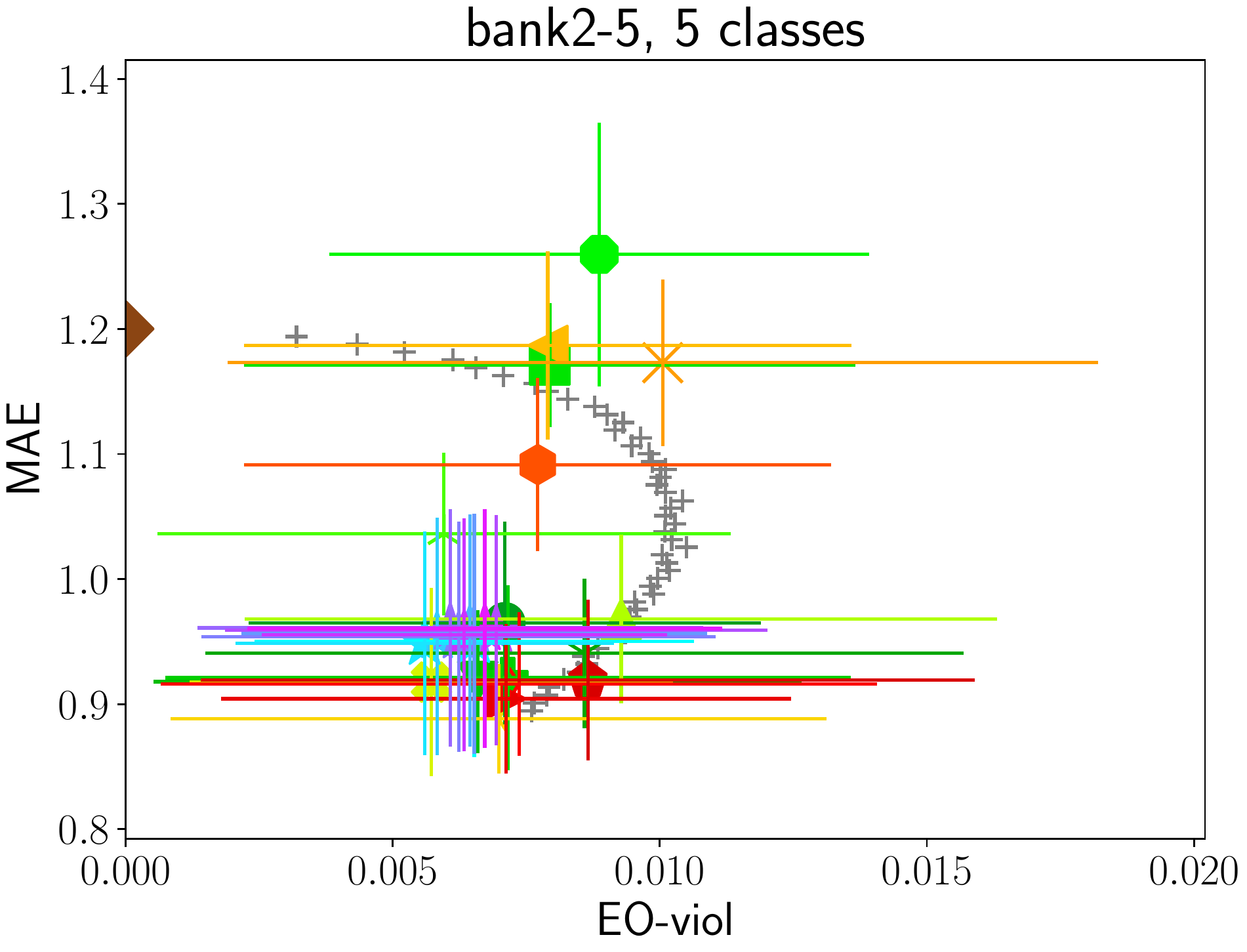}
    \hspace{\abstA}
    \includegraphics[scale=\scaleparameterA]{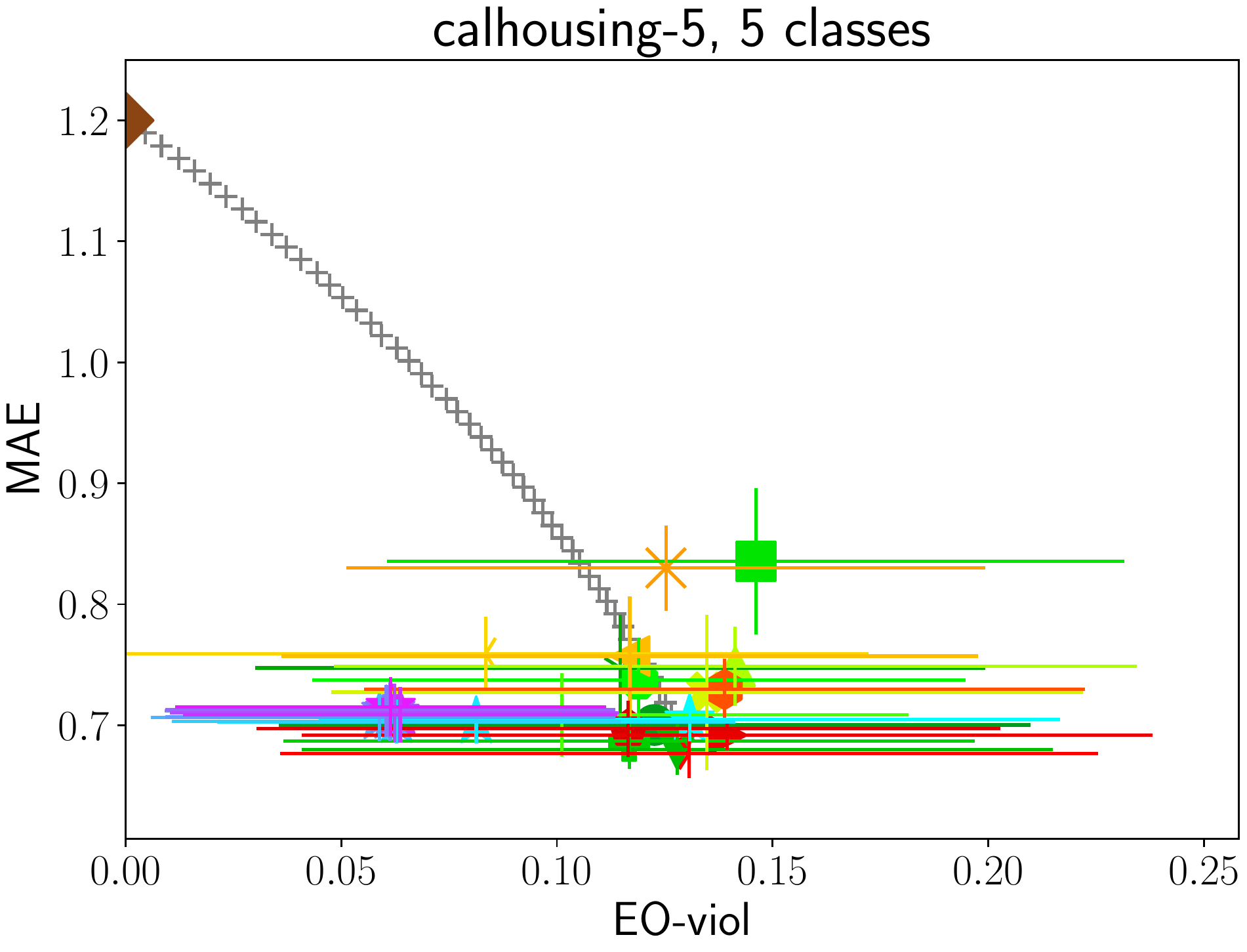}
    
    \includegraphics[scale=\scaleparameterA]{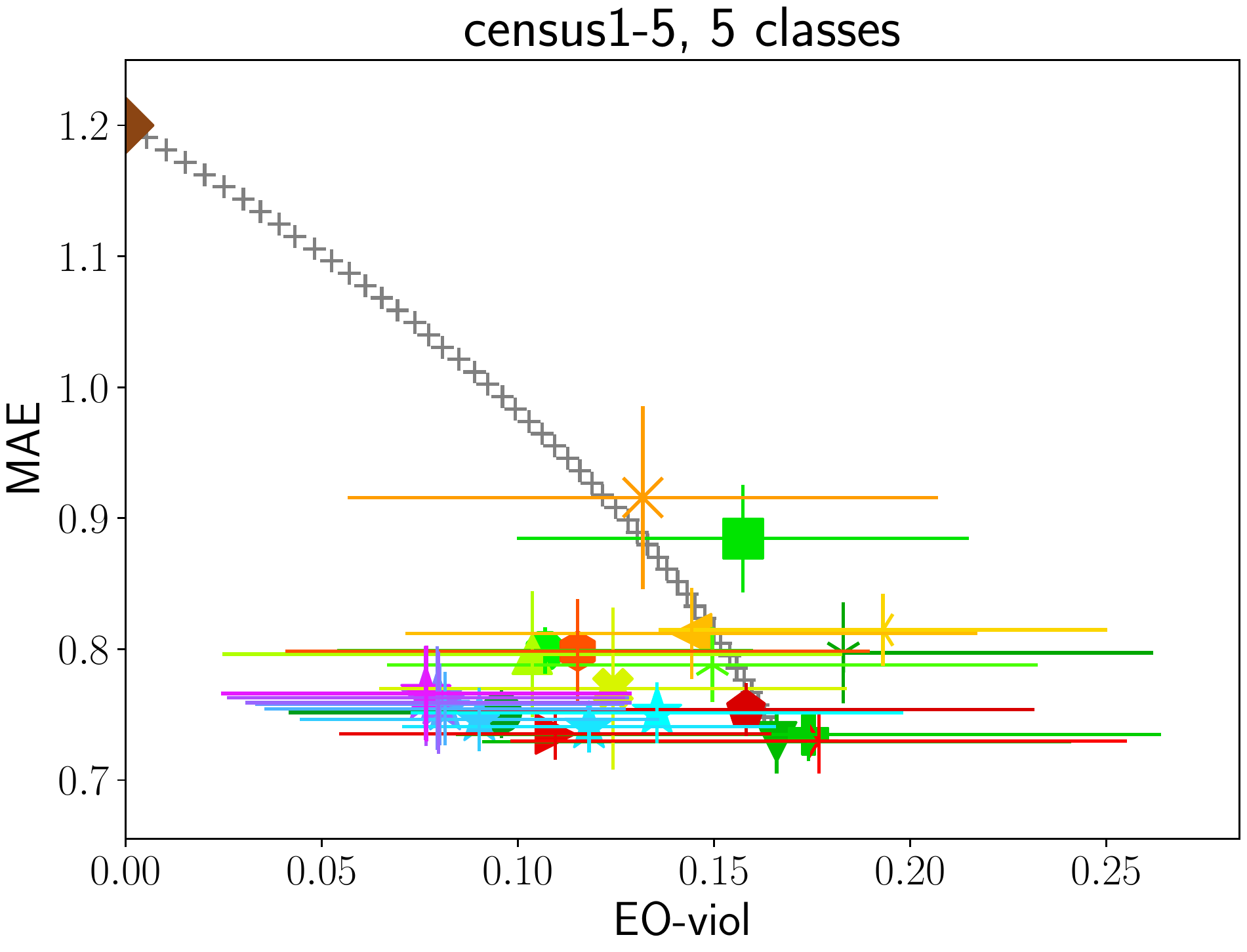}
    \hspace{\abstA}
    \includegraphics[scale=\scaleparameterA]{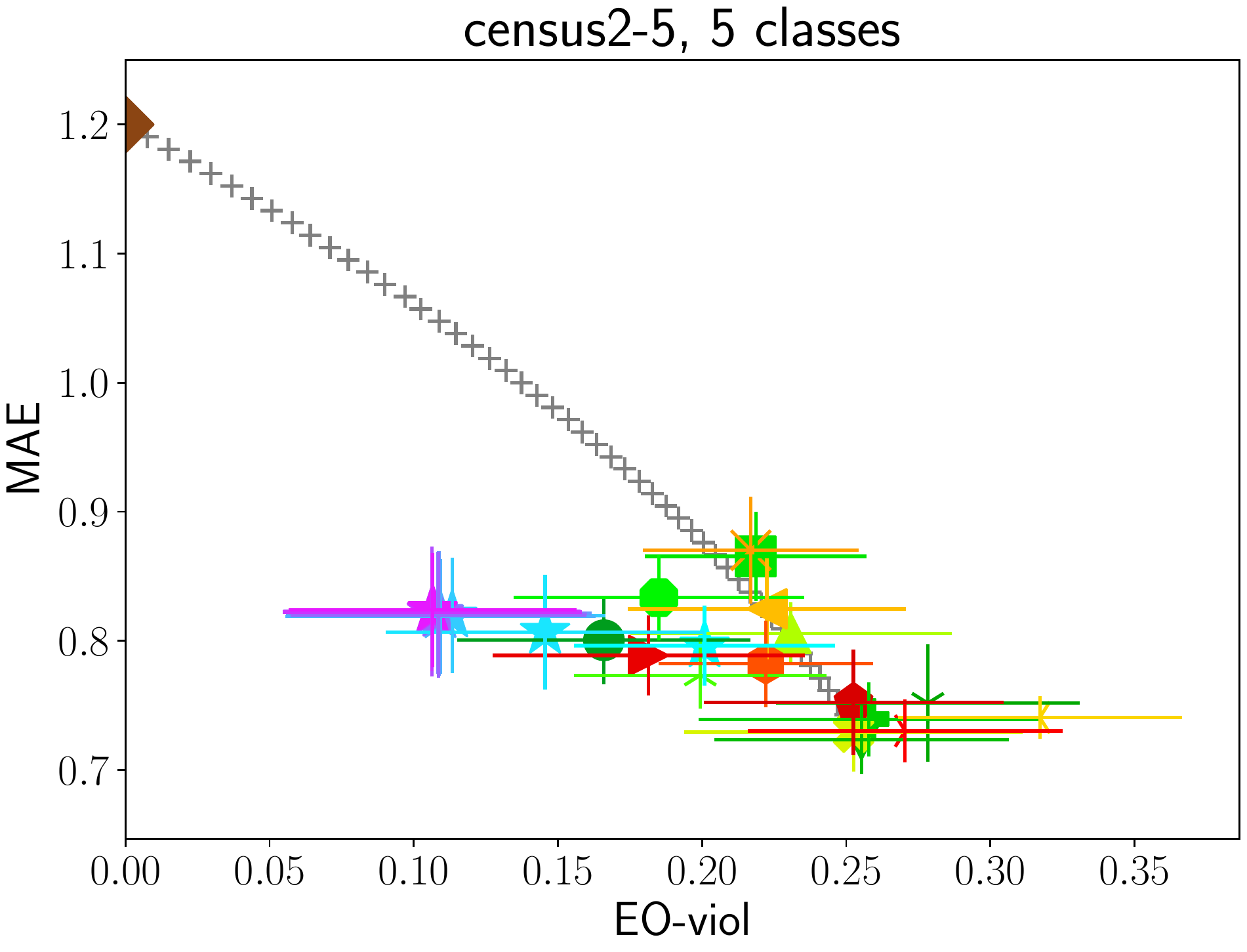}
    \hspace{\abstA}
    \includegraphics[scale=\scaleparameterA]{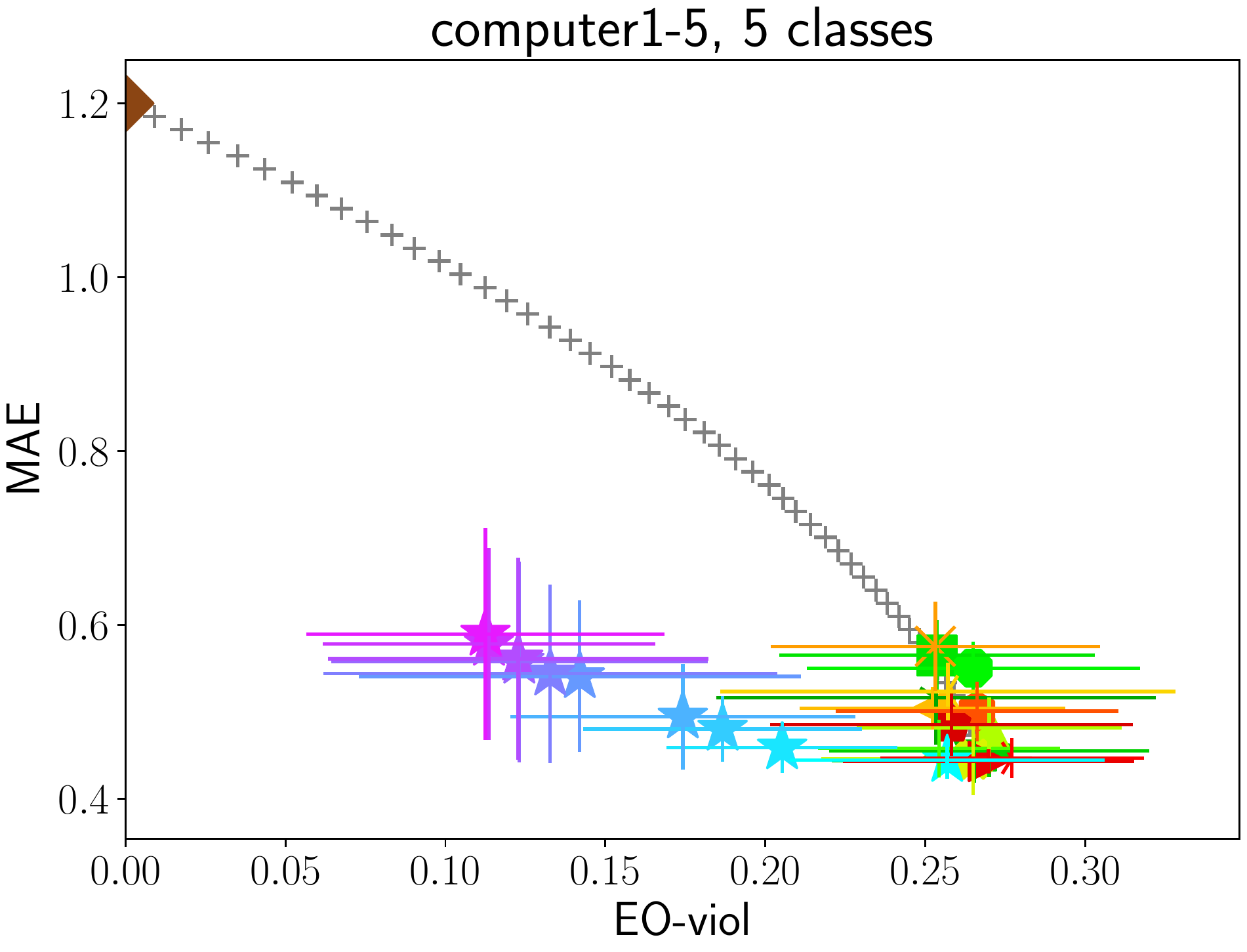}
    \hspace{\abstA}
    \includegraphics[scale=\scaleparameterA]{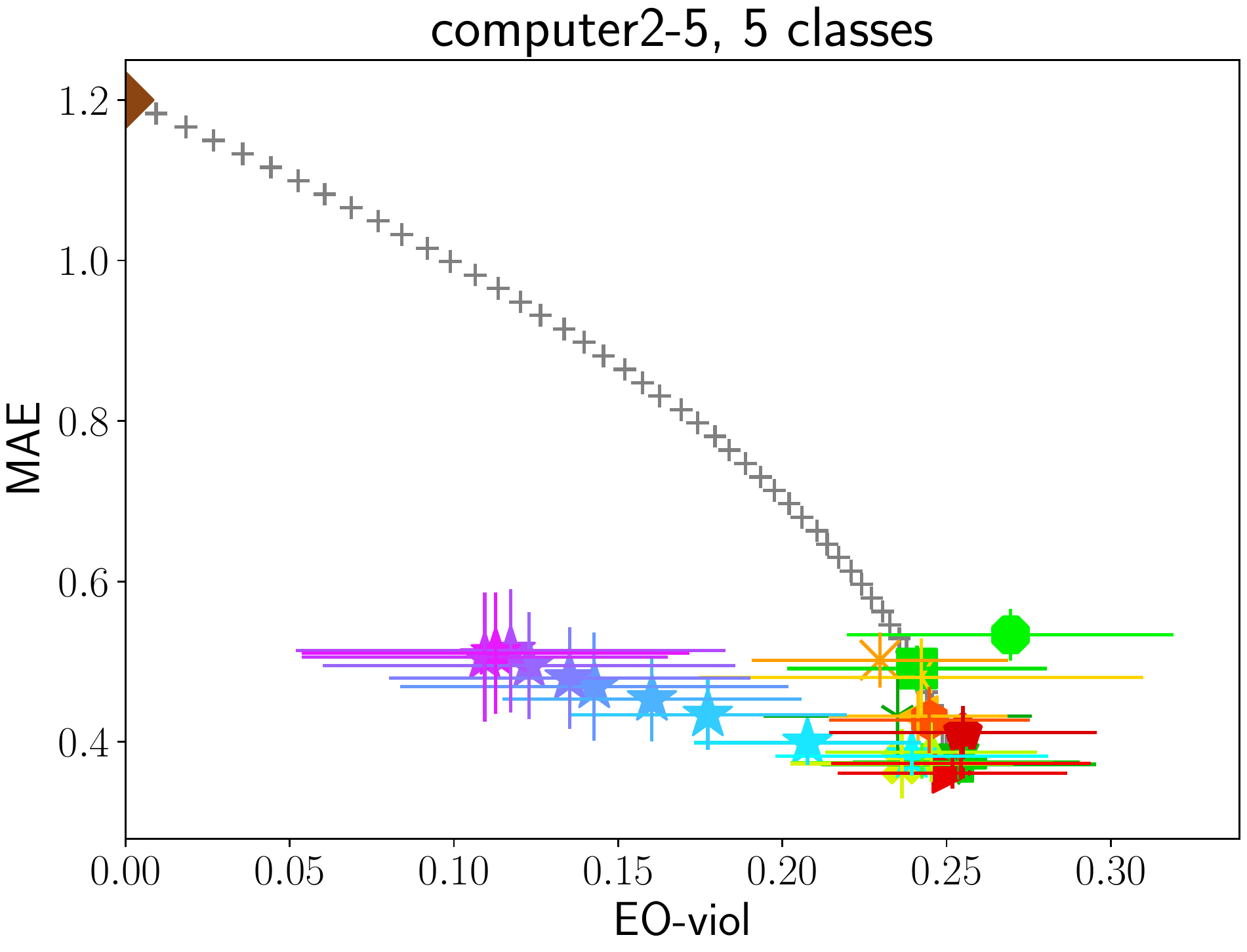}
    
    \includegraphics[scale=\scaleparameterA]{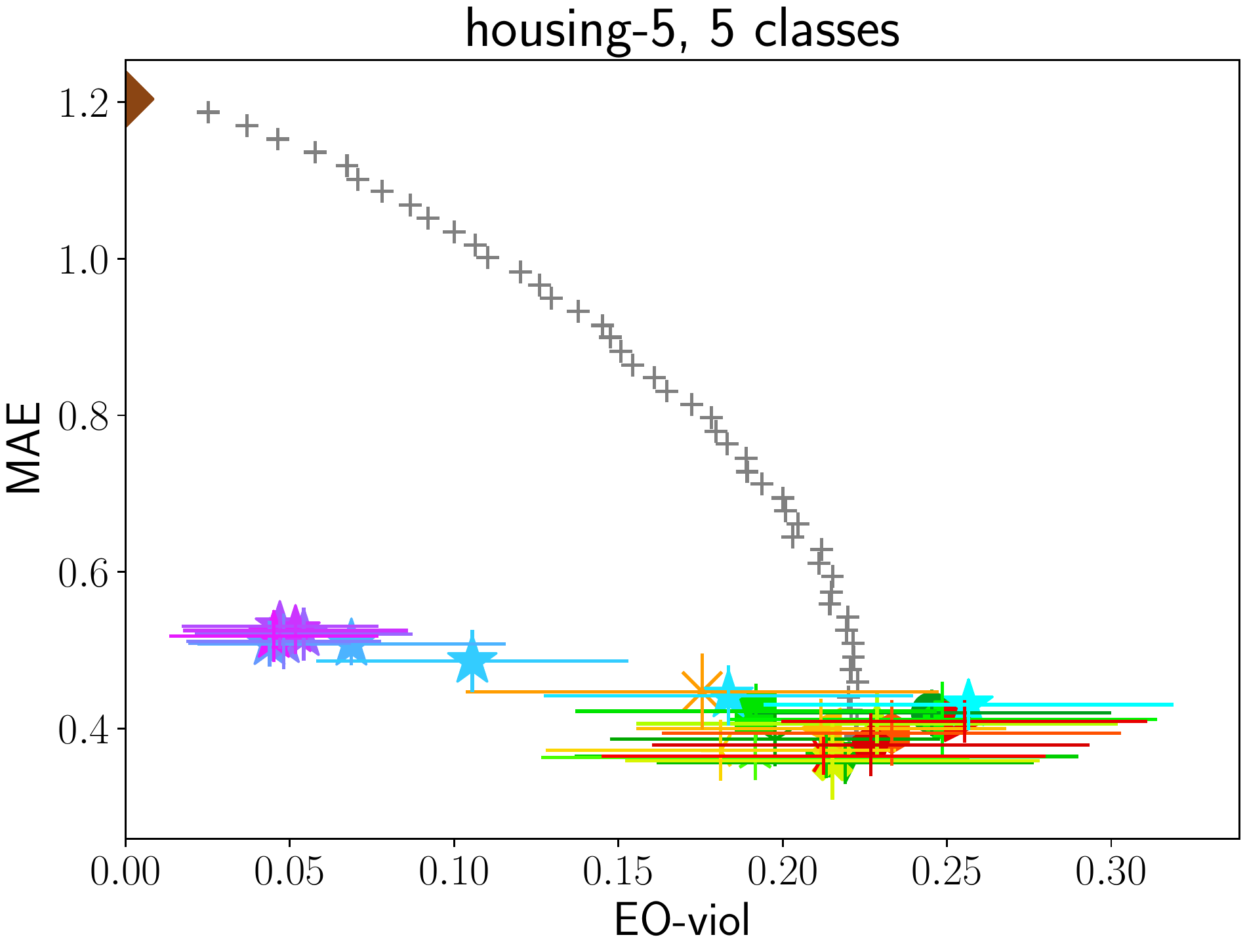}
    \hspace{\abstA}
    \includegraphics[scale=\scaleparameterA]{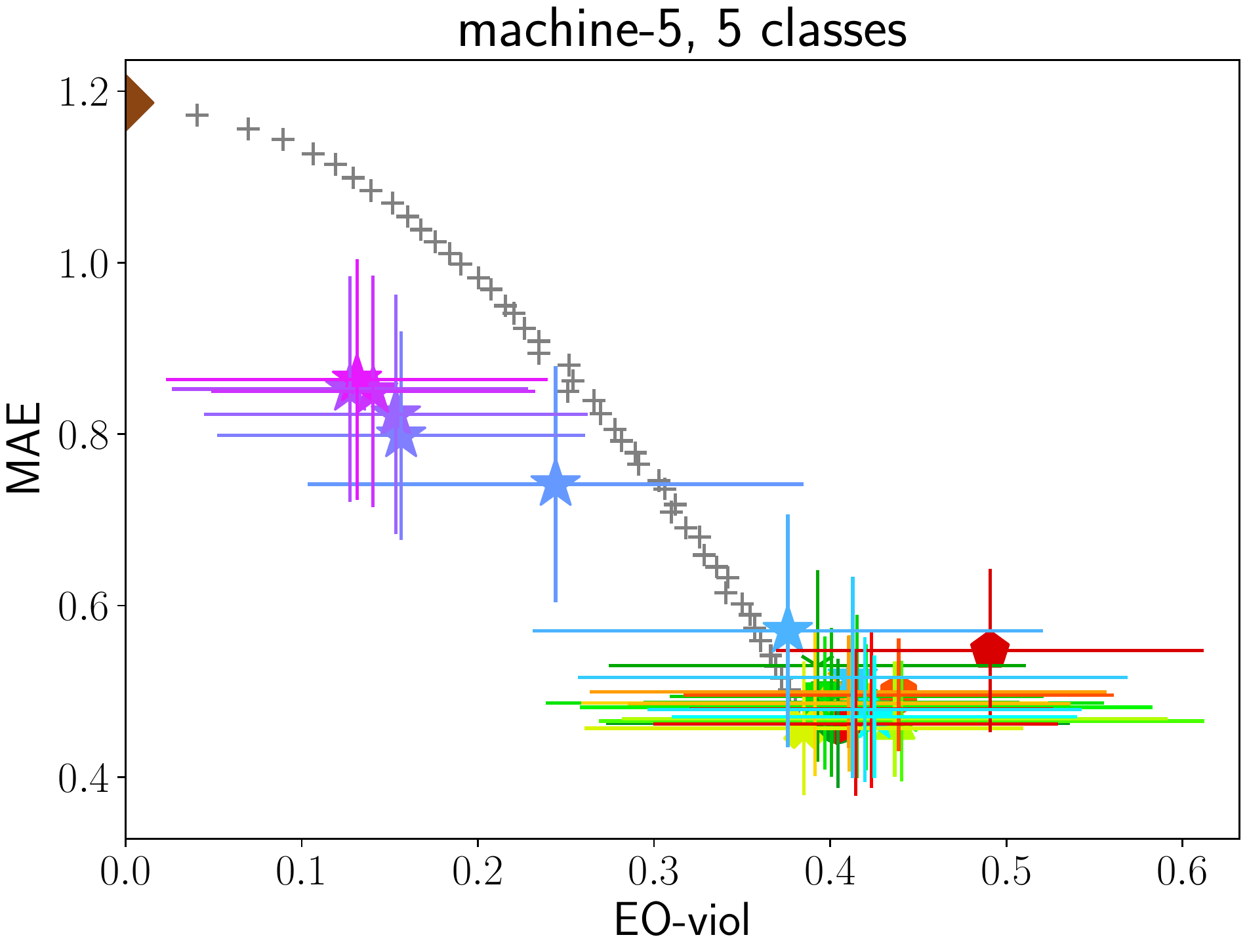}
    \hspace{\abstA}
    \includegraphics[scale=\scaleparameterA]{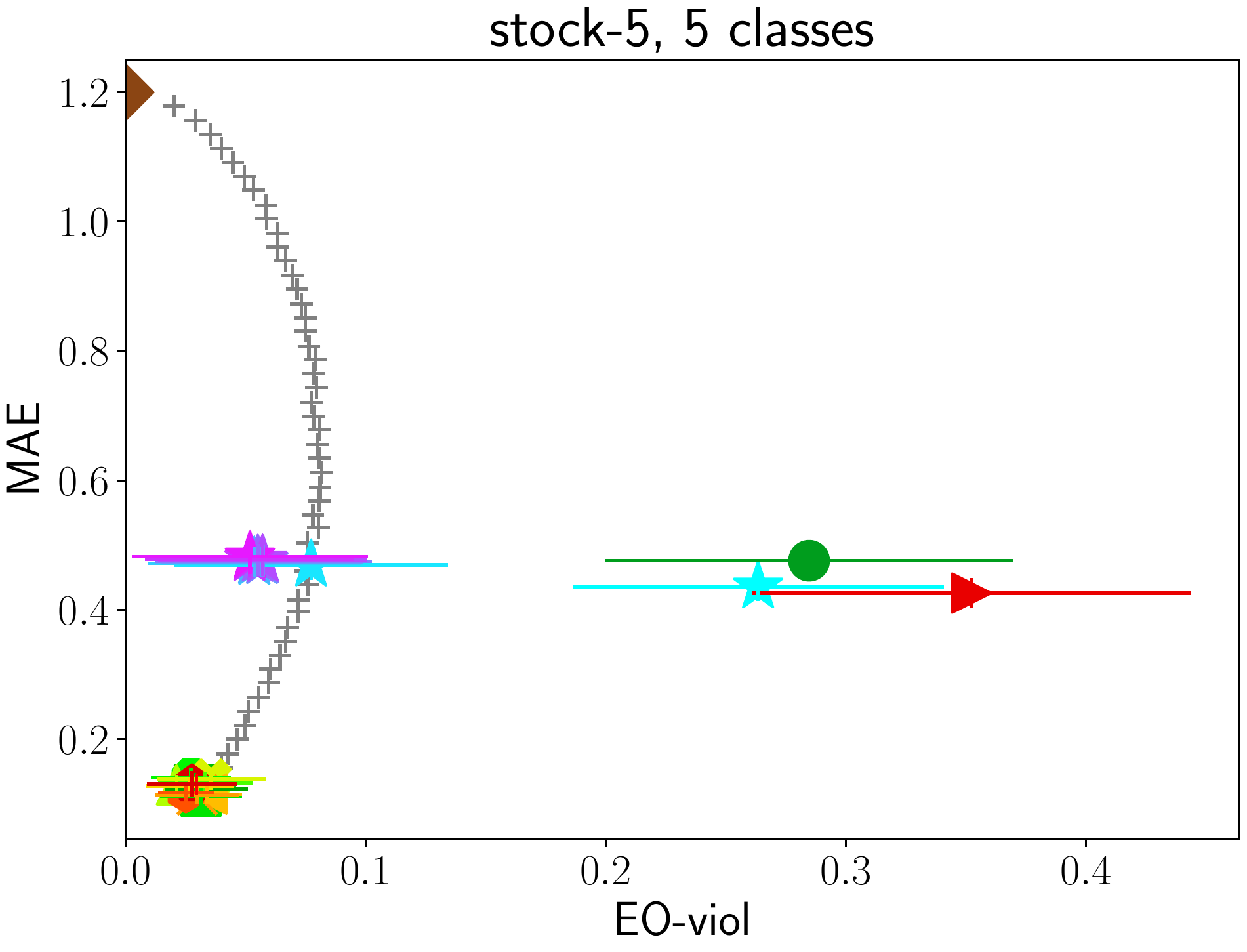}

    \caption{Experiments of Section~\ref{subsection_experiment_comparison} on the \textbf{discretized  regression datasets with 5 classes} when aiming for \textbf{pairwise EO}. The errorbars show the standard deviation over the 20 splits into training and test sets.}
    \label{fig:exp_comparison_APPENDIX_DISC_5classes_EO_with_STD}
\end{figure*}

\begin{figure*}
    
    \includegraphics[width=\linewidth]{experiment_real_ord/legend_big.pdf}

    \includegraphics[scale=\scaleparameterA]{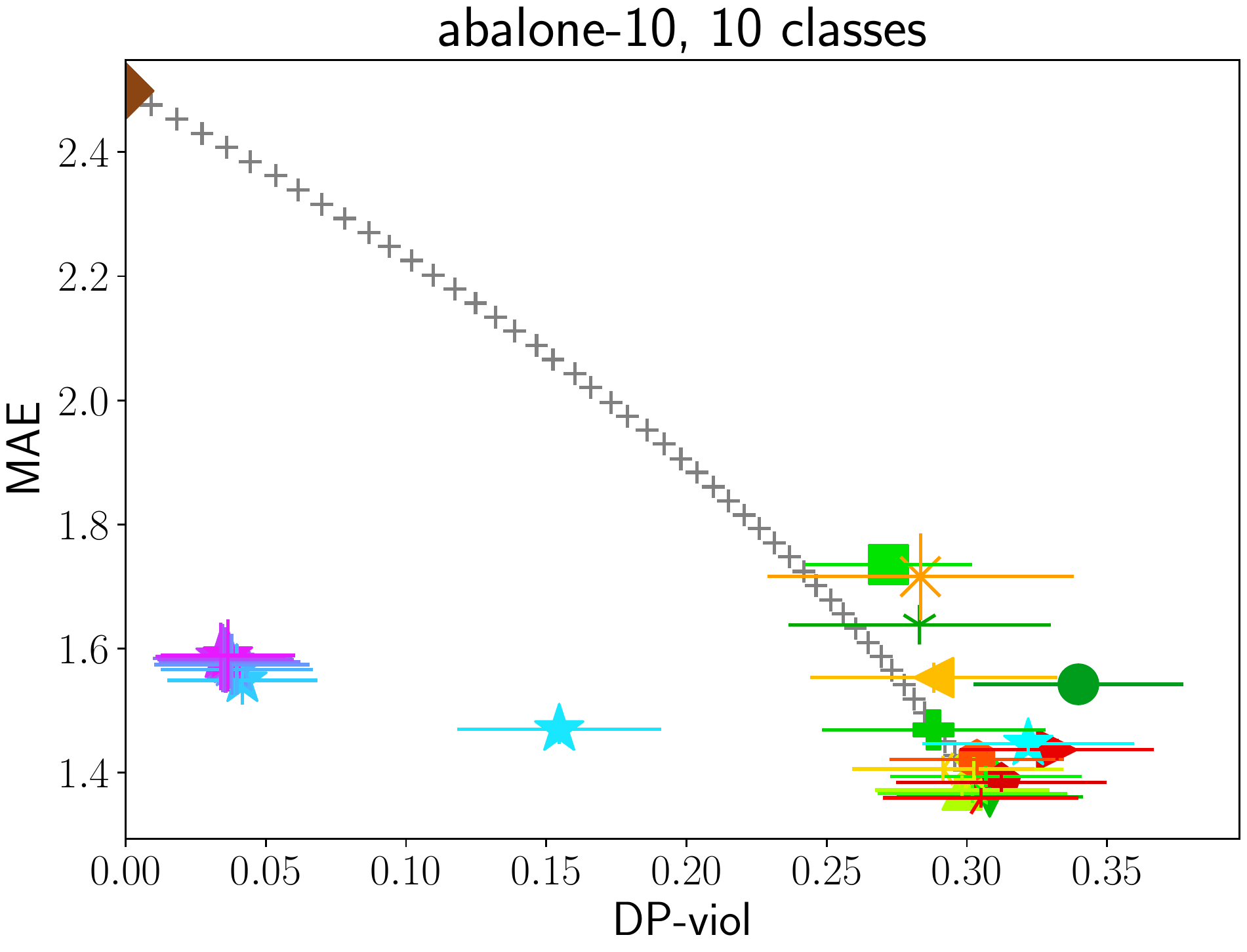}
    \hspace{\abstA}
    \includegraphics[scale=\scaleparameterA]{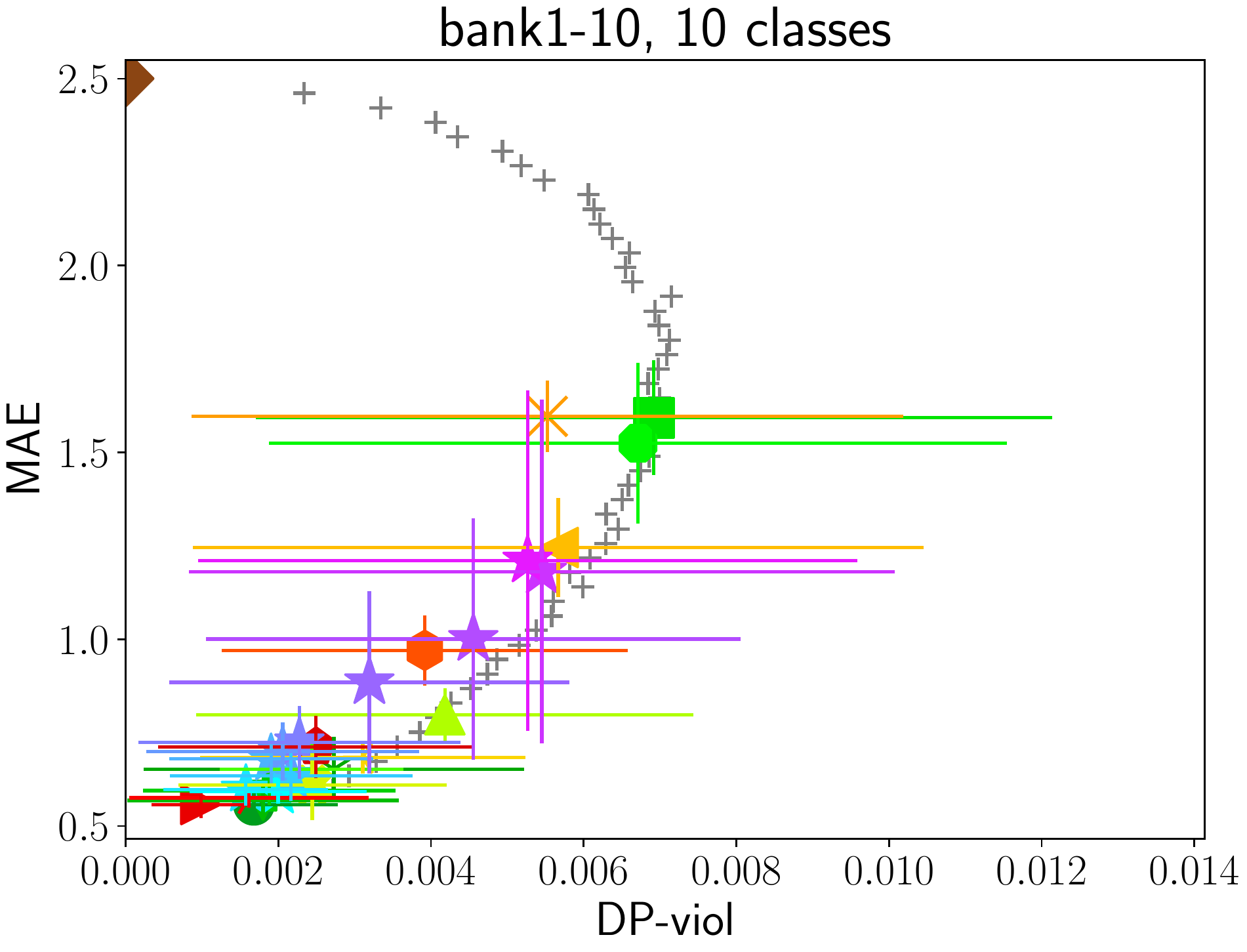}
    \hspace{\abstA}
    \includegraphics[scale=\scaleparameterA]{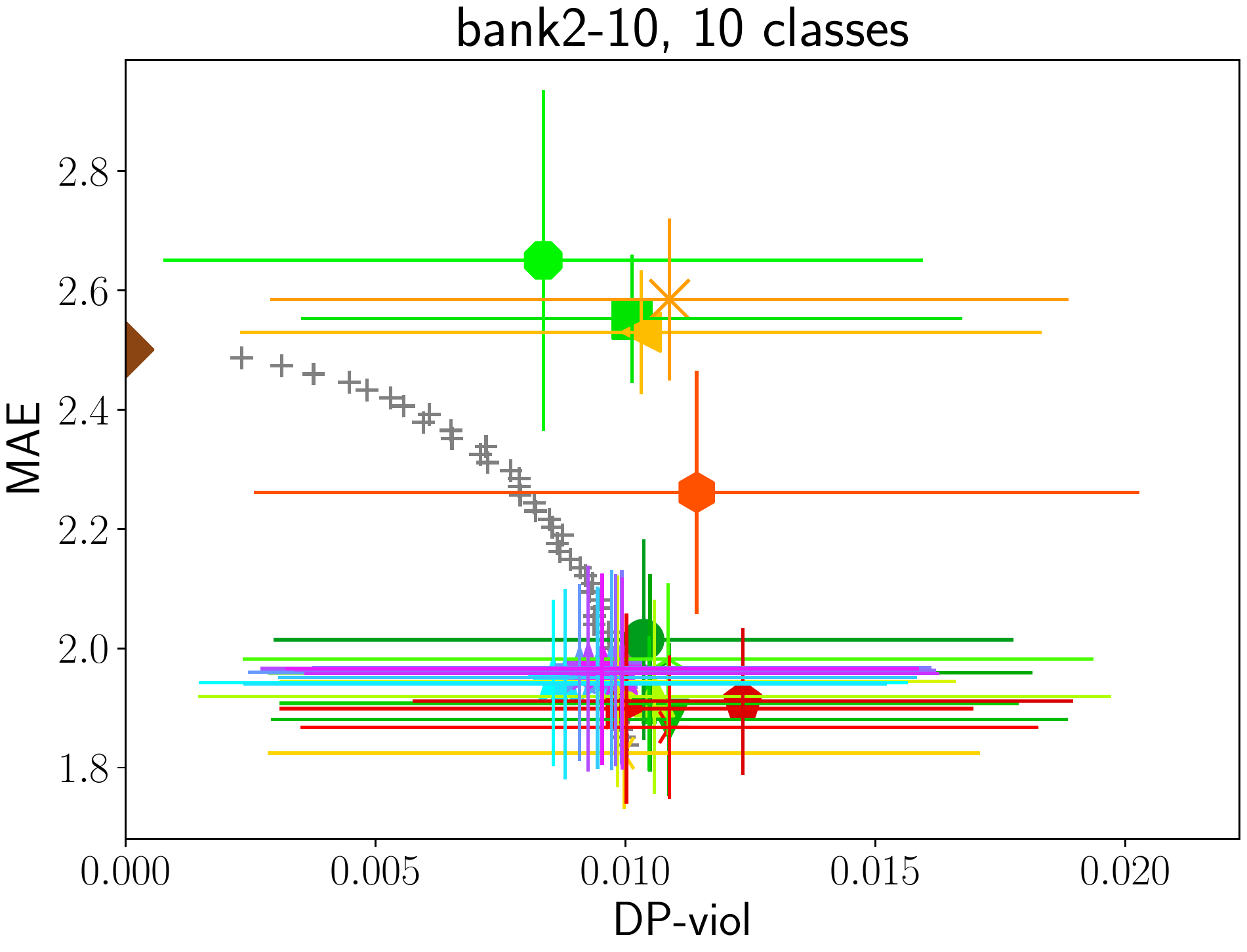}
    \hspace{\abstA}
    \includegraphics[scale=\scaleparameterA]{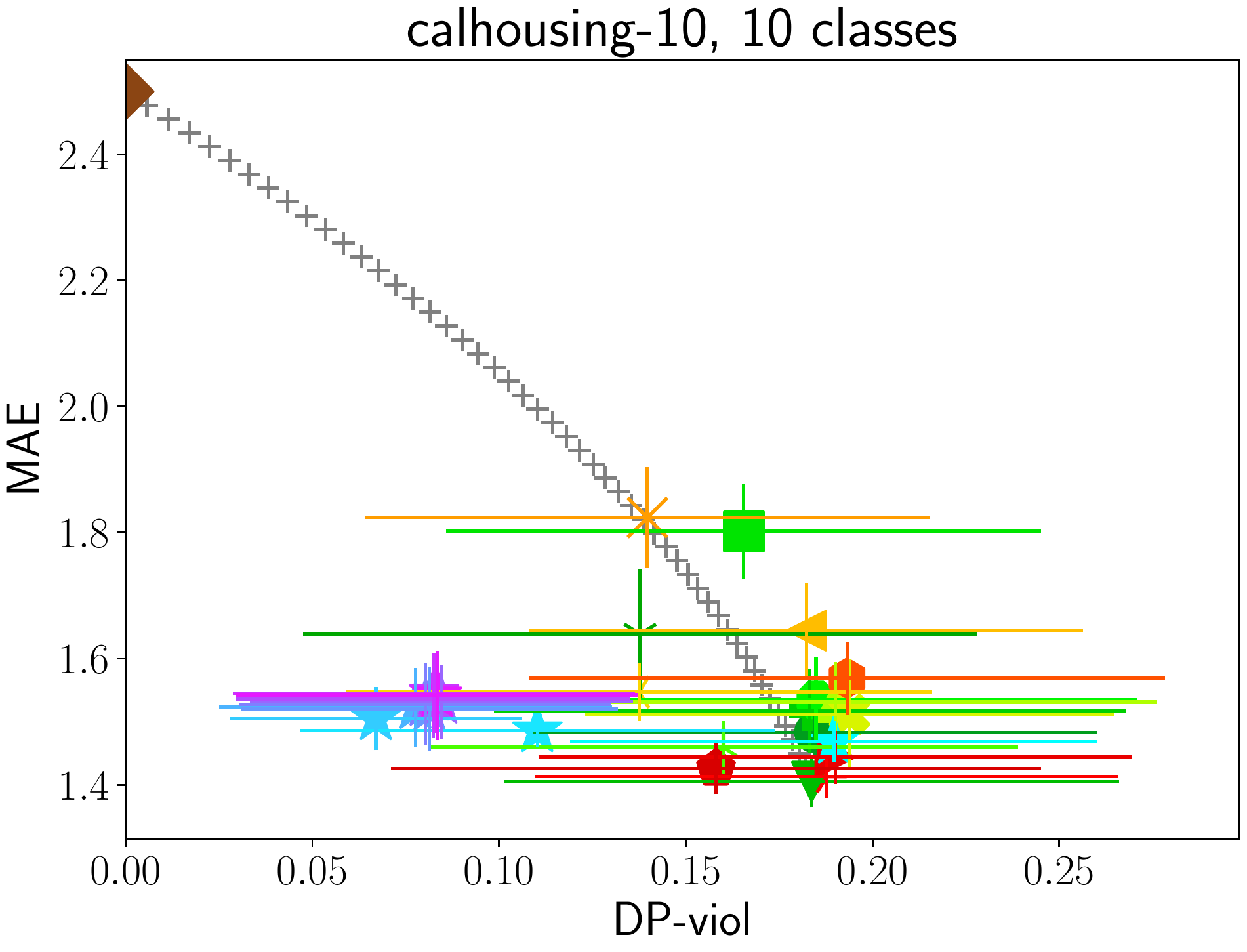}
    
    \includegraphics[scale=\scaleparameterA]{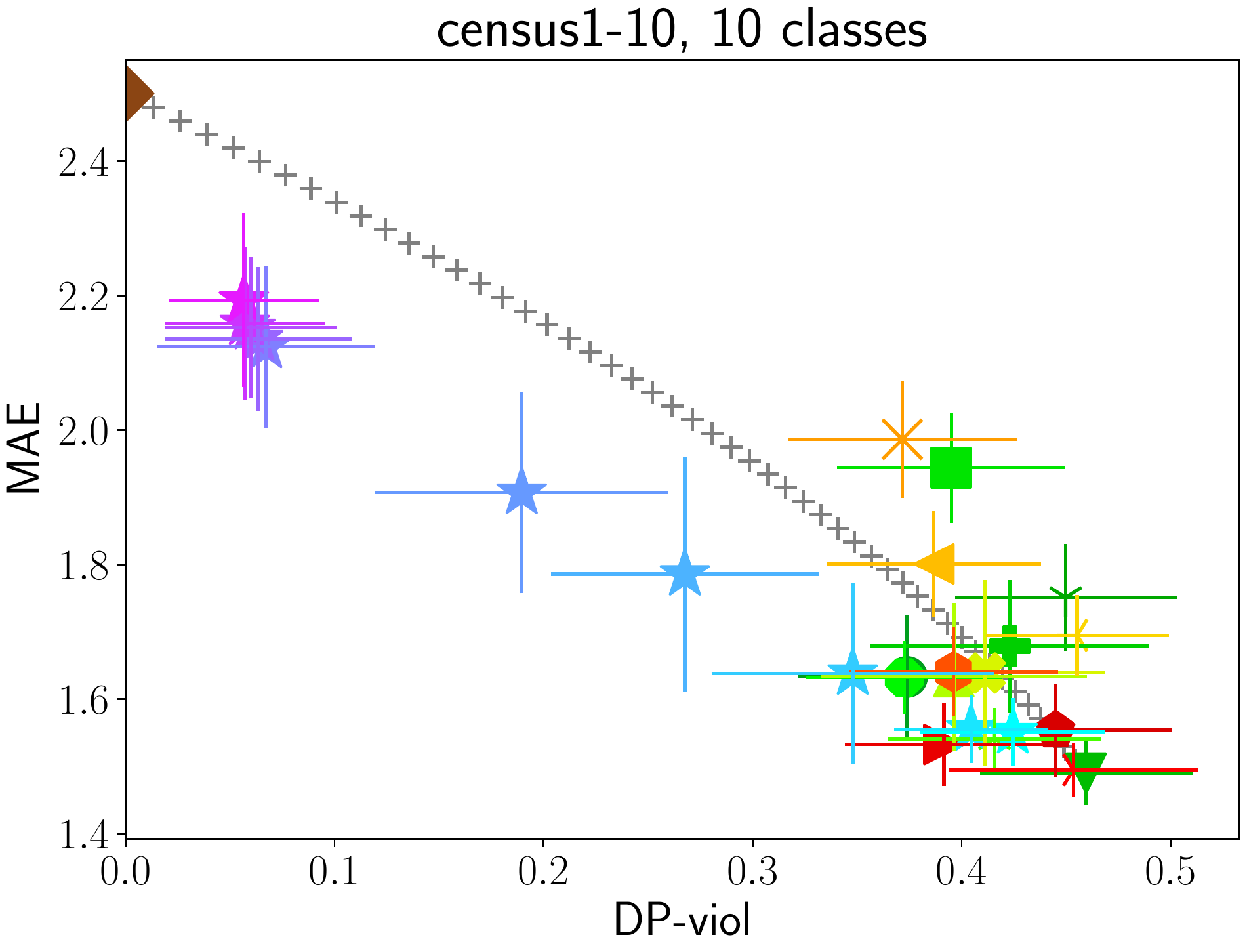}
    \hspace{\abstA}
    \includegraphics[scale=\scaleparameterA]{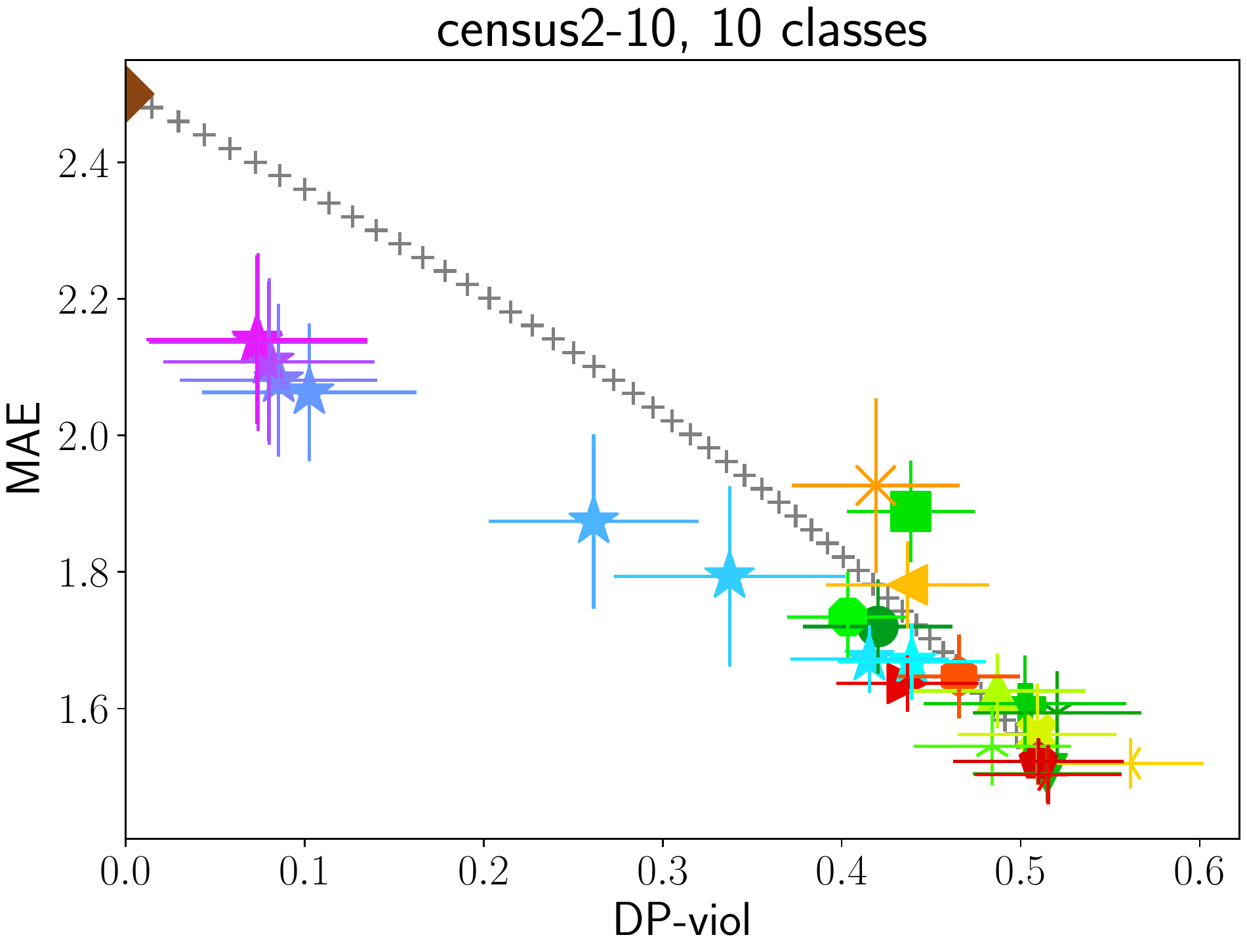}
    \hspace{\abstA}
    \includegraphics[scale=\scaleparameterA]{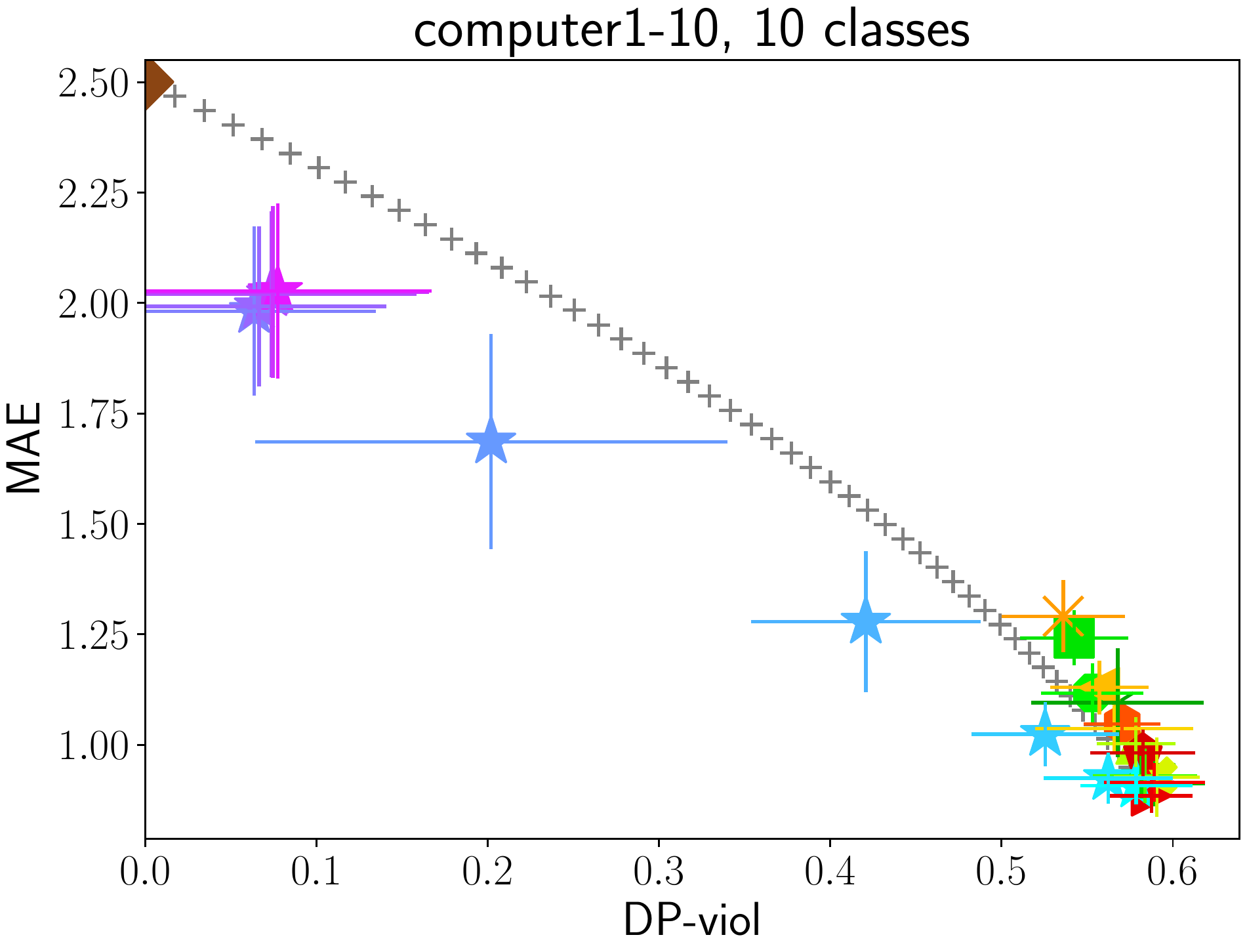}
    \hspace{\abstA}
    \includegraphics[scale=\scaleparameterA]{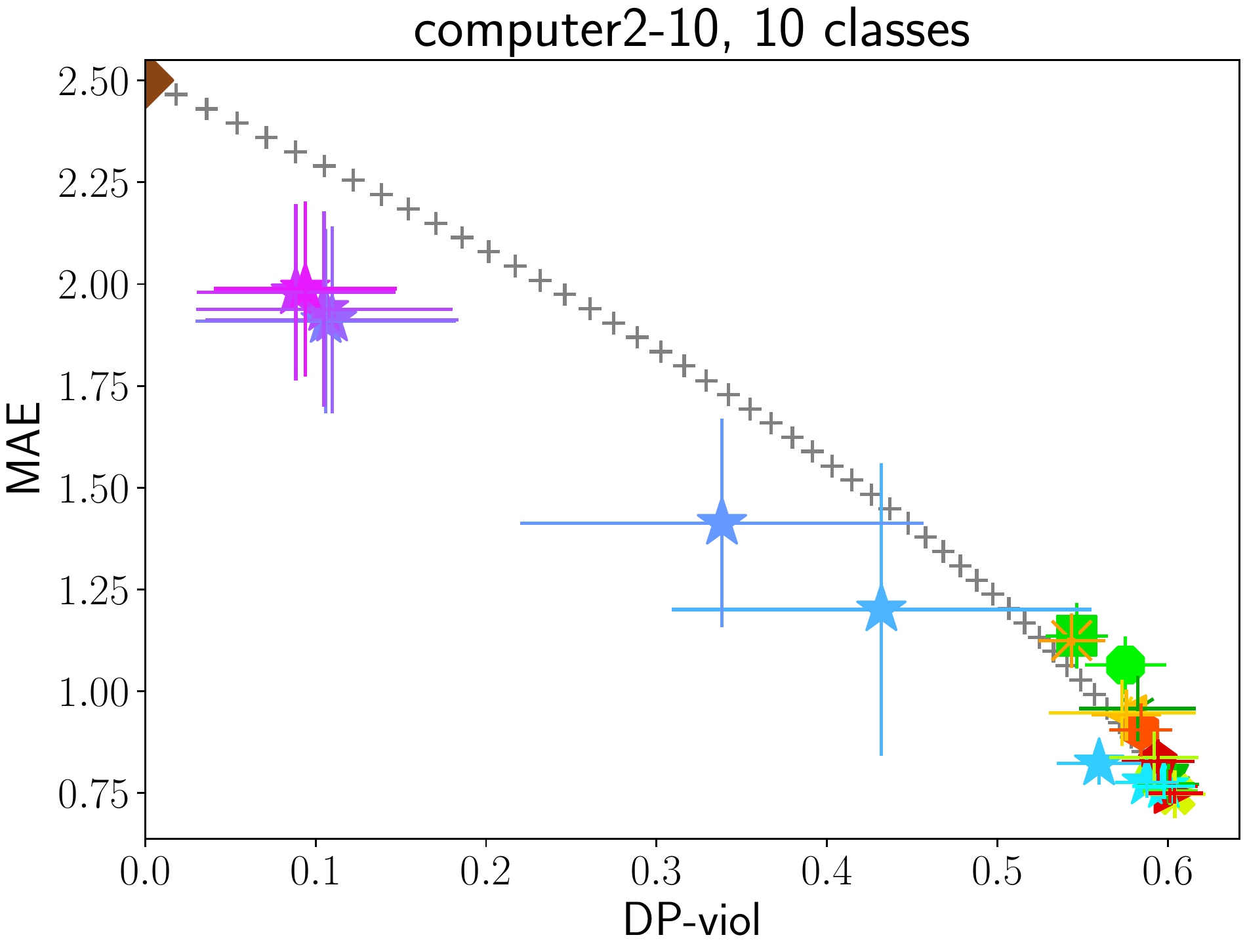}
    
    \includegraphics[scale=\scaleparameterA]{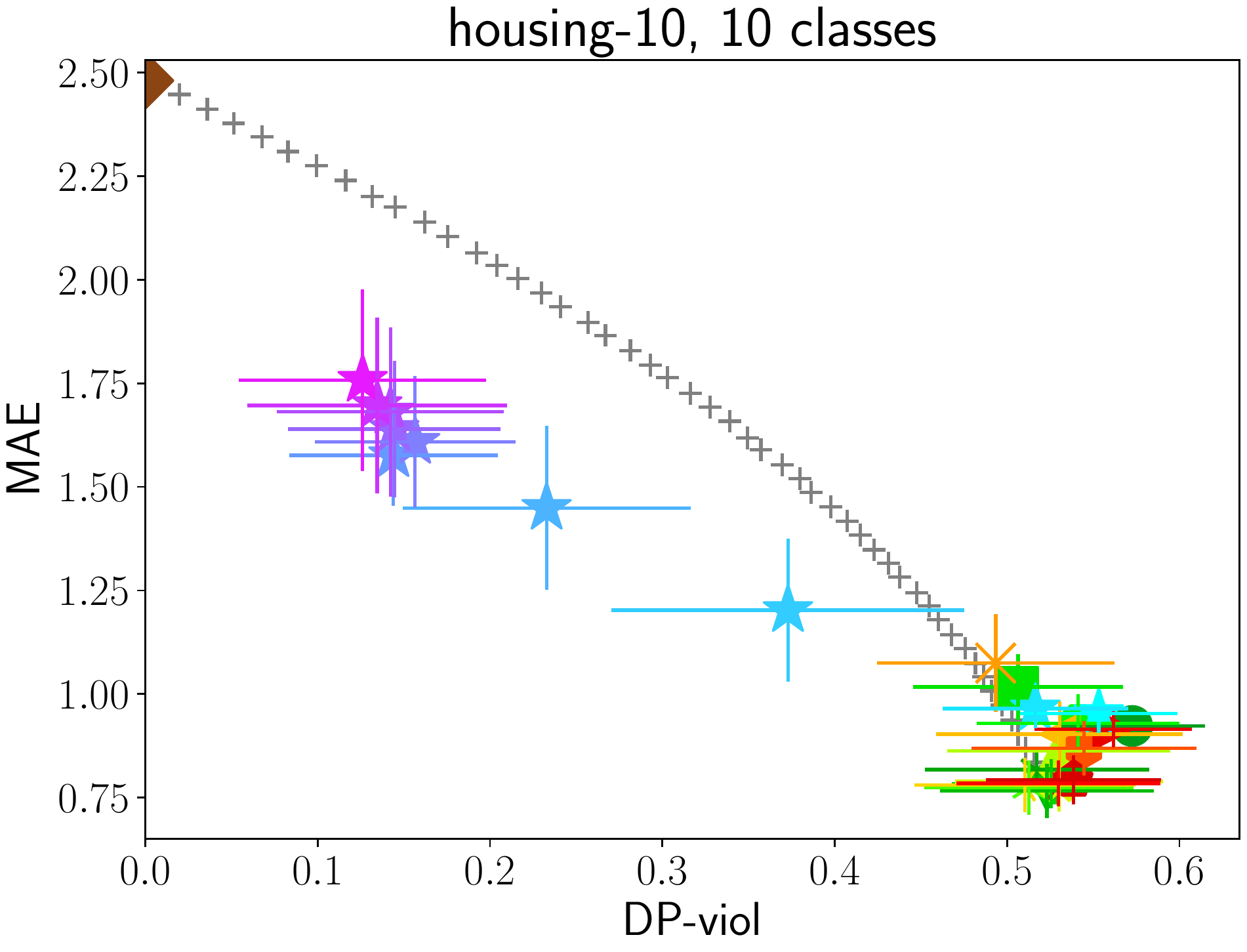}
    \hspace{\abstA}
    \includegraphics[scale=\scaleparameterA]{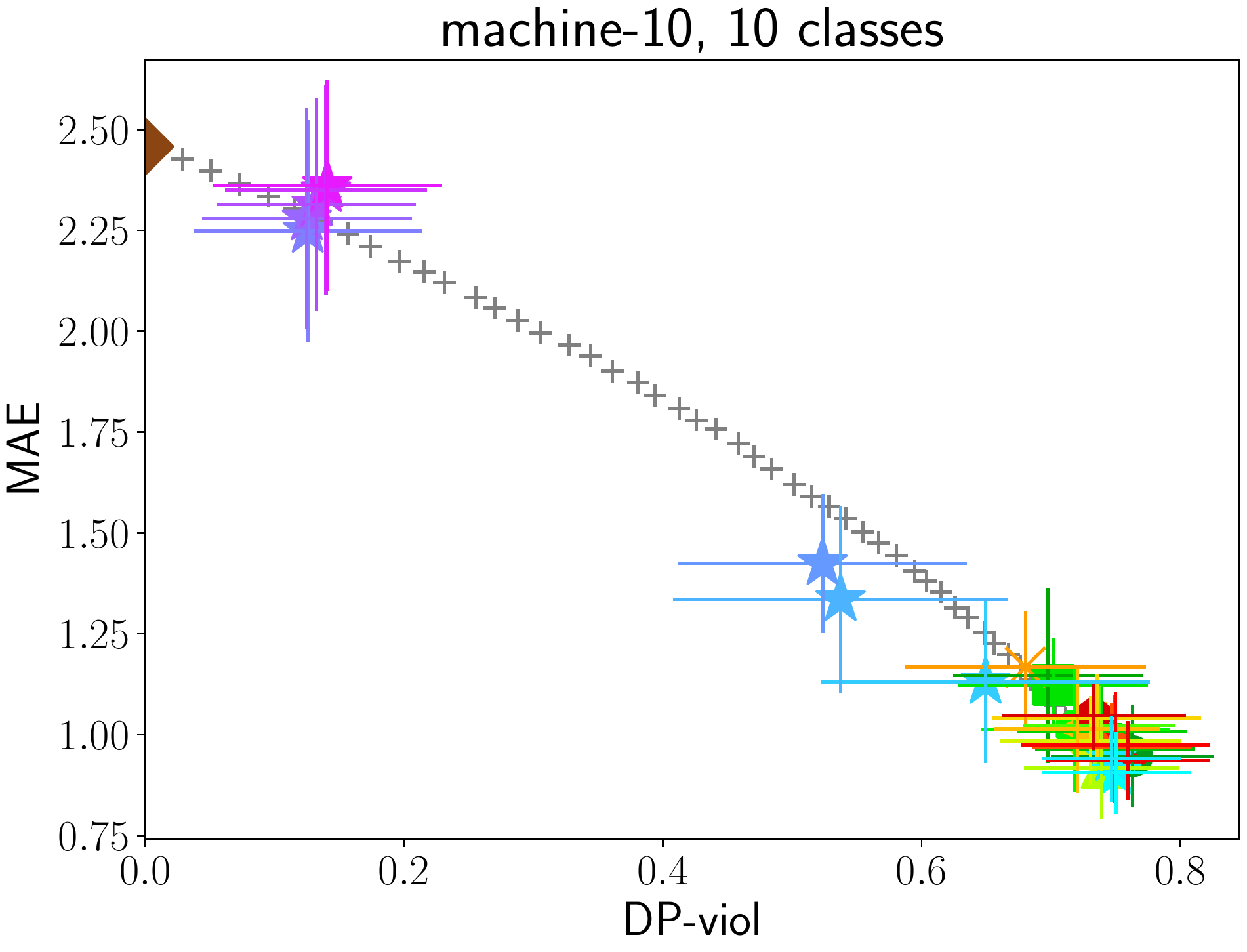}
    \hspace{\abstA}
    \includegraphics[scale=\scaleparameterA]{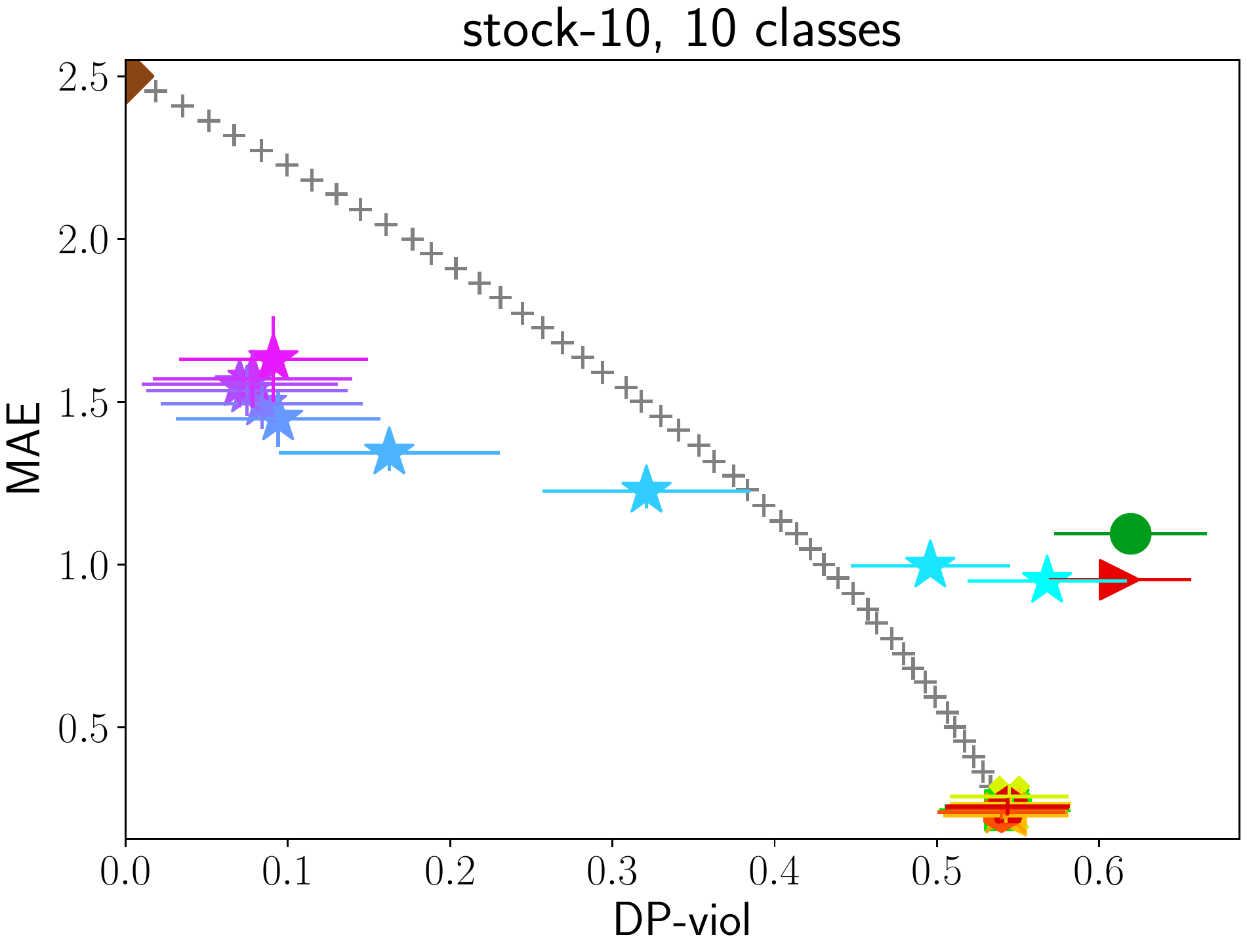}

    \caption{Experiments of Section~\ref{subsection_experiment_comparison} on the \textbf{discretized  regression datasets with 10 classes} when aiming for \textbf{pairwise DP}. The errorbars show the standard deviation over the 20 splits into training and test sets.}
    \label{fig:exp_comparison_APPENDIX_DISC_10classes_DP_with_STD}
\end{figure*}

\begin{figure*}
    
    \includegraphics[width=\linewidth]{experiment_real_ord/legend_big.pdf}

    \includegraphics[scale=\scaleparameterA]{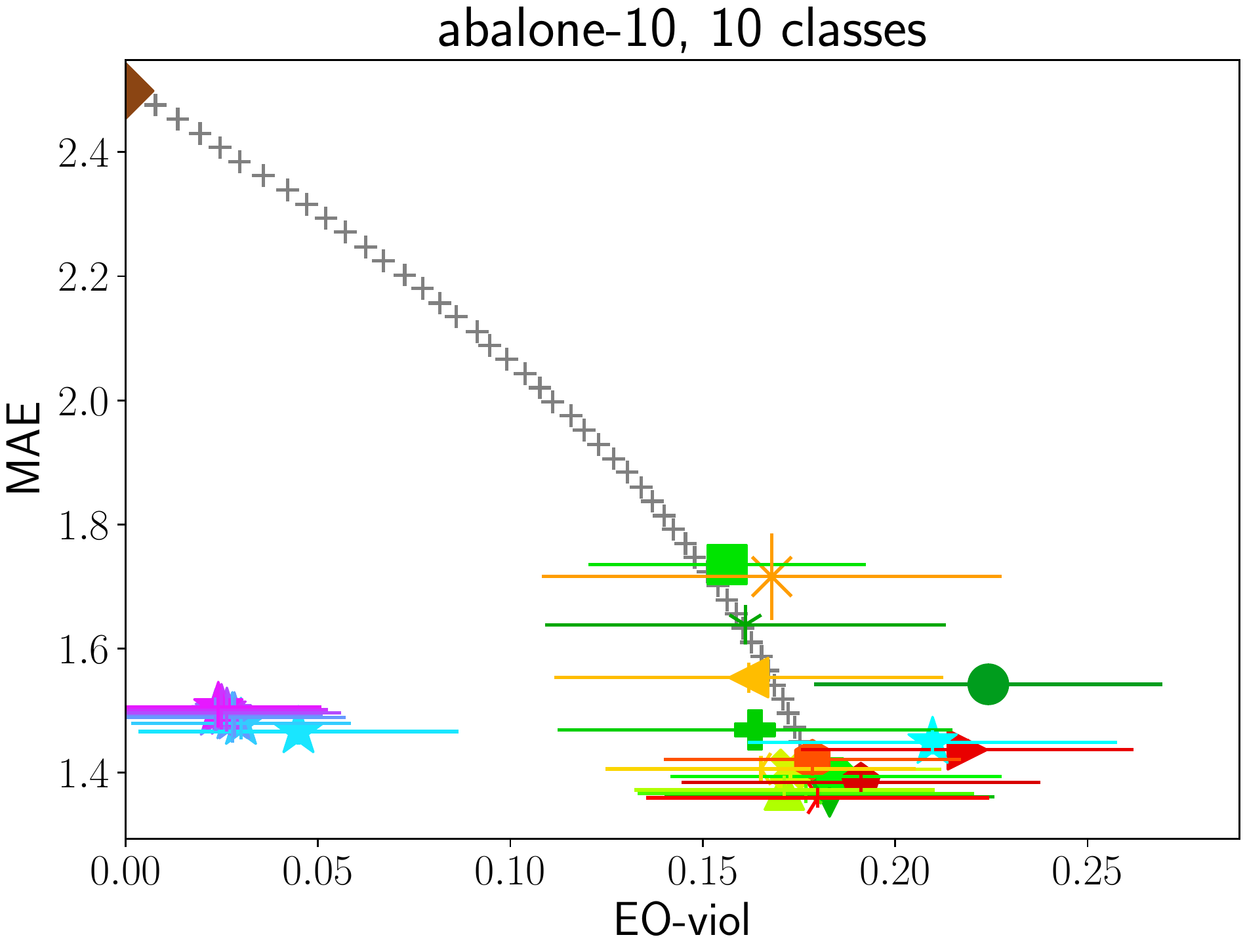}
    \hspace{\abstA}
    \includegraphics[scale=\scaleparameterA]{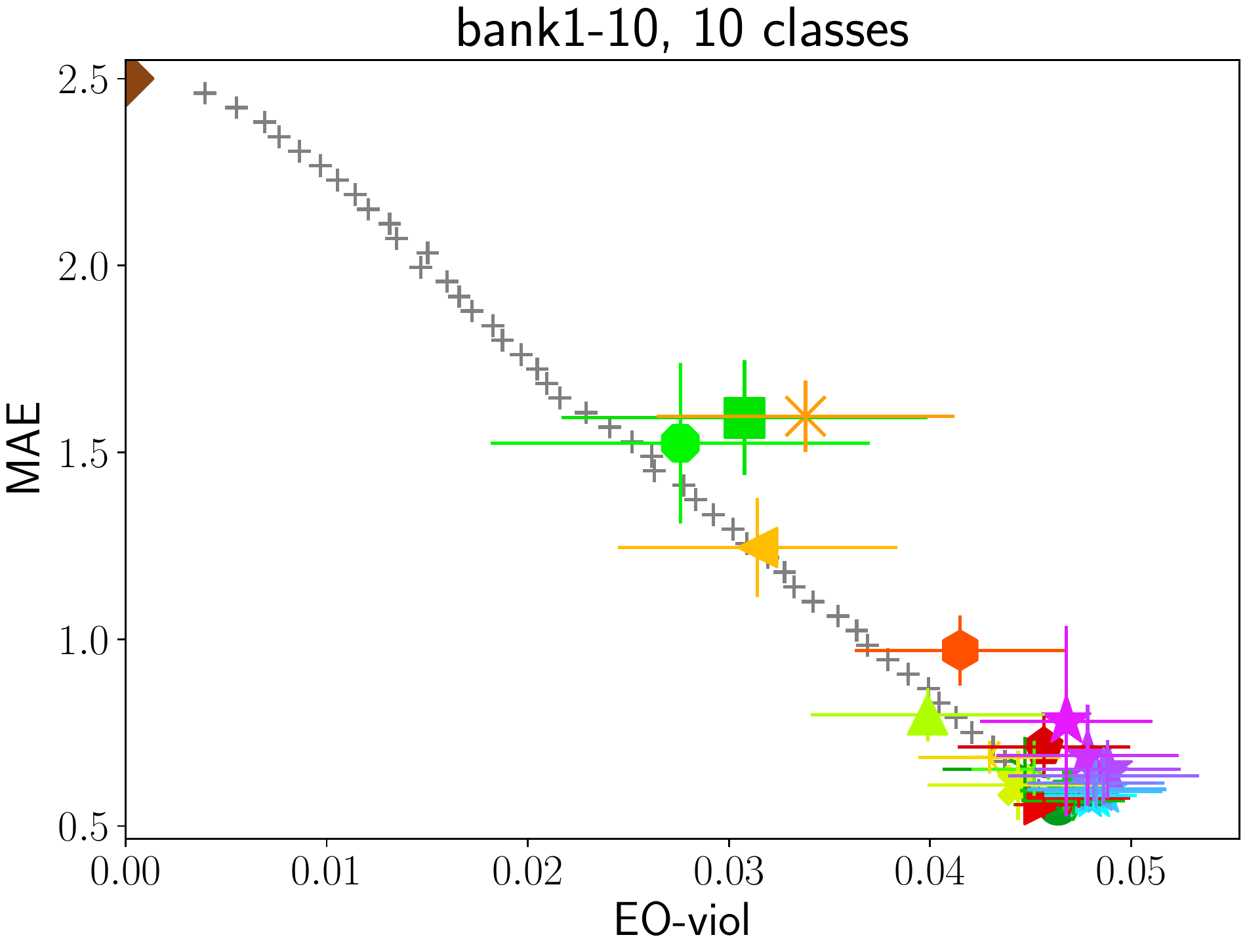}
    \hspace{\abstA}
    \includegraphics[scale=\scaleparameterA]{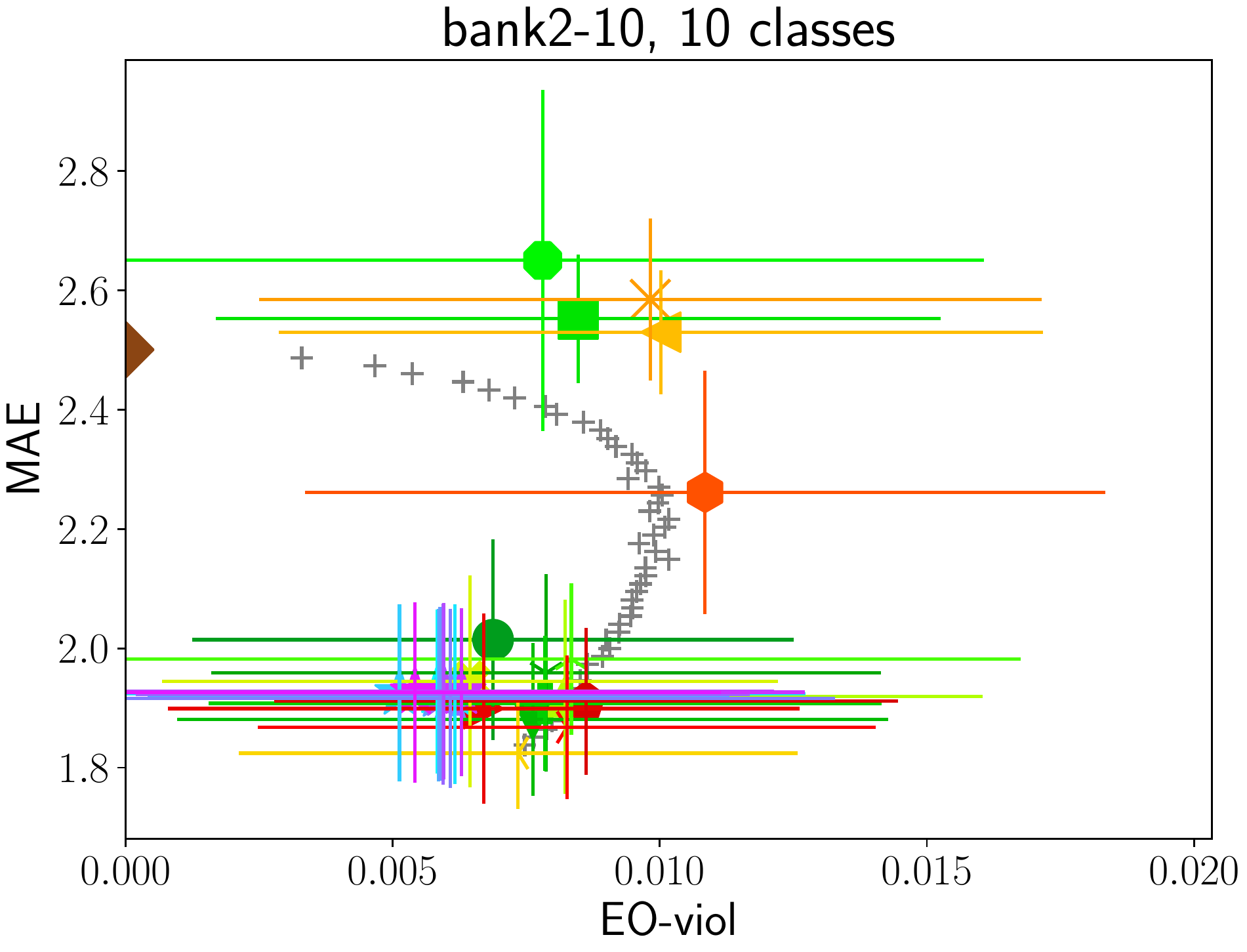}
    \hspace{\abstA}
    \includegraphics[scale=\scaleparameterA]{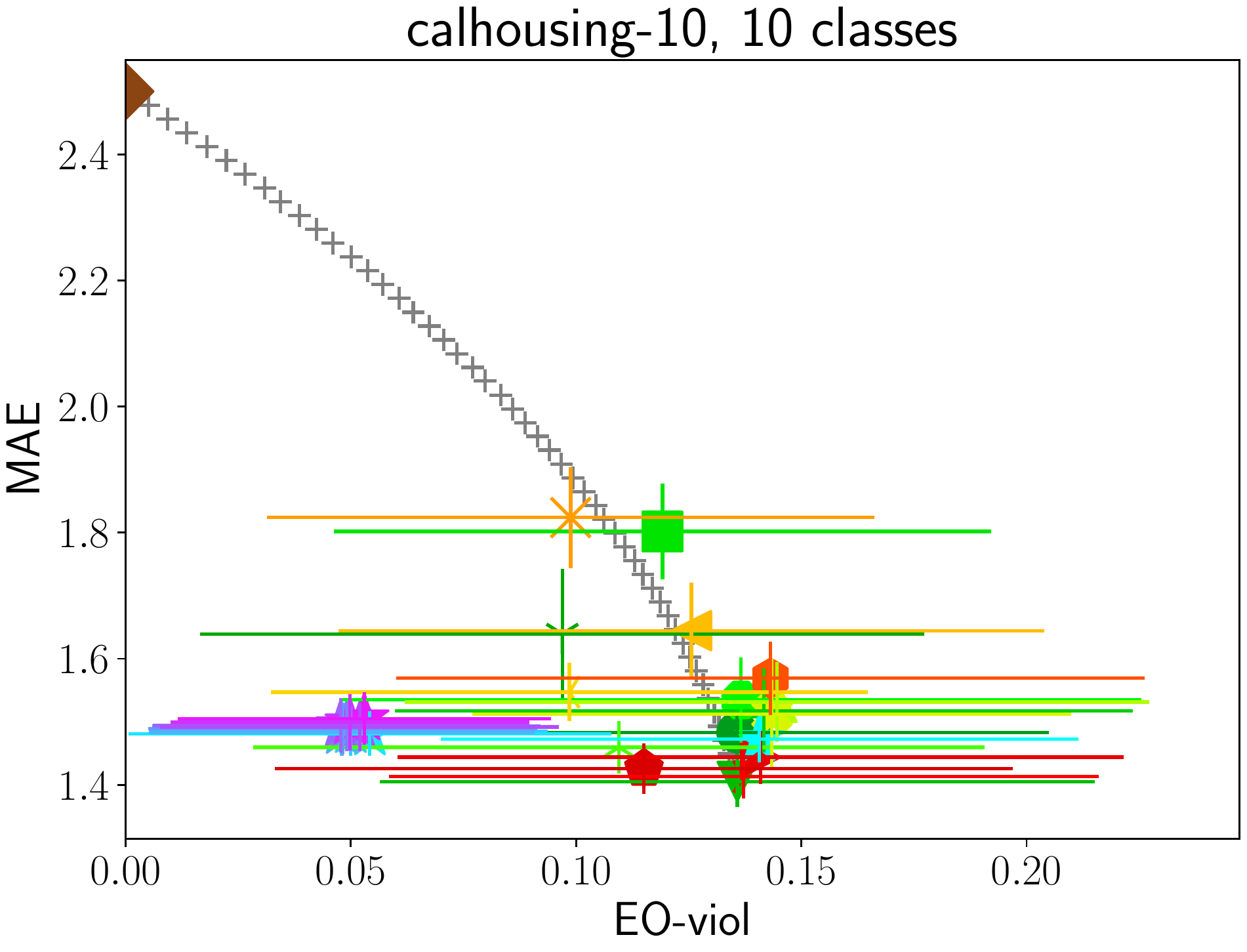}
    
    \includegraphics[scale=\scaleparameterA]{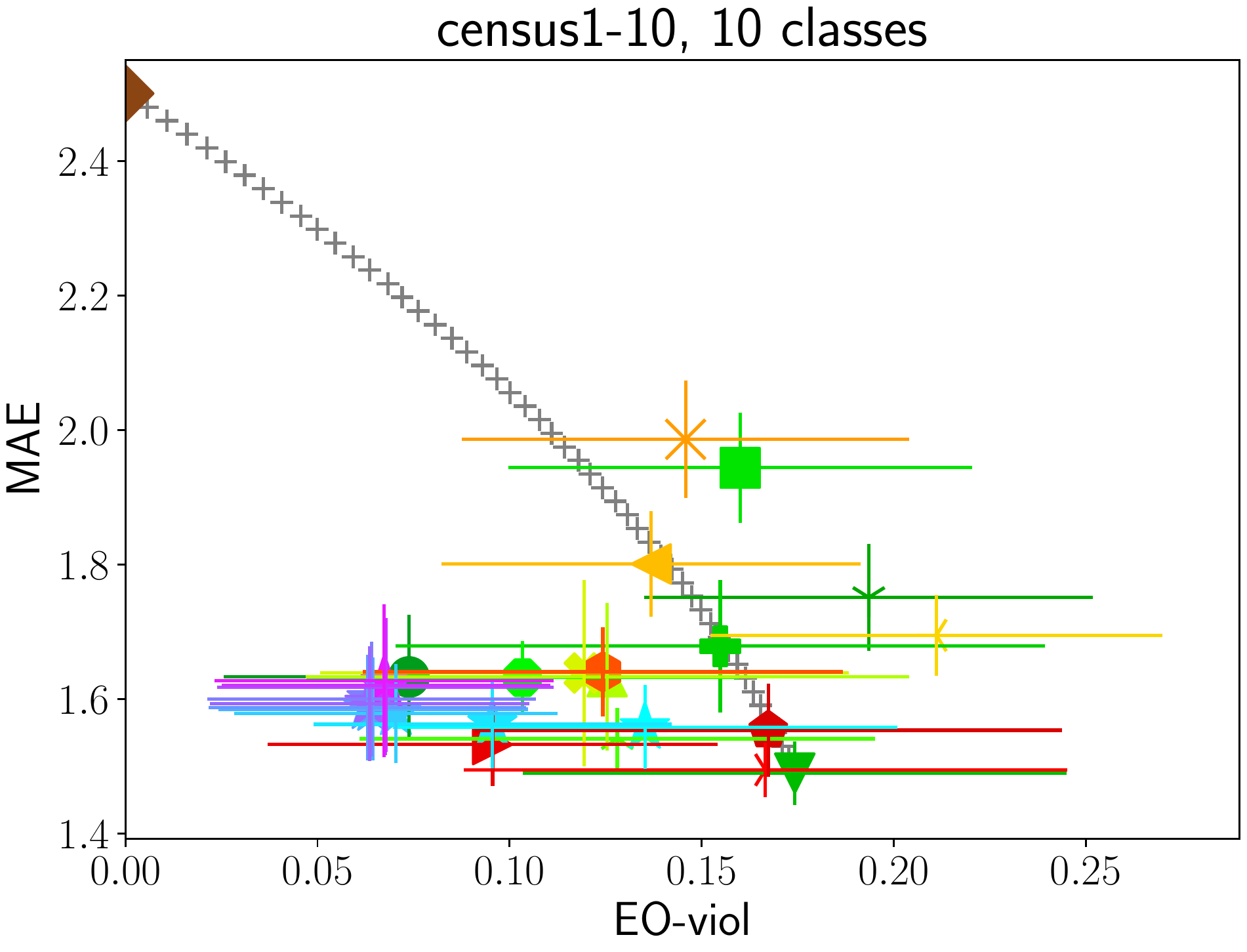}
    \hspace{\abstA}
    \includegraphics[scale=\scaleparameterA]{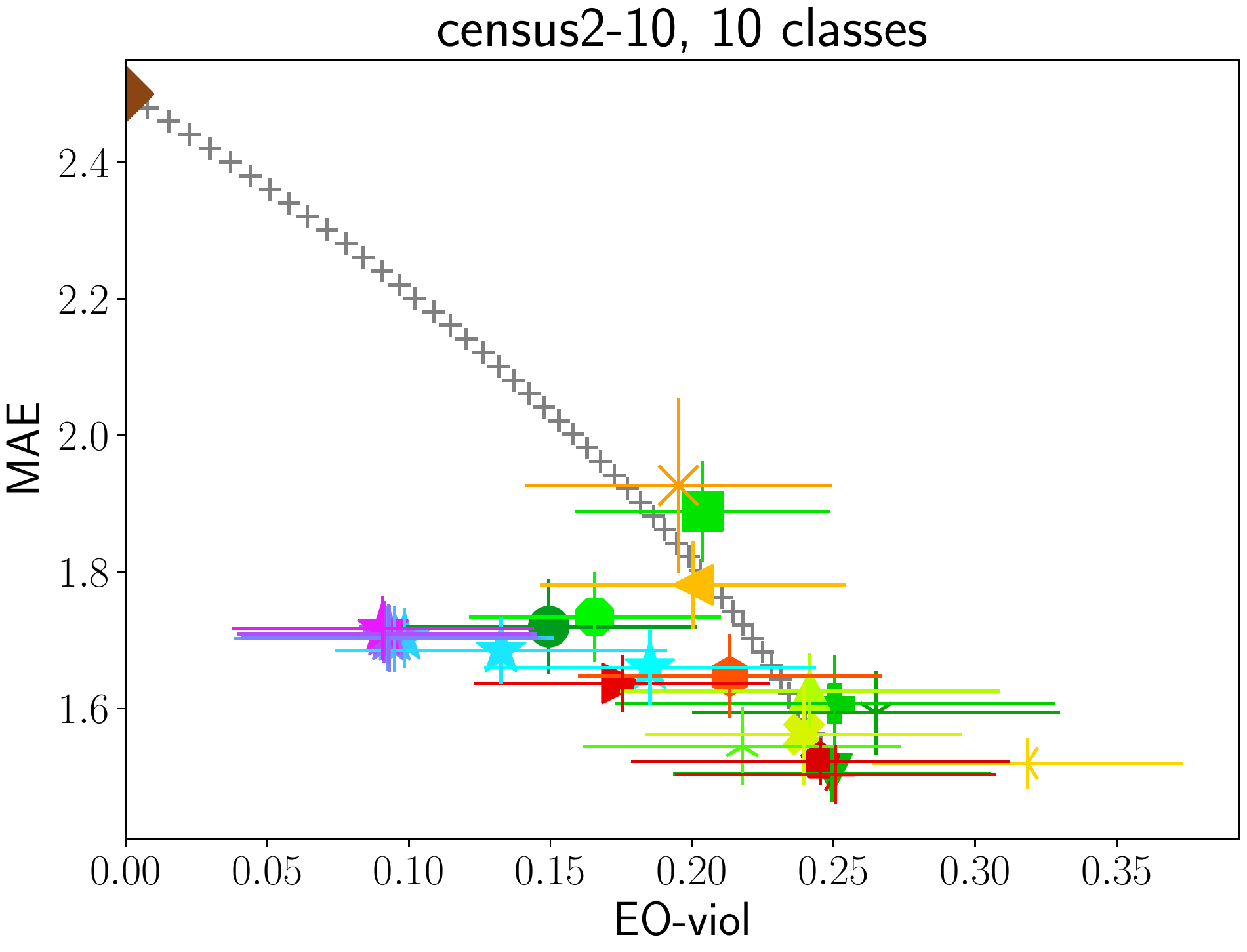}
    \hspace{\abstA}
    \includegraphics[scale=\scaleparameterA]{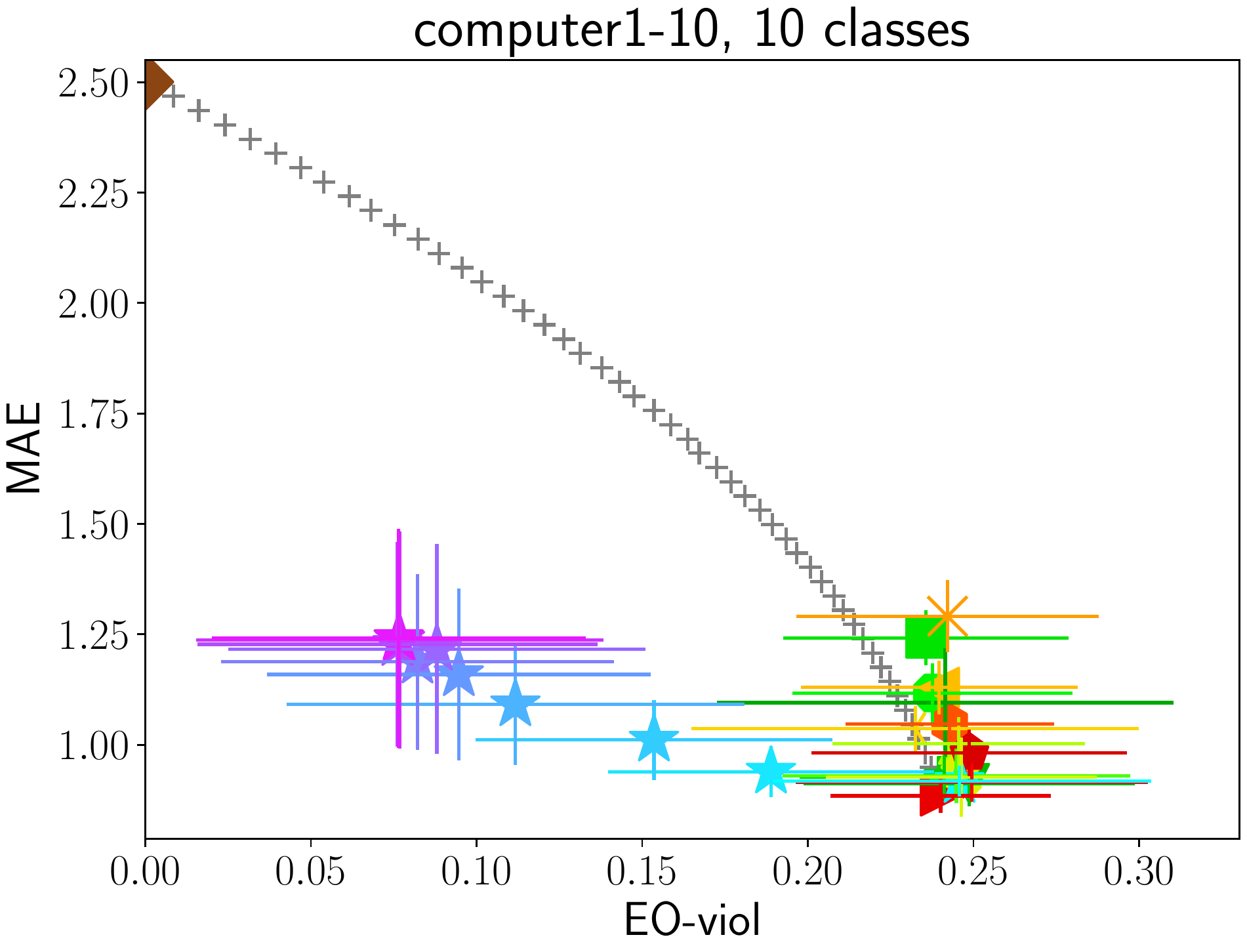}
    \hspace{\abstA}
    \includegraphics[scale=\scaleparameterA]{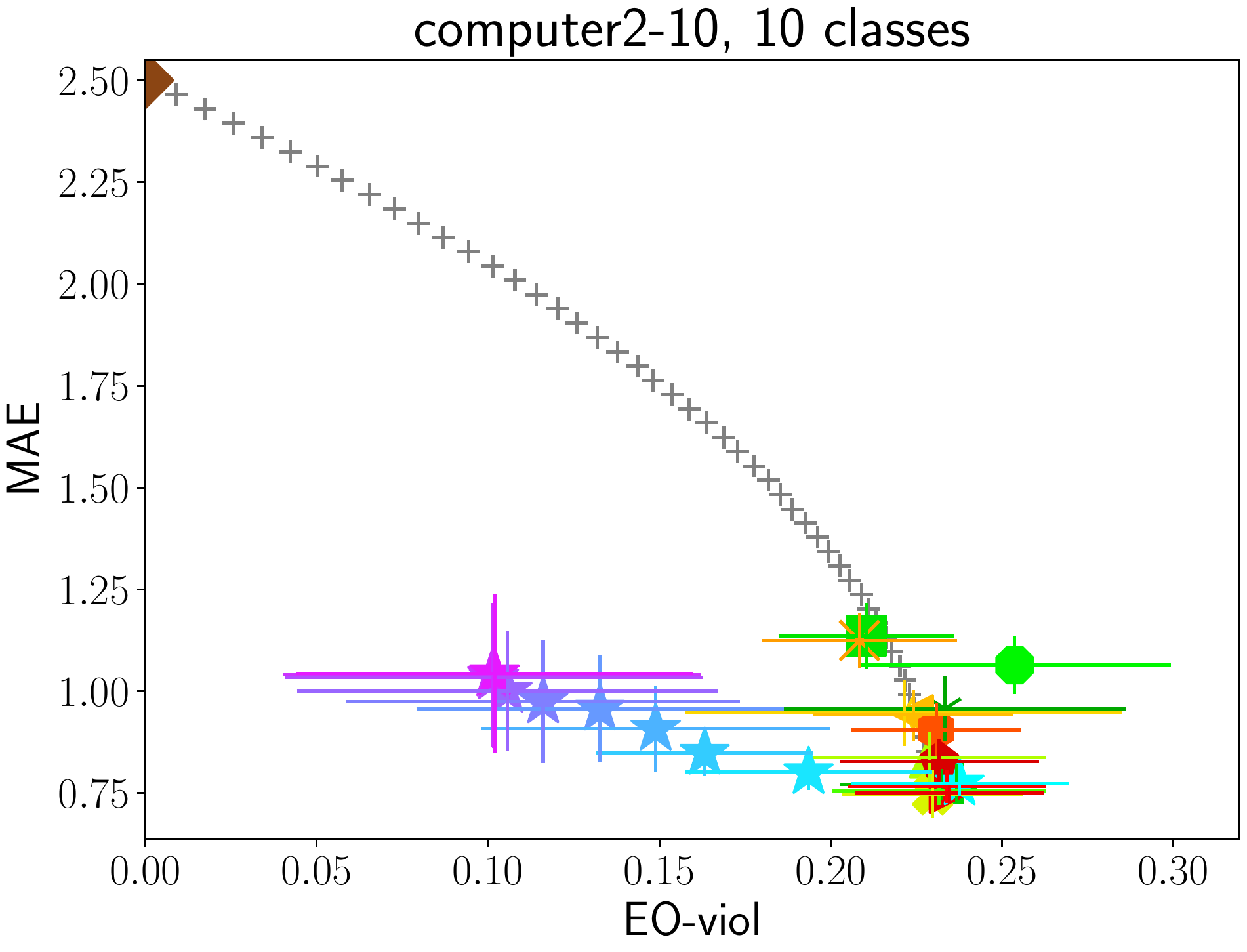}
    
    \includegraphics[scale=\scaleparameterA]{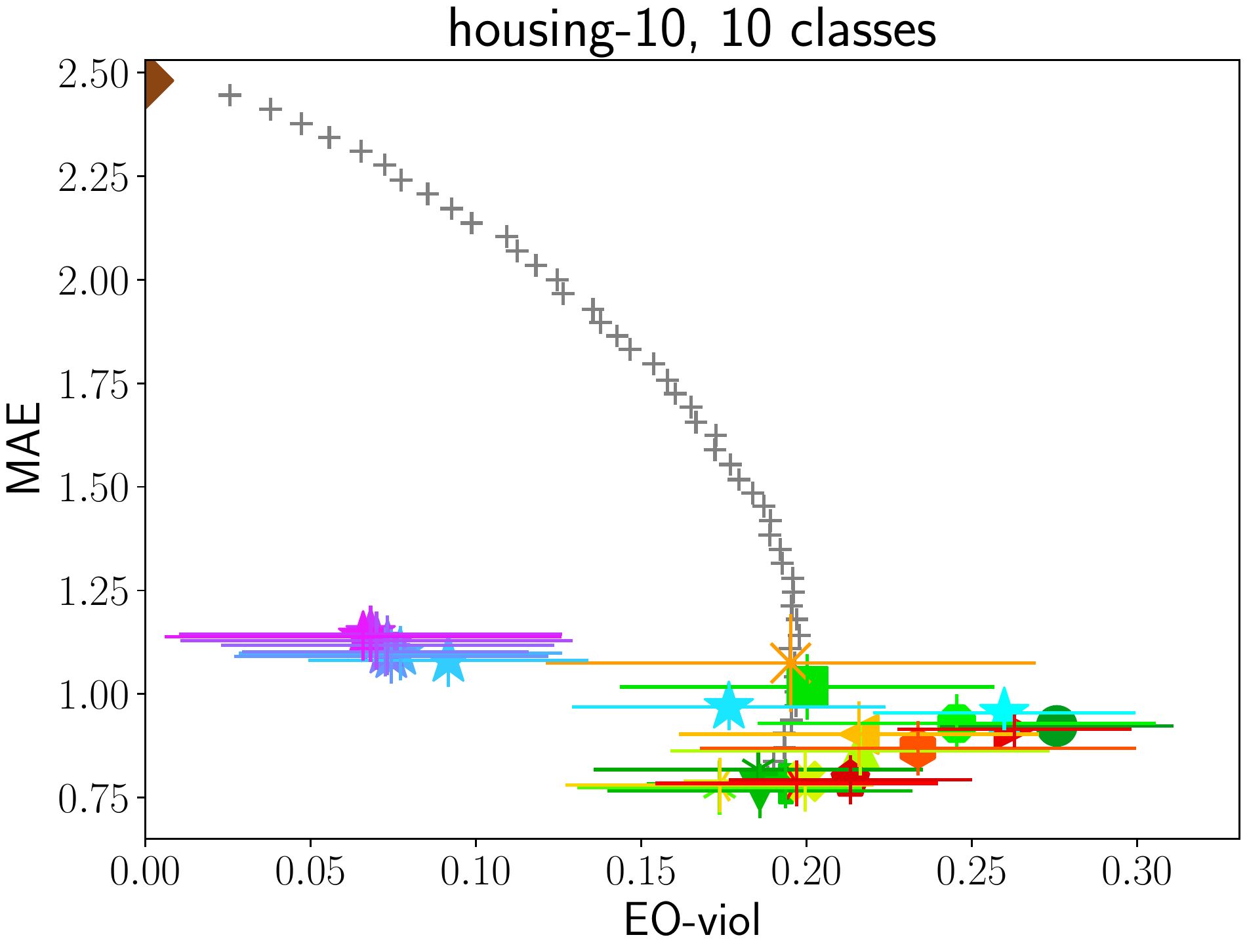}
    \hspace{\abstA}
    \includegraphics[scale=\scaleparameterA]{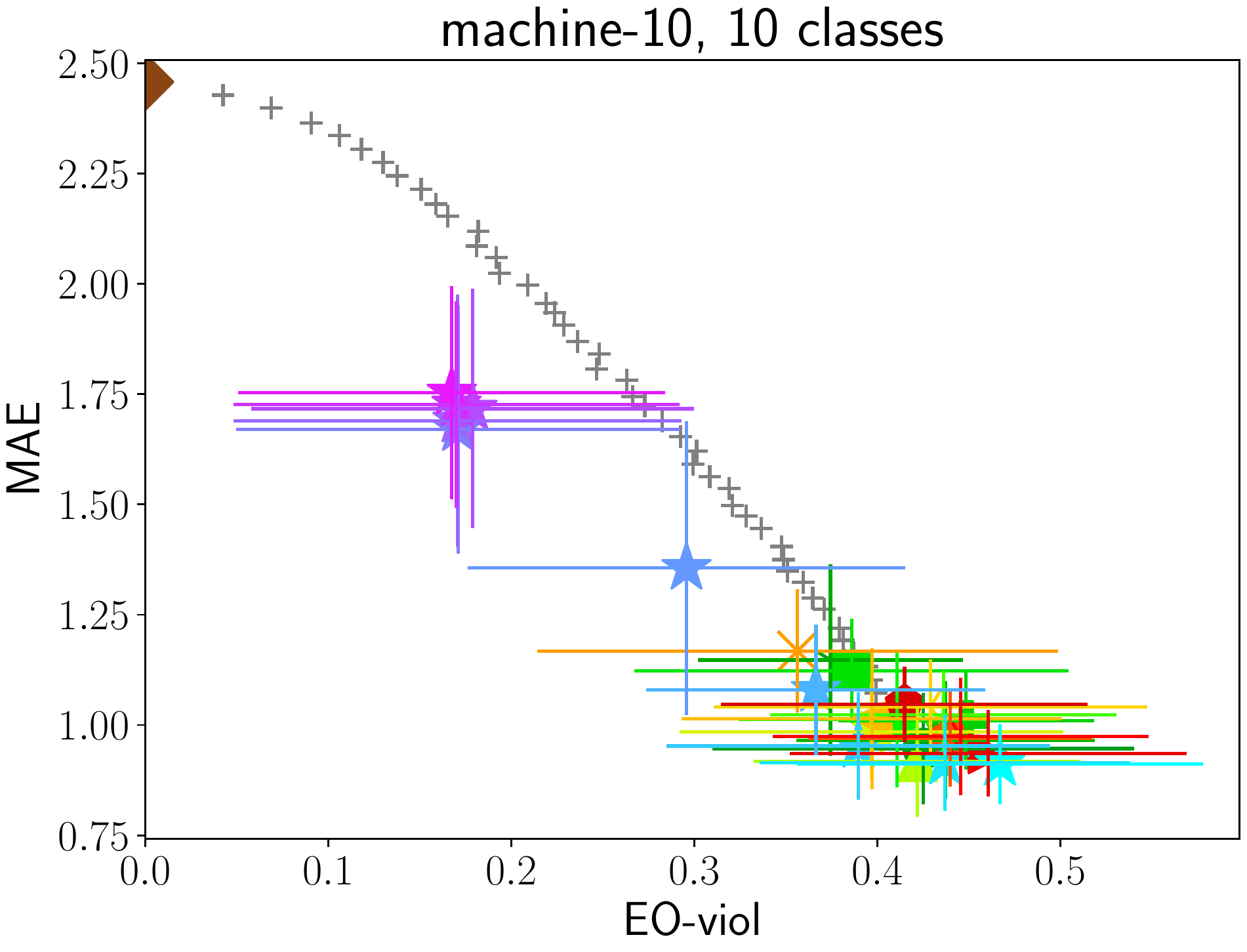}
    \hspace{\abstA}
    \includegraphics[scale=\scaleparameterA]{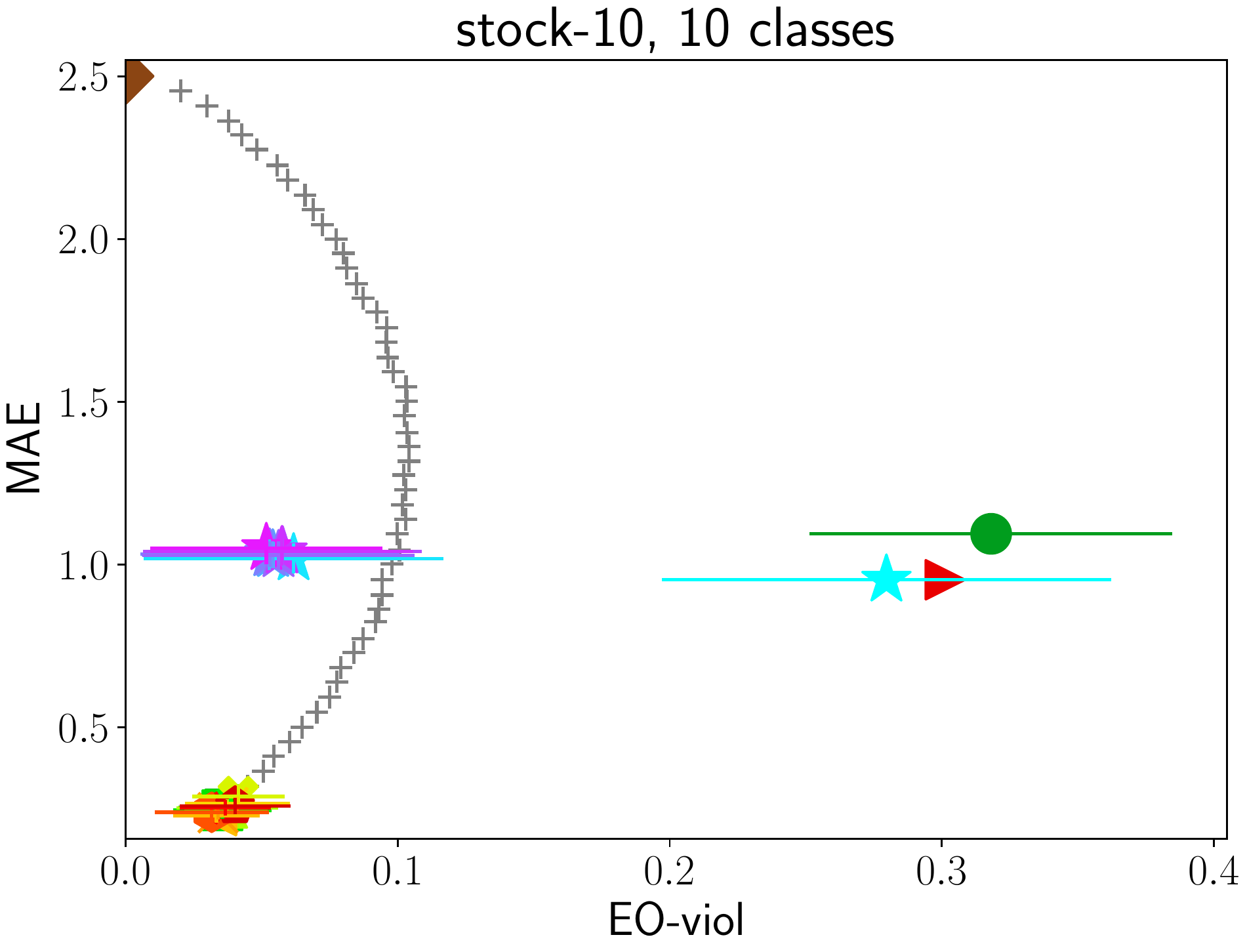}

    \caption{Experiments of Section~\ref{subsection_experiment_comparison} on the \textbf{discretized  regression datasets with 10 classes} when aiming for \textbf{pairwise EO}. The errorbars show the standard deviation over the 20 splits into training and test sets.}
    \label{fig:exp_comparison_APPENDIX_DISC_10classes_EO_with_STD}
\end{figure*}

\end{document}